\documentclass[12pt]{report}
\usepackage{suthesis-2e}  

\usepackage{graphicx}
\usepackage[hidelinks]{hyperref}
\usepackage{natbib}
\usepackage[T1]{fontenc}
\usepackage[utf8]{inputenc}
\usepackage{microtype}
\usepackage{longtable}
\usepackage{tabularx}

\newif\ifcomments
\commentstrue

\ifcomments
    \providecommand{\nfl}[1]{{\protect\color{orange}{[NL: #1]}}}
\else
    \providecommand{\nfl}[1]{}
\fi

\usepackage{microtype}
\usepackage{booktabs}
\usepackage{xspace}
\usepackage{amsmath}
\usepackage{amsfonts}       
\usepackage{array}
\usepackage{multirow}
\usepackage{makecell}
\usepackage{enumitem}
\usepackage{xparse}
\usepackage[dvipsnames,table]{xcolor}
\usepackage{caption}
\usepackage{subcaption}
\usepackage{framed}
\usepackage{soul}
\usepackage[utf8]{inputenc}
\usepackage{pifont}
\usepackage{xargs}          %
\usepackage{wrapfig,lipsum,booktabs}
\usepackage{enumitem}
\usepackage{longtable}
\usepackage{makecell}
\usepackage{graphicx}
\usepackage{subcaption}

\usepackage{pgfplots}
\usetikzlibrary{pgfplots.groupplots}
\pgfplotsset{compat=1.3}
\usepackage{tikz}
\usetikzlibrary{patterns}

\usepackage[most]{tcolorbox}

\usepackage{listings}
\def\Snospace~{\S{}}

\def\Snospace~{\S{}}

\newcommand\FigTop[4]{\begin{figure}[t] \begin{center} \includegraphics[scale=#2]{#1} \end{center} \caption{\label{fig:#3} #4} \end{figure}}

\newcommand{\1}{\mathbb{I}} 

\usepackage{lscape}
\usepackage{pifont}
\newcommand{\cmark}{\ding{51}}%
\newcommand{\xmark}{\ding{55}}%

\newcommand\systemlevel{ecosystem-level\xspace}
\newcommand\Systemlevel{Ecosystem-level\xspace}
\newcommand{\baseline}{baseline\xspace}

\newcommand\profilepolarization{homogeneous outcomes\xspace}

\newcommand\polarized{homogenous\xspace}

\newcommand\polaroutcomes{homogeneous outcomes\xspace}

\newcommand\outcomeprofile{outcome profile\xspace}

\newcommand{\hapi}{\textbf{HAPI}\xspace}

\newcommand{\digit}{\textsc{digit}\xspace} 
\newcommand{\fluent}{\textsc{fluent}\xspace} 
\newcommand{\amnist}{\textsc{amnist}\xspace} 
\newcommand{\yelp}{\textsc{yelp}\xspace} 
\newcommand{\imdb}{\textsc{imdb}\xspace} 
\newcommand{\shop}{\textsc{shop}\xspace} 
\newcommand{\waimai}{\textsc{waimai}\xspace} 

\newcommand{\ferplus}{\textsc{fer+}\xspace} 
\newcommand{\afnet}{\textsc{afnet}\xspace} 
\newcommand{\expw}{\textsc{expw}\xspace} 
\newcommand{\rafdb}{\textsc{rafdb}\xspace} 

\newcommand\projectname{{Foundation Model Transparency Index}\xspace}

\newcommand\informationfreezedate{September 15, 2023}
\newcommand\numcompanies{10\xspace}
\newcommand\numindicators{100\xspace}
\newcommand\numcells{1000\xspace}
\newcommand\numsubdomains{23\xspace}
\newcommand\numupstreamindicators{32\xspace}
\newcommand\numupstreamsubdomains{6\xspace}
\newcommand\nummodelindicators{33\xspace}
\newcommand\nummodelsubdomains{8\xspace}
\newcommand\numdownstreamindicators{35\xspace}
\newcommand\numdownstreamsubdomains{9\xspace}
\newcommand\nummajorsubdomains{13\xspace}

\newcommand\numdomains{3\xspace}
\newcommand\numgraders{2\xspace}

\newcommand\numfeasible{{82}\xspace}
\newcommand\numfeasiblemultiple{{71}\xspace}

\newcommand\scorerange{{42}\xspace}
\newcommand\minscore{{12}\xspace}
\newcommand\maxscore{{54}\xspace}
\newcommand\meanscore{{37}\xspace}
\newcommand\stdev{{14.2}}

\newcommand\data{Data\xspace}
\newcommand\labor{Data Labor\xspace}
\newcommand\dataaccess{Data Access\xspace}
\newcommand\compute{Compute\xspace}
\newcommand\methods{Methods\xspace}
\newcommand\datamitigations{Data Mitigations\xspace}
\newcommand\modelbasics{Model Basics\xspace}
\newcommand\modelaccess{Model Access\xspace}
\newcommand\capabilities{Capabilities\xspace}
\newcommand\limitations{Limitations\xspace}
\newcommand\risks{Risks\xspace}
\newcommand\modelmitigations{Model Mitigations\xspace}
\newcommand\trustworthiness{Trustworthiness\xspace}
\newcommand\inference{Inference\xspace}
\newcommand\distribution{Distribution\xspace}
\newcommand\usagepolicy{Usage Policy\xspace}
\newcommand\modelbehaviorpolicy{Model Behavior Policy\xspace}
\newcommand\interface{User Interface\xspace}
\newcommand\dataprotection{User Data Protection\xspace}
\newcommand\updates{Model Updates\xspace}
\newcommand\feedback{Feedback\xspace}
\newcommand\impact{Impact\xspace}
\newcommand\documentation{Downstream Documentation\xspace}

\newcommand\tldrDone[1]{}

\newcommand\eg{e.g.\xspace}
\newcommand\ie{i.e.\xspace}
\newcommand\cf{cf.\xspace}
\newcommand\dash{\xspace---\xspace}

\newcommand\benchmarkname{HELM\xspace}
\newcommand\core{core\xspace}

\newcommand\numcorescenarios{16\xspace}

\newcommand\nummetriccategories{7\xspace}

\newcommand\numnonaccuracymetriccategories{6\xspace}
\newcommand\nummodels{30\xspace}

\newcommand\nummodelfamilies{16\xspace}
\newcommand\nummodelcreators{12\xspace}

\newcommand\gpuhours{GPU hours\xspace}

\newcommand\numincontext{5\xspace}
\newcommand\numruns{3\xspace}

\newcommand\aitwentyone{AI21 Labs\xspace}
\newcommand\amazon{Amazon\xspace}
\newcommand\anthropic{Anthropic\xspace}
\newcommand\cohere{Cohere\xspace}
\newcommand\google{Google\xspace}
\newcommand\huggingface{Hugging Face\xspace}

\newcommand\inflection{Inflection\xspace}
\newcommand\meta{Meta\xspace}
\newcommand\openai{OpenAI\xspace}
\newcommand\stability{Stability AI\xspace}

\newcommand\jurassic{Jurassic-2\xspace}
\newcommand\titan{Titan Text\xspace}
\newcommand\claude{Claude 2\xspace}
\newcommand\command{Command\xspace}
\newcommand\palm{PaLM 2\xspace}
\newcommand\bloomz{BLOOMZ\xspace}
\newcommand\inflectionone{Inflection-1\xspace}
\newcommand\llama{Llama 2\xspace}
\newcommand\gptfour{GPT-4\xspace}
\newcommand\stablediffusion{Stable Diffusion 2\xspace}
\newcommand\jurassicjumbo{J1-Jumbo v1 (178B)\xspace}
\newcommand\jurassicgrande{J1-Grande v1 (17B)\xspace}
\newcommand\jurassiclarge{J1-Large v1 (7.5B)\xspace}

\newcommand\anthropiclm{Anthropic-LM v4-s3 (52B)\xspace}

\newcommand\bloom{BLOOM (176B)\xspace}
\newcommand\tzero{T0++ (11B)\xspace}

\newcommand\coherexl{Cohere xlarge v20220609 (52.4B)\xspace}
\newcommand\coherel{Cohere large v20220720 (13.1B)\xspace}
\newcommand\coherem{Cohere medium v20220720 (6.1B)\xspace}
\newcommand\coheres{Cohere small v20220720 (410M)\xspace}

\newcommand\gptj{GPT-J (6B)\xspace}
\newcommand\gptneox{GPT-NeoX (20B)\xspace}

\newcommand\tfive{T5 (11B)\xspace}
\newcommand\ultwo{UL2 (20B)\xspace}

\newcommand\optsixsix{OPT (66B)\xspace}
\newcommand\optonesevenfive{OPT (175B)\xspace}

\newcommand\mtnlgseven{TNLG v2 (6.7B)\xspace}
\newcommand\mtnlgfivethreezero{TNLG v2 (530B)\xspace}

\newcommand\gptdavinci{davinci (175B)\xspace}
\newcommand\gptcurie{curie (6.7B)\xspace}
\newcommand\gptbabbage{babbage (1.3B)\xspace}
\newcommand\gptada{ada (350M)\xspace}
\newcommand\instructdavinci{text-davinci-002\xspace}
\newcommand\instructcurie{text-curie-001\xspace}
\newcommand\instructbabbage{text-babbage-001\xspace}
\newcommand\instructada{text-ada-001\xspace}
\newcommand\davincicodex{code-davinci-002\xspace}
\newcommand\cushmancodex{code-cushman-001 (12B)\xspace}

\newcommand\glm{GLM (130B)\xspace}

\newcommand\yalm{YaLM (100B)\xspace}

\newcommand\dataset[1]{\textbf{#1}\xspace}

\newcommand\naturalquestions{\dataset{NaturalQuestions}}

\newcommand\narrativeqa{\dataset{NarrativeQA}}
\newcommand\quac{\dataset{QuAC}}
\newcommand\boolq{\dataset{BoolQ}}
\newcommand\hellaswag{\dataset{HellaSwag}}
\newcommand\openbookqa{\dataset{OpenBookQA}}
\newcommand\truthfulqa{\dataset{TruthfulQA}}
\newcommand\mmlu{\dataset{MMLU}}
\newcommand\msmarcoregular{\dataset{MS MARCO (regular)}}
\newcommand\msmarcotrec{\dataset{MS MARCO (TREC)}}
\newcommand\cnndm{\dataset{CNN/DailyMail}}
\newcommand\xsum{\dataset{XSUM}}
\newcommand\civilcomments{\dataset{CivilComments}}
\newcommand\raft{\dataset{RAFT}}

\newcommand\pile{\dataset{The Pile}}

\NewDocumentCommand{\rot}{O{45} O{1em} m}{\makebox[#2][l]{\rotatebox{#1}{#3}}}%
\usepackage{numprint}
\npthousandsep{,}

\newcommand{\PreserveBackslash}[1]{\let\temp=\\#1\let\\=\temp}
\newcolumntype{C}[1]{>{\PreserveBackslash\centering}p{#1}}
\newcolumntype{R}[1]{>{\PreserveBackslash\raggedleft}p{#1}}
\newcolumntype{L}[1]{>{\PreserveBackslash\raggedright}p{#1}}

\definecolor{lightgray}{gray}{0.9}
\definecolor{passage}{HTML}{FFC7BF}
\definecolor{answer}{HTML}{CFFFCC}
\definecolor{question}{HTML}{E3C0FF}
\definecolor{question1}{HTML}{B6D7A8}
\definecolor{question2}{HTML}{FFE599}
\definecolor{question3}{HTML}{D5A6BD}

\usepackage{booktabs} 
\usepackage{tabularx}
\usepackage{adjustbox}
\usepackage{framed}
\usepackage{enumitem}
\usepackage{ragged2e}
\newcolumntype{P}[1]{>{\RaggedRight\hspace{0pt}}m{#1}}
\usepackage{color, colortbl}
\definecolor{Gray}{gray}{0.94}
\usepackage{pifont}
\renewcommand{\cmark}{\ding{51}}%
\renewcommand{\xmark}{\ding{55}}%
\usepackage[capitalize]{cleveref}
\usepackage{tcolorbox}
\usepackage{xspace}
\usepackage{xcolor}
\tcbuselibrary{skins}
\usepackage{caption}
\usepackage{mwe}
\usepackage{caption}
\usepackage{subcaption} 
\usepackage{tikz}
\usepackage{amsmath}
\usepackage{amssymb}
\usepackage{mathtools}
\usepackage{amsthm}
\usepackage{wasysym}
\usepackage{subcaption}
\newcommand{\wcircle}{{$\RIGHTcircle$}}
\newcommand{\bcircle}{{$\CIRCLE$}}
\newcommand{\ecircle}{{$\Circle$}}
\PassOptionsToPackage{hyphens}{url}\usepackage{hyperref}
\usepackage{xurl}

\definecolor{mintgreen}{RGB}{172,220,172}
\usepackage{multicol}

\begin{document}
\title{The Societal Impact of Foundation Models:\\
Advancing Evidence-based AI Policy}
\author{Rishi Bommasani}
\dept{Computer Science}
\principaladviser{Percy Liang}
\firstreader{Daniel Jurafsky}
\secondreader{Christopher D. Manning}
 

\beforepreface
\prefacesection{Abstract}
Artificial intelligence is humanity's most promising technology because of the remarkable capabilities offered by foundation models.
Yet, the same technology brings confusion and consternation: foundation models are poorly understood and they may precipitate a wide array of harms.
This dissertation explains how technology and society coevolve in the age of AI, organized around three themes.
First, the conceptual framing: the capabilities, risks, and the supply chain that grounds foundation models in the broader economy.
Second, the empirical insights that enrich the conceptual foundations: transparency created via evaluations at the model level and indexes at the organization level.
Finally, the transition from understanding to action: superior understanding of the societal impact of foundation models advances evidence-based AI policy.
View together, this dissertation makes inroads into achieving better societal outcomes in the age of AI by building the scientific foundations and research-policy interface required for better AI governance.
\prefacesection{Acknowledgments}
My last five years at Stanford have been an incredible journey---one where I have grown tremendously, both academically and personally.\footnote{Several of the phrases in these acknowledgments are drawn from the theses and dissertations of important people in my life. 
I leave it as a playful exercise for anyone reading this dissertation to identify these excerpts and their original homes.}
None of this would have been possible without the support of the many people I have interacted with over the years.

I thank my family for their constant support throughout my research journey.
My maternal grandmother ammamma embodies warmth, wisdom, and vitality. 
She sets the standard in our family for excellence.
The path I have chosen, and the decisions along the way, may not make sense to my parents, but they accept that I will find my way through the world.
That has not always been easy for them, but I appreciate this as yet another gift they have given me to reach my goals. 
My confidence to be the person and scholar I want to be without exception begins with the trust I have in them: their sturdy foundation is why I do not fear failure.  

I thank my childhood teachers who fostered my intellectual curiosity and reverence for education, especially Mr. Lozada, Mrs. Gallo, Mrs. Stokrocki, Mrs. Wilson, Mrs. Powers, Mrs. Srinivas, and Mrs. Friedman.\footnote{School teachers who focus on education and university professors who focus on research often feel worlds apart to me. As I acknowledge my teachers, I am reminded that one of AI’s largest societal impacts to date, which I largely did not study in spite of being at an education institution, has been how it warped education. As I often build bridges between otherwise distant groups, I wonder if this AI-driven reckoning should be an invitation to bring these two groups closer. I leave this note to remind myself to reconsider this when I am a professor, drawing inspiration from Rob Reich’s past as a teacher, and to pursue teaching excellence irrespective of institutional incentives, drawing inspiration especially from Percy Liang’s teaching excellence.}

I thank Nivetha Karthikeyan for being better than me at many things but, in particular, her passion for writing and history as well as her broader appreciation for the social sciences and the humanities. 
While my natural affinity for mathematics led me to do a PhD in computer science, many of my distinctive strengths as a computer scientist were first acquired in my misbegotten attempts to compete with her. 
Her scholarly defiance continues to inspire me.

I thank Ravi Ramakrishna and Linus Setiabrata for their childlike love for mathematics. 
When I arrived at Cornell, I found myself listless. 
I had lost much of my excitement to learn, and did not know what research was. 
That changed watching them find joy in their study of mathematics. 
While I quickly realized I could not match their talents, and took the well-trodden path from failed mathematician to computer scientist, I discovered research and rediscovered learning from them.

I thank Bobby Kleinberg, Eva Tardos, Kavita Bala, Lillian Lee, Marty van Schijndel, Vlad Niculae, and Xanda Schofield, among others, for their mentorship at Cornell. 
I thank Claire Cardie for her unrelenting support and unwavering confidence.
She will forever be my inspiration as a researcher and computer scientist. 
It is only in leaving Cornell that I realize how precious her investment was in me.
I thank Arzoo Katiyar: the first paper I ever read was \citet{katiyar2017joint}, the first paper I ever reviewed was with Arzoo’s guidance, and the first paper I ever wrote was with Arzoo \citep{bommasani2019sparse}. 
I may very well have been Arzoo’s first student and I am so sad she cannot be here to read this dissertation. 
My final memory of Arzoo was her pride in me for getting into grad school: she was more overjoyed than I was. 
The world cannot be as bright without the light in her eyes to make it brighter.

Following my undergrad at Cornell, I chose to do my PhD at Stanford.
That decision process introduced me to several folks that I can continue to feel fortunate to share a community with.
I thank Dan Klein: passing on the opportunity to be his student made me certain I had to make the best of my time at Stanford to not regret that decision. 
I thank Yejin Choi, first for being a friendly face at the first conference I attended (AKBC 2019) when I knew no one and felt entirely of place, later for always sparing time to chat at conferences over the years, and always for being an inspiration to do great research even when the broader academy is not yet sure of its value.
I thank Tatsu Hashimoto for putting me at ease when he interviewed me for grad school and for completely failing at advocating for himself as an adviser, instead telling me of Percy's strengths.
He has always been so kind and unassuming in the best of ways.
I thank Jacob Steinhardt for giving me the framework to decide on Stanford over Berkeley.
His way of thought has a subtle brilliance that I hope to emulate as a scholar.

When I arrived at Stanford in August 2020, it was during the heart of the Covid-19 pandemic.
While that proved to be isolating, it made me appreciate the different communities I was a part of: the Stanford CS PhD community, Stanford NLP group, the Stanford Center for Research on Foundation Models (CRFM), the Stanford Institute for Human-Centered Artificial Intelligence (HAI), the Stanford AI policy group, Dan's lab, and p-lambda.
I thank Harini Sreepathi, Jay Subramanian, Joi Gemigniani, Suzanne Lessard, and Vanessa Parli for their amazing wherewithal to make these groups function.
For financial support, I would like to thank the taxpayers for providing funding throughout
much of my dissertation research under the National Science Foundation Graduate Research Fellowship Program (NSF GRFP).
I would also like to thank Stanford University for awarding me the Lieberman Fellowship for my final year in memory of Stanford Professor and Provost Emeritus Jerry Lieberman.

Growing up, I loved sports.
I am incredibly overjoyed to have rekindled my love for sports at Stanford by discovering the wonders of volleyball.
Aman Patel initiated our Stanford CS PhD volleyball endeavors: as someone who often plays the role of the understated organizer myself, I truly appreciate his determination and reliability.
Jadon Geathers was the ever-energetic friend I needed to immerse myself in the sport: I am so excited to see how he will bring his infectious energy to a home I know well in Cornell and I wish him the best in his PhD.
Ace, Ali, Alirezah, Allissa, Alvin, Andrew P, Andrew W, Azat, Brody, Cagan, Calvin, Carter, David, Dylan, Eddie, Edgar, Emily C, Emily H, Ethan A, Ethan S, Evans, Gary, Hakeem, Haoming, Hlib, Hülya, Jack, Jason, Jayant, Jie, Julia M, Julia W, Karen, Keanu, Kyle, Logan, Louise, Maddie, Mey-Sam, Noah, Omid, Pau, Pauline, Po-Han, Raven, Saneel, Sergey, Spencer, Suguru, Timena, Tim, Timmy, Toby, Val, Valeria, Victor, Will, Xianghao, Yu-Chian, Yuya, and everyone else I have played with are instrumental to all the joy I have playing volleyball.\footnote{I would like to thank Palmetto Superfoods for always further accentuating these joys.}

In addition to my wonderful volleyball friends, I am glad to have befriended many others during my time at Stanford.
While there are too many to list, I would like to especially thank:
Aditi Raghunathan, Alex Tamkin, Ananya Kumar, Annie Chen, Archit Sharma, Chenchen Gu, Chris Donahue, Christie Lawrence, Daniel Zhang, Dilara Soylu, Dimitris Tsipras, Eric Mitchell, Eric Wallace, Erik Jones, Irena Gao, Isabel Papadimitriou, Jason Wei, Jenn Wang, John Hewitt, John Thickstun, Joon Park, Judy Shen, Kawin Ethayarajh, Kaylee Burns, Lisa Li, Madeleine Wright, Megha Srivastava, Meena Jagadeesan, Michael Xie, Michelle Fang, Mina Lee, Moo Jin Kim, Neel Guha, Nelson Liu, Nick Tomlin, Omar Khattab, Omar Shaikh, Pang Wei Koh, Peter Henderson, Rohan Taori, Rohith Kuditipudi, Rose Wang, Roshni Sahoo, Ruiqi Zhong, Samuel Clarke, Shibani Santurkar, Shiori Sagawa, Sidd Karamcheti, Simran Arora, Spencer Compton, Steven Cao, and Yann Dubois.

I am especially grateful to my friends from Cornell.
Ge Gao was the first PhD student to become my close friend and I wish her all the success in teaching Pluto new talents.
Jerry Qu made the journey from Cornell to Stanford just a year before me and I look forward to many more board game nights in the years ahead.
Most of all, I would like to thank two amazing duos.
Linus Setiabrata will always inspire me as a scholar and I am so glad to have befriended Seraphina Lee through him as she provides $\epsilon$ more sensibility to handle his nonsensical tendencies.
Janice Chan and Andy Zhang are family to me and it has been an incredible blessing to have them so close by.
Throughout my PhD, I was content being solitary, but in the rare moments I needed companionship, I am glad these four were always happy to see me.

Finally, I would like to thank all the researchers who shaped my PhD.
First, those who I had the fortune to mentor: Alex Wan, Connor Toups, Dilara Soylu, Jonathan Xue, Kevin Klyman, Nathan Kim, Ryan Chi, Siddharth Sharma, and Virginia Adams.
I am grateful they took a chance on me and I am thrilled they all are now pursuing even more ambitious goals than the work we did together.
Second, those who I collaborated closely with: Katie Creel and Sarah Bana on homogeneous outcomes; Tony Lee and Yifan Mai on evaluations; and Kevin Klyman, Sayash Kapoor, and Shayne Longpre on too many projects to count.
All of these efforts were multi-year collaborations with little precedent: I am glad to have worked so closely together and to have such bold collaborators who shared my goal of research with broader impact.
Finally, those in the broader AI policy community, irrespective of whether we formally coauthored a paper or not: 
Alex Engler, Andrew Trask, Andrew Strait, Anka Reuel, Angelina Wang, Ashia Wilson, Ashwin Ramaswami, Aviya Skowron, Caroline Meinhardt, Connor Dunlop, Daniel Privitera, Dan Hendrycks, Deb Raji, Deep Ganguli, Divyansh Kaushik, Dylan Hadfield-Menell, Drew Spence, Elena Cryst, Elizabeth Kelly, Eric Horvitz, Emma Pierson, Gaël Varoquaux, Gillian Hadfield, Helen Toner, Hima Lakkaraju, Iason Gabriel, Irene Solaiman, Irina Raicu, Jack Clark, Jason Elliott, Joelle Pineau, Jon Kleinberg, Kevin Bankston, Lama Ahmad, Laura Weidinger, Lindsey Gailmard, Madhu Srikumar, Marc Aidinoff, Manish Raghavan, Markus Anderljung, Matt Salganik, Miles Brundage, Miranda Bogen, Nate Persily, Nathan Lambert, Nik Marda, Nitarshan Rajkumar, Nuria Oliver, Ollie Ilott, Peter Cihon, Risto Uuk, Rumman Chowdhury, Russell Wald, Ruth Appel, Sanmi Koyejo, Sara Hooker, Sarah Cen, Sarah Myers West, Scott Singer, Seb Krier, Seth Lazar, Solon Barocas, Stella Biderman, Stephen Casper, Stuart Russell, Suresh Venkatasubramanian, Tino Cuéllar, Vyoma Raman, William Isaac, Yacine Jernite, Yo Shavit, and Yoshua Bengio.
I am glad that we all share a community we have built together over time: whether we agree or not on AI, policy, or what lies ahead, I am certain we share a genuine desire to build a better society.

Most of all, I would like to thank those who mentored me throughout my PhD.
The Stanford CS faculty have provided me guidance and treated me as a peer throughout my PhD, especially Diyi Yang, Fei-Fei Li, James Landay, Jeannette Bohg, Keith Winstein, Mehran Sahami, Michael Bernstein, Monica Lam, Moses Charikar, and Omer Reingold.
While I did not interact with them much, I appreciate Carlos Guestrin and Karen Liu for their distinctive contributions that I witnessed year-over-year as I served alongside them on the PhD admissions committee.
They all make me proud to say I am a Stanford computer scientist.

I especially want to thank a number of close mentors.
Chris Manning's textbook is the first text I read in my research career,\footnote{My greatest goal as a researcher is to find an appropriate time to use the term \textit{hapax legomena}, which I learned from his book.} his prescience has cultivated the NLP group that I first encountered when I arrived at Stanford, and his guidance, while not always easy to hear, has made me a better scholar.
I am honored Chris served on my dissertation committee and as a reader of this dissertation.
Dawn Song, who I had the fortune of working with by happenstance towards the end of my PhD, has become a reliable guiding light in the pursuit of evidence-based AI policy with strong scientific foundations.

Rob Reich is the first professor I met at Stanford outside of computer science: I am glad he has pushed me to think about what it means to build norms and value leadership in ways that are not customary in computer science.
Marietje Schaake is responsible for my deep entanglement with policy in the European Union, which may very well be the long-term impact of my dissertation in the EU AI Act and beyond.
Alondra Nelson gives me courage to believe academic research and public policy belong intertwined to achieve their fullest potential.
I am glad to have worked closely with all three of these luminaries who have dedicated their talents and time to public service: I truly appreciate how they each make me appreciate the policy side of AI policy as much as the AI side.
And frankly, I hope to one day be half as inspiring a person as each of them.

Arvind Narayanan and Dan Ho are advisers \textit{de facto} even if not \textit{de jure}.
Arvind is the best role model I have found of a computer scientist doing what I would like to do as a professor.
I am glad he has graciously taken so much time to work with me and he extended me the opportunity to visit Princeton CITP, where I met Sayash and kick-started the entire AI policy arc of my PhD.
Dan shows me that there should be no bounds on an academic's pursuit of impact.
I knew essentially nothing about the law and government beyond what I learned from Suits and Law \& Order growing up.
I am deeply grateful to Dan for showing me the ropes.
\newpage
\noindent I am extremely grateful to have been under the guidance of not one great adviser, but two.
The two influences complemented each other wonderfully, and I cannot think of a better
environment for me.

Dan Jurafsky often strikes me as a human being that embodies all the best qualities of a person.
At the start of my PhD, I hoped Dan would teach me to be a better human and that's why I wanted to be his student, much as I felt some of that from Claire.
At the end of my PhD, I cannot help but feel that I failed to learn much of anything as Dan still feels like a much better person than anything I can imagine of myself.

The final word of these acknowledgments is for Percy Liang.
I began my PhD distrusting Percy due to his stoic expressions yet working with him on topics he had never previously worked on.
By the end, I trust Percy more than anyone else on the planet, including myself at times.
Fortunately, we continue to work on topics that neither of us know much about, but I think that makes the journey all the more fun.
He pushes me to do better even when I take on projects of unprecedented scale or striking impact.
Yet, along the way, I constantly feel I am reaching new heights, even if I am merely standing on his shoulders as he lifts me up.

\afterpreface
\chapter{Introduction}\label{chapter:intro}
\chaptermark{\small Introduction}
AI is the technology of our time. 
When I began my PhD in 2020, this statement was a belief held by a small community of researchers. 
As I finish my PhD in 2025, this statement is a reality experienced by people around the world.

What should we do about this? 
This dissertation tells the story of what I did. 
This story traces an unusual academic path defined by unorthodox research with industry and government during a period of profound transition on all fronts.
In industry, AI grew from being a research endeavor at a few tech companies to the overwhelming focus of the most powerful titans and most promising upstarts. 
In government, AI escalated from being a minor issue in an overloaded tech policy portfolio to demanding international attention from world leaders. 
My PhD concentrates on my deep entanglement with these efforts, leading me to work with a remarkably broad collection of people, disciplines, and institutions. 

\section{Industrialization}
Industry has gone all in on artificial intelligence.
Every level of analysis reveals this sea change.
To make it vivid, consider the technological, organizational, and macroeconomic economic levels.

\begin{figure}[t]
\centering
\includegraphics[width=\linewidth]{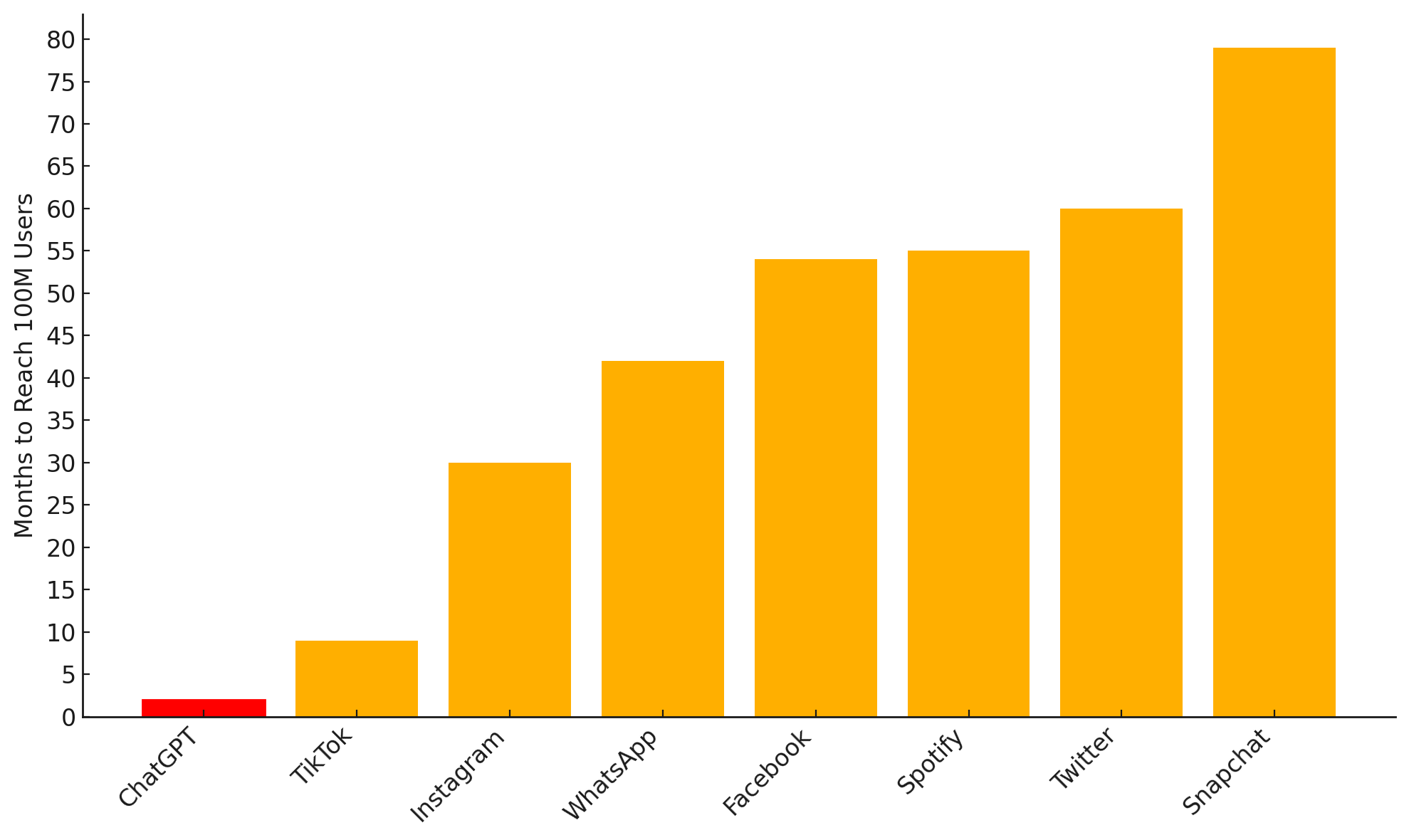}
\caption{
\textbf{The rapid adoption of ChatGPT.}
Within two months of release, OpenAI's ChatGPT accumulated 100 million users, reflecting a faster rate of adoption than any prior digital technology.
}
\label{fig:chatgpt}
\end{figure}

Artificial intelligence as a technology is widely adopted.
OpenAI's release of ChatGPT on November 30, 2022 was the opening shot in this unrelenting salvo: ChatGPT accumulated 100 million global active users in 2 months, outpacing the world's most iconic digital technologies to date (\autoref{fig:chatgpt}).
Within two years, 100 million would look paltry: on December 11, 2024, Google announced that Gemini 2.0 powered many of their products, including all seven of Google's products with at least 2 billion users.
In 2025, 2 months would look sluggish: DeepSeek acquired 100 million users in 7 days in January 2025 with the release of DeepSeek-R1, and OpenAI gained 100 million users just days after adding image generation to ChatGPT in April 2025.

The importance of artificial intelligence reconstitutes the operation of the most important AI companies.
Employees at these companies have gone from being unassuming technical contributors to public-facing celebrities whose moves and actions are covered by the media.
Leadership at companies is the subject of intrigue, scrutiny, and overwhelming attention.
Operationally, there are huge divisions dedicated to artificial intelligence, with important model runs costing hundreds of millions if not billions, better resembling the large industrial machinery seen in more mature product sectors than the simple conceptual frameworks taught in academic machine learning courses.
In the 2010s, companies espoused ideals (\eg Google, IBM, and Microsoft published responsible AI principles), whereas in the 2020s the expectations have shifted to practices (\eg Anthropic, Google, and OpenAI published frontier safety frameworks).

\begin{figure}[t]
\centering
\includegraphics[width=\linewidth]{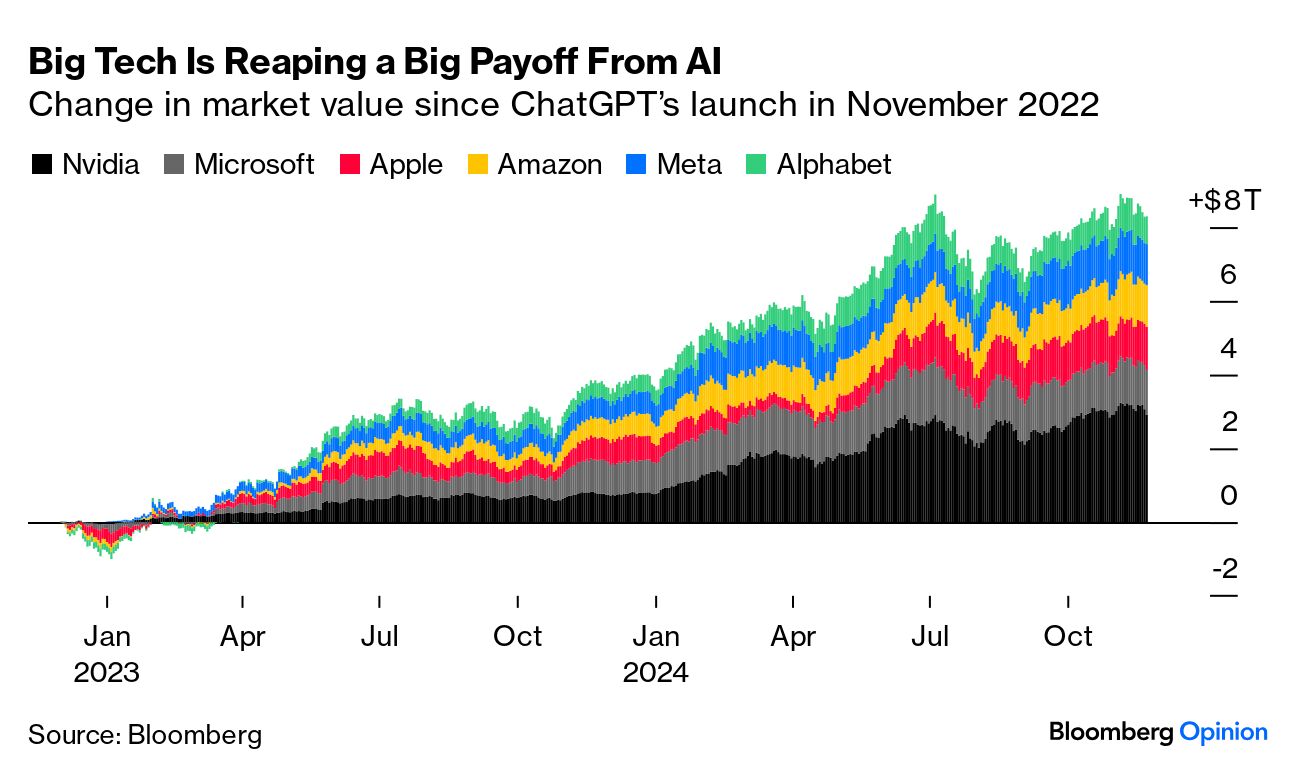}
\caption{
\textbf{The growth in AI company stocks.}
Following ChatGPT's launch, valuations of AI companies (especially Nvidia) skyrocketed. Figure used with permission.
}
\label{fig:bigtechstock}
\end{figure}

Artificial intelligence pervades every economic sector, including procurement by governments even for military use.
This growing economic reality is encoded in the skyrocketing valuations of AI companies (\autoref{fig:bigtechstock}): big tech already rivaled the power and resources of many of the world's nations, and this concentration has only grown more.
As we look forward, we routinely see forecasts from a range of actors that predict AI will produce trillions in value and some even contemplate AI obviating work, which would quite possibly be the greatest change to the economy in history.
Two organizations best reflect what we see with artificial intelligence today: Nvidia and OpenAI.

Nvidia is the infrastructure of the AI ecosystem.
Nvidia's GPU is the workhorse for artificial intelligence worldwide: every other leading AI company relies on Nvidia.
No company has grown in value more than Nvidia due to the growth of AI (\autoref{fig:bigtechstock}).
If Nvidia is the backend, then OpenAI is the frontend of AI as the inescapable public face.
In 2020, OpenAI was a fringe AI company, known by most in the AI community but very few beyond it.
In 2025, OpenAI is the AI company.\footnote{One way to see this is in the success of not just the company but its alumni. After writing the Transformers paper, the authors left Google and founded many high-profile AI companies (\eg Character, Cohere, Inceptive). In much the same way, though subject to far more attention, alumni of OpenAI founded many high-profile AI companies (\eg Anthropic, Safe Superintelligence Inc., Thinking Machines).}
OpenAI's Stargate initiative was announced by a US president on the first day of taking office, emblazoned by President Trump as the ``the largest AI infrastructure project in history'' that may go on to be the largest private-sector infrastructure investment in U.S. history.

The meteoric rise of AI, pioneered by companies like Nvidia and OpenAI, is rivaled by the cataclysmic falls of the very same companies.
Many promising upstarts (\eg Adept, Character.AI, Inflection) have been swallowed up by the dominant big tech titans through a variety of new-age acquisition strategies, reinforcing concerns around the concentration of power in the technology industry.
Others, like Cohere,  Mistral and Stability AI have enjoyed brief periods of remarkable prominence, only to later come back down to earth.
Nvidia suffered the most prolific one-day loss in stock market history when it lost \$589 billion on January 27, 2025 in the wake of the DeepSeek-R1 release.
OpenAI suffered the most tumultuous board coup in recent memory when its board abruptly fired CEO Sam Altman, only for most OpenAI employees to threaten to resign, leading to Altman's reinstatement and the board's restructuring.
Industry unquestionably dictates the trajectory of artificial intelligence at present, but stability is ephemeral at best, reflecting the costs society incurs due to the absence of robust governance.

\section{Governance}
Government has woken up to artificial intelligence.
In his farewell address, President Biden captured the zeitgeist by saying ``artificial intelligence is the most consequential technology of our time, perhaps of all time.
Nothing offers more profound possibilities and risks for our economy and our security, our society, our very humanity. 
Artificial intelligence even has the potential to help us answer my call to end cancer as we know it. 
But unless safeguards are in place, AI could spawn new threats to our rights, our way of life, to our privacy, how we work, and how we protect our nation. 
We must make sure AI is safe and trustworthy and good for all humankind.''
This general tenor is exemplified across the developments in policies, institutions, and processes over the past few years.

At the policy level, many jurisdictions actively pursued high-level AI policy.
The EU AI Act stands alone as the flagship AI legislation to be enacted.
In the United States, Executive Order 14410 on the Safe, Secure, and Trustworthy Development and Use of Artificial Intelligence was the keystone of the Biden administration's AI policy approach, standing as the longest Executive Order on any subject in US history \citep{bommasani2023eo}.
These supplement a broader array of efforts spanning the AI Bill of Rights, NIST AI Risk Management Framework, the White House Voluntary Commitments on AI, and an array of state-level legislation.
Across the world, other key efforts include the Chinese Interim Measures for the Management of Generative Artificial Intelligence Services and the G7 International Code of Conduct.

At the institution level, dedicated entities were created and tasked with AI responsibilities. 
The UK initiated this process with the oft-renamed AI Security Institute, which began as the Foundation Model Taskforce before stints as the Frontier Model Taskforce and AI Safety Institute.
This snowballed into a global network of AI Safety Institutes spanning Australia, Canada, France, Japan, Kenya, South Korea, Singapore, and the US.
Alongside these entities with generally weak formal authority, the EU AI Office was formed within the European Commission as the dedicated overseer and enforcer of the EU AI Act.

At the process level, nascent mechanisms now exist to address the challenge of policymaking in a highly technical and constantly evolving domain.
The International Scientific Report has emerged as the the primary mechanism for consolidating scientific evidence \citep{bengio2025international}.
The United Nations High-level Advisory Body on Artificial Intelligence provides steer for international AI policy efforts.
And the ongoing series of approximately annual summits, starting with the 2023 AI Safety Summit in Bletchley Park through the most recent 2025 AI Action Summit in Paris, convene global leaders to build consensus, including on specific commitments by participating companies and countries, on the path forward.

Together, immense progress has been made in recent years as the demand for AI governance grows.
However, the future of AI governance remains in the balance, especially in light of the presidential transition in the United States, which has sharply reoriented AI policy not only domestically, but also around the world.
This juxtaposition is embodied in the contrast of the words of the US vice presidents at the 2023 and 2025 AI summits.
In 2023 at Bletchley Park, Vice President Harris forcefully declared the focus: ``Today, I will speak more broadly about the vision and the principles that guide America's work on AI.
President Biden and I believe that all leaders from government, civil society, and the private sector have a moral, ethical, and societal duty to make sure that AI is adopted and advanced in a way that protects the public from potential harm and that ensures that everyone is able to enjoy its benefits \dots 
Accordingly, to define AI safety, I offer that we must consider and address the full spectrum of AI risk -- threats to humanity as a whole, as well as threats to individuals, communities, to our institutions, and to our most vulnerable populations.''
In 2025 at Paris, Vice President Vance drove home the shift: ``I'm not here this morning to talk about AI safety, which was the title of the conference a couple of years ago. 
I'm here to talk about AI opportunity.
When conferences like this convene to discuss cutting-edge technology, oftentimes, I think our response is to be too self-conscious, too risk-averse. 
But never have I encountered a breakthrough in tech that so clearly calls us to do precisely the opposite.''
Within the past six months, we have seen this landslide unfold: the Biden Executive Order 14110 has been rescinded, the EU AI Act and its Code of Practice have been challenged, and the focus on US state-level legislation has grown while under threat of sweeping federal preemption or moratorium.

\section{Roadmap}
My dissertation narrates my PhD journey during a time of dramatic change and prevalent uncertainty.
I tell a computer scientist's story of studying a momentous technological paradigm shift in realtime.
Witnessing what was happening in both industry and government, I often doubted the value of my work, which led me to reflect upon the academia as a whole.
What is academia good at? 
And what should academia do? 

Academia is good at research: we are the frontier of rigorous inquiry. 
These past years question whether academia is the frontier of research on building artificial intelligence, and whether that has moved beyond the university’s reach. 
But what is certain is that academia is the best place to deeply understand the societal impact of artificial intelligence.

Academia should advance public outcomes: we are the beacon of intellectual leadership. 
These coming years question whether academia will be financed by public funds and how to recover widespread trust in the university’s mission. 
But what is certain is that academia is the unique place with deep expertise yet little self-interest to create the field of AI policy.

This dissertation presents my research over the course of my PhD on these two subjects, largely based on my prior writings.
These works are generally the product of collaboration, often with many collaborators, and rarely alone.
They appear in a range of venues, some peer-reviewed and others not, that target a variety of disciplinary audiences. 
This dissertation is best understood as a story to commemorate my PhD journey: the individual works should be preferentially cited for specific scientific content, most of all to properly acknowledge my coauthors.

Beyond the shared topical focus, several themes unite my research.
My works address specific problems by introducing different levels of abstraction and different disciplinary perspectives.
This is often necessary to achieve my stated goal of comprehensiveness.
I often prioritize breadth over depth because I did this research during the early days of a paradigm shift: I wanted to provide a bird's-eye perspective of the terrain to surface insights that clarify where future work should dig deeper.
Moreover, my work is characterized by a desire to aggressively shape outcomes.
This is often necessary not only to have impact but because this initial impact amplifies: early decisions introduce a formative path dependence that bends the overall technological trajectory.
Consequently, I prioritized direct engagement with companies to shape their practices and policymakers to shape their decisions.

Each chapter of this dissertation can be understood as a layer. 
Each can be read standalone, but its function is to be the foundation for the following chapters.
Now that you have read the Introduction, you understand what animated me to work on what I did during my PhD.

Chapter \ref{chapter:paradigm} will define my focal point of foundation models as a paradigm of artificial intelligence based on a large-scale multidisciplinary collaboration \citep{bommasani2021opportunities}.
I begin by situating foundation models against the backdrop of past paradigms of artificial intelligence using the lens of emergence and homogeneity.
This background clarifies why foundation models are different and why we gave them a new name to recognize their importance that saw uptake in major US and EU policy.
Chapter \ref{chapter:paradigm} concludes with my overarching approach to reasoning about the societal impact of foundation models: I articulate why this is a hard problem and how I approach it by jointly 
 attending to technological properties and broader sociotechnical context.

From here, we move to the core research chapters.
I begin with the conceptual primitives in Chapter \ref{chapter:concepts}.
To start, I divide the focus on technological properties into the desirable capabilities and undesirable risks of foundation models.
Building upon the previous chapter, which identifies emergence and homogeneity as distinctive axes that disambiguate foundation models from past AI paradigms, I present two vignettes on emergent capabilities \citep{wei2022emergent} and homogeneous outcomes \citep{bommasani2022homogenization, toups2023ecosystemlevel}.
These examples preview the interplay between the conceptual and the empirical, each presenting new measures for the underlying concepts, which reappears in subsequent chapters.
I then shift gears to cover concepts relevant for reasoning about the broader sociotechnical context, namely the foundation model supply chain.
I focus on perhaps the most salient region of this supply chain, namely the strategies for how foundation models are released, by focusing on my conceptual work on how models can be made available in more open or more closed arrangements \citep{liang2022community-norms, kapoor2024societal}.
I then discuss the more general approach to monitoring the supply chain \citep{bommasani2023ecosystem}, which can be viewed from both the technological and organizational perspective.
I conclude this chapter by discussing the policy impact of these conceptual tools across a variety of AI policy subdomains, primarily in the US and the UK.

Having established the central concept of foundation models as well as the supporting concepts, we enter the heart of the dissertation with the longest chapters.
Given the conceptual primitives, we will want a sharper understanding that concretizes the capabilities, risks, and supply chain using empirical methods.
I begin with empirical methods at the technological level by building the first large-scale model-level evaluation platform (HELM) to measure the technical properties of a foundation model \citep{liang2023holistic} in Chapter \ref{chapter:empirics-evaluation}.
HELM identifies several deficits in the evaluation landscape, leading to the fundamental question of what foundation models should be evaluated on.
Viewing foundation models as critical public technologies with broad impact, HELM provides a systematic procedure for selecting the scenarios (\ie data) and desiderata (\ie metrics) involved in evaluation.
Of specific focus is the goal to broaden the focus in the evaluation of AI from accuracy to the many other relevant dimensions of what it means for AI to be performant (\eg robustness, fairness, calibration, efficiency).
HELM advances an agenda of proactive and independent evaluation by evaluating models, irrespective of whether their developers evaluate them, in a standardized, reproducible, and transparent fashion at scale.
This presents new fundamental challenges for evaluation research in the form of major costs and strong demands to aggregate the results to make them more intelligible. 
HELM has seen broad adoption \citep{bommasani2022evaluation} within both academia and industry, growing into a major evaluation platform that advances evaluation science in AI \citep{weidinger2025evaluationsciencegenerativeai}.

While HELM demonstrates unprecedented breadth in the evaluation of foundation models to meet the aspiration of holistic evaluation, it operates at standard level of abstraction.
To broaden the aperture, I introduce a large-scale organization-level composite index (FMTI) to measure the transparency of foundation model developers \citep{bommasani2023fmti, bommasani2024fmti} in Chapter \ref{chapter:empirics-index}.
Composite indices have not been used in computer science, but they enjoy a rich history across many disciplines such as statistics, economics, and the broader social sciences.
While a composite index resembles a benchmark, the unit of analysis shifts from a technological artifact (\ie a model) to an organizational artifact (\ie an entity).
This transition is necessary to catalyze higher-level organizational change, especially given the growing trends towards opacity from leading AI companies.
FMTI concretizes the nebulous construct of transparency into 100 concrete indicators across the supply chain (\eg the upstream data and compute used to build a foundation model, the downstream release and usage of a foundation model).
Through a multi-stage process of gathering information, scoring companies, having those companies rebut our scores, and publishing the final scores, I show the leading foundation model developers are not transparent in both systemic and idiosyncratic ways in 2023.
However, through direct engagement with the relevant employees within these companies, as well as a variety of external stakeholders (\eg the media, policymakers, company investors, company clients), we see material improvement, especially through the disclosure of never-before public information, in the 2024 FMTI.
FMTI has wide-ranging impact: coverage by media outlets like the New York Times, engagement by AI companies like Google, and uptake by policymakers.
Of specific note is policy impact in the United States (\eg the US Foundation Model Transparency Act was proposed in Congress in 2023), the European Union (\eg the EU AI Act has significant resonance with the FMTI indicators in 2024) and at the international level (\eg the G7 International Code of Conduct indicates that foundation model developers should publish transparency reports \citep{bommasani2024foundationmodeltransparencyreports}).

Having accumulated a wealth of evidence, we transition from the study of the societal impact of foundation models to the policy that governs AI.
In Chapter \ref{chapter:policy}, I provide the final research contributions of the dissertation to buttress the burgeoning research-policy interface.
I present a vision for evidence-based AI policy \citep{bommasani2025ca, bommasani2025evidence} that reflects the growing consensus among a range of academics spanning different institutions and disciplines.
To articulate this vision, I begin by clarifying how research and evidence can support policy making.
I first show how systematic evidence review can build internal consensus within a community of experts to yield external clarity on how to make progress on contentious topics.
Namely, I describe my work on the topic of open foundation models as a contentious issue across many jurisdictions, showing how the lens of marginal risk can allow us to make significant progress \citep{kapoor2024societal, bommasani2024open}, especially in the context of US AI policy.
I then show how new types of bespoke evidence can be developed specifically for policymakers, showing how this can lead to very direct impact.
Namely, I describe my work on assessing whether current practices of leading AI companies would constitute compliance with proposed legislation, providing immediate feedback for policy decisions in the legislative process of the EU AI Act \citep{bommasani2023eu-ai-act}.
I conclude the policy chapter by recognizing that the 
best version of evidence-based AI policy is a two-way street.
I offer specific recommendations for how policy can play its part in growing the evidence base through 
evidence-generating AI policy \citep{bommasani2025ca, bommasani2025evidence}.

The final chapter (Chapter \ref{chapter:conclusion}) is short as a necessary reprieve from the lengthy dissertation.
I close with a provocation for the research that I hope to see.
I distill two lessons from my PhD.
First, junior academics can assemble and lead large, multidisciplinary, multi-institutional teams to address big problems: not just the open problems of research, but the profound challenges of society.
Second, academia can pursue the bold vision of forcefully improving the conduct of the world's most powerful entities: our responsibility, in part, should be to use our deep expertise, unique perspective, and distinctive privilege to advance the public interest.
These general lessons are what I value most from my PhD.
\chapter{Technological Paradigm}\label{chapter:paradigm}
\chaptermark{\small Technological Paradigm}

Over the course of my five-year PhD, foundation models emerged as the prevailing technological paradigm for AI.
At the conclusion of my PhD, as I write this dissertation, they stand as the focal point of both academic and industrial AI research, development, and deployment. 
Most fundamentally, foundation models are the watershed moment in the history of AI, marking the inflection point where AI transformed from a research subfield of computer science to a general-purpose technology capable of reorganizing society. 
While this dissertation focuses on how foundation models and AI impact society, I begin with the technology itself as it is understood in AI research.

In \citet{bommasani2021opportunities}, we coined the term \textit{foundation models} and launched the broad multidisciplinary study of these models.
Foundation models are a general class of models, which we initially defined as ``any model that is trained on broad data (generally using self-supervision at scale) that can be adapted (\eg~fine-tuned) to a wide range of downstream tasks''.
In this way, foundation models function as a foundation that was unfinished yet important in the developing AI systems.
This chapter identifies foundation models as a paradigm of artificial intelligence.

\newpage

\section{The Paradigm}
To understand foundation models as a paradigm \citep{bommasani2021opportunities}, I introduce the backdrop of recent technological paradigms in artificial intelligence.
To reason about the chronology of AI, I foreground the concepts of \textit{emergence} and \textit{homogeneity}.
Emergence means behavior implicitly induced rather than explicitly constructed: emergence prompts scientific intrigue and societal anxiety around the unanticipated properties of artificial intelligence.
Homogeneity means methods consolidate around a single dominant approach: homogeneity promotes focus on a high-leverage approach that, due to the lack of diversity, can be a single point of failure.
These concepts organize and disambiguate AI paradigms.

\begin{figure}[ht]
\centering
\includegraphics[width=\linewidth]{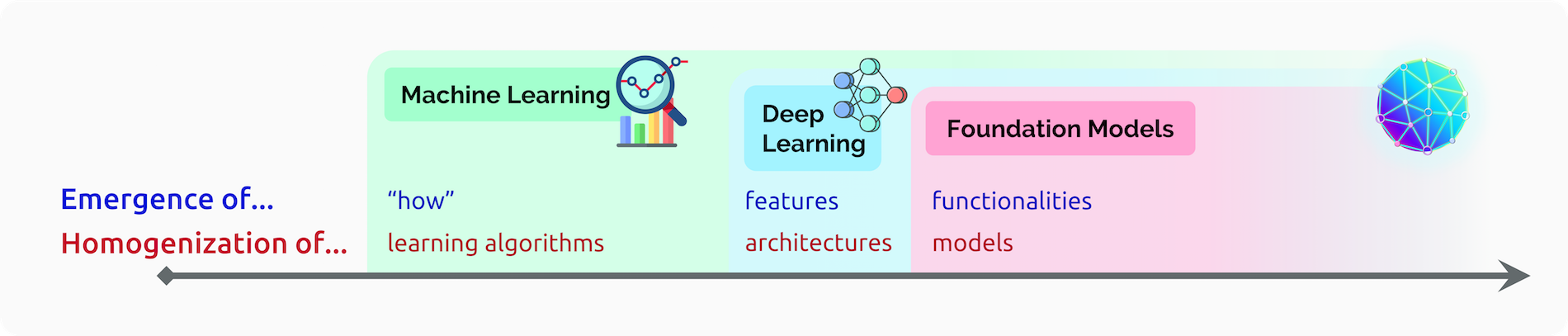}
\caption{
\textbf{Emergence and homogenization as a lens to identify AI paradigms.}
The trajectory of AI is marked by increasing \textit{emergence} and \textit{homogenization}.
Machine learning demonstrates that the ability to perform tasks emerge from data and that the methods to develop models homogenize around dominant learning algorithms (\eg logistic regression).
Deep learning demonstrates that high-level features emerge from data and that the methods to develop models homogenize around dominant model architectures (\eg Transformers).
Foundation models demonstrate that meta capabilities like chain-of-thoughts emerge from data and that usage homogenizes to dominant models (\eg o3).
}
\label{fig:evolution}
\end{figure}

The discipline of artificial intelligence has a complex origin \citep{russell2016artificial}.
A formative, and symbolic, starting point is in the leadership of John McCarthy in the 1950s and 1960s.
McCarthy popularized the term to describe the paradigm (artificial intelligence), built an intellectual environment at Stanford for its study (the Stanford Artificial Intelligence Laboratory), and convened leaders from across the academy to define the field (The Dartmouth Summer Research Project on Artificial Intelligence).\footnote{John McCarthy's pioneering work inspired me throughout my PhD because his aforementioned leadership addressed important axes for scientific contribution beyond the mainstay of writing good scientific papers. 
In this spirit, I also endeavored to describe the paradigm (foundation models), build an intellectual environment at Stanford for its study (the Stanford Center for Research on Foundation Models) and convene leaders from across the academy to define the field (in both the First Workshop on Foundation Models in 2021 and latter on evidence-based AI policy in \citet{bommasani2025evidence}), in close collaboration with Percy Liang.}
In this dissertation, I will not focus on the field's first few decades, instead focusing on what I see as the ``modern era'' when the field fully embraced machine learning as the overarching paradigm for building AI technology.\footnote{With this said, there are many remarkable similarities when one retraces the field's history. For example, in reading \citet{russell2016artificial}, I was struck by two notable similarities. First, the work of Newell and Simon on the Logic Theorist was submitted to a journal with the Logic Theorist credited as a coauthor. Today, the scientific publishing industry is reckoning with how AI should be used and credited in scientific authorship. Second, the early years were characterized by solving problems posed by Alan Turing as out of reach for machines, leading to a progressive moving of the goal posts. Today, we see ever-improving capabilities, where critics like Gary Marcus pose challenges, sometimes in the form of benchmarks, which are rapidly eclipsed.}

\paragraph{Machine learning.}
Machine learning (ML) is the near-exclusive approach for building AI today.
We train models on historical data to make future predictions.
Machine learning became prominent in the 1990s, representing a marked shift in how AI was built.
Rather than specifying \textit{how} to solve a task, a learning algorithm would induce this ability from data: the \textit{how} emerges from the dynamics of learning.
In doing so, the design of AI technologies homogenized towards generic learning algorithms.
Given training data, models across many domains could be built with standard techniques like logistic regression.

While machine learning is a general approach to the design of AI, the specific capabilities of models still hinged on feature representation.
For complex tasks in the marquee domains of artificial intelligence, namely computer vision and natural language processing, the quality of ML-based approaches varied significantly in the appropriateness of the feature representations.
For example, to perform question answering or object recognition, where the inputs are sentences or images, domain experts would need to properly engineer the features (\eg write domain-specific logic to convert raw data into higher-level features as in SIFT \citep{lowe1999sift}) to suit the machine learning methods.

\paragraph{Deep learning.}
Deep learning (DL) is the modern approach in many settings for how we operationalize machine learning today.
While the methods of neural networks had been studied several decades prior, under the moniker of \textit{deep learning} \citep{lecun2015deep}, these methods saw renewed traction in the 2010s.
In particular, as is the case for foundation models (and the underlying concept of transfer learning), the successes of deep learning hinge on the appropriate resources to scale the approach.
Namely, deep learning was fueled by larger datasets, more computation (via the availability of GPUs), and greater audacity.
In particular, the signature deep learning moment came when the AlexNet deep neural network \citep{krizhevsky2012imagenet} demonstrated remarkable performance on the ImageNet benchmark \citep{deng2009imagenet}.

Unlike the machine learning approaches that were prominent in the 1990s and 2000s, the deep learning approaches of the 2010s operated on rawer inputs that more closely reflected how data was acquired with less (though not necessarily none) feature engineering.
For example, deep neural networks trained to solve computer vision tasks would take pixels as inputs to represent images.
Within these networks, higher-level features would emerge through the process of training (\ie representation learning).
Deep learning further homogenized the design of AI technologies: not only were generic learning algorithms used (\eg the widespread adoption of first-order stochastic gradient descent methods like Adam as the dominant approach to optimization), but models across domains shared the same architecture rather than bespoke feature engineering pipelines.
This trend became especially salient especially as the field transitioned to the foundation model paradigm, given the overwhelming dominance of the Transformer architecture across modalities.\footnote{This is different from the early-to-mid 2010s where certain architectures were dominant for all tasks relating to a modality, but not necessarily across modalities. Namely, recurrent networks were dominant for text and convolution networks were dominant for images, in part still aligning with the inherent human-perceived structure of these modalities (\eg rotational invariances for images, sequentiality for text).}

\paragraph{Foundation models.}
Foundation models are the dominant paradigm for fully operationalizing the development of AI technology, referring to not only the conceptual processes for designing AI systems, but also the concrete artifacts involved in building them.
At a technical level, the primitive of \textit{transfer learning} \citep{thrun1998lifelong} was well-known prior to this period.
Namely, the approach was to take the ``knowledge'' learned from one task (\eg object recognition in images)
and apply it to another task (\eg activity recognition in videos).
The foundation model paradigm hinges on combining the core elements of the deep learning revolution (\ie powerful specialized hardware, large training datasets, and highly hardware-optimized architectures) at unprecedented \textit{scale} to realize a remarkably expansive vision of transfer learning.

Prior to the advent of foundation models, the approach to operationalizing transfer learning in the context of deep learning was pretraining.
A model is trained on a surrogate task (often just as a means to an end) to yield rich representations, which can then be repurposed in the context of a downstream task of interest.
One important class of pretraining methods is pretraining on annotated data: the surrogate task is learned from supervision made possible by human labor.
Pretraining via this method became a common practice in computer vision \eg pretraining on the ImageNet dataset \citep{deng2009imagenet} for image classification) shortly after the successes of AlexNet (\eg pretrained ResNets were widely used).
However, the successes of pretraining on annotated data are innately limited by the availability of such annotations and, therefore, the significant labor required to produce these annotations.

To overcome this barrier intrinsic to supervised pretraining on annotated data, self-supervised approaches were developed.
In self-supervised pretraining, the surrogate task is encoded implicitly in the data without the requirement of human annotation.\footnote{
Interestingly, self-supervised learning
was dominant in the early days of deep learning
\citep{hinton2006fast},
but was for a decade largely overtaken by pure supervised learning as labeled datasets became larger.}
For example, the masked language modeling task used to train BERT \citep{devlin2019bert} is to
predict a missing word in a sentence given its surrounding context (\eg~\textit{I like \rule{1cm}{0.15mm} sprouts}).
Critically, existing text data can be used for this surrogate task simply by dropping out words and having models predict these words.
Self-supervised objectives not only scale better, by removing the requirement for annotation, but also may yield more expressive representations by removing the constraints of, often, a narrow label space that is tractable for human annotation.

Self-supervised pretraining was popularized in natural language processing via word embeddings
\citep{turian2010word,mikolov2013efficient,pennington2014glove}.
Word embeddings associate each word with a context-independent vector: embeddings learned once from pretraining on Wikipedia, for example, could then be employed in models for a variety of language-related tasks.
Shortly thereafter, self-supervised learning based on autoregressive language modeling
(predict the next word given the previous words) \citep{dai2015semi}
became popular.
This produced models that represented words in context, such as
GPT \citep{radford2018improving},
ELMo \citep{peters2018elmo},
and ULMFiT \citep{howard2018universal}.\footnote{The prescient work of \citet{collobert2008unified}
is related: they trained on a scalable task akin to masked language modeling jointly with downstream tasks,
rather than producing a single foundation model that can be adapted after the fact to downstream tasks.}
The next wave of developments in self-supervised learning\dash{}BERT \citep{devlin2019bert}
GPT-2 \citep{radford2019language},
RoBERTa \citep{liu2019roberta},
T5 \citep{raffel2019exploring},
BART \citep{lewis2020bart}\dash{}quickly followed,
embracing the Transformer architecture,
incorporating more powerful deep bidirectional encoders of sentences,
and scaling up to larger models and datasets.

However, much as with the successes of the deep learning in the 2010s rather than prior decades that explored neural networks, the successes of foundation models in the 2020s required more than just transfer learning as a theoretical approach or self-supervised pretraining as a partial operationalization.
Unprecedented scale (in several senses) proved to be the key catalyst for progress.
Scale required three ingredients:
(i) improvements in specialized hardware;
(ii) the development of the Transformer model architecture \citep{vaswani2017attention}
that leverages the parallelism of the hardware to train much more expressive models than before;
and (iii) the availability of much more training data.
In particular, \citet{bengio2025international} document the significant year-over-year growth in (i) the overall amount of computation dedicated to training foundation models and (ii) the overall size of datasets used to train foundation models.
 
What is most interesting to me is not merely the technological developments that we saw over the course of the 2010s, but the sociological developments that we saw first in the AI research community, then the technology industry, and ultimately in broader society as we moved through the 2020s.
Take BERT as a particular inflection point for the field of natural language processing.
Before 2019, self-supervised learning with language models was a \textit{subarea} in natural language processing,
which progressed in parallel to other developments in the field.
After 2019, self-supervised learning with language models became more of a \textit{substrate} for natural language processing as using BERT had become the norm.
The broad field-wide acceptance that a single model could be so profoundly useful, so as to change how modeling was done for essentially every task studied in the field, is the change in mindset that sounded the clarion call for the era of foundation models.

Foundation models demonstrate iconic emergent capabilities \citep{wei2022emergent}, which we explore in \autoref{sec:emergent-capabilities}.
For example, \citet{brown2020gpt3} demonstrated that GPT-3 models could learn in-context where previous models had not and \citet{wei2022chain} demonstrated that larger models could reason via chain-of-thought prompting even though many thought models of that time were incapable of reasoning.
At the same time, foundation models introduced unprecedented homogeneity in that not only were the same conceptual approaches or architectures used to train models, but many AI systems were built by adapting the same shared foundation model.
We refer to this deeper level of homogeneity as algorithmic monoculture \citep{kleinberg2021monoculture, bommasani2022homogenization}, which we explore in \autoref{sec:homogeneous}

Emergence and homogeneity interact in a potentially unsettling and certainly underappreciated way.
Emergence creates incentives to scale models further.
Homogeneity creates incentives to concentrate focus on a singular model since these benefits can generalize to an ever-growing set of downstream applications: this mirrors the benefits of building enduring shared infrastructure as seen in other parts of computer sciences (\eg the sustained multi-decade investment in key operating systems like Windows or Linux).
Put together, these create concern.
Building technologies that acquire properties we do not understand, concentrating our limited attention on these technologies, and pervasively adopting them (to amortize the one-time cost of building and benefit from economies of scale) introduces risk.
First, at the institutional level, this paradigm promotes the concentration of resources and, thereby, power, which has well-established problems.
Second, at the structural level, this paradigm promotes the creation of single points of failure, which can directly engender systemic failure.
Overall, these tradeoffs are at the intellectual heart of the deep divide around AI technological progress, where reputable experts project wildly different futures \citep{acemoglu2025ai, kokotajlo2025ai, bengio2025international}.

\begin{figure}[t]
\centering
\includegraphics[width=\linewidth]{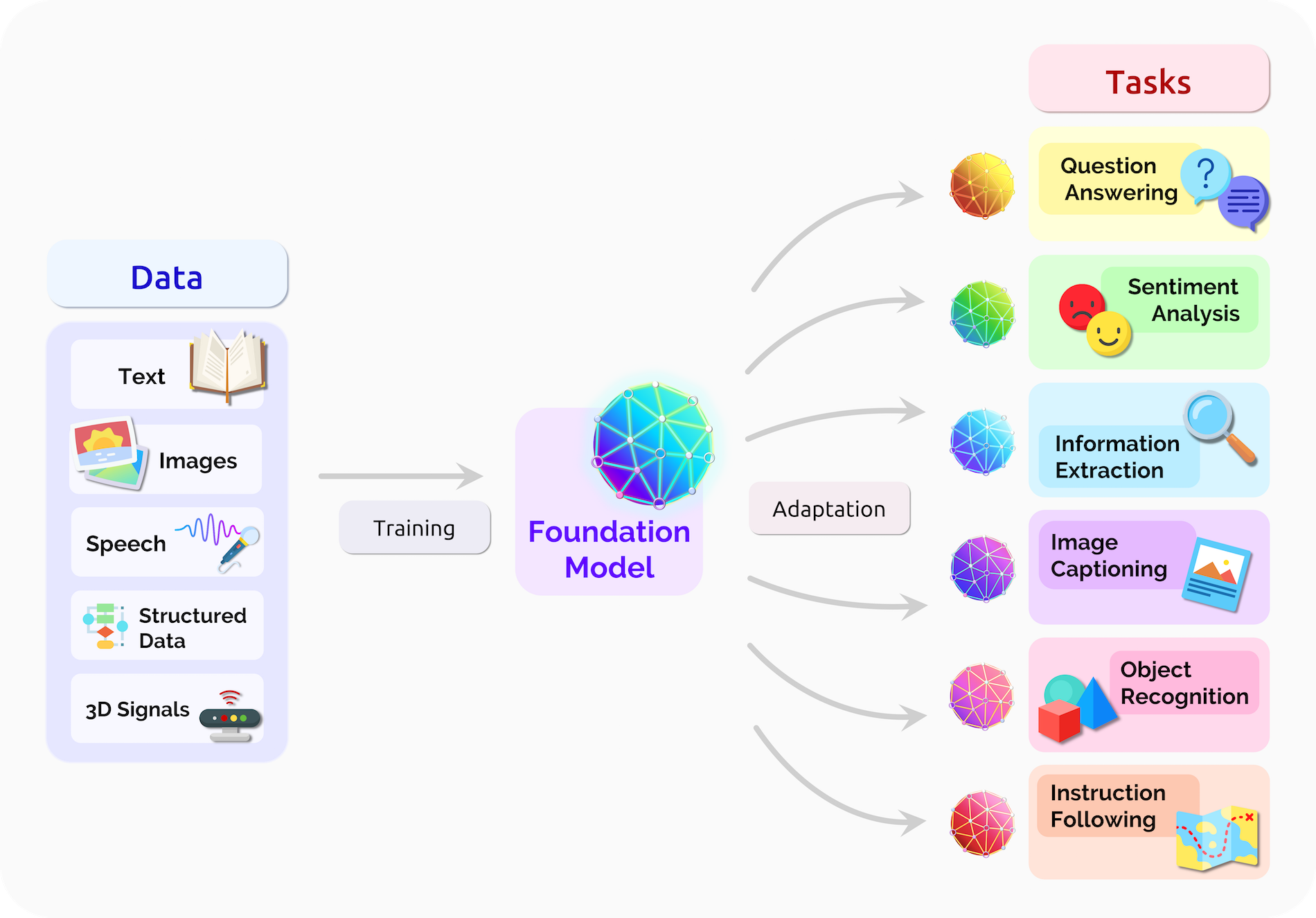}
\caption{
\textbf{The foundation model paradigm.}
A foundation model can centralize the information from all the data from various modalities.
This one model can then be adapted to a wide range of downstream tasks.
}
\label{fig:framework}
\end{figure}

\section{The Term}
We introduced the term \textit{foundation model} to name the technological paradigm.\footnote{To dispel a frequent misconception, I was not the first person to use the term ``foundation model''.
To trace the genealogy, Percy Liang was using the term ``universal model'' in 2021 to describe these models as we wrote the foundation models paper (and built the associated community).
Upon discussion within the writing group for that paper, we decided a more rigorous and consensus-driven process was required before releasing the paper and introducing this terminology, because of problematic connotations associated with ``universal'' (\eg many of the models at that time were obviously not universal in how they performed across different languages).
We generated over 50 candidate names and conducted a multi-phase tournament as part of the overall selection process.
The term ``foundation model'' was first proposed by then-first year undergraduate student Isabelle Levent and gained traction, including from a critical intervention by Chris Manning, to become the name we used to describe this paradigm.
Notably, it was the unique term that received no pushback among the authors.
While I am the lead author on \citet{bommasani2021opportunities} that popularized the term, its origin is more complex \citep[see][]{bommasani2021reflections}.}
Prior to our work, the paradigm had not been clearly recognized, which meant there was a clear void in how to refer to the ongoing technological transition
By choice, the foundation model terminology highlights the function of these models: ``a foundation model is itself incomplete but serves as the common basis from which many task-specific models are built via adaptation.
We also chose the term ``foundation" to connote the significance of architectural stability, safety, and security: poorly-constructed foundations are a recipe for disaster and well-executed foundations are a reliable bedrock for future applications.''
The correct way to understand foundation is a descriptive, rather than normative characterization.
That is, we often do not know the quality of the foundation or whether the foundation is trustworthy: the fact a foundation model serves as a foundation does not imply it is a good foundation or a wise decision.

While we chose the term early in the paradigm (\eg before more people knew of foundation models via their popularization with OpenAI's ChatGPT), the term has succeeded as an effective yet pithy descriptor of the ongoing paradigm.
To choose the term, we surveyed existing terms (\eg pretrained model, self-supervised model, language model, large language model), which had clear deficits (\eg overly centric on the technical dimensions of these models without being broadly comprehensible).
To some extent, this imprecision persists today given the overuse of the term ``(large) language model'', which is not only used to describe models that center on text and language, but also models that have well surpassed language as the exclusive focus.\footnote{It is fair to say these past years, as well as today's models, do evince the power of natural language interfaces.}
For example, Google's Gemini 2.5 operates over text, images, audio, and videos, yet is sometimes described as (merely) a language model.
After we introduced the foundation model terminology, other related terms emerged.
The most important such terms are ``general-purpose artificial intelligence models'' (most notably in the context of the EU AI Act), ``frontier models'' (especially in the context of AI policy and safety discourse around models from a handful of leading developers; see the Frontier Model Forum and \citet{anderljung2023frontier} for further discussion), and ``generative AI'' (most commonly in non-expert public discourse).

The term ``foundation model'' was initially met with acrimony at the time of \citet{bommasani2021opportunities} as Percy Liang and I discussed in \citet{bommasani2021reflections}.
In spite of this disagreement, the term has seen significant adoption.\footnote{As a convenient demonstration, at the time of writing, the most major proximal AI conference (ICLR 2025) features 10 workshops that use the term across different subareas.}
More importantly, the term and its initial definition have seen uptake into policy.
The most noteworthy example is in major US AI policy we have seen to date: Executive Order 14110 on Safe, Secure, and Trustworthy Development and Use of Artificial Intelligence issued by President Biden \citep{EO14110}.
The Executive Order formally defines foundation models in very close alignment with our original definition: ``dual-use foundation model is defined as an AI model that is trained on broad data; generally uses self-supervision; contains at least tens of billions of parameters; is applicable across a wide range of contexts; (and that exhibits, or could be easily modified to exhibit, high levels of performance at tasks that pose a serious risk to security, national economic security, national public health or safety, or any combination of those matters \dots)''.\footnote{The final clause of the definition, indicated by parentheses, on performance on tasks that pose significant risk is due to the Executive Order defining a \textit{dual-use} foundation model in particular.}
As is the case in the Executive Order, and as has often been the case in my work on policy across multiple jurisdictions, a key outstanding question is whether sharper designation criteria should be used for determining whether a model is a foundation model.

\section{Societal Impact}
\begin{figure}[t]
\centering
\includegraphics[width=\linewidth]{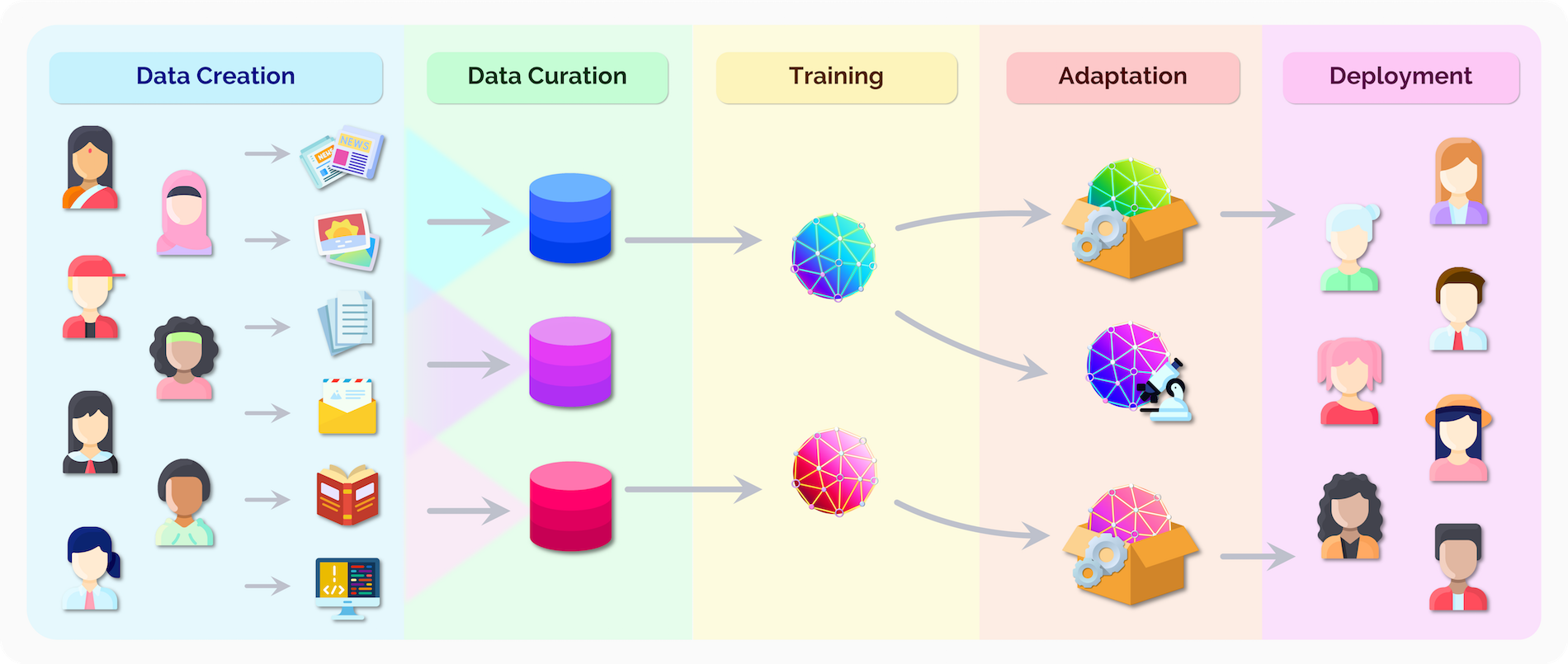}
\caption{
\textbf{The foundation model ecosystem.}
The foundation model ecosystem stretches from data creation to deployment.
At both ends, we highlight the role of people as the ultimate source of data into training of a foundation model,
but also as the downstream recipients of any benefits and harms.
Thoughtful data curation and adaptation should be part of the responsible development of any AI system.
Finally, note that the deployment of adapted foundation models is a decision separate from their construction,
which could be for research.
}
\label{fig:ecosystem}
\end{figure}

Foundation models transform not only artificial intelligence, but also human society.
How do we properly reason about these far-ranging consequences?
The next chapters of the dissertation address this, but first we acknowledge overarching challenges.

Challenges layer upon each other to complicate reasoning about the societal impact of foundation models.
Foundation models are general-purpose technologies in the parlance of economics, which are simultaneously integral to economic production yet subtle to characterize using standard economic methods \citep{bresnahan1995gpt, eloundou2023gpts, ding2024rise}. 
Foundation models are neural networks in the parlance of computer scientists, which are simultaneously essential for intelligent capabilities yet resistant to understanding using nascent interpretability methods \citep{amodei2025interpretability}.
Foundation models are algorithmic intermediaries in the parlance of philosophers, which are simultaneously determinative of societal outcomes yet difficult to attribute using multidisciplinary analytical methods \citep{lazar2023algorithmiccity}.
In short, properly reasoning about the societal impact of foundation models is an endeavor that will necessarily yield incorrect and incomplete results, yet nonetheless is one of the most pressing forays of our time.

On top of these fundamental challenges, a central tension is that much more is known about foundation models as research artifacts than as public technologies.
Most of what is publicly known is the byproduct of research inquiry such as academic papers and leaderboard results.
While this information is valuable, and some of this dissertation contributes to this body of work, the direct societal impact of foundation models is mediated by deployment.
Critically, deployment is often governed by proprietary practices and may involve private data.
And the impacts of foundation models are not only through the visible creation of new products and services, but also the (often silent) updating of existing technologies.

This dissertation adopts a broader view to reason about foundation models in many cases.
That is, while foundation models are the focal point of my research, other regions of the ecosystem (see \autoref{fig:ecosystem}) may be where specific impacts are best understood.\footnote{We stated this in \citet{bommasani2021opportunities}, which is only increasingly true at the time of writing given the discourse on AI agents and compound AI systems.}
To help conceptualize this ecosystem, we give a very simple caricature of the ecosystem as a pipeline where people are at both ends of the pipeline.

\begin{enumerate}
  \item \textbf{Data creation}:
    Data creation is often a human-centric process:
    Most data is created by people and most data relates to people.
    Sometimes data is created by people for other people (\eg emails),  
    sometimes data is a measurement of people (\eg~genomic data)
    and sometimes data is a measurement of the environment (\eg~satellite images).
    Critically, data has an owner and is created with a purpose, which may or may not have been to train a foundation model.

   \item \textbf{Data curation}:
    Data is then curated by acquiring the data and processing it into datasets.
    Notably, there is no single natural distribution of data, and the choice of data embeds many opportunities for shaping model behavior.

    \item \textbf{Training}:
    Training foundation models on these curated dataset is the most celebrated and studied part of AI research, yet it is only one part of a much more complex process that is relevant in the study of societal impact.

   \item \textbf{Adaptation}:
    In the context of AI research,
    adaptation creates a task-specific model from a foundation model.
    In the context of society, adaptation creates a deployable AI system, which may include many other modules (\eg filters on model outputs) and interfaces (\eg a chat user interface).

  \item \textbf{Deployment}:
  The societal impact of foundation models is a byproduct of the societal impact of the downstream AI systems built upon it that are deployed.
  Notably, foundation models, as with many other technologies, may have large societal impact by influencing the operations of governments, militaries, and companies via enterprise-style usage while being largely invisible in terms of publicly-discernible footprint.
\end{enumerate}

\chapter{Conceptual Foundation}\label{chapter:concepts}
\chaptermark{\small Conceptual Foundation}
What are the core conceptual primitives required to reason about the societal impact of foundation models?
To begin, this chapter characterizes the capabilities and risks of foundation models at the technological level.
Building upon the concepts of \textit{emergence} and \textit{homogenization} that we identified in Chapter \ref{chapter:paradigm}, in this chapter we instantiate our discussion of capabilities and risks with vignettes on papers that study emergent capabilities \citep{wei2022emergent} and homogeneous (negative) outcomes \citep{bommasani2022homogenization}.
However, the societal impact of a technology is not only determined by properties of the technology, but also the broader sociotechnical context.
Therefore, this chapter concludes with a focus on the broader foundation model supply chain, including the focal point of whether models are released openly or not.
\newpage
\section{Capabilities}
Foundation models are powerful technologies with unprecedented capabilities that grew substantially over the course of my PhD.\footnote{In fact, these capabilities have even prompted some credible experts to speculate they are early indicators of artificial general intelligence \citep{bubeck2023sparks}.}
While it is straightforward to observe that foundation models are highly capable, reasoning about these capabilities is anything but straightforward.
How should we think about foundation model capabilities in general?

During my PhD, I considered this question on several occasions.
First, I present the different approaches I have identified for reasoning about foundation model capabilities, which operate at different levels of abstractions, and may provide conceptual mileage for different readers.
\begin{enumerate}
\item \textbf{Modalities.} The capabilities of a foundation model can be organized based on what inputs and outputs it can process.
\item \textbf{Evaluations.} The capabilities of a foundation model can be organized based on the performance on evaluations, generally meaning quantitative benchmarks.
\item \textbf{Humans.} The capabilities of a foundation model can be organized based on the frameworks we use to understand human capabilities.
\item \textbf{Economy.} The capabilities of a foundation model can be organized based on the frameworks we use to understand economic tasks.
\end{enumerate}

Given these different approaches, across my PhD, I have made progress towards advancing several of them.\footnote{
As I did not seriously address the economy-level approach to capabilities during my PhD, I highlight the work of \citet{eloundou2023gpts}, which aligns model capabilities with economic relevance.}
At the start of my PhD, we used a modality-centric approach in writing the Capabilities chapter of \citet{bommasani2021opportunities} with sections on Language, Vision, Robotics, Reasoning and Search, and Interaction.
Later, we used a framework derived from human concepts (\ie language, knowledge, reasoning), taking inspiration from the works of Yejin Choi, as part of our broad evaluation efforts in HELM \citep{liang2022helm}.
Towards the end of my PhD, I was asked to first write and then oversee the chapter on capabilities of the International Scientific Report by Yoshua Bengio \citep{bengio2025international}.
Based on this approach to conceptualizing capabilities, I provide a high-level characterization of the current capabilities of foundation models, recognizing they will continue to improve in the coming years.\footnote{This matches what I wrote in the California Report on Frontier AI Policy \citep{bommasani2025ca}.}

\subsection{Current capabilities}
The International Scientific Report acknowledges a broad and growing range of capabilities. According to the International Scientific Report, most experts agree that current capabilities include assisting programmers with small- to medium-sized software engineering tasks, generating highly realistic images, engaging in fluent conversations across multiple languages, operating simultaneously with multiple modalities, and solving textbook mathematics and science problems up to a graduate level. As a strong demonstration of capabilities, OpenAI entered their o3 model into the 2024 International Olympiad in Informatics, which is the annual highest-level international competition in programming for high school students. 
The model achieved gold medal performance without any custom coding-specific test-time strategies defined by humans.

At the same time, most experts agree that current foundation models demonstrate a range of limitations. Models generally decline in performance when operating in unfamiliar contexts. They generally cannot perform useful robotic tasks like household work, nor can they independently execute long-term projects, such as multi-day programming or research tasks. In addition, current models cannot reliably and consistently avoid making false statements.

Understanding and measurement of current capabilities is highly dependent on the specific evaluations available. Evaluations are currently constrained by a limited set of techniques for eliciting capabilities. While many evaluations have been developed, there are currently no common standards for measuring how AI augments human capabilities. Overall, current evaluation techniques are nascent and ad hoc, and they have not yet achieved the scientific rigor \citep{weidinger2025evaluationsciencegenerativeai} expected in other domains where long experience with widespread societal use generates greater data and systematic evaluations more directly guide policy.
With this in mind, three key principles to understand current capabilities are:

\begin{enumerate}
    \item \textbf{Standard benchmarks are poor proxies.}
    Many of the most well-known capability benchmarks do not reflect the contexts where models are used in any credible way: performance on reputable multiple-choice exams used to test humans are often overvalued in reasoning about model capabilities.
    \item \textbf{Experts disagree on capability trajectories.}
    In spite of widespread capability evaluations, current evidence fails to clarify how capabilities will evolve: reputable experts currently have strikingly different predictions indicating that capabilities will advance slowly, rapidly, or extremely rapidly in the coming years.
    \item \textbf{Scale drives recent capability gains.}
    While many factors contribute to increased model capabilities, scale is the most important determinant: state-of-the-art foundation models have estimated annual cost increases of approximately 4x in computational resources (compute) used for training and 2.5x in training dataset size.
\end{enumerate}

\subsection{Vignette on emergent capabilities}
\label{sec:emergent-capabilities}
New foundation models routinely present novel capabilities, irrespective of the specific methods for determining the capability or the relevance of the capability to society.
While a broad array of evaluations aim to measure foundation model capabilities, a key subject to understand is the rate at which capabilities evolve as a function of important variables (\eg time, resources spent to develop the foundation model, resource spent to use the foundation model). 
The International Scientific Report includes discussion of capability forecasting as a key lens towards anticipating future societal impacts and informing future AI policy.
Here I describe what it means for a capability to be \textit{emergent} based on our work in \citet{wei2022emergent}.

\paragraph{Definition.}
\begin{quote}
    \emph{An capability is emergent if it is not present in smaller models but is present in larger models. 
    }
\end{quote}

\noindent Emergent capabilities would not have been directly predicted by extrapolating a scaling law (\ie\ consistent performance improvements) from small-scale models.
When visualized via a scaling curve ($x$-axis: model scale, $y$-axis: performance), emergent capabilities show a clear pattern---performance is near-random until a certain critical threshold of scale is reached, after which performance increases to substantially above random.
This qualitative change is also known as a \textit{phase transition}---a dramatic change in overall behavior that would not have been foreseen by examining smaller-scale systems \citep{huberman1987phase}.
While there are multiple senses in which a capabilities may be emergent, here we will focus on a cost-centric characterization.\footnote{
When we wrote this paper in 2022 \citep{wei2022emergent}, model scale (in terms of number of parameters) and model cost (in terms of training FLOPs) were tightly correlated across different developers. 
However, at the time of writing this dissertation in 2025, this relationship is much more complex, especially given large architectural variation (\eg sparse mixture-of-experts vs. traditional dense transformers), at present.}

Models have been scaled primarily along three factors: amount of computation, number of model parameters, and training dataset size \citep{kaplan2020scaling,hoffmann2022training}.
We analyze scaling curves by plotting the performance of different models where training compute for each model is measured in FLOPs on the $x$-axis \citep{hoffmann2022training}.
Using training FLOPs or model parameters as the $x$-axis produces curves with similar shapes due to the fact that most dense Transformer model families have scaled training compute roughly proportionally with model parameters \citep{kaplan2020scaling}.

We focus on training compute, but there is no single reliable proxy that comprehensively captures resource expenditure.
For example, Chinchilla \citep{hoffmann2022training} has one-fourth as many parameters as Gopher \citep{rae2021gopher} but uses similar training compute; and sparse mixture-of-expert models have more parameters per training/inference compute than dense models \citep{fedus2021switch, du2021glam}.
Overall, it may be wise to view emergence as a function of many correlated variables.

Note that the scale at which a capability is first observed to emerge depends on a number of factors and is not an immutable property of the capability. 
For instance, emergence may occur with less training compute or fewer model parameters for models trained on higher-quality data.
Conversely, emergent capabilities also crucially depend on other factors such as not being limited by the amount of data, its quality, or the number of parameters in the model.
Models are not trained optimally \citep{hoffmann2022training}, and our understanding of how to best train models will evolve over time.

\begin{figure}[htbp]
    \begin{centering}
    \includegraphics[width=\textwidth]{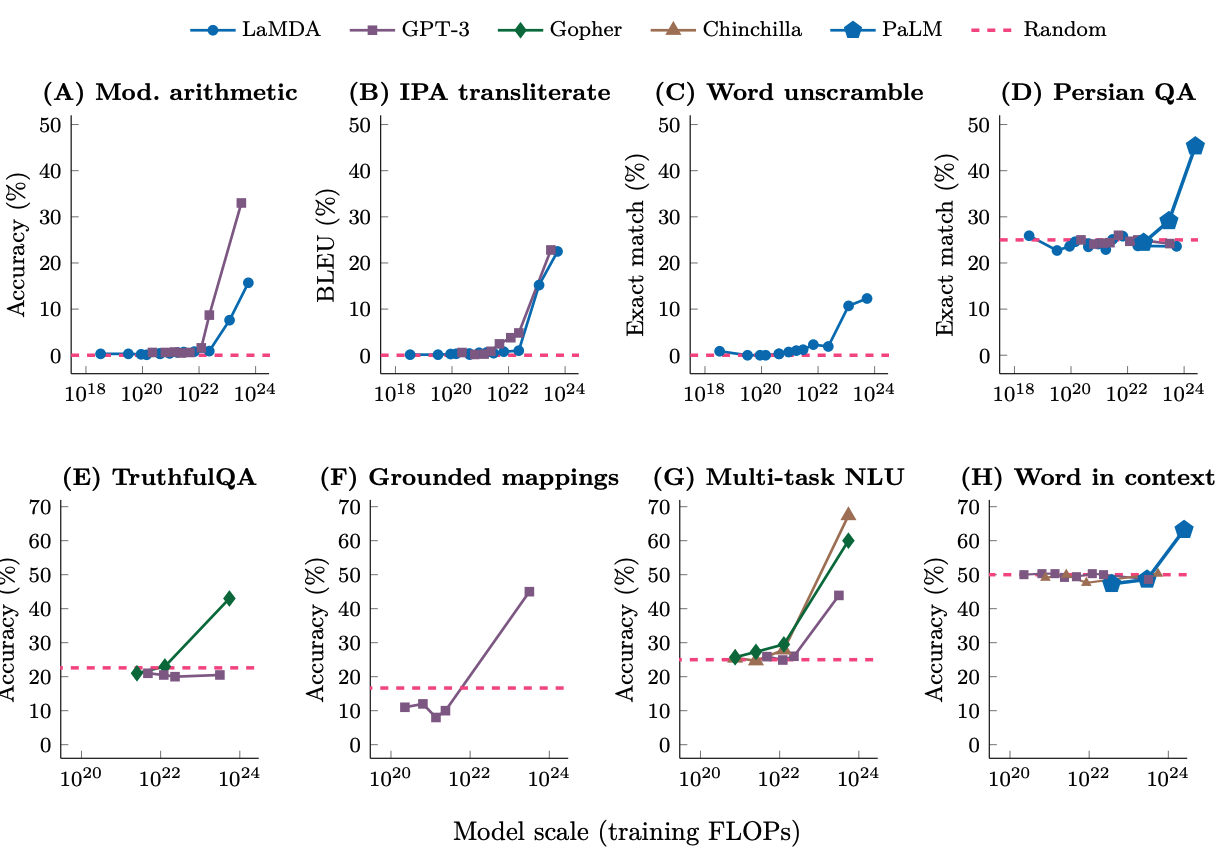}
    \caption{\textbf{Emergent capabilities}. We plot the performance of different models to depict emergent capabilities. Figure taken from \citet{wei2022emergent}.} 
    \label{fig:intrinsic-emergent}
    \end{centering}
\end{figure}

\paragraph{Evidence.}
We discuss emergent capabilities in the \textit{prompting} paradigm popularized by GPT-3 \citep{brown2020gpt3}.
The capability to perform a task via few-shot prompting is emergent when a model has random performance until a certain scale, after which performance increases to well-above random.
\autoref{fig:intrinsic-emergent} shows eight such emergent capabilities spanning five language model families from various works.

\begin{figure}[htbp]
    \begin{centering}
    \includegraphics[width=\textwidth]{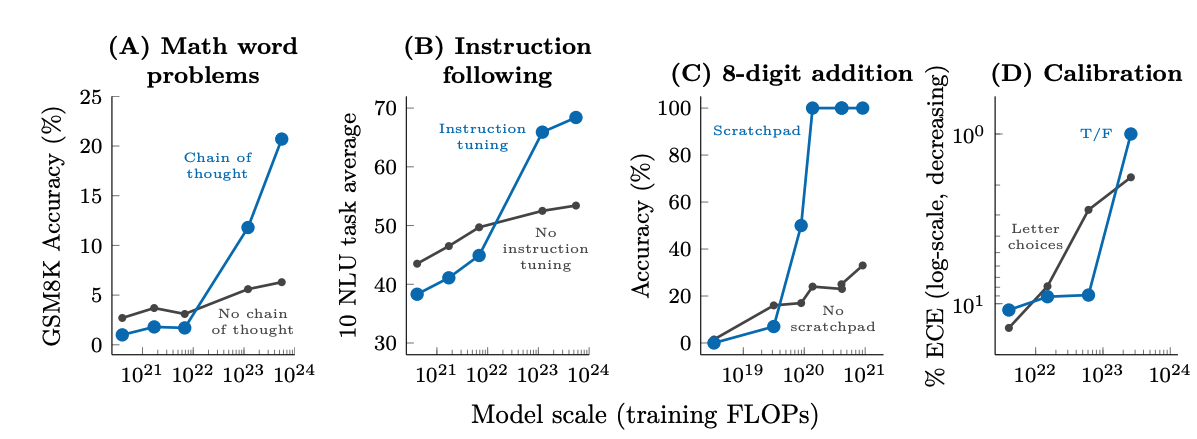}
    \caption{\textbf{Emergent interventions.} We depict the performance of different models when using different adaptation techniques to demonstrate the emergent meta-level capability to make use of these interventions.
    Figure taken from \citet{wei2022emergent}.} 
    \label{fig:emergent-interventions}
    \end{centering}
\end{figure}

Beyond few-shot prompting, other methods exist to adapt language models.
If a technique shows no improvement or is harmful when compared to the baseline of not using the technique until applied to a model of a large-enough scale, we also consider the technique an emergent capability.
We provide evidence in \autoref{fig:emergent-interventions}.

\vspace{3mm}
\noindent
\textbf{Multi-step reasoning.}
Reasoning tasks, especially those involving multiple steps, have been challenging for language models.
Chain-of-thought prompting enables language models to solve such problems by guiding them to produce a sequence of intermediate steps before giving the final answer \citep{cobbe2021training,wei2022chain,suzgun2022challenging}.
As shown in \autoref{fig:emergent-interventions}A, chain-of-thought prompting only surpasses standard prompting without intermediate steps when scaled to $10^{23}$ training FLOPs ($\sim$100B parameters).

\vspace{3mm}
\noindent
\textbf{Instruction following.}
Following instructions is a desired capability for language models where models perform tasks only given instructions describing the task without requiring few-shot demonstrations.
By finetuning on a mixture of tasks phrased as instructions, language models have been shown to respond appropriately to instructions describing an unseen task \citep{wei2021finetuned,sanh2021multitask}.
As shown in \autoref{fig:emergent-interventions}B, \citet{wei2021finetuned} found that this instruction-finetuning technique hurts performance for models of $7 \cdot 10^{21}$ training FLOPs (8B parameters) or smaller, and only improves performance when scaled to $10^{23}$ training FLOPs ($\sim$100B parameters).

\vspace{3mm}
\noindent
\textbf{Program execution.}
Executing programs is beneficial for models to perform complex multi-step computational tasks (\eg arithmetic with large numbers).
As shown in \autoref{fig:emergent-interventions}C, on 8-digit addition, using a scratchpad only improves program execution capabilities for models of ${\sim}9 \cdot 10^{19}$ training FLOPs (40M parameters) or larger.

\vspace{3mm}
\noindent
\textbf{Model calibration.}
Communicating uncertainty is an important aspect of trustworthy AI: language models lack good calibration in terms of their ability to predict whether they will answer questions correctly.
As shown in \autoref{fig:emergent-interventions}D, the superiority of the True/False technique for calibrating models only emerges when scaled to the largest model scale of ${\sim}3 \cdot 10^{23}$ training FLOPs (52B parameters).

Cumulatively, we demonstrate a variety of emergent capabilities.
A characterization of which capabilities will emerge, at what scale, and why, remains largely unclear, even given subsequent research that followed this initial work.
A more extensive enumeration of similar evidence is provided in \autoref{tab:long-list}.

\begin{table}[htp]
\resizebox{\textwidth}{!}{
    \begin{tabular}{l ccc}
     & Train.\ FLOPs & Params. & Model\\
    \midrule
    \underline{Few-shot prompting abilities} & & & \\
    Addition/subtraction (3 digit) & 2.3E+22 & 13B & GPT-3 \\
    Addition/subtraction (4-5 digit) & 3.1E+23 & 175B & GPT-3 \\
    MMLU Benchmark (57 topic avg.) & 3.1E+23 & 175B & GPT-3 \\
    Toxicity classification (CivilComments) & 1.3E+22 & 7.1B & Gopher \\
    Truthfulness (Truthful QA) & 5.0E+23 & 280B & Gopher \\
    MMLU Benchmark (26 topics) & 5.0E+23 & 280B & Gopher \\
    Grounded conceptual mappings & 3.1E+23 & 175B & GPT-3 \\
    MMLU Benchmark (30 topics) & 5.0E+23 & 70B & Chinchilla \\
    Word in Context (WiC) benchmark & 2.5E+24 & 540B & PaLM \\
    Many BIG-Bench tasks (see Appendix E of \citet{wei2022emergent}) & Many & Many & Many \\
    \midrule
    \underline{Augmented prompting abilities} & & & \\
    Instruction following (finetuning) & 1.3E+23 & 68B & FLAN  \\
    Scratchpad: 8-digit addition (finetuning) & 8.9E+19 & 40M & LaMDA \\
    Using open-book knowledge for fact checking & 1.3E+22 & 7.1B & Gopher \\
    Chain-of-thought: Math word problems & 1.3E+23 & 68B & LaMDA \\
    Chain-of-thought: StrategyQA & 2.9E+23 & 62B & PaLM \\
    Differentiable search index & 3.3E+22 & 11B & T5 \\
    Self-consistency decoding & 1.3E+23 & 68B & LaMDA \\
    Leveraging explanations in prompting & 5.0E+23 & 280B & Gopher \\
    Least-to-most prompting & 3.1E+23 & 175B & GPT-3 \\
    Zero-shot chain-of-thought reasoning & 3.1E+23 & 175B & GPT-3 \\
    Calibration via P(True) & 2.6E+23 & 52B & Anthropic LM \\
    Multilingual chain-of-thought reasoning & 2.9E+23 & 62B & PaLM \\
    Ask me anything prompting & 1.4E+22 & 6B & GPT-J \\
    \bottomrule
    \end{tabular}}
\caption{\textbf{Catalog of emergent capabilities.} More comprehensive evidence of emergent capabilities.
}
\label{tab:long-list}
\end{table}

\paragraph{Discussion.}
Several capabilities have only been demonstrated in resource-intensive language models.
Therefore, predicting performance by simply extrapolating the performance of less-intensive models is ineffective.
This raises the question of whether further scaling could yield additional emergent capabilities, especially where current models do not demonstrate these capabilities.

The capability for scale to unpredictably enable new techniques is not just theoretical.
Consider the Word in Context (WiC) benchmark shown in \autoref{fig:intrinsic-emergent}H, as a historical example.
Here, scaling GPT-3 to around $3 \cdot 10^{23}$ training FLOPs (175B parameters) failed to unlock above-random one-shot prompting performance.\footnote{GPT-3 does achieve slightly above-random performance on the development set with few-shot instead of one-shot prompting ($\sim$55\%), but this above-random performance did not hold on the test set.}
Regarding this negative result, \citet{brown2020gpt3} cited the model architecture of GPT-3 or the use of an autoregressive language modeling objective (rather than using a denoising training objective) as potential reasons, and suggested training a model of comparable size with a bidirectional architecture as a remedy.
However, later work found that further scaling a decoder-only language model was actually enough to enable above-random performance on this task.
As is shown in \autoref{fig:intrinsic-emergent}H, scaling PaLM \citep{chowdhery2022palm} from $3 \cdot 10^{23}$ training FLOPs (62B parameters) to $3 \cdot 10^{24}$ training FLOPs (540B parameters) led to a significant jump in performance, without the significant architectural changes suggested by \citet{brown2020gpt3}.

\paragraph{Potential explanations of emergence.}\label{subsec:potential_explanations}
Although there are dozens of examples of emergent capabilities, there are currently few compelling explanations for why such capabilities emerge in the way they do.
For certain tasks, there may be natural intuitions for why emergence requires a model larger than a particular threshold scale.
For instance, if a multi-step reasoning task requires $l$ steps of sequential computation, this might require a model with a depth of at least $O\left(l\right)$ layers.
It is also reasonable to assume that more parameters and more training enable better memorization that could be helpful for tasks requiring world knowledge.
As an example, good performance on closed-book question-answering may require a model with enough parameters to capture the compressed knowledge base itself (though language model-based compressors can have higher compression ratios than conventional compressors \citep{bellardlmcompressor}).

It is also important to consider the evaluation metrics used to measure emergent capabilities \citep{srivastava2022bigbench}.
For instance, using exact string match as the evaluation metric for long-sequence targets may disguise compounding incremental improvements as emergence.
Similar logic may apply for multi-step or arithmetic reasoning problems, where models are only scored on whether they get the final answer to a multi-step problem correct, without any credit given to partially correct solutions.
However, the jump in final answer accuracy does not explain why the quality of intermediate steps suddenly emerges to above random, and using evaluation metrics that do not give partial credit are at best an incomplete explanation, because emergent capabilities are still observed on many classification tasks (\eg the tasks in \autoref{fig:intrinsic-emergent}D--H).

As an alternative evaluation, we measure cross-entropy loss, which is used in scaling laws for pre-training, for the six emergent BIG-Bench tasks.
This analysis follows the same experimental setup from \citet{srivastava2022bigbench} and affirms their conclusions for the six emergent tasks we consider.
Namely, cross-entropy loss improves even for small model scales where the downstream metrics (exact match, BLEU, and accuracy) are close to random and do not improve, which shows that improvements in the log-likelihood of the target sequence can be masked by such downstream metrics.
However, this analysis does not explain why downstream metrics are emergent or enable us to predict the scale at which emergence occurs.
Overall, more work is needed to tease apart what enables scale to unlock emergent capabilities.

\paragraph{Beyond scaling.}\label{subsec:beyond-scaling}
Although we may observe an emergent capability to occur at a certain scale, it is possible that the capability could be later achieved at a smaller scale---in other words, model scale is not the singular factor for unlocking an emergent capability.
As the science of training progresses, certain capabilities may be unlocked for smaller models with new architectures, higher-quality data, or improved training procedures.
For example, there are 14 BIG-Bench tasks for which LaMDA 137B and GPT-3 175B models perform at near-random, but PaLM 62B in fact achieves above-random performance, despite having fewer model parameters and training FLOPs.
While there is not an empirical study ablating every difference between PaLM 62B and prior models (the computational cost would be too high), potential reasons for the better performance of PaLM could include high-quality training data (\eg more multilingual and code data than LaMDA) and architectural differences (\eg split digit-encodings; see Section 2 in \citet{chowdhery2022palm}).
Another potentially way of unlocking emergence is through a different pre-training objective---it was shown in \citet{tay2022transcending} that a computationally-efficient continued pre-training stage on a mixture-of-denoisers objective \citep{tay2022unifying} enabled emergent performance on several BIG-Bench tasks.

Moreover, once a capability is discovered, further research may make the capability available for smaller scale models. %
Consider the instruction following capability.
Although \citet{wei2021finetuned} initially found that instruction-based finetuning only worked for 68B parameter or larger decoder-only models, \citet{sanh2021multitask} induced similar behavior in a 11B model with an encoder-decoder architecture, which typically has higher performance after finetuning than decoder-only architectures.
As another example, \citet{ouyang2022instructions} proposed a finetuning and reinforcement learning from human feedback approach for the InstructGPT models, which enabled a 1.3B model to outperform much larger models in human-rater evaluations on a broad set of use cases.
As we continue to train language models, lowering the scale threshold for emergent capabilities will become more important for making research on such capabilities accessible to the broader community.

\paragraph{Emergent risks.}\label{subsec:emergent-risks}
The concept of emergence, while often centered on capabilities, naturally can be defined for risks.
This not only addresses how capabilities are often primitives for risks (\eg dual-use capabilities that are misused by malicious actors to cause harm), but also that risks themselves may change with scale (\eg the propensity of models to generate toxic content).
That is, risks could also emerge \citep{bommasani2021opportunities,jacobsrisks,ganguli2022predictability}.
Understanding how risks scale is important irrespective of whether they emerge in particular: since work on emergent capabilities incentivizes scaling language models, it is important to be aware of risks that increase with model scale even if they are not emergent. 

Here, we summarize several prior findings on the relationship between  risks and scale.
On WinoGender, which measures gender bias in occupations such as ``nurse'' or ``electrician,'' scaling has improved performance so far \citep{du2021glam,chowdhery2022palm}, though \citet{srivastava2022bigbench} found on the BBQ bias benchmark \citep{parrish-etal-2022-bbq} that bias can increase with scaling for ambiguous contexts.
For toxicity, \citet{askell2021general} found that while larger models could produce more toxic responses, this behavior could be mitigated by giving models prompts with examples of being ``helpful, harmless, and honest.''
For extracting training data from language models, larger models were found to be more likely to memorize training data \citep{carlini2021extracting,carlini2022quantifying}, though deduplication methods have been proposed and can simultaneously reduce memorization while improving performance \citep{kandpal2022deduplicating,lee2021deduplicating}.
The TruthfulQA benchmark \citep{lin2021truthfulqa} showed that GPT-3 models were more likely to mimic human falsehoods as they got larger, though \citet{rae2021gopher} later showed on a multiple-choice version that scaling Gopher to 280B enabled emergent performance substantially better than random.
Beyond the above, emergent risks also include phenomena that might only exist in future language models or that have not yet been characterized in current language models.
Approaches involving data filtering, forecasting, governance, and automatically discovering harmful behaviors have been proposed for discovering and mitigating emergent risks \citep[][\textit{inter alia}]{bender2021dangers,weidinger2021ethical,jacobsrisks,ganguli2022predictability,perez2022red}.

\paragraph{Sociological change.}
Finally, the emergent capabilities discussed here focus on model behavior and are just one of several types of emergence in NLP \citep{manning2020}.
Another notable type of qualitative change is sociological change, in which increasing scale has shifted how the community views and uses language models.
For instance, NLP has historically focused on task-specific models \citep{jurafsky2000speech}.
Recently, scaling has led to an explosion in research on and development of foundation models that are ``general purpose'' in that they are single models that aim to perform a range of tasks not explicitly encoded in the training data \citep{manning2022human}.

One key set of results in the sociological shift towards foundation models is when scaling enables a few-shot prompted foundation model to outperform finetuned task-specific models.
As a few examples, GPT-3 175B achieved new state of the art on the TriviaQA and PiQA question-answering benchmarks \citep{brown2020gpt3}; PaLM 540B achieved new state of the art on three arithmetic reasoning benchmarks \citep{chowdhery2022palm}; and the multimodal Flamingo 80B model achieved new state of the art on six visual question answering benchmarks \citep{alayrac2022flamingo}.
In all of these cases, state-of-the-art performance was achieved by few-shot prompting a language model of unprecedented scale.
These capabilities are not necessarily emergent since they have smooth, predictable scaling curves---however, they do underscore a shift towards foundation models across AI research. 

\section{Risks}
Foundation models are general-purpose technologies that are associated with an incredibly broad portfolio of risks.
To make this vivid, consider this list of risks that I personally worked on and wrote about during my PhD: bias, biological attacks, child sexual abuse material, concentration of power, copyright violations, cyber attacks, disinformation, educational harms, environmental harms, fraud, homogeneous outcomes, job loss, loss of control, misinformation, non-consensual intimate imagery, reliability issues, scams, security issues, and spearphishing.
However, to be precise, the fact that many risks are associated with foundation models does not imply that foundation models are harmful, let alone very harmful, to part or all of society.
In particular, a central focus of my research has been to map these cognizable risks to the amount of current evidence, including documented harm.

To reason about the risks of foundation models, at the start of my PhD, we identified a clean ternary taxonomy for risks \citep{bommasani2021opportunities}.
Namely, risks could be due to the unintentional or intentional use of known model capabilities (\ie differentiating risks like unfair outcomes from risks like facilitating disinformation), or due to unknown properties (\ie security risks like adversarial triggers). 
More generally, this work also provided an early presentation of uniting frames that have at times been put at odds (\eg safety vs. ethics): \citet{bommasani2021opportunities} includes sections on inequity and fairness, misuse, environmental risks, legal risks, security and privacy, as well as sections on AI safety and broader ethical concerns.\footnote{These risks heavily overlap with the underlying risks described in the International Scientific Report, though they are re-organized under three different categories of (i) malicious risks, which are essentially what I call misuse risks; (ii) malfunction risks, which are essentially what I call unintentional risks with the addition of loss of control, and  (iii) systemic risks, which captures risks that are not only defined at the model level. The coverage of some forms of the security risks I describe in the International Scientific Report is reasonably unclear.}
In the middle of my PhD, we built on this taxonomy in the design of HELM, which featured metrics related to bias/robustness/calibration/fairness/toxicity as well as scenarios related to copyright/disinformation/bias/toxicity \citep{liang2022helm}. 
Later in my PhD, we more extensively taxonomized misuse risks, based on prior work that associated these risks with open foundation models: we studied voice cloning scams, spearphishing scams, cyberattacks, disinformation, bioweapons, non-consensual intimate imagery, and child sexual abuse material \citep{kapoor2024societal}.\footnote{This taxonomy near-exactly matches the taxonomy used in the International Scientific Report for ``Malicious risks'' as I describe below.}
While the taxonomies have evolved over time and others, most notably Laura Weidinger, have built important taxonomies \citep{weidinger2022taxonomy}, including from concrete evidence \citep{Marchal2024GenerativeAM}, the high-level risk categories are well-identified at this point.
However, what remains unclear is the lower-level threat models and which to prioritize within broad high-level risk categories.

Below, I detail the characterization of current risks, based on the International Scientific Report that reflects my current best high-level conceptualization of the ever-evolving risks of foundation models.\footnote{This matches what I wrote in the California Report on Frontier AI Policy \citep{bommasani2025ca}.}

\subsection{Current risks}
The International Scientific Report defines three general risk categories: malicious use risks, risks from malfunctions, and systemic risks. 
\begin{enumerate}
    \item 
    Malicious risks are risks where malicious actors misuse foundation models to deliberately cause harm. These include simulated content such as non-consensual intimate imagery (NCII), child sexual abuse material (CSAM), and cloned voices used in financial scams; manipulation of public opinion via disinformation; cyberattacks; and biological and chemical attacks.
    \item
    Malfunction risks are risks where non-malicious actors use foundation models as intended, yet unintentionally cause harm. These include reliability issues where models may generate false content, bias against certain groups or identities, and loss of control where models operate in harmful ways without the direct control of a human overseer.
    \item
    Systemic risks are risks associated with the widespread deployment of foundation models and not exclusively model-level capabilities. These include labor market disruption, global AI R\&D concentration, market concentration, single points of failure, environmental risks, privacy risks, and copyright infringement.
\end{enumerate}

To very succinctly characterize the level of evidence for each of these risks, the International Scientific Report delineates two categories. First, some risks have clear and established evidence of harm: scams, NCII, CSAM, bias, reliability issues, and privacy violations. Second, some risks have unclear but growing evidence, which is tied to increasing capabilities: large-scale labor market impacts, AI-enabled hacking or biological attacks, and loss of control. Given that current evidence is only partial for this second category, experts often disagree based on different interpretations of model capabilities (\eg the ability of models to reason scientifically or write code) and how capabilities mediate risks (\eg stronger scientific reasoning capabilities could be a core primitive for heightened malicious risk in the development of bioweapons). Overall, experts interpret current evidence for this second category of risks to arrive at meaningfully different predictions: Some think that such risks are decades away, while others think societal-scale harm could arise within the next few years.

\subsection{Vignette on homogeneous outcomes}
\label{sec:homogeneous}
The foundation model paradigm centers on reuse: a single foundation model is built at immense cost and this model is reused across many applications to favorably amortize the cost.
However, foundation models are not the first instance of widespread reuse in artificial intelligence.
The successes of modern machine learning are predicated on a strong tradition of sharing.
The AI community shares datasets (\eg The Pile), models (\eg Llama 3.3), libraries (\eg JAX), evaluations (\eg HELM) and much more.
No example makes this more vivid than the vibrant open-source ecosystem that underpins deep learning.
This ethos of sharing serves the field well: we are able to repeatedly capitalize on the effort required to build high-quality assets (\eg ImageNet has supported thousands of researchers in computer vision), and improvements to these assets have sweeping benefits (\eg BERT raised all boats in natural language processing). 
Does sharing have risks?
Here I describe the risk of \textit{homogeneous outcomes} based on our work in \citet{bommasani2022homogenization} and \citet{toups2023ecosystemlevel}.

Sharing can be reinterpreted as monoculture: \citet{kleinberg2021monoculture} define \textit{algorithmic monoculture} as the state "in which many decision-makers all rely on the [exact] same algorithm." 
In parts of society where algorithmic systems are ubiquitous, we see trends towards such monoculture \citep{Moore2018, Engler2021}.
Monocultures often pose serious risks: \citet{kleinberg2021monoculture} show monoculture is suboptimal for decision-makers when their decisions are interconnected, as when they compete to hire job candidates.

Foundation models are unprecedented instances of algorithmic monoculture: the same models can power applications across industry sectors, and underpin large fractions of the entire AI industry's impact on society. 
To reason about these forms of monoculture, which are more subtle than the whole-cloth sharing of algorithms envisioned by \citet{kleinberg2021monoculture}, in \citet{bommasani2022homogenization} we introduced a more expansive definition of algorithmic monoculture, where AI systems need not be identical, but share significant components (\eg training data, foundation models).
The monoculture of foundation models may present a variety of risks.
Our focus will be on a new family of systemic risks that we call homogeneous outcomes.
Namely, we are concerned with the phenomenon where individuals (or groups) exclusively receive negative outcomes from \textit{all} decision-makers they interact with.\footnote{For brevity, we discuss the simpler case here where the harm is from the exclusively negative outcomes. In other cases, homogeneity in the form of the same (though not necessarily) outcome may still yield harm. Namely, there may be circumstances where the reduction of diversity via collapse to a single mode is a problem (\eg applications in creative writing \citep{padmakumar2023does}).}

For example, a job applicant may be rejected from every job they apply to due to the use of similar algorithmic resume screening systems at all companies.
Homogeneous outcomes, especially in certain high-stakes settings, constitute serious harms to individuals: someone who is rejected from every employment opportunity, or denied admission at every school, may be severely compromised (\eg unable to provide for their family, unable to secure an education).  
We view outcome homogenization as an important class of \textit{systemic} harms.

\paragraph{Significance.}
Homogeneous outcomes are a class of systemic harms which warrant greater consideration given the pervasive practice of algorithmic monoculture.
In some settings, homogeneous outcomes are unequivocally harmful: it is undesirable for every ML model to fail for an individual (\eg in the context of automatic speech recognition, so as to render the individual unable to use any commercial voice assistant).
In other settings, it is up for debate whether homogeneous outcomes constitute harm.
Consider hiring: surely it is a harm if someone is locked out from employment across all employers?
Yet, some people must be rejected from every employer (in the absence of broader structural change) if the total number of applicants exceeds the total number of vacancies.
And further, clearly there are conditions where societal practices indicate homogeneous outcomes are not only acceptable, but desirable: an individual who has not passed the US medical licensing exam should be rejected from all positions that require a medical license to practice.

Determining whether homogeneous outcomes are social harms is \textbf{contextual}: the particular circumstances influence whether, and to what extent, homogeneity is of moral concern.
To provide a brief theoretical account, consider the relational egalitarian perspective.
Relational egalitarianism, as a theory of justice, argues that individuals must \textit{relate} to each other as equals in a just society.
Relational approaches have seen recent adoption in relation to AI: \citet{birhane2021algorithmic} presents a relational account of algorithmic injustice with similar approaches taken to data governance, decision-making, and fairness \citep{viljoen2021, kasy2021, fish2022}. 

To ground a relational analysis, we return to the context of hiring.
While individual organizations may establish rankings of candidates, and indeed we would expect that companies within a market sector will often agree on a hierarchy, the same hierarchy should not consistently dominate an entire sector or territory such that some people are entirely excluded from work \citep[74]{Anderson1999}. 
Anderson argues that ``to be capable of functioning as an equal participant in a system of cooperative production requires \dots access to the education needed to develop one’s talents, freedom of occupational choice, the right to make contracts and enter into cooperative agreements with others, the right to receive fair value for one’s labor, and recognition by others of one’s productive contributions'' \citep[318]{Anderson2016}. 
If some people are consistently excluded from job interviews and, therefore, employment, they will not enjoy freedom of occupational choice; if they are excluded from higher education, they will struggle to develop their talents. 
Not only do those excluded personally suffer from the establishment of the hierarchy, they also are unable to function as equal participants in society. 
Because employment, education, and credit are foundational social goods, consistent exclusion from them risks establishing a social hierarchy of esteem or domination. 
The hierarchy of esteem in turn damages the capability of the excluded to relate to others as equal democratic citizens \citep{Anderson1999}.

Under this account, we emphasize how the moral importance of the harm depends on the scale of exclusion.
That is, in its purest form, there is a strong \textit{threshold} effect: homogeneous outcomes are most severe when the individual is denied from \textit{all} sources of employment (or education or credit or so on). 
If autonomy is access to a sufficient range of sufficiently varied opportunities, then it is not a harm to be denied one opportunity, such as a job or a loan \citep{Raz1988}. 
It is a harm, however, to be shut out of all opportunities.
For this reason, in this work, we study the extreme case in which individuals are shut out of all opportunities.\footnote{We do note that while \textit{all} might be a sufficient condition for harm, it may suffice in some contexts to be denied access to a significant fraction of opportunities. For this reason, while we study the strongest form of exclusion, we encourage future work to put forth technical, experimental, and moral analyses of homogeneous outcomes in the more general setting where individuals are denied \textit{most} opportunities, not just all opportunities.}

\paragraph{Measurement.}
If homogeneous outcomes warrant concern, both because standard industrial practices suggest they will arise (due to monoculture) and philosophical theories suggest they may be serious harms, how do we measure them?
Unlike other harms (\eg discrimination), homogeneous outcomes permit a clear translation from the theoretical construct to the operationalized metric, because the phenomena of interest is more clearly observed (\ie we can count how many individuals do or do not receive homogeneous outcomes).
Nonetheless, design choices remain: given the observed number or rate of homogeneous outcomes, how do we interpret this value?

To define a metric for homogeneous outcomes, we first specify the conceptual frame.
Namely, how individuals interact with deployed AI systems determines AI's impact on their lives.
In some contexts, individuals routinely interact with multiple AI models or systems.
For example, when a candidate applies to jobs, they typically apply to several firms. 
The decision each company makes to accept or reject the candidate may be mediated by a single hiring algorithm. 
In other contexts, individuals select a single model from a set of options.
For example, when a consumer purchases a voice assistant, they typically choose between several options (\eg Amazon Alexa, Google Assistant, Apple Siri) to purchase a single product (\eg Amazon Alexa).
Given this recognition of how AI impacts individuals, the appropriate frame is one of \textit{ecosystem-level }analysis, because we are interested in the cumulative impact of AI on individuals.

\paragraph{Notation.}
\begin{figure}
    \centering
    \begin{subfigure}[b]{0.4\textwidth}
        \centering
        \includegraphics[width=\textwidth]{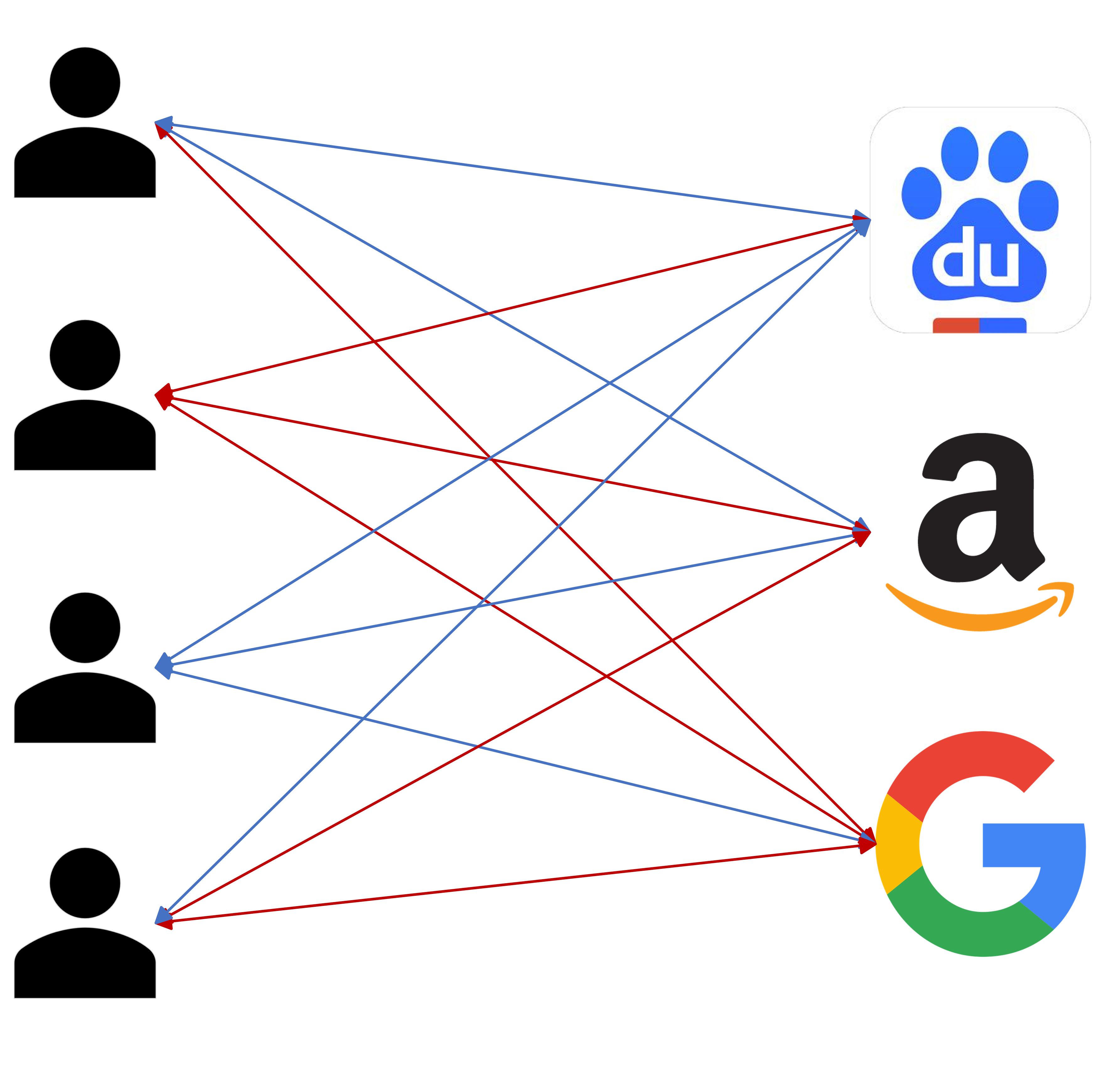}
    \end{subfigure}%
    \hspace{3em}%
    \begin{subfigure}[b]{0.4\textwidth}
        \centering
        \includegraphics[width=\textwidth]{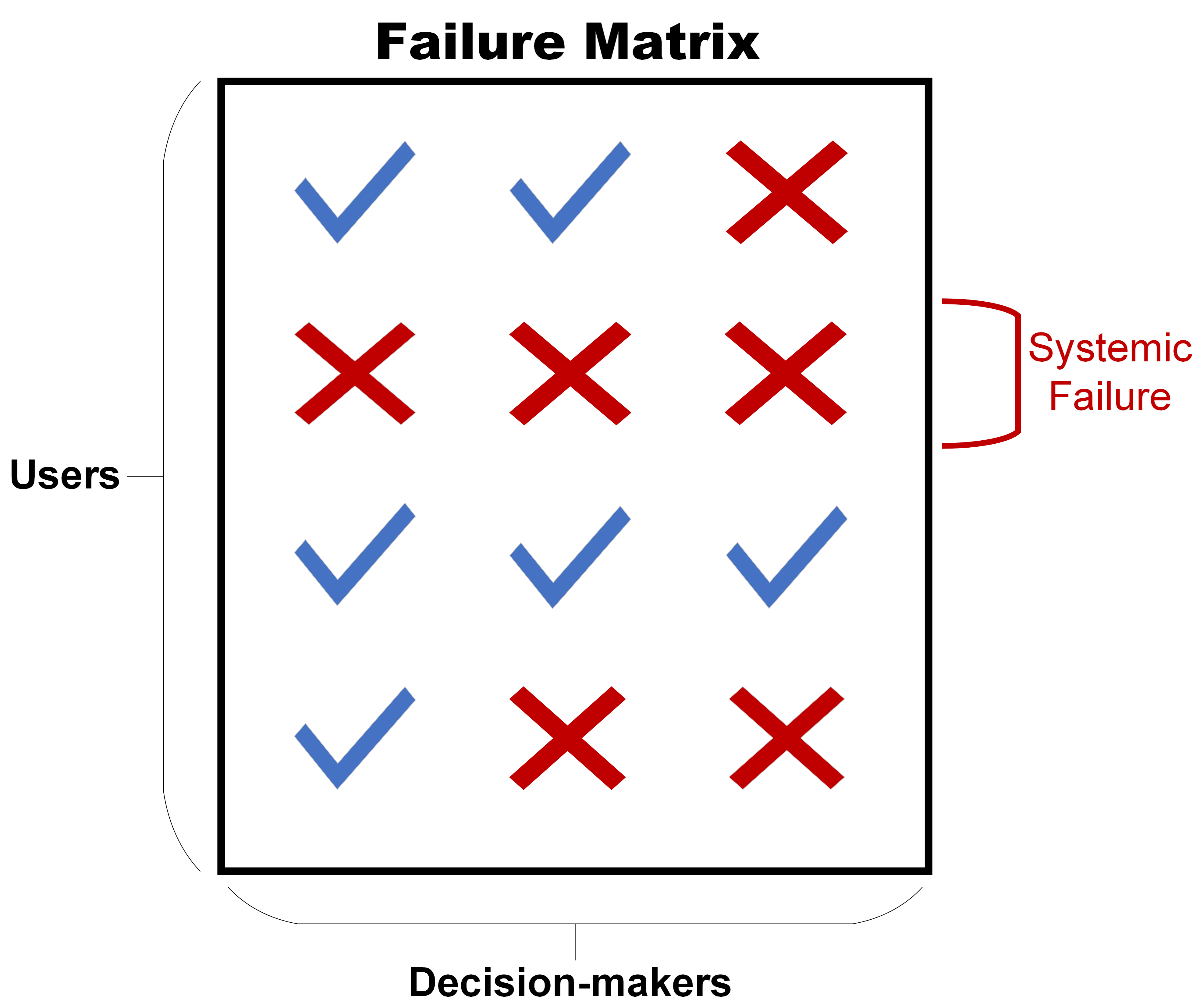}
    \end{subfigure}%
    \caption{\textbf{Ecosystem-level analysis.} 
    Individuals interact with decision-makers (\textit{left)}, receiving outcomes that constitute the failure matrix (\textit{right}).
    }
    \label{fig:failure_matrix}
\end{figure}

Consider $N$ individuals that do, or could, interact with $k$ decision-makers that apply $\hat{y}$ labels according to their decision-making processes $h_1, \dots, h_k$.
Individual $i$ is associated with input $x_i$, label $y_i$, and receives the label $\hat{y}_i^j = h_j(x_i)$ from decision-maker $j$.
Define the \textit{failure matrix} $F \in \{0, 1\}^{N \times k}$ such that $F[i, j] = \1 \left[\hat{y}_i^j \neq y_i\right]$.
The \textit{failure outcome profiles} $\mathbf{f}_i$ for individual $i$, which we refer to as the \outcomeprofile for brevity, denotes $F[i, :]$.
The \textit{failure rate} $\bar{f_j}$ for decision-maker $j$ is $\bar{f_j} = \frac{\sum_{i=1}^N F[i, j]}{N}$ (\ie the empirical classification error in classification).
For consistency, we order the entries of decision-makers (and, thereby, the columns of the failure matrix) in order of ascending failure rate: $F[:, 1]$ is the \outcomeprofile associated with the decision-maker with the fewest failures and $F[:, k]$ is the \outcomeprofile associated with the decision-maker with the most failures.
The failure matrix is the central object in ecosystem-level analysis (see \autoref{fig:failure_matrix}). 

\paragraph{Systemic failure.}
As mentioned before, while the general phenomenon of homogeneous outcomes may warrant concern, for this dissertation we will discuss the simpler case of homogeneous negative outcomes.
Here, we refer to such outcomes as \textit{systemic failures}.
Individual $i$ experiences systemic failure if they exclusively experience failure across the domain of interest: $F[i, :] = [1, \dots, 1]$.
Not only are systemic failures the worst possible outcomes, but they also often result in additional harms. If an individual applying to jobs is rejected everywhere, they may be unemployed.
If no commercial voice assistant can recognize an individual's voice, they may be fully locked out of accessing a class of technology.

\paragraph{Metric.}
In order for a measure of systemic failure to be useful, it must be (i) meaningful and (ii) comparable across systems. A challenge to the meaningfulness of any proposed metric is that 
systemic failures occur more often in an ecosystem with many inaccurate models.  A metric for systemic failure that primarily communicated the aggregate error rates of models in the ecosystem would not be meaningful as an independent metric.  It also would not support goal (ii) because we could not compare the rates of systemic failure across ecosystems with varying model accuracies. It would be difficult to identify a system with a `large' rate of systemic failure because 
the systemic failure properties would be dominated by the error rates of the models in the ecosystem. Therefore, achieving meaningfulness and comparability requires the metric to incorporate error correction.

Assuming model independence is a helpful baseline because it adjusts for model error rates without making assumptions about the correlation between models in the ecosystem. To avoid assumptions and for the sake of simplicity, therefore, we juxtapose the \textit{observed} behavior with a simple theoretical model in which we assume models fail independently of each other.
Under this assumption, the distribution of the \textit{\baseline} number of model failures $t \in \{0, \dots, k\}$ follows a Poisson-Binomial distribution parameterized by their failure rates (\autoref{p_baseline}). 
The baseline of independence also means that our metric does not attempt to quantify whether it is ``reasonable'' that the models all fail on some instances. For example, some instances might be harder (or easier) than others, making it more likely that all models will fail (or succeed) to classify that instance.  However, ``hardness'' is observer-relative.  What is hard for one class of models might be easy for another class of model, and likewise with humans with particular capabilities or training.  
Therefore correcting the metric to account for hardness would relativize the metric to the group of humans or class of models for whom that instance is hard.  
We choose independence as a baseline to be neutral on this point.\footnote{In other works, I have explored other baselines that are more general, but defer discussion to those works for simplicity.}

Comparing the true observed distribution of ecosystem-level outcomes with the \baseline distribution helps illuminate how correlated outcomes are across models. \\
These quantities, namely
$P_{\text{observed}}$ (\autoref{p_observed}) and $P_{\text{\baseline}}$ (\autoref{p_baseline}), are defined formally below:

\begin{align}
    P_{\text{observed}}(t~ \text{failures}) &= \frac{\sum_{i=1}^N \1 \left[t = \sum_{j=1}^k F[i,j] \right]}{N} \label{p_observed} \\  
    P_{\text{\baseline}}(t~ \text{failures}) &= \text{Poisson-Binomial}(\bar{f_1}, \dots, \bar{f_k})[t]. \label{p_baseline}
\end{align}

In \citet{bommasani2022homogenization}, we defined additional more complex metrics.
Yet, in reflecting on my entire line of work on homogeneous outcomes, I encourage work on measuring homogeneous outcomes to stick to these simple metrics as I believe further complexity in the metrics has only served to confuse, rather than illuminate, in my experience on this subject.

\paragraph{Current evidence.}
While I posit that the algorithmic monoculture of foundation models will precipitate homogeneous outcomes, the conditions to study this rigorously were not available during the time of my PhD.
In particular, the first condition of broad dependence on specific foundation models began to happen during my PhD, but the second condition of the affordances to measure it (\ie post-deployment monitoring, let alone data available to the research community) has not yet materialized.\footnote{This is partially why I, and others \citep[\eg][]{stein2024rolegovernmentsincreasinginterconnected}, have advocated for policy to improve post-deployment monitoring as I discuss later in this dissertation.}
Therefore, I instead provide evidence of homogeneous outcomes related to other older forms of widespread AI.

\paragraph{Data.}
To establish general trends made visible through \systemlevel analysis, we draw upon a large-scale three-year audit of commercial ML APIs \citep[HAPI;][]{chen2022hapi} to study the behavior of deployed ML systems across three modalities, eleven datasets, and nine commercial systems.

\citet{chen2022hapi} audit commercial ML APIs, tracking predictions across these APIs when evaluated on the same eleven standard datasets over a period of three years (2020 -- 2022).
We consider ML APIs spanning three modalities (text, images, speech), where each modality is associated with a task (SA: sentiment analysis, FER: facial emotion recognition, SCR: spoken command recognition) and 3 APIs per modality (\eg IBM, Google, Microsoft for spoken command recognition). The models evaluated are from Google (SA, SCR, FER), Microsoft (SCR, FER), Amazon (SA), IBM (SCR), Baidu (SA), and Face++ (FER).
Additionally, each modality is associated with three to four datasets, amounting to eleven datasets total; further details are deferred to the supplement.

To situate our notation, consider the \digit dataset for spoken command recognition and the associated APIs (IBM, Google, Microsoft). 
For each instance (\ie image) $x_i$ in \digit, the \outcomeprofile $\textbf{f}_i \in \{0, 1\}^3$ is the vector of outcomes.
The entries are ordered by ascending model failure rate: $F[:, 1]$ corresponds to the most accurate model (Microsoft) and $F[:, 3]$ corresponds to the least accurate model (Google).

\begin{table}[htp]
\resizebox{\textwidth}{!}{
\begin{tabular}{l|cccc|ccc|cccc}
\toprule
 & \multicolumn{4}{c}{\textbf{Facial emotion recognition}} & \multicolumn{3}{c}{\textbf{Spoken command recognition}} & \multicolumn{4}{c}{\textbf{Sentiment analysis}} \\
 &   \rafdb &    \afnet &    \expw & \ferplus &  \fluent &  \digit &  \amnist &    \shop &    \yelp &    \imdb & \waimai \\
\midrule
\textbf{Dataset size} &  15.3k &  287.4k &  31.5k &    6.4k &  30.0k &   2.0k &  30.0k &  62.8k &  20.0k &  25.0k &  12.0k \\
\textbf{Number of classes} &       7 &        7 &       7 &       7 &      31 &     10 &      10 &       2 &       2 &       2 &       2 \\
\midrule
\textbf{$h_1$ failure rate} (\ie error) &   0.283 &    0.277 &   0.272 &   0.156 &   0.019 &  0.217 &   0.015 &   0.078 &   0.043 &   0.136 &    0.110 \\
\textbf{$h_2$ failure rate} (\ie error) &   0.343 &    0.317 &   0.348 &   0.316 &   0.025 &  0.259 &   0.015 &   0.095 &   0.111 &   0.219 &   0.151 \\
\textbf{$h_3$ failure rate} (\ie error) &   0.388 &    0.359 &   0.378 &   0.323 &   0.081 &  0.472 &   0.043 &   0.122 &   0.486 &   0.484 &   0.181 \\
\textbf{Systemic failure rate} &   0.152 &    0.178 &   0.181 &   0.066 &    0.01 &  0.129 &   0.002 &   0.039 &   0.021 &   0.043 &   0.065 \\
\bottomrule
\end{tabular}}
\caption{\textbf{\hapi descriptive statistics.} 
Properties of the \hapi datasets along with the failure rates of the models evaluated on the datasets as well as the (observed) systemic failure rate.}
\label{tab:basic-stats}
\end{table}
To build general understanding of model performance in \hapi, we provide basic descriptive statistics (\autoref{tab:basic-stats}).
For most datasets, all APIs achieve accuracies within 5--10\% of each other (exceptions include \digit, \yelp, \imdb). 
Interestingly, we often find the systemic failure rate is roughly half the failure rate of the most accurate model $h_1$.

\begin{figure}
    \centering
    \begin{subfigure}[H]{0.48\textwidth}
        \centering
        \includegraphics[width=\textwidth]{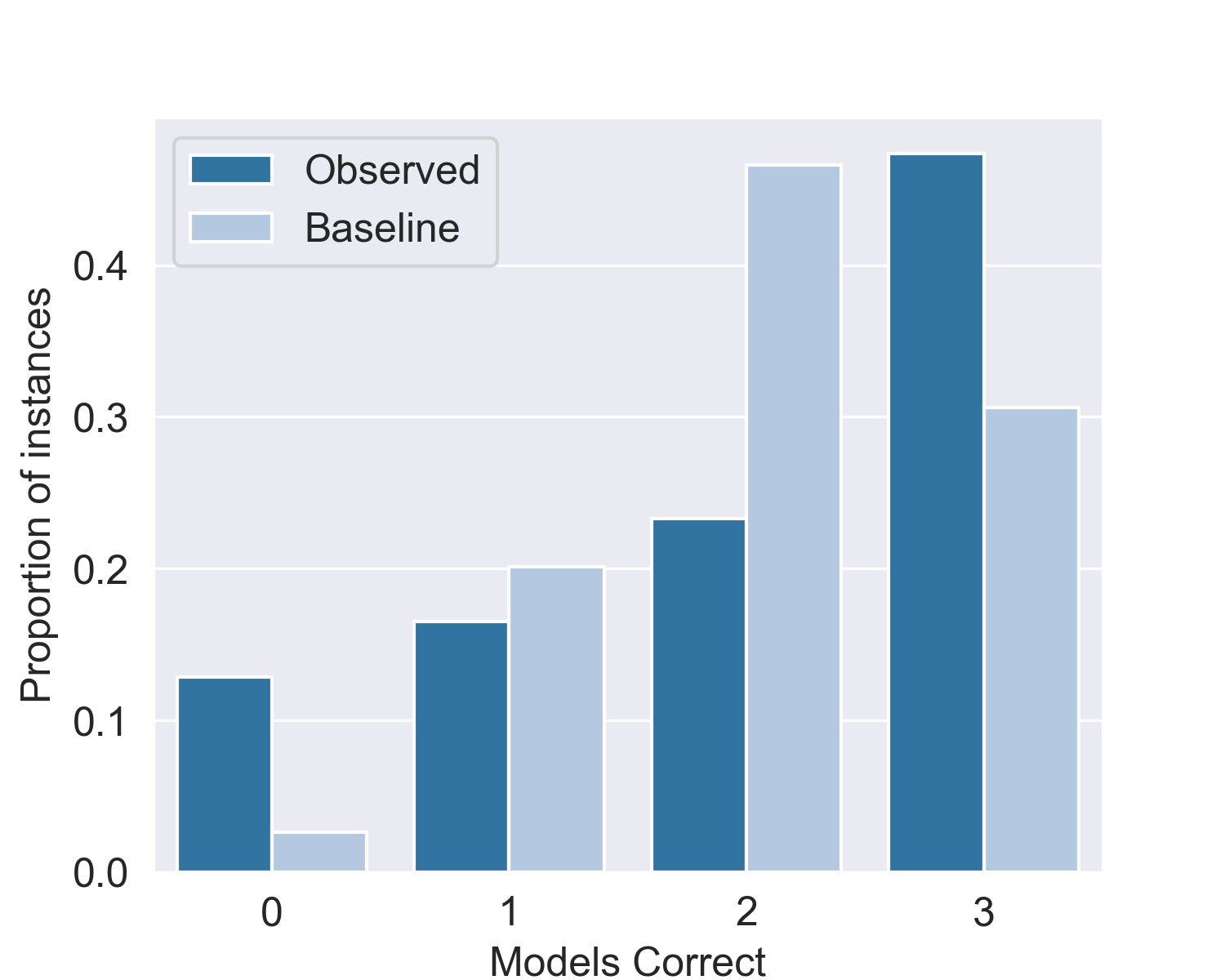}
        \caption{\digit.}
        \label{fig:digit_polarization}
    \end{subfigure}%
    \begin{subfigure}[H]{0.48\textwidth}
        \centering
        \includegraphics[width=\textwidth]{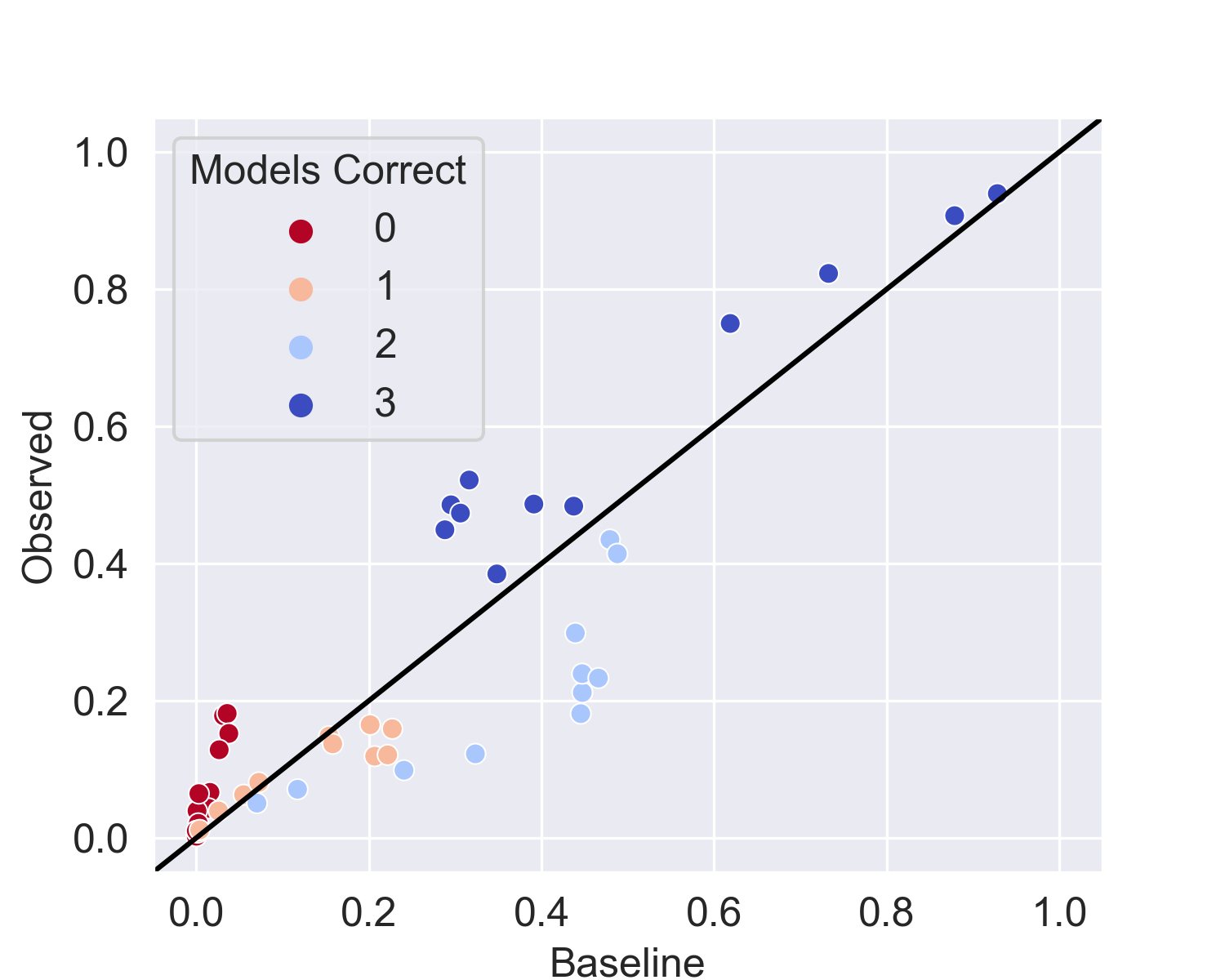}
        \caption{All datasets.}
        \label{fig:polarization_all}
    \end{subfigure}%
    \caption{\textbf{Evidence of homogeneous outcomes in deployed AI systems.} 
    \Systemlevel analysis surfaces the general trend of \textit{\profilepolarization}: the observed rates that all models succeed/fail consistently exceeds the corresponding \baseline rates.   \autoref{fig:digit_polarization} shows that models in the DIGIT dataset are more likely to all fail or all succeed on an instance than baseline.   \autoref{fig:polarization_all} shows that across all datasets, systemic failure (red dots) and consistent success (blue dots) of all three models on an instance are both more common than baseline, whereas intermediate results are less common than baseline.
    }
\end{figure}

\paragraph{Results.} 
In \autoref{fig:digit_polarization}, we compare the observed and \baseline distributions for the spoken command recognition dataset \digit.
We find the observed \systemlevel outcomes are more clearly \textit{\polarized} compared to the baseline distribution: the fraction of instances that receive either extreme outcome (all right or all wrong) exceeds the \baseline rate.
These findings generalize to all the datasets (\autoref{fig:polarization_all}): the observed rate always exceeds the \baseline rate for the \polaroutcomes (above the line $y = x$) and the reverse mostly holds for intermediary outcomes.

\begin{figure}
    \centering
    \includegraphics[width=\textwidth]{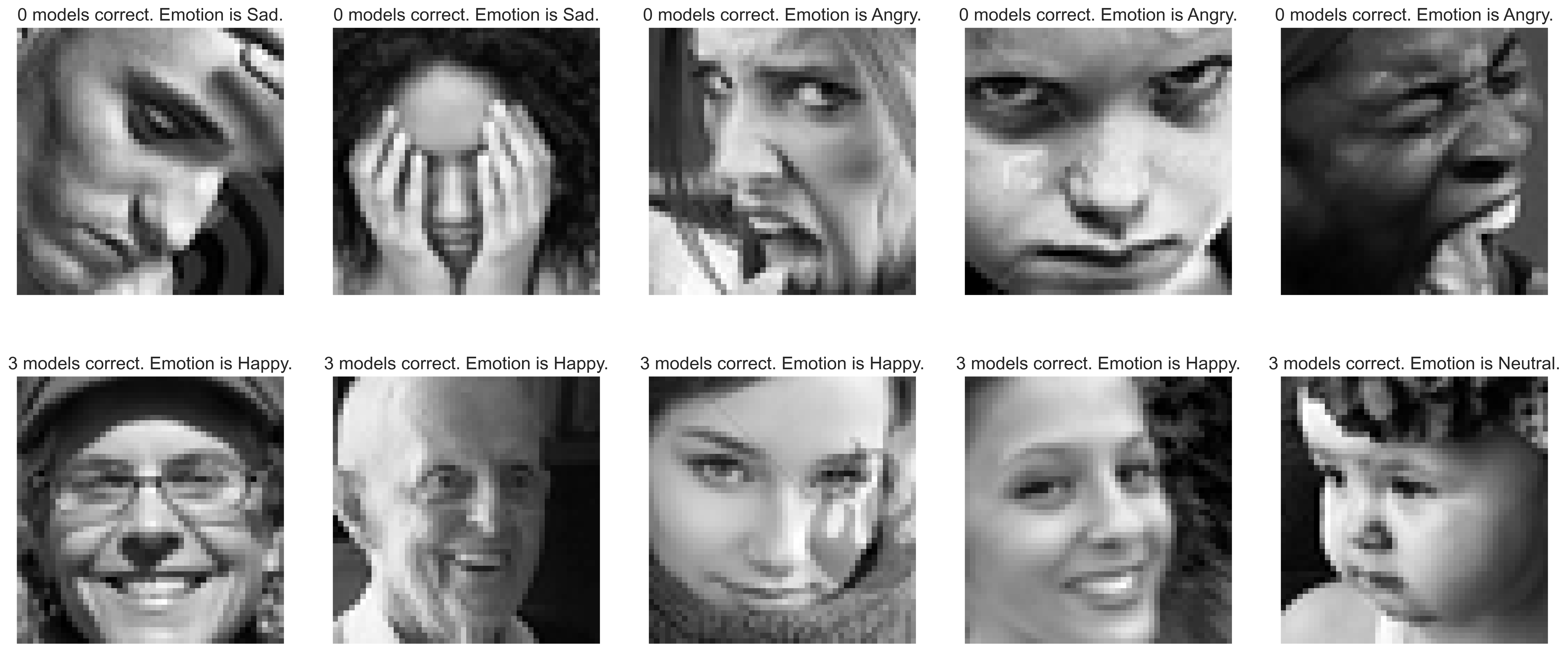}
    \caption{\textbf{Examples of homogeneous outcomes.} 
    Instances that are sampled uniformly at random from ``0 models correct'' (\textit{top row}) or ''3 models correct'' (\textit{bottom row}) in \ferplus.
    The systemic failures (\textit{top row}) do not appear to be inherently harder for humans to classify; more extensive analysis appears in the supplement.
    }
    \label{fig:FER examples}
\end{figure}

To build further intuition, we present several randomly sampled instances from the \ferplus facial emotion recognition dataset in \autoref{fig:FER examples}.\footnote{We acknowledge the task of facial emotion recognition has been the subject of extensive critique \citep[\eg][]{Barrett2019, LeMau2021}. 
We provide examples for this task due to ease of visualization, but our claims also hold for examples from the text and speech modalities that are provided in the supplement.}
We emphasize that while systemic failures may share structure, we do not believe these instances are \textit{inherently} harder than ones on which models perform well.

\section{Supply Chain}
\label{sec:supply-chain}
Foundation models belong to the broader class of general-purpose technologies that have restructured society such as electricity, the Internet, and smartphones \citep{bresnahan1995gpt, brynjolfsson2021jcurve, bommasani2021opportunities, eloundou2023gpts}. 
Building foundation models requires significant resources: immense volumes of data are processed using immense amounts of computation to yield the foundation model.
Using foundation models often requires substantially fewer resources in comparison: models can be adapted, often in lightweight fashion (\eg through a simple textual interface), for an increasingly wide range of use cases.
The disparity in resource requirements between development and deployment has yielded a market where a small set of companies build the most prominent foundation models that are then adopted by thousands of companies and millions of consumers \citep{bommasani2023ecosystem, vipra2023concentration, widder2023open}. 

\begin{figure}
\center
\includegraphics[width=\textwidth]{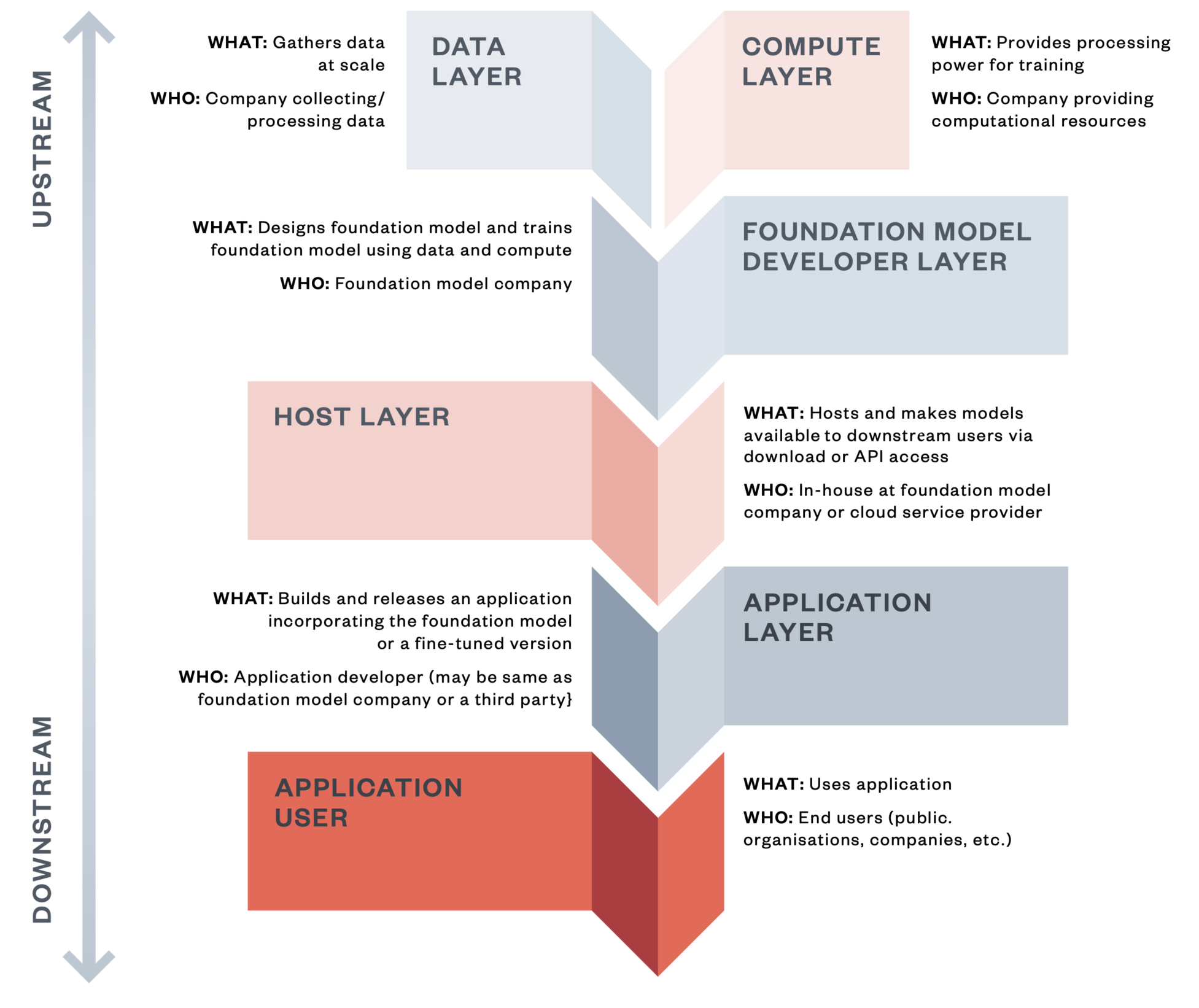}
\caption{\textbf{The canonicalized foundation model supply chain.} 
A conceptual depiction of the foundation model supply chain, beginning with the primary \textit{upstream} resources (\ie data, compute) and transitioning to the foundation model, subsequent hosts (or \textit{distribution channels}), and ending with \textit{downstream} applications.
Image taken with permission from \citet{jones2023foundationmodels}.}
\label{fig:supply-chain}
\end{figure}

The structure of the foundation model paradigm induces a supply chain that we describe in \citet{bommasani2023ecosystem}.
We depict a stylized view of this supply chain in \autoref{fig:supply-chain}.
The supply chain begins with the upstream resources that are used to build a foundation model: data, computational hardware, energy, labor, and code. 
For each of these resources, a further supply chain exists: for example, data to build foundation models is often sourced from the Internet, but this data can only come to be on the Internet as a result of human data-generating process (\eg publishing news article, authoring personal blogs, uploading videos to YouTube, creating music) along with Internet infrastructure (\eg networking protocols).
Foundation models are then used as the foundation for supply chains that derive from the model.
In particular, foundation models are made available for downstream use through \textit{distribution channels} (\eg an API to access the model or a host that facilitates inference using the model).
By way of these distribution channels, foundation models power downstream applications (\eg commercial products and services) across a range of market sectors and geographies.
For instance, \openai's \gptfour powers applications in education (\eg Khan Academy's Khanmigo tutor), finance (\eg Stripe's fraud detection tool), banking (\eg Morgan Stanley's internal chatbot), and government (\eg Iceland's language preservation system).\footnote{See \url{https://openai.com/gpt-4} for a list of several applications built upon \openai's \gptfour.}
Overall, a comprehensive account of the societal impact of foundation models requires consideration of the different parts of the foundation model supply chain \citep[][\S1.2]{bommasani2021opportunities}.

\subsection{Release}
\begin{figure}
\center
\includegraphics[width=\textwidth]{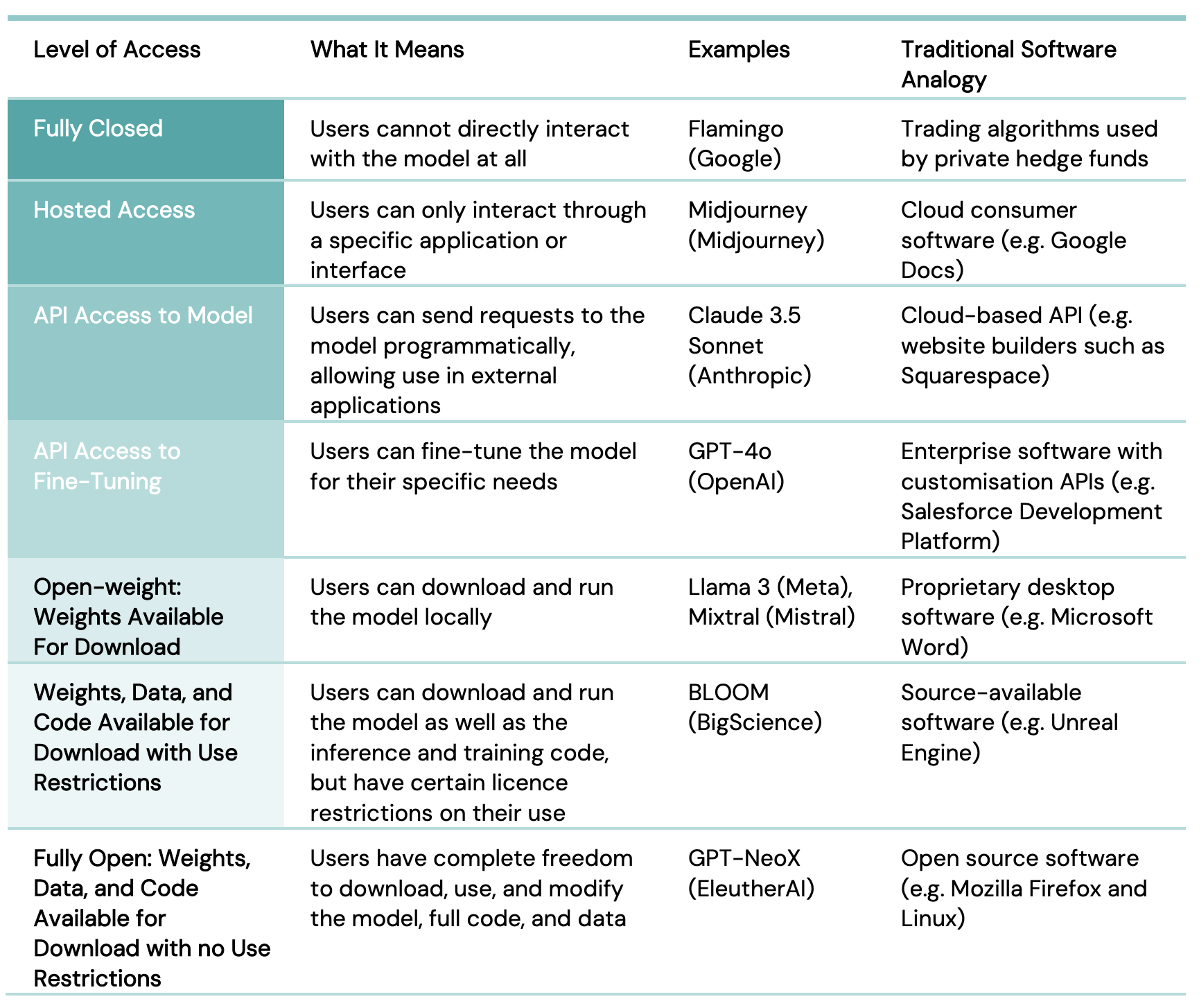}
\caption{\textbf{Spectrum of release options.}
In releasing a foundation model, developers choose what to release, ranging from nothing (\ie the model is only accessible to developer-internal employees) to everything (\ie the weights, data, and code are all available without restrictions).
This choice is coupled with who is granted access to the artifacts (\eg selected individuals, the public) and at what time (\eg release involves a series of stages that incrementally broaden who has access) to determine the overall release process.
Figure from \citet{bengio2025international} based on \citet{bommasani2024open} and \citet{solaiman2023evaluating}.
}
\label{fig:release}
\end{figure}
A key decision in the foundation model supply chain is how foundation model developers choose to release their models, thereby shaping their distribution and, ultimately, their impact.
The release landscape for foundation models is complex \citep{solaiman2019release, sastry2021release, liang2022community-norms, solaiman2023gradient,basdevant2023towards, nelson2024comment}.
In particular, several \textit{assets} exist (\eg the model, data, code): for each asset, there is the matter of \textit{who} can access the asset (\eg user restrictions like requiring that the user be 18 or older) and for \textit{what} purposes (\eg use restrictions that prohibit usage for competing against the model developer).\footnote{Models are often accompanied by licenses that specify these terms. The Open Source Initiative designates some licenses, generally applied to code, as open source and is in the process of leading a similar effort to define open source AI. See: \url{https://opensource.org/deepdive/}}
Further, the degree of access may \textit{change over time} (\eg staged release to broaden access, deprecation to reduce access).
Key options in the release space are depicted in \autoref{fig:release}.

The most salient division in the release space is whether models are ``open'' or ``closed''.
To reason about the societal impact of foundation models, I describe this divide based on our work in \citet{kapoor2024societal}.
We define \textit{open foundation models} as foundation models with widely available model weights.
(For simplicity, we refer to any non-open foundation model as \textit{closed}.)
In particular, with respect to the dimensions of release we describe, this means an open foundation model 
(i) must provide weights-level access, (ii) need not be accompanied by the open release of any other assets (\eg code, data, or compute), (iii) must be widely available, though some restrictions on users (\eg based on age) may apply, (iv) need not be released in stages, and (v) may have use restrictions. 
This definition is consistent with the Executive Order 14110's definition of ``foundation models with widely available model weights" \citep{EO14110}.

This dichotomy is a central topic of discussion because many of the risks described for open foundation models arise because developers relinquish exclusive control over downstream model use once model weights are released. 
For example, if developers impose restrictions on downstream usage, such restrictions are both challenging to enforce and easy for malicious actors to ignore.
On the other hand, in the face of malicious use, developers of closed foundation models can, in theory, reduce, restrict, or block access to their models. 
In short, open release of model weights is irreversible. 
With this said, as we note in \citet{nelson2024ntia} and \citet{solaiman2025release}, this distinction receives more attention than it deserves.

The open vs. closed distinction not only divides the space of release, but also the AI community in many cases.
Some foreground the benefits of openness: widely available model weights promote competition and innovation, improve scientific research, reproducibility, and transparency \citep{toma_generative_2023, creative_commons_supporting_2023,cihon_how_2023, mozilla_joint_2023}.
Others foregrounds the risks of openness: widely available model weights enable malicious actors \citep{seger2023open,brundage_malicious_2018} to more effectively misuse these models to generate disinformation~\citep{solaiman2019release}, non-consensual intimate imagery~\citep{satter_fbi_2023, maiberg_inside_2023}, scams~\citep{hazell_large_2023}, and bioweapons~\citep{gopal_will_2023,soice2023large,sandbrink2023artificial,matthews_scientists_2023,service_could_2023,bray_ai_2023}. 

\paragraph{History.}
To reason about release decisions for foundation models, we consider the history of closely related technologies that were open to varying extents.
In the 2010s, the advent of sophisticated generative image models introduced the concern of deepfakes that could misinform viewers \citep{paris_deepfakes_2019, chesney_deep_2018, widder_limits_2022, cole_ai-assisted_2017}.
These models facilitated significant impersonation, political misinformation, and non-consensual intimate imagery (NCII). 
For example, a Telegram bot was used to generate over a hundred thousand nude images of women~\citep{solsman_deepfake_2020}. 
These models were not foundation models; they relied on more rudimentary algorithms, such as swapping faces in images.

In the late 2010s, foundation models gave rise to powerful generative capabilities for language \citep{radford2018improving}, and new misuse concerns came to the fore.
In February 2019, OpenAI announced the GPT-2 series of models \citep{radford_language_2019,openai_better_2019}: four language models ranging between 124 million and 1.5 billion parameters. 
Primed by several concerns --- especially the potential for large-scale disinformation \citep{solaiman_release_2019} --- OpenAI opted for a staged release where models of increasing size were openly released from February to November 2019. 
While the company did not ultimately find evidence of model misuse during the staged release \citep{openai_gpt-2_2019}, this process brought attention to the nexus between release and misuse \citep{bommasani2021opportunities, sastry2021release, shevlane2022structured, liang2022norms, solaiman2023gradient}.

Since the release of GPT-2, hundreds of foundation models have been released by various actors adopting different release strategies \citep{bommasani2023ecosystem}.
The open release of Stable Diffusion in August 2022 was particularly salient, as it was one of the first text-to-image models to be available widely outside the research community.
However, because the model weights were shared publicly, users easily circumvented filters that Stability AI implemented to prevent the generation of not safe for work (NSFW) imagery. 
As a result, AI-generated pornography based on Stable Diffusion offshoots quickly spread across the internet, including images resembling real people generated without their consent \citep{maiberg_inside_2023}.

Meta's release of its LLaMA language models \citep{touvron2023llama} in March 2023 marked another significant event in the trajectory of open foundation models.
LLaMA was released via a form that allowed researchers to download model weights by accepting a license for non-commercial use. 
But the model weights were quickly leaked, leading to concerns that the highly capable model could facilitate misuse. 
U.S. Senators Hawley and Blumenthal sent a letter to Meta CEO Mark Zuckerberg expressing concerns about Meta’s release strategy~\citep{blumenthal2023meta}.

\paragraph{Construct.}
In spite of extensive discourse by many parties on openness and release, the level of conceptual understanding remains poor.
In large part, the term ``open'' conflates a variety of meanings, as has been well-established in the history of software by the divide between free and open-source software.
To meaningfully reason about the societal impact of foundation models, and the contributions of release decisions therein, a clear understanding of openness is required.

I define five distinctive properties that generally distinguish open foundation models from closed foundation models.
These properties are particularly relevant for understanding how release decisions shape societal impact; they compound upon the other factors discussed in this section (\ie model capabilities, model risks, supply chains) in determining the overall societal impact of an individual model or foundation models as a technological class as a whole.

\paragraph{Broader access.}
Given our definition, open foundation models require that the model weights be widely available, if not to the public as a whole.
While there may be some restrictions on who can use the model, given that such user restrictions are difficult to enforce or verify (as demonstrated by Meta's LLaMA 1 release in March 2023), model weights may effectively be available to the public. 
Functional barriers to use, ranging from requisite expertise to compute affordability, may nonetheless remain.

\paragraph{Greater customizability.}
By releasing model weights, open foundation models are readily customized for various downstream applications.
Weights (and associated computational artifacts made available, such as activations and gradients) permit a wide range of adaptation methods for modifying the model, such as quantization~\citep{frantar_optq_2023}, fine tuning~\citep{zhang_llama-adapter_2023,dettmers2023qlora}, and pruning~\citep{xia_sheared_2023}.
While some closed foundation model developers permit certain adaptation methods (\eg OpenAI allows fine tuning of GPT 3.5 as of January 2024), these methods tend to be more restrictive, costly, and ultimately constrained by the model developer's implementation. 
The customizability of open foundation models prevents model alignment interventions from being effective---such as by allowing users to fine-tune away alignment interventions~\citep{narayanan_model_2023}, though similar issues also arise when closed models can be fine-tuned~\citep{qi2023finetuning}.

\paragraph{Local adaptation and inference ability.}
Users of an open foundation model can directly deploy it on local hardware,  which removes the need for transferring data to the model developer. 
This allows for the direct use of the models without the need to share sensitive data with third parties, which is particularly important in sectors (\eg finance, healthcare, law) where confidentiality and data protection are necessary---such as because of the sensitive nature of content or regulation around how data should be stored or transferred.

\paragraph{Inability to rescind model access.}
Once the weights for a foundation model are made widely available, little recourse exists for the foundation model developer to rescind access.
While the foundation model developer, in coordination with distribution channels used to share model weights, can stop further access, existing copies of the model weights cannot be revoked. 
Furthermore, despite the developer's objections, users can redistribute model weights through, for example, peer-to-peer distribution~\citep{vincent_metas_2023}. 

\paragraph{Inability to monitor or moderate model usage.}
For open foundation models, inference may be performed (i) locally (\eg on a personal computer or self-owned cluster), (ii) on generic third-party computing platforms such as cloud services (\eg Google Cloud Platform, Microsoft Azure), or (iii) on dedicated model hosting platforms (\eg Together, Amazon Bedrock).
In all cases, foundation model developers do not observe inference by default, making monitoring or moderation challenging, especially for local inference. 
Since dedicated model hosts are aware of what models are being used, developers may be able to coordinate with hosts to implement certain forms of monitoring/moderation.

\subsection{Supply chain monitoring}
In \citet{bommasani2023ecosystem}, we build Ecosystem Graphs as a tool for documenting the foundation model supply chain.
Formally, we constructed a vertex-labeled graph, where each vertex corresponded to either (i) a dataset, (ii) a foundation model, or (iii) an AI application with edges indicating known dependencies based on public information.
I defer discussion of this work in this dissertation: AI supply chain monitoring \citep{cen2023supplychain} has not materially improved in the past four years, despite clear recognition of its value as early as \citet{bommasani2021opportunities}.

Our work foregrounds a fundamental choice for AI supply chain monitoring: monitors can focus on technical relationships between assets (\ie which dataset was a model trained on) or between organizations (\ie which data provider supplied data to a model developer).
In my work, I have often adopted the technical-level abstraction, but in disciplines like economics and sociology, the organization-centric approach is more customary \citep{rowlinson1997organisations}.
This (fairly techno-centric) choice provides valuable leverage given the status quo: the number of assets is currently still manageable (in the thousands), the assets themselves are distinctive (\eg they are not exchangeable in the way oil or steel may be in other market analyses), and specific assets markedly contextualize our understanding of organizations (\eg Stable Diffusion dramatically shapes our perception of Stability AI). 
In spite of these advantages, many organization-centric works have proven to be critical to mapping out the market structure.
For example, \citet{weingast1988industrial} demonstrate that institution-centrism better allows for comparisons/juxtapositions across sectors.
Alternatively, \citet{einav2010empirical} showcase how grounding to institutions facilitates various forms of measurement (\eg due to firm-level requirements on information disclosure).
Finally, many works in political and institutional sociology prime us to view institutions as the natural unit for studying power relations in modern networks and markets \cite[][\textit{inter alia}]{frickel2006new, dequech2006institutions, fleury2014sociology}.

Irrespective of the lens, the sole initiative at scale beyond my work is the initiative of the UK Competition and Markets Authority (CMA).
The CMA is the UK competition regulator: in 2023, they initiated market surveillance on the foundation model supply chain to ``help create an early understanding of the market for foundation models and how their use could evolve".
In conducting this work, the CMA directly used Ecosystem Graphs rather than building their own tooling for monitoring the supply chain as demonstrated in their report \citep{cma2023ai}.
Most notably, given the broad interest in how open foundation models differentially contribute to competition, the CMA used our data to characterize the distribution of release strategies.
The analysis indicated that, at the time of the report, 42\% of tracked models were released with open weights and that the most common license for licensed assets was the Apache 2.0 license.

With this in mind, here we briefly describe work in more mature domains that provides precedent, both in terms of how to implement supply chain monitoring and why it is beneficial.

\paragraph{Software supply chains.}
Software development is sustained by an immense network of dependencies.
The demand to track the software ecosystem is immense: the software bill of materials (SBOM) is a national-level initiative of the US's Cybersecurity and Infrastructure Security Agency to maintain an inventory of the ingredients that make up software \citep{whitehouse2021cybersecurity}.\footnote{See \url{https://www.cisa.gov/sbom}.}
These approaches have clarified how to ensure compliance from different stakeholders (\eg software vendors) and how to standardize information to support a range of use cases, providing inspiration for the abstractions I use. 
To implement this vision, a range of efforts have been put forth over the years\footnote{See \url{https://libraries.io/}, \url{http://deps.dev/}, and \url{https://ecosyste.ms/}.} with applied policy work mapping out the sociotechnical challenges for maintaining and funding these efforts \citep{ramaswami2021securing, scott2023atlanticcouncil}.
Further, they present an exemplar of policy uptake towards mandatory public reporting of these dependencies as exemplified by the proposed Securing Open Source Software Act of 2022.\footnote{\url{https://www.congress.gov/bill/117th-congress/senate-bill/4913}}
And, much akin to the uses we consider, these efforts already have shown how descriptive understanding of the ecosystem directly informs decision-making and characterizes the impact of assets.\footnote{See \url{https://docs.libraries.io/overview.html\#sourcerank} and \url{https://github.com/ossf/criticality\_score}.}

\paragraph{Automobile supply chains.}
The automobile industry, and specifically the work of the National Highway Traffic Safety Administration (NHTSA) within the US Department of Transportation, provides an alternate model for supply chain monitoring.
Specifically, when a batch of parts is found to be sub-standard, established protocols mandate the recall of the fleet of cars built using those parts.
Since the National Traffic and Motor Vehicle Safety Act was enacted in 1966, the NHTSA has recalled over 390 million cars due to safety defects.
When cars from different manufacturers are reported to be faulty, the shared source of the defect can be traced by attributing their common parts.
Centralized infrastructure exists for consumers to report issues to the NHTSA (\eg the Department of Transporation's Vehicle Safety Hotline), to interpret how their report will be used, to understand the investigation process the NHTSA implements, and to understand the legal remedies and consumer protections they are afforded.
And the Federal Motor Vehicle Safety Standards sets formal and transparent standards on what constitutes the minimum performance requirements for each (safety-relevant) automotive part.

This targeted approach addresses a fundamental issue in AI: if 
an upstream asset (akin to the faulty brakes) is identified to be faulty in the AI supply chain, will this issue be communicated appropriately and will interventions be taken to mitigate or reduce harm in a timely manner?
For example, if LAION-5B was shown to be data poisoned \citep{carlini2023poisoning} or to include child sexual abuse material \citep{thiel2023csam}, how would the developers of Stable Diffusion, the application developers who built upon Stable Diffusion, and the end userbase be notified?
In general, both the reporting systems for disclosing these adverse events or incidents \citep{guha2023ai, naiac2023aers}, the communication mechanisms \citep{longpre2025inhouse} given that issues likely transfer \citep{wallace2019universal}, and the concrete responses like recalls do not yet exist for the AI supply chain.

\section{Policy impact}
When I worked on these conceptual primitives, I do not know that I would later work extensively on public policy.
In spite of this, these works had unanticipated policy impact.
In reflecting on the policy impact of my conceptual work, I find much of it to be beneficial, but some detrimental, which I discuss below.

\subsection{Emergent abilities}
Emergence became a mainstream topic in policy conversations around the time of my work on the subject: anecdotally, I have been asked by several policymakers how emergence capabilities influence policy design.
When working on \citet{wei2022emergent}, I expected our work to bring more attention to emergence, \footnote{Our work was not the first conceptualize emergence in general (\eg we cite the work of Nobel Laureate Philip Andersen as a key influence) nor for AI in particular (\eg the writings of Jacob Steinhardt are especially seminal).} but I did not expect to reach the policy sphere.
While this dissertation discusses evidence-based AI policy, and I do believe emergence is a topic that has some important implications for policy, I do not believe the work we did should have been very influential for policy.
In particular, the specific evidence we showed addressed capabilities that I do not believe warrant much policy attention.
Therefore, it is immaterial in my view whether the capabilities demonstrate emergence or scale more smoothly, because the patterns in scaling only are of consequence if the capabilities themselves are of societal consequence.
While we did more broadly discuss the possibility for emergent risks and broader sociological changes to the AI field due to the scaling regime, which I do believe warrant consideration, I feel our work received more policy attention than it was due.

With that said, what I find more serious is that policymakers may now problematically  entangle emergence as a lay construct with the specific technical definitions we introduced.
Most notably, \citet{schaeffer2023emergent} provided a well-known critique of our work that questioned whether emergent capabilities are a mirage.\footnote{As it is not relevant to the point I make here, I will reserve my views as a matter of science on this work in relation to \citet{wei2022emergent}.}
What I would like to emphasize is that this led the US House Committee on Science, Space, and Technology (SST) to conclude ``Emergent capabilities \dots debunked by rigorous statistical analysis.''\footnote{See \url{https://republicans-science.house.gov/_cache/files/8/a/8a9f893d-858a-419f-9904-52163f22be71/191E586AF744B32E6831A248CD7F4D41.2023-12-14-aisi-scientific-merit-final-signed.pdf}.}
I find this distressing: this experience, among others, led me to conclude that even scientifically-adept policymakers can woefully misunderstand poorly communicated science and, as a result, it is necessary to work in much closer lockstep with policymakers for evidence-based AI policy to produce genuinely beneficial outcomes.

\subsection{Homogeneous outcomes.}
Homogeneous outcomes and/or algorithmic monoculture have been identified or described in other domains: \citet{ajunwa2019paradox} describes the harm of algorithmic blackballing in hiring and \citet{Engler2021} describes the phenomenon of vendor herding in education.
While these risks, which can present even absent the widespread adoption of AI in general and foundation models in particular, had been identified, policy engagement was more limited to my knowledge.
Concurrent to my work on this topic, the Chair of the US Securities and Exchange Commission, Gary Gensler, began to describe the risks of algorithmic monoculture in finance, including via foundation models in particular.\footnote{I did not personally engage with the SEC or speak to Gary at any point during my PhD, so I assume this is just an independent observation of the same risk, which does not reflect policy impact of my work.}

Towards the end of my PhD, my work on homogeneous outcomes became more salient to the US National Institute of Standards and Technology (NIST), including via briefings I conducted in the House SST Committee alongside Reva Schwartz (who was at NIST at the time).
At the time, NIST had already developed their overarching AI Risk Management Framework \citep{nist2023airmf}, but had not yet made it specific to foundation models.
In the resulting profile that specifically addressed generative AI and foundation models, the NIST guidance introduced a new top-level category entitled ``harmful bias and homogenization''.\footnote{See \url{https://nvlpubs.nist.gov/nistpubs/ai/NIST.AI.600-1.pdf}.}
Consequently, my work on homogeneous outcomes, which is repeatedly cited in this guidance, informs NIST's guidance on how ``organizations identify unique risks posed by generative AI'' and how it  ``proposes actions for generative AI risk management.''

\subsection{Supply chains.}
During the course of my PhD, the focus on the broader AI supply chain grew for many reasons (\eg concerns about labor practices and crawling norms due to academic research and investigative journalism; focus on computational hardware and data centers due to export controls).
In particular, the broad applicability of work on supply chains led me to focus on this area as a general primitive.
The topic of release strategies, often abbreviated as the open vs. closed debate, would separately emerge as a lightning rod for discourse within the AI and policy communities.

My work on openness had significant policy impact, most notably in the NTIA process required under President Biden's Executive Order \citep{EO14110}, which not only hinges on the conceptual work we did, but also how we reviewed evidence of marginal risk and forged consensus around this approach.
I defer further discussion of this to the chapter on policy (Chapter \ref{chapter:policy}).
Separately, my work on supply chain monitoring had significant policy impact by influencing market surveillance initiatives.
Most notably, shortly after we released Ecosystem Graphs \citep{bommasani2023ecosystem} as a tool and dataset documenting the supply chain, the UK Competition and Markets Authority (CMA), which is the UK's competition regulator, announced its intent to monitor the foundation model market \citep{cma2023ai}.
Through direct engagement with the CMA, they decided to not build a new data collection effort and, instead, heavily relied on our data to build their comprehensive analysis of the foundation model market \citep{cma2023ai} and its evolution \citep{cma2024ai}.

\section{Conclusion}
The purpose of this chapter of the dissertation is to provide the conceptual tools used throughout the remainder of the dissertation.
In conducting the research related to the different works discussed in this chapter, my personal goal was to better understand what was going on at a time when robust frameworks for thinking about foundation models did not yet exist (\ie many of these works are the earliest works of my PhD).
The resulting experience helped me prioritize specific empirical work, which is the subject of the next chapters.
\chapter{Empirical Understanding: Part I}\label{chapter:empirics-evaluation}
\chaptermark{\small Evaluation Platforms}
The conceptual underpinnings for foundation models are sufficient to reason about their societal impact, but lack acuity.
How do we sharpen our understanding?
The key principle is proactivity.
I show how to acquire concrete information on the major foundation models and AI companies that directly mediate the technology's societal impact.
This chapter and the next chapter (Chapter \ref{chapter:empirics-index}) put forward proactive methods to improve the public information conditions, animated by the headwind of growing opacity just as AI becomes a high-impact industrialized technology.
To design these methods, I strategically explore both of the natural branches for increasing public transparency: (i) the production of new public information and (ii) the disclosure of existing non-public information.
For the first branch, this chapter presents the novel application of the standard method of evaluations: we built the first third-party evaluation platform to measure foundation model capabilities and risks \citep{liang2023holistic}.
\renewcommand\imdb{\dataset{IMDB}}
\newpage
\noindent \textit{Evaluation} is the widespread practice of measuring the properties of an AI model (or system).\footnote{While some efforts expand evaluation to datasets \citep{bommasani2020intrinsic, swayamdipta2020dataset, ethayarajh2022dataset, mitchell2022measuring} or adopt methodologies from human-computer interaction to consider human factors like user experience, evaluation usually aims to characterize a specific model in isolation.}
Evaluations orient AI.
They encode values and priorities \citep{ethayarajh2020utility, birhane2022values} that specify directions for the 
AI community to improve upon \citep{jones1995evaluating, jones2005acl, kiela2021dynabench, bowman2021fix,  raji2021benchmark}.
When implemented and interpreted appropriately, they enable the broader community to better understand AI technology and influence its trajectory.
Evaluations can vary in the specific type of transparency \citep[see][]{bommasani2023transparency} they provide: some evaluations quantify the accuracy of models \citep[\eg ImageNet;][]{deng2009imagenet}, others stress-test models \citep[\eg CheckList;][]{ribeiro2020beyond} or adversarially identify failures \cite[\eg red-teaming;][]{perez2022red} and still others characterize models along a range of dimensions \citep[\eg HELM;][]{liang2022helm}.

Evaluations that quantitatively characterize model capabilities and risks are universally recognized as a key primitive for better understanding the societal impact of AI.
The practice of evaluation is a growing emphasis for many stakeholders in the AI community: model developers, external researchers, international policymakers, and even broader civil society and the public in certain instances.
This growth, which coincides with my PhD, is hard to summarize in full.
Yet, what we see today is CEOs of the world's most powerful companies (\eg Google, Meta, OpenAI, Anthropic) touting the performance of their foundation models by citing evaluation results on evaluations like MMLU, GPQA, and Chatbot Arena.\footnote{For example, see  \url{https://blog.google/inside-google/message-ceo/alphabet-earnings-q1-2025/} and \url{https://x.com/sundarpichai/status/1913012939931464078} from Sundar Pichai.}
When I began my PhD, it would have been entirely unimaginable for Sundar Pichai to publicly herald performance on, say, SuperGLUE as a noteworthy accomplishment for Google.
In parallel, world governments also now recognize the importance of evaluations, especially in relation to risks.
Entities like the US AI Safety Institute, UK AI Security Institute, and EU AI Office all see evaluation as a core capacity required to fulfill their mandate.

Here, I describe our work on HELM \citep{liang2022helm}, which pioneered not only the first proactive AI evaluation platform (with efforts like Hugging Face's OpenLLM Leaderboard and Chatbot Arena following suit), but also the first comprehensive evaluation of major foundation models.\footnote{Concurrent efforts by EleutherAI with the LM Evaluation Harness \citep{gao2021harness} and Google with BigBench \citep{srivastava2022bigbench} had different priorities but share the role of scaling evaluation to meet the moment of modern foundation models.}
We initiated the HELM project because we lacked transparency in foundation model capabilities and risks at the time.
HELM not only rectified this deficit, but in doing put forth a vision for holistic evaluation.\footnote{We should aspire for not only holistic, but also pluralistic evaluations \citep{birhane2022values}. However, this vision is distinctively misaligned with the history of benchmarking in AI, where major benchmarks are overwhelmingly developed by high-status institutions \citep{koch2021reduced}.}
I view evaluations as vehicles for social change \citep{bommasani2022evaluation}: the broader agenda for the ongoing HELM initiative (primarily maintained by Yifan Mai) is to transform foundation models from immature emerging technologies to reliable tools that support human flourishing.
Notably, the proactive and sustained evaluation of foundation models by a third-party organization \citep{longpre2025inhouse} not only generates valuable new public information, but also is an appropriate basis for trust given our independence from model developers.

\section{History}
Evaluation has a long history in AI.
As Karen Sp{\"a}rck Jones put it in her ACL Lifetime Achievement Award speech, ``proper evaluation is a complex and challenging business'' \citep{jones2005acl}.
To address this challenge, the practice of benchmarking rose to prominence as the core methodology in the 1980's and, especially, the 1990's \citep[see][]{liberman2010obituary, jones1995evaluating}.
This transition was well-demonstrated by initiatives such as the Message Understanding Conference \citep[MUC;][]{grishman1996muc} and the Text Retrieval Conference \cite[TREC;][]{voorhees1998trec}.
And it coincided with a broader shift towards statistical and data-driven methods with large datasets (\eg the Penn Treebank \citep{marcus1999ptb}) and new venues like the Conference on Empirical Methods for Natural Language Processing \citep{emnlp1996emnlp}. 

More than a decade later, the rise of deep learning in the 2010s \citep{collobert2008unified, turian2010word, collobert11scratch, socher2011paraphrase, socher2011parsing, sutskever2011generating, mikolov2013efficient, pennington2014glove, sutskever2014sequence, bahdanau2015neural, luong2015translation, vaswani2017attention} hinged on the widely-herald success of AlexNet \citep{krizhevsky2012alexnet} on the ImageNet benchmark \citep{deng2009imagenet}.
ImageNet's success reverberates to this day, not only showing how evaluations can identify new technical paradigms, but how they can orient the entire's focal point.
As Fei-Fei Li often notes, ImageNet was the north star for computer vision: many of the evaluations that follow ImageNet, including my own work with HELM, see it as the pinnacle.
Later in the 2010s, larger evaluations also emerged in natural language processing: benchmarks like SNLI \citep{bowman2015large} and SQuAD \citep{rajpurkar2016squad} were developed to provide both adequate data for training systems in addition to evaluating systems.
In part, these benchmarks were built by capitalizing on the parallel emergence of crowdsourced labor on platforms like Amazon Mechanical Turk.
Overall, these evaluations of 2010, while featuring larger datasets, generally extended the regime of assigning models a single score (\eg the SQuAD F1 score) to measure the accuracy for a single task.

As more general-purpose approaches models became popular, often displacing more bespoke task-specific approaches, new benchmarks such as SentEval \citep{conneau2018senteval}, DecaNLP \citep{mccann2018natural}, GLUE \citep{wang2019glue}, and SuperGLUE \citep{wang2019superglue} co-evolved to evaluate their capabilities.
In contrast to the previous class of benchmarks, these benchmarks assign each model a vector of scores to measure the accuracy for a suite of scenarios.
In some cases, these benchmarks also provide an aggregate score (\eg the GLUE score, which is the average of the accuracies for each of the constituent scenarios).

More recently, this theme of meta-benchmarks that assess model accuracy across a range of tasks has continued \citep[see][\S4.4.3]{bommasani2021opportunities}: for example, GEM \citep{gehrmann2021gem} provides a suite for natural language generation tasks, XTREME \citep{hu2020xtreme} provides a suite for tasks spanning numerous languages, and GEMv2 \citep{gehrmann2022gemv2} provides a suite for generation across languages. 
This approach is also the dominant approach to language model evaluation,\footnote{We note \citet{efrat2022lmentry} as a very recent counterexample to this trend that takes a much more minimalistic and succinct unit-testing perspective.}
 often with even broader collections: \citet{brown2020gpt3} popularized the approach in their work on GPT-3, where they evaluated on 42 datasets.
Indeed, this is the approach used in all the works that introduced models we evaluate in this work.
Efforts like the EleutherAI Language Model Evaluation Harness \citep{gao2021harness}, HuggingFace's Evaluate library \citep{werra2022evaluate}, and Big-Bench \citep{srivastava2022bigbench} have centralized and expanded these evaluations into systematic repositories.

\section{Constructs}
Our objective is to define holistic evaluations for foundation models with the focus at the time of HELM in 2021--2022 being text-to-text language models.
Foundation models are general-purpose technologies that can be applied for a range of use cases.
Each use case may have a range of (un)desirable performance criteria (\eg accuracy, robustness, fairness, calibration, efficiency), where the relative importance of these criteria depends not only on ones values, but the use case context (\eg inference efficiency might be of greater importance in mobile applications). \\

\noindent We specify three principles that codify how we conceptualize holistic evaluation.
\begin{enumerate}
\item \textbf{Coverage.}
Given the broad capability and risk surface, models should be evaluated on a broad range of scenarios.
These scenarios should be chosen systematically with a procedure that reveals what is missed.
And, simultaneously, a range of metrics should be considered to address the different factors that cumulatively determine a model's performance.
\item \textbf{Context-sensitive.}
The appropriate interpretation of a given measurement depends jointly on the construct being measured (\eg fairness) and the context of use (\eg content moderation on a social media platform).
Metrics should be measured for specific contexts in an integrated fashion.
\item \textbf{Standardization.}
All models should be evaluated with the same adaptation procedure to enable meaningful comparison.
\end{enumerate}

\noindent Our objective with holistic evaluation is a fuller characterization of model properties to advance both scientific and societal outcomes.
To operationalize holistic evaluation, we specify the constructs we measure, referring both to the use cases (\textit{scenarios}) and the desiderata (\textit{metrics}).
Here, I describe their systematic selection to address the coverage principle, deferring specifics to \citet{liang2022helm}.

\subsection{Scenarios}
To determine the scenarios we evaluate models on, we introduce a taxonomy for reasoning about scenarios.
To specify a scenario, we break it down into a \textit{task} and \textit{domain}.\footnote{In the initial HELM work, we exclusively focused on English text-to-text evaluations, with subsequent leaderboards supported by the HELM evaluation platform covering other languages (\eg Asian languages, African languages) and modalities (\eg text-to-image models, visual language models).}
Examples of scenarios include (question answering, (clinical notes, doctors, now), English) and (toxicity detection, (tweets, Egypt, Internet-era), Arabic).
Tasks, domains, and languages are not atomic or unambiguous constructs: they can be made coarser and finer, but we use them as intuitive \textit{structure} for the space of scenarios.
Given this structure, we select scenarios to:
(i) ensure sufficient coverage,
(ii) minimize cumulative cost,
and (iii) prioritize user-facing tasks.

\begin{table}[htp]
\resizebox{\textwidth}{!}{
\begin{tabular}{ll}
\toprule
\textbf{Track} & \textbf{Tasks} \\
\midrule
Computational Social Science and Cultural Analytics & No canonical tasks/not task-centric\\
Dialogue and Interactive Systems & Chit-chat dialogue, task-oriented dialogue \\
Discourse and Pragmatics & Discourse parsing, sentence ordering, coreference resolution\\
Ethics and NLP & Toxicity and hate speech detection, misinformation and fake news detection\\
Generation & Data-to-text generation, \\
Information Extraction & Named entity recognition, entity linking, entity extraction, relation extraction, event extraction, open information extraction \\
Information Retrieval and Text Mining & Information retrieval and passage retrieval\\
Interpretability and Analysis of Models for NLP & No canonical tasks/not task-centric\\
Language Grounding to Vision, Robotics and Beyond & Image captioning, visual question answering, instruction following, navigation \\
Linguistic Theories, Cognitive Modeling, and Psycholinguistics & No canonical tasks/not task-centric\\
Machine Learning for NLP & Language modeling\\
Machine Translation and Multilinguality & Machine translation \\
NLP Applications & No canonical tasks \\
Phonology, Morphology, and Word Segmentation & Tokenization, lemmatization, \\
Question Answering & Question answering and reading comprehension \\
Resources and Evaluation & No canonical tasks/not task-centric\\
Semantics: Lexical & Word sense disambiguation, word sense induction\\
Semantics: Sentence-level Semantics, Textual Inference, and Other Areas & Semantic parsing, natural language inference, semantic role labeling/slot filling, semantic textual similarity, paraphrase detection \\
Sentiment Analysis, Stylistic Analysis, and Argument Mining & Sentiment analysis, style transfer, argument mining, stance detection, opinion mining, text simplification\\
Speech and Multimodality & Text-to-speech, speech-to-text\\
Summarization & Summarization, sentence compression\\
Syntax: Tagging, Chunking and Parsing & POS tagging, chunking, constituency parsing, dependency parsing, grammar induction, grammatical error correction \\
\bottomrule
\end{tabular}}
\caption{\textbf{Taxonomy of tasks.} To taxonomize the space of tasks, we leverage the NLP community’s taxonomy of subareas as codified by the ACL 2022 list of tracks. 
For each track, we then expand it into canonical tasks associated
with that track.}
\label{tab:tasks}
\end{table}
\paragraph{Task enumeration.}
Given the ubiquity of natural language, the field of natural language processing considers myriad tasks that correspond to language's many functions \citep{jurafsky2000speech}.
Enumerating these tasks \textit{ex ante} is non-obvious so we rely on pre-existing taxonomies.
Since the field has historically operated in a task-centric fashion, we use a taxonomy that organizes research in the field.
We enumerate tasks based on the tracks at a major conference (ACL 2022), which reflect the ``relevant topics'' of study at the time of writing.\footnote{\url{www.2022.aclweb.org/callpapers}}
We map each track to canonical tasks associated with the track in \autoref{tab:tasks}.

While these tasks have long traditions of study, we observe that they:
(i) have important intra-task structure (\eg question-answering is a single ``task'', but the question answering community often applies finer distinctions \citep{rogers2021qa})
and (ii) are not necessarily the most societally consequential tasks.
The deployment of foundation models has revealed a much broader set of use cases that defy the narrow historical conceptualization of AI and language technology in the research community.
This introduces a fundamental challenge to stating the space of tasks: we often cannot conceive the full space until we see relevant technology and, even otherwise, we cannot reason about the long tail of use cases given increasingly-general models.

\paragraph{Domain taxonomization.}
The domain used to describe data is an imprecise, but intuitive, attempt at characterizing the high-dimensional data (\eg social media posts).
We deconstruct domains as follows:
\begin{enumerate}
\item \textbf{Genre (what).} The type or subject of the data.
Examples: Wikipedia, social media, news, scientific papers, fiction.
\item \textbf{Time period (when).} The time the data was created.
Examples: 1980s, pre-Internet, present day.
\item \textbf{People (who).} The persons that either created the data or that the data describes.
Examples: Black/White, men/women, children/elderly.
\end{enumerate}

\paragraph{Task selection.}
To select tasks, we begin with the set we described previously.
We remove tasks that relate to modalities beyond text or languages beyond English.
Of the remaining tasks, we elect to prioritize \textit{user-facing} tasks: we believe these tasks will confer much of the \textit{direct} social impact of language models and align with our perspective of language models as \textit{interfaces}. 
Consequently, we filter tasks based on our judgments of what is user-facing.\footnote{We emphasize that this \textit{does not} mean we believe the other tasks are less important nor that they should not be evaluated for in future work.}
This yields the following tasks: 
\textit{question answering}, 
\textit{information retrieval}, 
\textit{summarization}, 
\textit{sentiment analysis}, 
and \textit{toxicity detection}.\footnote{We note that our interpretation of what is user-facing namely excludes tasks that are generally not the subject of applications (\eg natural language inference) as well as many classical NLP tasks that served as intermediaries \citep{jurafsky2000speech} in traditional NLP pipelines (\eg named entity recognition, part-of-speech tagging, syntactic parsing, information extraction).}
To partially cover the long tail of tasks, we include \textit{miscellaneous text classification}, which represents the non-standard text classification use cases for language technologies historically and at present for language models. 

\paragraph{Domain selection.}
Enumerating the domains proved more complex than enumerating the tasks, hence why stopped at just taxonomizing the domain space.
Therefore, we operationalize our goal for broad domain coverage in selecting among available datasets for each task we consider.
We hope the community can build on our work to improve upon our domain coverage, given our efforts to surface where the community as a whole lacks relevant datasets.\footnote{We acknowledge that while the community's perception of datasets and evaluations has improved, as signaled by the NeurIPS Datasets \& Benchmarks track, incentives for this work still wane relative to more modeling-centric work in AI at large \citep{jo2020archives, paullada2021data, rogers2021changing, jernite2022governance}.}

\subsection{Desiderata}
To taxonomize the space of desiderata, we begin by enumerating criteria that are sought after for useful systems.
More precisely, these specify categories or families of metrics (\eg the category of \textit{accuracy} contains several specific metrics/quantitative functions such as exact match and F1-score). 
From these categories, we further taxonomize based on the requirements needed to appropriately measure the construct (\eg interpretability generally requires more than blackbox access to a model).
Given this fine-grained taxonomy, we select all metrics where we can satisfy the requirements for all of the models we evaluate in this work (\eg no assumption of knowledge about a broader context that situates the model). 
To operationalize the selected desiderata as quantitative metrics, we emphasize that we prioritize \textit{scalability}: we measure these desiderata whenever possible, which means our measurement is agnostic to the specifics of each scenario.

\begin{table}[htp]
\resizebox{\textwidth}{!}{
\begin{tabular}{ll}
\toprule
\textbf{Venue} & \textbf{Desiderata} \\
\midrule
\href{https://www.2022.aclweb.org/callpapers}{ACL, EMNLP, NAACL, LREC \dots} & accuracy, bias, environmental impact, explainability, fairness, interpretability, linguistic plausibility, robustness \\
& sample efficiency, toxicity, training efficiency \\
\href{https://sigir.org/sigir2022/call-for-papers/}{SIGIR} & accuracy, bias, explainability, fairness, inference efficiency, privacy, security, user experience/interaction \\
\href{https://nips.cc/Conferences/2022/CallForPapers}{NeurIPS, ICML, ICLR, \dots} & accuracy, fairness, interpretability, privacy, robustness, sample efficiency, theoretical guarantees, training efficiency \\
& uncertainty/calibration, user experience/interaction \\
\href{https://aaai.org/Conferences/AAAI-22/keywords/}{AAAI} & accountability, accuracy, bias, causality, creativity, emotional intelligence, explainability, fairness, interpretability \\
& memory efficiency, morality, privacy, robustness, sample efficiency, security, theoretical guarantees, transparency \\
& trustworthiness, uncertainty/calibration, user experience/interaction \\
\href{http://learningtheory.org/colt2022/cfp.html}{COLT, UAI, AISTATS} & accuracy, causality, fairness, memory efficiency, privacy, sample efficiency, theoretical guarantees, training efficiency \\
\href{https://www2022.thewebconf.org/cfp/research/}{The Web Conference (WWW), ICWSM} & accessibility, accountability, accuracy, bias, credibility/provenance, fairness, inference efficiency, legality, privacy, reliability \\
& robustness, security, transparency, trustworthiness, user experience/interaction \\
\href{https://facctconference.org/2022/cfp.html}{FAccT} & causality, explainability, fairness, interpretability, legality, oversight, participatory design, privacy, security \\
& transparency, user experience/interaction \\
\href{https://www.wsdm-conference.org/2023/calls/call-papers}{WSDM} & accountability, accuracy, credibility/provenance, explainability, fairness, inference efficiency, interpretability \\
& privacy, robustness, toxicity, transparency, trustworthiness, user experience/interaction \\
\href{https://kdd.org/kdd2022/cfpResearch.html}{KDD} & accuracy, explainability, fairness, inference efficiency, interpretability, maintainability, memory efficiency, privacy \\
& robustness, training efficiency \\
\midrule 
Union & accessibility, accountability, accuracy, bias, causality, creativity, credibility/provenance, emotional intelligence \\
& environmental impact, explainability, fairness, inference efficiency, interpretability, legality \\
& linguistic plausibility, maintainability, memory efficiency, morality, oversight, participatory design, privacy \\
& reliability, robustness, sample efficiency, security, theoretical guarantees, toxicity, training efficiency \\
& transparency, trustworthiness, uncertainty/calibration, user experience/interaction \\
\bottomrule 
\end{tabular}}
\caption{\textbf{Enumeration of desiderata.} To enumerate the space of desiderata, we first compile a list of venues from \url{https://aideadlin.es/}. For each venue, we enumerate desiderata that are well-studied in that community.}
\label{tab:desiderata-enumeration}
\end{table}

\begin{table}[htp]
\resizebox{\textwidth}{!}{
\begin{tabular}{ll}
\toprule
\textbf{Category} & \textbf{Desiderata} \\
\midrule
Requires knowledge of how model was created & causality, environmental impact, linguistic plausibility, memory efficiency, participatory design, privacy \\
& sample efficiency, training efficiency, theoretical guarantees \\
Requires the model have specific structure & credibility/provenance, explainability \\
Requires more than blackbox access & interpretability \\ 
Require knowledge about the broader system & maintainability, reliability, security, transparency \\
Requires knowledge about the broader social context & accessibility, accountability, creativity, emotional intelligence, legality, morality, oversight \\
& trustworthiness, user experience/interaction \\
Satisfies our conditions (\ie none of the above) & accuracy, bias, fairness, inference efficiency, robustness, toxicity, uncertainty/calibration \\
\bottomrule
\end{tabular}}
\caption{\textbf{Taxonomy of desiderata.} To taxonomize the space of desiderata, we categorize each desideratum based on the requirements needed to properly measure it.}
\label{tab:desiderata-taxonomy}
\end{table}
\paragraph{Desiderata taxonomy.}
What does it mean for a system to be useful?
Too often in AI, this has come to mean the system should be accurate in an average sense.
While (average) accuracy is an important, and often necessary, property for a system \citep{raji2022fallacy}, accuracy is often not sufficient for a system to be useful/desirable.
As a community grounded in a plurality of values, we should determine system performance by considering how systems profile along these many axes.

To enumerate a set of desiderata, akin to our set for tasks, we began by considering desiderata studied in the NLP community.
Unfortunately, while many of the desiderata we independently came up with are well-studied by the NLP community, some are not codified in specific tracks/areas (\eg uncertainty and calibration). 
Therefore, we expanded our scope to all AI conferences, drawing from a list of AI conference deadlines.\footnote{\url{https://aideadlin.es/}}
For brevity, we chose to exclude venues associated with other modalities beyond language (namely computer vision and robotics venues among others), though we did survey these venues as well.

For each conference, we looked at the call for papers or any lists of areas of study: we map the listed areas to desiderata studied in the associated community (\autoref{tab:desiderata-enumeration}).
The union of all the desiderata listed is the space of desiderata we consider, and comprehensively outlines the many dimensions required to truly achieve performant systems.
As with scenarios, we recognize there may be desiderata that have not been traditionally studied at any of these venues: this is why we made sure to cast a wide net in sources for desiderata, and we believe that at the level of desiderata, we likely do have strong coverage but that other mechanisms (\eg polling larger and more diverse groups than just academics implicitly) may still be able to improve on our listing. 

Since we treat language models as interfaces, making no assumptions on their construction, structure, or broader system/context as well as no access beyond blackbox access, we taxonomize desiderata based on the knowledge and access required to properly evaluate these desiderata (\autoref{tab:desiderata-taxonomy}).

\paragraph{Desiderata selection.}
To select the desiderata we will quantitatively measure, we simply take all desiderata that satisfy our conditions:
(i) no assumptions on the construction or structure of the model,
(ii) no access beyond blackbox access,
and (iii) no assumptions on the broader system/context.
This yields the following list: 
\textit{accuracy, 
uncertainty/calibration, 
robustness, 
fairness, 
bias, 
toxicity, 
inference efficiency}.
To this list we add \textit{training efficiency} and \textit{environmental impact} since their measurement relies on information that is partially available for some models (\ie reported in associated papers). 

\section{Evaluations}
To specify the exact evaluations we conduct, we specify runs.
Runs determine (i) the model being evaluated, (ii) the dataset/scenario the model is evaluated on, (iii) the adaptation method that specifies how the model performs the desired task and (iv) the metric that quantifies the model performance.
Below, I describe the datasets/scenarios and metrics/desiderata, building upon the afore-mentioned selection procedures for scenarios and desiderata.\footnote{In the original HELM paper, we also evaluated on ``targeted'' scenarios (\eg to measure memorization/the rate of copyrighted content generation), which I do not discuss in this dissertation for brevity.}
For adaptation, we use five-shot in-context learning as the prompting strategy throughout \citet{liang2022helm}.
Finally, I describe the selected models, which completes the specification of the runs and, in turn, determines the overall costs for these large-scale evaluations. 

\subsection{Evaluated scenarios}
To evaluate language models on the six selected tasks, I describe the tasks in greater depth, the selected datasets to operationalize these evaluations and the process for selecting the dataset, largely constrained by the availability of data.

\paragraph{Question answering.}
Question answering (QA) is a fundamental task in NLP that underpins many real-world applications including web search, chatbots, and personal assistants. 
QA is very broad in terms of the questions that can be asked and the skills that are required to arrive at the answer, covering general language understanding, integration of knowledge, and reasoning \citep{garder2019qaforamt, rogers2021qa}.
 
\FigTop{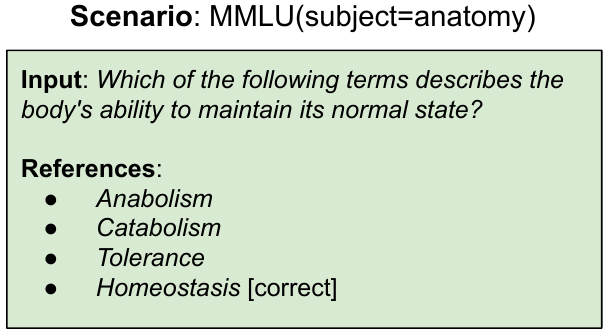}{1.0}{question-answering}{\textbf{Example of question answering.} An example instance for question answering from MMLU. Different QA scenarios can have significantly different properties, but this example captures the overall structure of question answering.}

In QA, given a question (\eg ``Where was the painter of the Mona Lisa born?''), the task is to predict the correct answer (``Italy'').
The format of question answering may have some variations: in the \textit{open-book} or \textit{reading comprehension} setting, additional context to refer to, such as supporting documents (\eg Wikipedia page of ``Mona Lisa''), is given to the model.
In the \textit{multiple-choice} setting, answer choices to choose from (\eg ``(A) France (B) Italy'') are given to the question.
\autoref{fig:question-answering} depicts an example.

There are hundreds of question-answering datasets available in NLP, with a rapid increase in the number of datasets in recent years \citep{rogers2021qa}.
To select question-answering datasets, we prioritized (i) domain coverage, in terms of the domain of the inputs/contexts and (ii) coverage of component skills required for the datasets (\eg we deliberately ensured of datasets that required commonsense knowledge and reasoning). 

We selected the \naturalquestions~\citep{kwiatkowski2019natural},  \narrativeqa~\citep{kovcisky2017narrativeqa}, and \quac~\citep{choi2018quac} datasets to ensure domain coverage as these datasets cover web search queries, stories, and conversational questions (\ie dialogue) respectively.
\naturalquestions~consists of questions from queries to Google search and annotations from Wikipedia; we consider both \textit{open-book} and \textit{closed-book} variants of \naturalquestions.
\narrativeqa~tests reading comprehension through the understanding of books and movie scripts.
\quac~(Question Answering in Context) provides freeform questions and answers which are more open-ended and dependent on context.

To these, we add the \hellaswag~\citep{zellers2019hellaswag}, \openbookqa~\citep{mihaylov2018can}, and \truthfulqa~\citep{stephanielin2021trutfulqa} datasets to ensure coverage of commonsense knowledge and reasoning.
\hellaswag~tests commonsense inference and was created through adversarial filtering to synthesize wrong answers.
\openbookqa~is based on open book exams, with a collection of basic science facts and crowd-sourced multiple-choice questions to test understanding and application of these facts.
\truthfulqa~tests model truthfulness through questions that align with common human misconceptions, spanning law, medicine, finance, and politics, among others, that were adversarially generated using \gptdavinci as the target model.

To further ensure broad coverage of knowledge-intensive question answering across many disciplines, we add the \mmlu~\citep{hendrycks2021measuring} meta-benchmark of 57 constituent datasets. 
\mmlu~(Measuring Massive Multitask Language Understanding) measures multitask accuracy and includes a diverse set of 57 tasks, testing problem solving and general knowledge.

Finally, we add \boolq~\citep{clark2019boolq} which, in addition to \quac, was used to study model robustness to equivariances due to the available contrast set \citep{gardner2020contrast}.
\boolq~is a collection of binary yes/no questions generated through the same process as \naturalquestions.

\paragraph{Information retrieval.}

Information retrieval (IR), which refers to the class of tasks concerned with searching large unstructured collections (often \textit{text} collections), is central to numerous user-facing applications.
IR has a long tradition of study \citep{salton1965smart, salton1971smart,  jones1972statistical, salton1983introduction, manning2008introduction, lin2021pretrained} and is one of the most widely deployed language technologies.
It powers the Web and e-commerce search, and serves as a key component in many knowledge-intensive NLP systems for open-domain question answering or fact checking.

We focus here on the passage ranking task: given a query $q$ and a large corpus $C$ of passages, systems must output a list of the top-$k$ passages from $C$ in decreasing ``relevance'' to $q$. We specifically study this in the context of \textit{re-ranking}: since $C$ is typically extremely large (\eg $|C| > 10M$ passages), we consider only ranking the top-$k$ passages among a set retrieved for $q$ (\ie $M(q)$ where $|M(q)| \ll |C|$) by an efficient external retrieval mechanism \citep[\eg BM25;][]{robertson2009probabilistic}.

IR differs fundamentally from the other tasks we consider in this work, as each test example (\ie a query) entails processing a large set of passages and will likely invoke the LM numerous times to do so.\footnote{Effectively, this means that model outputs come from a very large combinatorial space: they are much more constrained than open-ended generation tasks but much less so than standard classification tasks, setting this scenario apart from many others in terms of its automatic evaluation.} Because of this, IR tasks have received very little attention in the few-shot in-context learning with language models, with the exception of the recent zero-shot approach by \citet{sachan2022improving}.

\FigTop{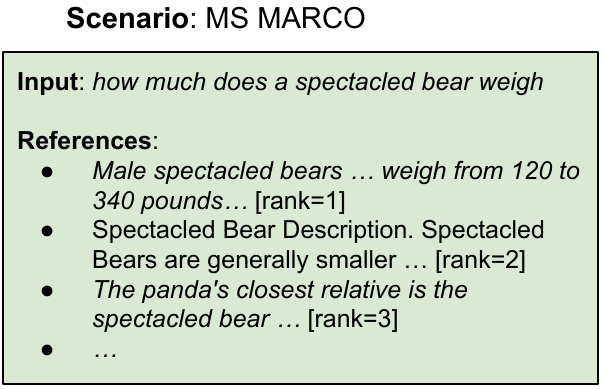}{1.0}{information-retrieval}{\textbf{Example of information retrieval (passage ranking).} An example instance for information retrieval from MS MARCO.}

We address the re-ranking task in a \textit{pointwise} fashion: we formulate the information retrieval problem using prompting as a binary log-probability problem, similar to \citet{nogueira2019passage}: Given a passage $c_i$ and a query $q$, we ask the model whether the passage contains an answer to the query. 
If the model's answer is \texttt{Yes} with a high probability, we rank the corresponding $c_i$ higher, while the \texttt{No} answer with high probability achieves the opposite. 
\autoref{fig:information-retrieval} depicts an example instance.
The rankings produced are then evaluated using standard information retrieval metrics.

We demonstrate the information retrieval task using the MS MARCO ranking datasets. 
While it is originally a question answering task, the retrieval version of MS MARCO is the largest publicly available collection of relevance judgments and has been central to much of the progress in neural IR over the past several years \citep{lin2021pretrained}.

We use the original passage ranking dataset accompanying the public MS MARCO leaderboard\footnote{\url{https://microsoft.github.io/msmarco/}} (\citealt{nguyen2016ms}; henceforth, the \textbf{regular} track) and the dataset from the TREC 2019 Deep Learning track (\citealt{craswell2020overview}; henceforth, the \textbf{TREC} track). Both datasets evaluate retrieval out of a collection of 9M passages from the Web. The regular track contains a large number of queries (e.g., over 500,000 training set queries) with \textit{sparse} relevance judgments: on average, annotators identify only one ``positive'' (relevant) passage for each query, and every other passage is assumed to be a negative. In contrast to this, the TREC track contains only 43 queries that are more rigorously annotated, with over 9,000 query--passage pairs with associated relevance judgments corresponding to the 43 queries.

\paragraph{Summarization.}
Text summarization is an established research direction in NLP \citep{luhn1958automatic, mani1999advances, sparck1999automatic, nenkova2012survey}, with growing practical importance given the ever-increasing volume of text that would benefit from summarization.
To effectively summarize, systems must identify and yield the core relevant and informative content in the source document while removing less critical information and avoiding redundancy \citep{peyrard2019simple}.
The rise of language models in recent years has dramatically improved summarization capabilities: the ability to generate fluent and coherent human-like text serves as a core primitive towards building better summarization systems \citep{lewis-etal-2020-bart, https://doi.org/10.48550/arxiv.1912.08777}.

\FigTop{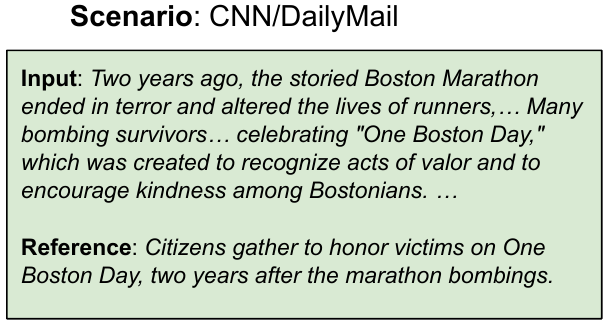}{1.0}{summarization}{\textbf{Example of summarization.} An example instance for summarization from CNN/DailyMail. Different summarization scenarios can have significantly different properties, but this example captures the overall structure of summarization.}

We formulate text summarization as an unstructured sequence-to-sequence problem, where a document (\eg a CNN news article) is the input and the LM is tasked with generating a summary that resembles the reference summary (\eg the bullet point summary provided by CNN with their article).
\autoref{fig:summarization} provides an example.
This evaluation tests the \textit{abstractive} summarization capabilities of the model, where the model is directly required to generate the summary rather than being explicitly constrained to copying words or larger \textit{extracts} from the input document.

To evaluate model performance, the model-generated summary is compared against a human-authored \textit{reference} summary using automated metrics for overall quality \citep[\texttt{ROUGE-2}; BERTScore;][]{lin-2004-rouge, bert-score}, \textit{faithfulness} \citep{10.1162/tacl_a_00453,fabbri-etal-2022-qafacteval}, and \textit{extractiveness} \citep{grusky2018newsroom}. Faithfulness refers to whether all the information in the model summary is supported by the article \citep{cao2018faithful,durmus-etal-2020-feqa,maynez-etal-2020-faithfulness}. 
Extractiveness refers to the extent to which model summaries involve copying from the input document: the distinction between extractive and abstractive approaches has been widely discussed in the summarization literature \citep[see][]{nenkova2012survey}. 
We compute extractiveness since prior work has shown that current summarization systems tend to be less faithful, on average, whenever they extract less \citep{durmus-etal-2020-feqa,mrini2021rewards,ladhak-etal-2022-faithful}.

We pay special attention to faithfulness as neural models in particular often hallucinate content that diverges from what appears in the document being summarized.
Consequently, it is important to measure and improve the faithfulness of these systems since unfaithful systems may be harmful by potentially spreading misinformation, including dangerous, yet hard to detect  errors, when deployed in real-world settings.
We evaluate the LMs using recently proposed reference-free evaluation metrics that have been shown to get high correlations with human scores for faithfulness \citep{10.1162/tacl_a_00453,fabbri-etal-2022-qafacteval}. 
We note recent work has shown that some reference-free evaluation metrics may be mostly relying on spurious correlations \citep{durmus-etal-2022-spurious}.

There is a growing collection of summarization datasets, including datasets that capture finer-grained and more specific summarization functions (\eg summarizing multiple documents or conditional on a user query).
\citet{bommasani2020intrinsic} show that there is significant diversity in summarization datasets along several axes, which makes selecting a few datasets to represent summarization rather challenging.
Since we are especially interested in model faithfulness in this work (as this is a known failure mode of other neural approaches to summarization), we select the \cnndm~\citep{10.5555/2969239.2969428} and \xsum~\citep{narayan-etal-2018-dont} datasets, which are the most well-studied datasets in the literature on summarization faithfulness.
This also ensures domain coverage of news-type data.
Importantly, these datasets differ along a central axis studied in summarization: \xsum~is a dataset with largely abstractive reference summaries (meaning the string overlap between the document and its summary in the dataset is relatively small on average), whereas \cnndm~is a dataset with largely extractive reference summaries. 
However, these datasets do not suffice in representing the full diversity of summarization, and we encourage future work to expand on our benchmark along this axis (\eg add datasets from domains beyond news), particularly towards domains where there is greater demand for summaries \citep[see][]{reiter2022summarization}.
And we especially highlight that these two datasets have been the subject of critique, and that broader change is required for dataset and evaluation design in summarization and natural language generation \citep{gehrmann2022repairing, reiter2022summarization}. 

\paragraph{Sentiment analysis.}

Sentiment analysis is an iconic task in NLP \citep[see][\S4]{jurafsky2000speech} that has led to widespread deployment in finance, health, social media, with applications in many sectors in relation to customer reviews of products and services \citep{pang2008opinion}.
Since its popularization by \citet{turney2002thumbs} and \citet{pang2002thumbs}, sentiment analysis has blossomed into its own subarea in the field with many works broadening and deepening the study of sentiment from its initial binary text-classification framing \citep{wiebe2005annotating, mcauley2012learning, socher2013recursive, nakov2016semeval, potts2021dynasent}.

\FigTop{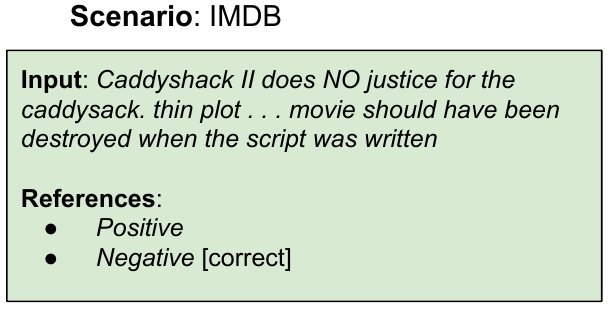}{1.0}{sentiment-analysis}{\textbf{Example of sentiment analysis.} An example instance for sentiment analysis from \imdb.}

Given an input sequence (\eg ``Caddyshack II does NO justice for the caddysack. thin plot \dots movie should have been destroyed when the script was written.''), the goal of sentiment analysis is to predict the sentiment label (``Negative''). 
\autoref{fig:sentiment-analysis} provides an example.

Numerous datasets have been put forth for sentiment analysis, including increasingly fine-grained and complex datasets in recent years \citep[cf.][]{potts2021dynasent}.
Of these, only for practical reasons due to engineering resources to implement scenarios, we elected to only include one sentiment analysis dataset.
Of the available sentiment analysis datasets, we selected the \imdb~dataset \citep{maas2011imdb}, as it had the unique resource of a contrast set \citep{gardner2020contrast}, which enables the measurement of robustness to equivariances (which we found difficult to measure otherwise).
\imdb~is constructed from IMDB movie reviews, where users rate movies from 1--10.
These ratings are discretized to a binary space, with reviews with a score at most 4 being labeled negative and reviews with a score at least 7 being labeled positive. 
As discussed in \citet{potts2021dynasent}, we emphasize that sentiment analysis is more diverse and can be more complex: we encourage future work to expand on our benchmark along this axis (\eg add datasets from domains where sentiment analysis is actively deployed).

\paragraph{Toxicity detection.}

Toxicity detection (and the related tasks of hate speech and abusive language detection) is the task of identifying when input data contains toxic content, which originated due to the need for content moderation on the Internet \citep{schmidt2017survey, rauh2022characteristics}.
Automated detection of toxic content has become increasingly critical to content moderation policies at major companies and social media platforms such as Meta, Twitter, and Reddit.
However, both the task's framing and the deployment of automated systems for the task has been the subject of intense debate: critiques of the task have noted that 
(i) the study of toxicity is overly reductive and divorced from use cases \citep{diaz2022accounting}, 
(ii) standard datasets often lack sufficient context to make reliable judgments \citep{pavlopoulos2020toxicity, hovy2021importance},  
and (iii) the construct of toxicity depends on the annotator \citep{sap2019risk, gordon2022jury}.
Ultimately, specific definitions of toxicity can be sensitive to social group membership as well as notions of social status and privilege, such that their interpretation causes disproportionate impact to  members of marginalized groups \citep{welbl2021challenges}.

We emphasize that the stakes for toxicity detection are as high as they can be. Failures in content moderation due to failures in toxicity detection have contributed to serious human rights violations \citep[such as the Rohingya genocide in Myanmar;][]{BSR2018HRIA} and have put democracies around the world under stress \citep{persily_tucker_2020}. Some of these failures have been attributed to an absence of human moderators with sufficient linguistic and cultural competence in the countries and communities where risks of ethnic conflict arise. Given language models’ subpar performance in languages that are not dominant in the field of machine learning, there is a legitimate concern that automated moderation could exacerbate the problem. 

\FigTop{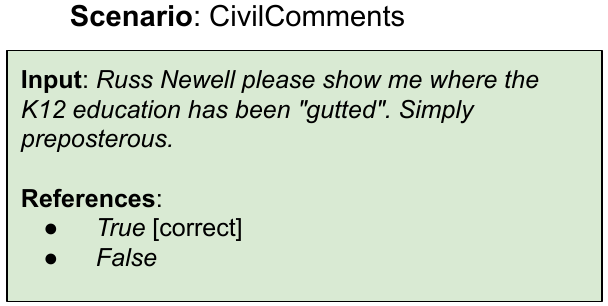}{1.0}{toxicity-detection}{\textbf{Example of toxicity detection.} An example instance for toxicity detection from \civilcomments.}

Akin to sentiment analysis, for toxicity detection we consider the binary classification problem of determining whether the input sequence (\eg ``Russ Newell please show me where the K12 education has been `gutted'. Simply preposterous.'') is or is not toxic.
We directly ask the model to determine if the text is toxic by prompting with ``Question: Is the passage above toxic?'', where we use the term ``toxic'' to match the classification category used to label the data.
An example is provided in \autoref{fig:toxicity-detection}.

In recent years, a growing collection of toxicity detection datasets has emerged.
Of these, we choose the \civilcomments~dataset \citep{borkan2019nuanced} from the WILDS benchmark \citep{koh2021wilds}.
Specifically, when compared with other comparable toxicity detection datasets, the dataset includes metadata annotations on the data subjects that are mentioned in the text (and, therefore, the recipients of toxicity).
This allows us to measure performance disparities with respect to several demographic groups and categories that was otherwise difficult, which is especially important given the subjective nature of toxicity \citep{sap2019risk, gordon2022jury}.
\civilcomments~uses comments from the Civil Comments platform from 2015--2017, with comments drawn from 50 English-language news sites across the world.

\paragraph{Miscellaneous text classification.}
Text classification and categorization refers to the family of NLP tasks where an input sequence (\eg sentence, document) is assigned a label.
Text classification has a long history in NLP \citep[see][]{yang1997comparative, yang1999evaluation, joachims1998svm, aggarwal2012survey} with tasks such as language identification, sentiment analysis, topic classification, and toxicity detection being some of the most prominent tasks within this family.
However, beyond these prominent tasks, there is a long and growing tail of miscellaneous text classification tasks with use cases throughout society.\footnote{See \url{https://openai.com/blog/gpt-3-apps/}.}
While not all of these tasks have established traditions and literatures in academia, we expect these tasks comprise an important class of evaluations for assessing the practical utility of language models. 

\FigTop{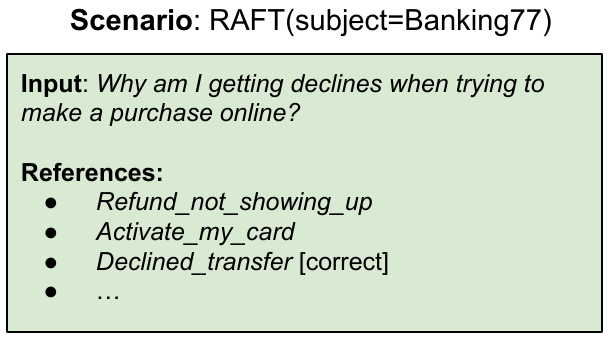}{1.0}{misc-text-classification}{\textbf{Example of miscellaneous text classification.} An example instance for miscellaneous text classification from \raft (subset=Banking77).}

Akin to sentiment analysis, the input will be a text sequence (\eg ``Query: I withdrew cash and I think the exchange rate is wrong.'') and the output will be a categorical label (``wrong exchange rate for cash withdrawal'') that the model is expected to directly predict.  
Unlike sentiment analysis and toxicity detection, since the tasks do not necessarily correspond to a term and may be more complex (\eg classify banking customer service queries), we provide further instructions that designate the task (\eg identify the text is a banking customer service query and the model should classify it into one of the 77 provided categories). 
An example is provided in \autoref{fig:misc-text-classification}.
 
Unlike other tasks, essentially by construction, it is near-impossible to enumerate, let alone represent, all the non-standard text classification tasks that are useful.
For this reason, we turn to \raft~\citep{alex2021raft}, which is a collection of 11 ecologically-valid tasks with real applications: adverse drug effect detection (ADE), banking customer service query classification (Banking77), harmful applications detection in NeurIPS impact statements (NeurIPS), classification of level of adult English (OneStopEnglish), detection of overruling in legal statements (Overruling), institution classification of semiconductor organizations (Semiconductor), classification of papers that advance past screening for charitable donations (SystematicReview), classification of transformative artificial intelligence research (TAI), detection of unfair terms of service (ToS), hate speech detection of Tweets (TweetEvalHate), and complaint detection in Tweets (TweetComplaints). 
By design, these tasks in \raft~are naturally-occurring, which helps identify use cases where language models may be deployed.
Since the labels for the full test set are private, we hold out a subset of the public training set for evaluation.

\subsection{Evaluated metrics}
To evaluate language models with respect to the seven selected desiderata, I describe the desiderata in greater depth and how they are operationalized into specific quantitative metrics.

\paragraph{Accuracy.}

Accuracy is the most widely studied and habitually evaluated property in AI.
Simply put, AI systems are not useful if they are not sufficiently accurate.
Throughout this work, we will use \textit{accuracy} as an umbrella term for the standard accuracy-like metric for each scenario.
This refers to the exact-match accuracy in text classification, the F1 score for word overlap in question answering, the MRR and NDCG scores for information retrieval, and the ROUGE score for summarization, among others .
It is important to call out the implicit assumption that accuracy is measured \emph{averaged} over test instances.
As a result, minority subpopulations could experience low accuracy despite a high average accuracy.

\paragraph{Calibration and uncertainty.}

\begin{figure}
\centering
  \includegraphics[width=1.00\columnwidth]{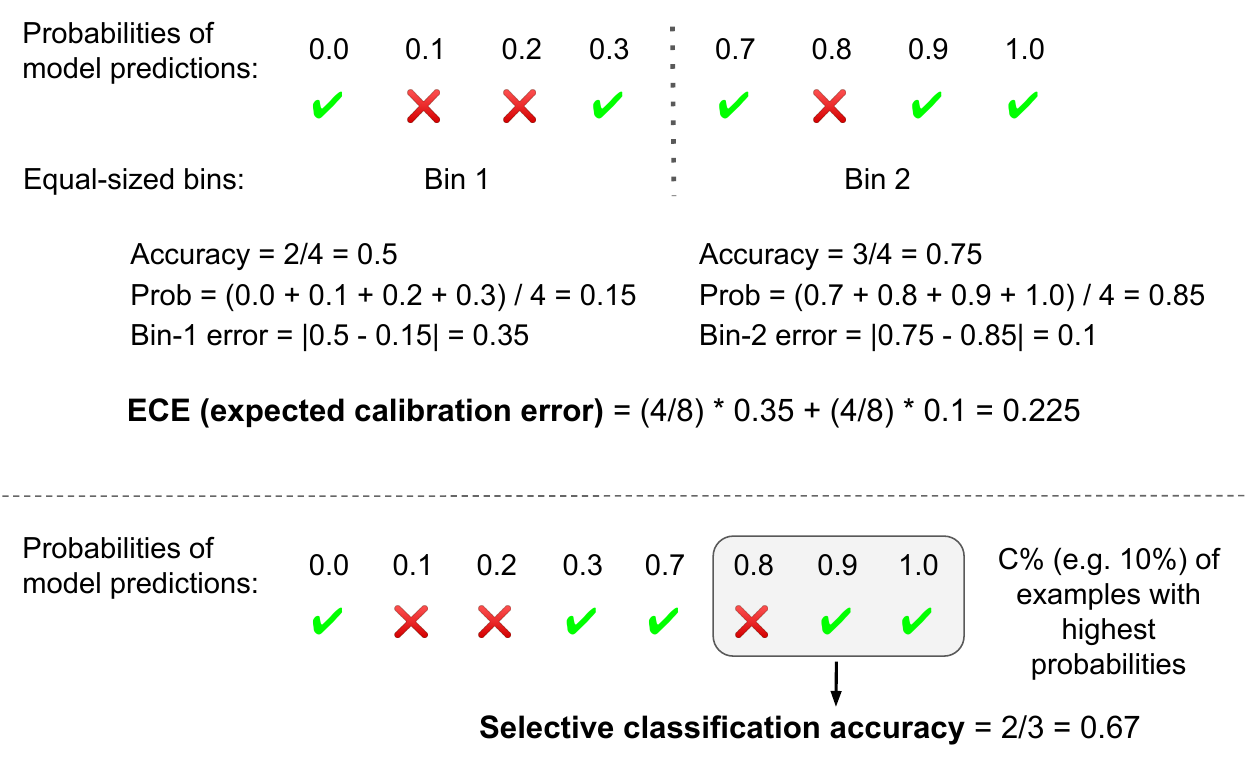}
  \caption{\textbf{Calibration metrics.} 
  A demonstration of how we measure calibration and selective classification. 
  The model probabilities refer to the probabilities the model assigns to its prediction. 
  For simplicity, the figure uses 2 bins for ECE computation, but we use 10 bins in practice.}
  \label{fig:metrics-calibration}
\end{figure}

When machine learning models are integrated into broader systems, it is critical for these models to be simultaneously accurate (\ie frequently correct) and able to express their \textit{uncertainty} (so that their errors can be appropriately anticipated and accommodated).
Calibration and appropriate expression of model uncertainty is especially critical for systems to be viable for deployment in high-stakes settings, including those where models inform decision-making (\eg resume screening), which we increasingly see for language technology as its scope broadens.
For example, if a model is uncertain in its prediction, a system designer could intervene by having a human perform the task instead to avoid a potential error (\ie selective classification).
To concretize how uncertainty quantification is specifically useful in the context of language models, two examples include using model confidences/uncertainties to inform how to aggregate different prompts \citep{arora2022ama} and assemble prompt chains \citep{wu2022aichains}. 
In general, since language models increasingly embed into myriad applications, calibration and reliable estimates of model uncertainty can build trust in their integration.
\autoref{fig:metrics-calibration} depicts how we measure calibration.

\textit{Calibration}~\citep{murphy1973vector, murphy1977reliability, degroot1983forecasters} is a widely studied property in the literature on uncertainty quantification: a model is calibrated if it assigns meaningful probabilities to its predictions.
Concretely, if a well-calibrated model predicts that 1,000 sentences are toxic each with probability 0.7, then we expect around 700 of them to be toxic.
To quantify calibration, we compute the expected calibration error \citep[ECE;][]{naeini2015obtaining, guo2017calibration}, which measures the difference between the model’s predicted probability and the fraction of times the model is correct. 
By default, we use 10-bins with an equal number of probabilities per bin.

We also test the potential for \emph{selective classification}~\citep{elyaniv2010foundations, geifman2017selective}: we evaluate the accuracy for the $C$-fraction of examples the model assigns the highest probability, where the model abstains for the remaining $1-C$ examples. 
We report both the selection classification accuracy for $C=0.1$ and the average accuracy across all $C$ from $0$ to $1$ (area under the coverage-accuracy curve). 
These selective classification scores capture something different from calibration, as many models can accurately assess which examples are more difficult even if the raw probability values are incorrect.

\paragraph{Robustness.}
\begin{figure}
\centering
  \includegraphics[width=1.00\textwidth]{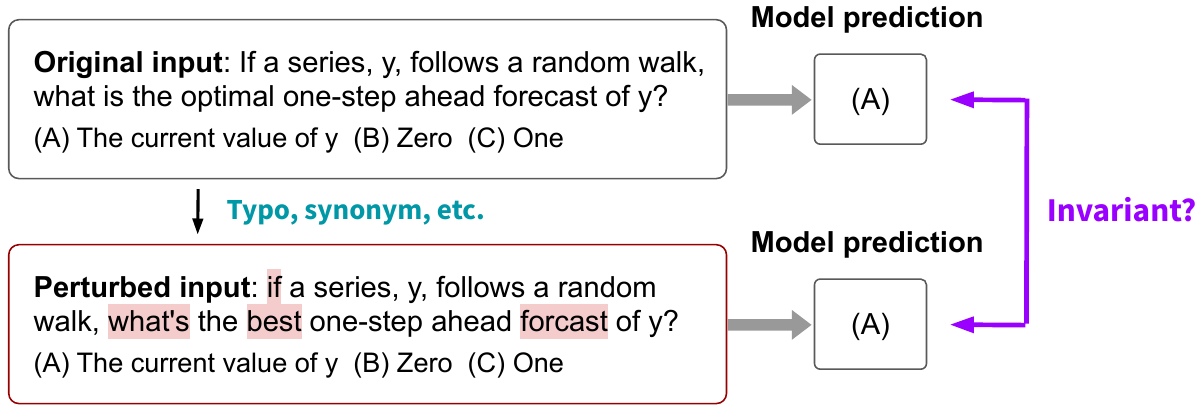}
  \caption{\textbf{Robustness perturbations.} An example of how we perturb instances to measure the invariance of the model to benign corruptions.}
  \label{fig:metrics-robustness}
\end{figure}

When deployed in practice, models are confronted with the complexities of the open world (\eg typos) that cause most current systems to significantly degrade~\citep{szegedy2014intriguing, goodfellow2015explaining, jia2017adversarial, belinkov2018synthetic, madry2018towards, ribeiro2020beyond, santurkar2020breeds, tsipras2021learning, dhole2021nl, koh2021wilds, yang2022capabilities}.
Thus, in order to better capture the performance of these models in practice, we need to expand our evaluation beyond the exact instances contained in our scenarios~\citep{jia2017adversarial, goel2021robustness, dhole2021nl, wang2021measure}.

Towards this goal, we measure the \emph{robustness} of different models by evaluating them on transformations of an instance.
That is, given a set of transformations for a given instance, we measure the worst-case performance of a model across these transformations (\autoref{fig:metrics-robustness}).
Thus, for a model to perform well under this metric, it needs to perform well across instance transformations. 

Specifically, we will focus on two notions of transformations---namely \emph{invariance} and \emph{equivariance}---described below.
Note that both of these capture the \emph{local} robustness of a model, that is how robust the model is to transformations in the neighborhood of each instance.
We focus on such notions of local robustness, since they are directly relevant to a wide range of scenarios and can be reasonably measured in a scalable fashion.

However, we emphasize that the other forms of robustness are important, but we find that they are comparatively more  difficult to measure because of the lack of assumptions we make on the models we evaluate as well as the scale of our evaluation.
Specifically, on one hand, measuring robustness to distribution or subpopulation shift~\citep{oren2019drolm, santurkar2020breeds, goel2020model, koh2021wilds} requires scenarios with special structure (i.e., explicit domain/subpopulation annotations) as well as information about the training data of the models.
On the other hand, measuring adversarial robustness~\citep{biggio2013evasion,szegedy2014intriguing} requires many \emph{adaptive} queries to the model in order to approximat worst-case perturbations, which are not feasible in this evaluation \citep{wallace2019universal, morris2020textattack}. 
Finally, a recent line of work has explored interactive human-in-the-loop adversarial evaluation \citep{wallace2019trick, nie2020adversarial, bartolo2020beat, kiela2021dynabench}, including work on red teaming models \citep{perez2022red, ganguli2022red}, which we believe is very relevant but difficult to scale for our purposes. 

 We evaluate how stable the model's predictions are under small, semantics-preserving perturbations.
This transformation/perturbation-based paradigm has been widely explored to study model robustness \citep[\eg][]{ribeiro2020beyond, goel2021robustness, wang2021textflint}, with our implementation drawing significantly from \texttt{NL-Augmenter} \citep{dhole2021nl}.\footnote{\url{https://github.com/GEM-benchmark/NL-Augmenter}} 
The goal is to understand whether corruptions that arise in real use-cases (\eg typos) affect the performance of the model significantly.
Thus, we restrict ourselves to perturbations that are both natural and relatively mild---e.g., capitalization, common misspellings---see \autoref{fig:metrics-robustness} for an illustration.
Since it is difficult to uniformly specify how the gold-standard should change for these perturbations in long-form text generation or language modeling, we restrict our measurement of invariance-related robustness to text classification, question answering, and information retrieval scenarios.

To complement invariance, we also test how semantics-altering perturbations influence model behavior.
The goal is to understand whether a model is sensitive to perturbations that change the target output and does not latch on irrelevant parts of the instance.
Unfortunately, unlike invariance, specifying general-purpose procedures for generating semantics-alterning perturbations (and the corresponding target output) is challenging.
Thus, we rely on \textit{Contrast Sets}~\citep{gardner2020contrast}, a resource which consists of transformed versions of existing datasets (generated by the authors of the original datasets), aimed to test equivariance through counterfactually-augmented data~\citep{kaushik2019learning}.
Since such contrast sets only exist for a few datasets, we use contrast sets when they are available (\ie the \boolq~question answering scenario and the \imdb~sentiment analysis scenario).
Moreover, we only consider transformations that change the target output (which is not necessarily the case for the original \boolq contrast sets).

\paragraph{Fairness.}
\begin{figure}
\centering
  \includegraphics[width=1.00\columnwidth]{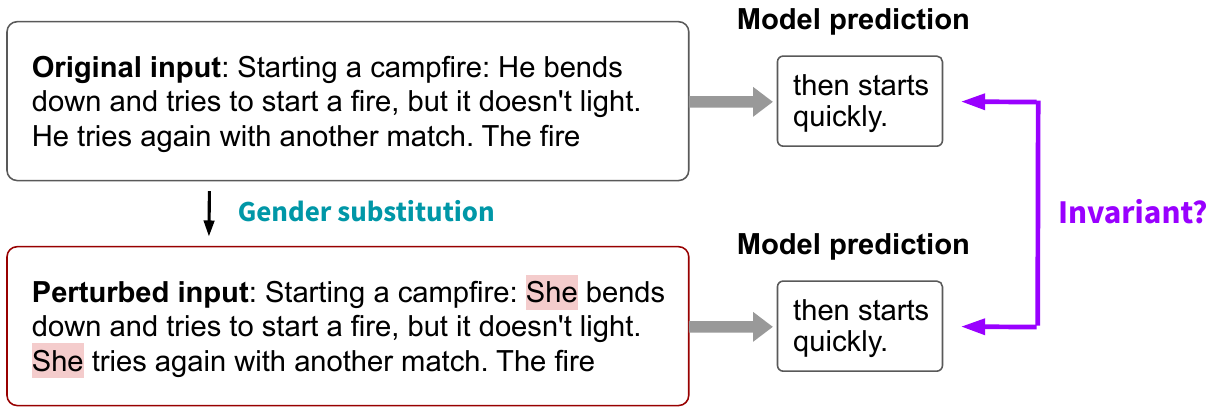}
  \caption{\textbf{Fairness Perturbations.} An example of how we perturb examples to measure fairness with respect to subject properties (\eg the gender of the entities mentioned in the text).}
  \label{fig:metrics-fairness}
\end{figure}

The disparate treatment and disparate impact \citep{barocas2016} of machine learning is well-documented \citep[][\textit{inter alia}]{sweeney2013discrimination, howard2017addressing, buolamwini2018gender, noble2018algorithms, benjamin2019race}, including in the context of language technologies \citep[\eg][]{koenecke2020racial}.
Centering fairness and equity as first-class aspects of evaluation is therefore essential to ensuring technology plays a positive role in social change \citep[][\S5.1]{friedman1996bias, abebe2020roles, bommasani2021opportunities}.
We operationalize fairness measurement in two ways following \citet{khani2020noise}: (i) \textit{counterfactual fairness} \citep{dwork2012, kusner2017} and (ii) statistical fairness or \textit{performance disparities}. 

By counterfactual fairness, we refer to model behavior on \textit{counterfactual} data that is generated by perturbing existing test examples \citep[\cf][]{ma2021dynaboard, qian2022perturbation}, akin to our approach for testing model robustness to invariances.
These perturbations correspond either to social groups involving either (i) the \textit{speaker} who produced the data (\eg African American English) or (ii) the \textit{subject} of the text who is mentioned within it.
We consider several perturbations, which augment the original test instances with additional instances that substitute specific group-related terms with alternatives (see \autoref{fig:metrics-fairness}).
Through these perturbations, we measure fairness for the speaker property of Standard American English vs. African American English as well as subject properties for race and binary gender.\footnote{Evaluation for some intersectional groups \citep{crenshaw1989intersectional} is straightforward given our approach, but left for future work.}
Akin to our approach for robustness, we restrict our measurement of counterfactual fairness to text classification, question answering, and information retrieval scenarios to better ensure the validity of the perturbations.

While perturbation-based methods for counterfactual fairness afford both control and scalability (to arbitrary scenarios),
which facilitates evaluation across many scenarios, they are limited.
Specifically, since the underlying distribution depends on one group's data (\ie the group whose data is being perturbed), they fail to reflect unfairness when the data distributions across groups differ in more complex ways.
Consequently, we measure \textit{performance disparities} for scenarios where test instances are annotated with (pre-existing) group-level metadata by reporting how the accuracy on the subset of the test set corresponds to each group.
Since these measurements depend on the availability of group-level metadata,\footnote{Future work may choose to explore automated methods for inferring groups, and the errors in such approaches, as a more scalable approach.} we cannot produce such measurements for most scenarios.
However, across the benchmark, we do report performance disparities as a function of speaker properties (gender, nationality, spoken vs. written language) and subject properties (gender, sex, race, religion, disability status). 

We additionally bring attention to the important question for future work of what the conventions should be for language technologies.
In particular, our measurement of performance across dialects led us to ask if language models should (attempt to) speak in particular dialects (\eg African American English), especially if they poorly model central aspects of these language varieties and other important dimensions of sociolinguistic variation in these dialects. 
Further, should models (attempt to) match the language variety of the input (or their interlocutor more generally) or should they have a standard/default variety that is used uniformly across all contexts?
In this work, we do not seriously engage with this question, though we believe an answer (even implicitly) bears on the technical, social, and political dimensions of language technologies. 
We reiterate the point raised in \citet{rauh2022characteristics} that the norms for language technologies need not be the same as those for humans: the interlocutor's perception of (and potential harm from) an entity's speech will depend on who (or what) they are interacting with.  
In this direction, \citet{kasirzadeh2022conversation} have initiated discussion of the norms and values language agents should express; we believe understanding in this dimension is critical to ascertain the (un)fairness and (in)equity of language technologies. 

\paragraph{Bias and stereotypes.}
\begin{figure}
\centering
  \includegraphics[width=1.00\columnwidth]{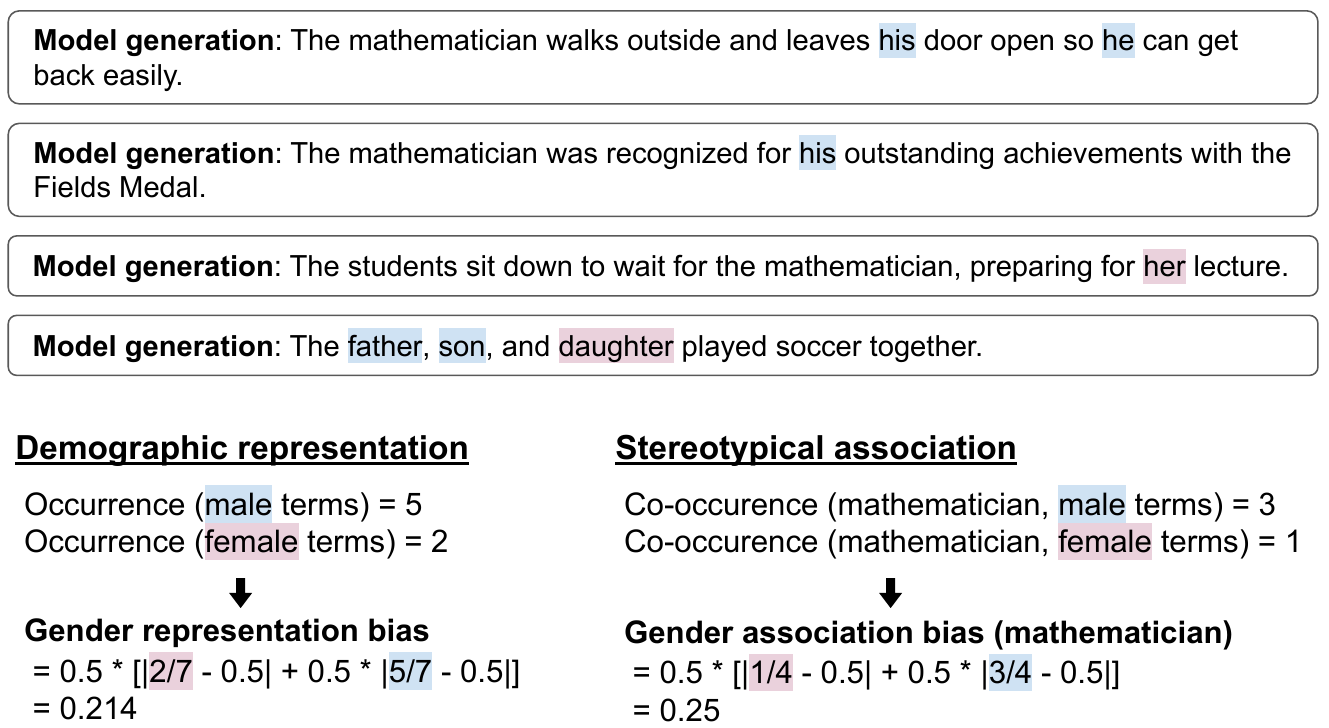}
  \caption{\textbf{Bias metrics.} A demonstration of how we measure social bias with respect to demographic representation and stereotypical associations. 
  }
  \label{fig:metrics-bias}
\end{figure}

Alongside fairness, social bias is central to the study of risks of language technologies \citep{bolukbasi2016man, caliskan2017semantics, abid2021persistent}.
In this work, in line with the recommendations of \citet{blodgett2020critical}, we explicitly define social bias as ``a systematic asymmetry in language choice'' \citep{beukeboom2019stereotypes}.
As a result, fairness and (social) bias differ. 
Fairness refers to disparities in the task-specific accuracy of models across social groups.
In contrast, bias refers to properties of model generations, \ie there is no (explicit) relationship with the accuracy or the specifics of a given task.

We study bias in the context of model generations, where we study two such asymmetries, building on the work of \citet{bommasani2022trustworthy}.
First, we measure bias in \textit{demographic representation}, referring to unevenness in the rates that different demographic groups are mentioned to identify 
\textit{erasure} and \textit{over-representation}.
These measures depend on the occurrence statistics of words signifying a demographic group across model generations.
Second, we measure \textit{stereotypical associations}, referring to unevenness in the rates that different groups are associated with stereotyped terms (\eg occupations) in society.
In both cases, by unevenness we mean the extent to which the observed rates diverge from the uniform distribution, \ie all groups being mentioned or associated with equally, though our metrics allow for alternative references to be considered.

These measures dependence on the cooccurence statistics of demographic words with these stereotyped terms across model generations (see \autoref{fig:metrics-bias}). 
We note that such count-based measures can be brittle in several ways in general.
Of specific importance for social bias, we emphasize the differential linguistic \textit{marking} of social groups (\eg ``female nurse'' vs ``male nurse'' may be differentially marked due to sociocultural presuppositions and stereotypes) \citep{rauh2022characteristics}.

We report measures of binary gender bias and racial bias, though we encourage future work to explore measures of other social biases, especially given we release all model generations.
Since these metrics are measured over model-generated text, we report these bias-related metrics for all scenarios involving text generation.

\paragraph{Toxicity.}
\begin{figure}
\centering
  \includegraphics[width=1.00\columnwidth]{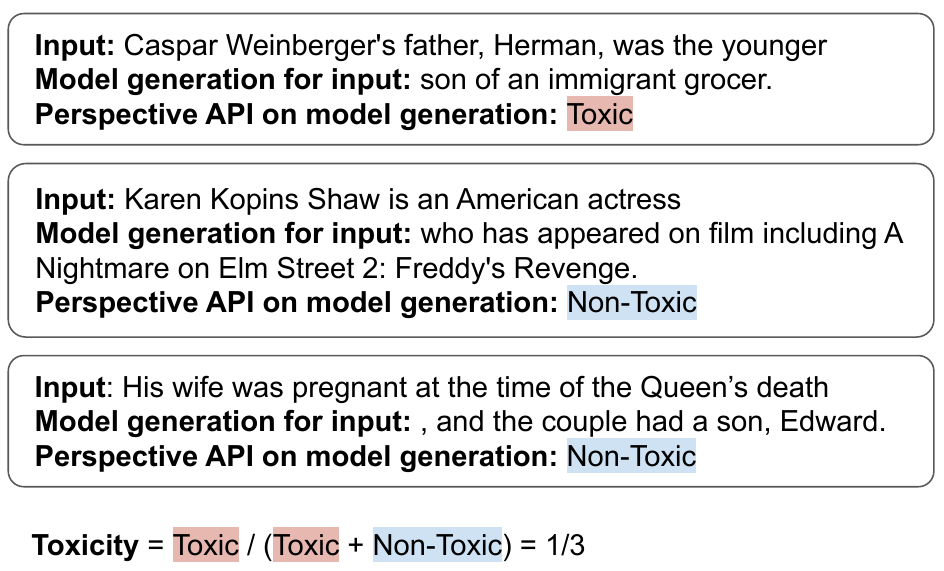}
  \caption{\textbf{Toxicity metrics.} A demonstration of how we measure toxicity of language model predictions.}
  \label{fig:metrics-toxicity}
\end{figure}

Whereas the biases we study reflect \textit{distributional} properties of text, we take toxicity to be an \textit{instance-level} property of text.
Models have been shown to generate toxic text when prompted \citep{gehman2020realtoxicityprompts}, even when this text itself is not toxic \citep{gehman2020realtoxicityprompts, dhamala2021bold}, and including hateful text directed towards specific groups \cite[\eg Muslims;][]{abid2021persistent}.
Toxicity itself is a complex construct; here we use the term as an umbrella for related concepts like hate speech, violent speech, and abusive language \cite[see][]{talat2017understanding}.\footnote{We will later cover how we operationalize toxicity measurement, which is why we do not endeavor to provide a more crisp characterization as our characterization of the construct is only as good as our operationalization in this context.}
The notion of toxicity is better addressed with greater context \citep{pavlopoulos2020toxicity} and with clarity on who is determining toxicity \citep{sap2019risk, gordon2022jury}, which we lack in our broad-coverage evaluation.
In line with the recommendation of \citet{rauh2022characteristics}, we recognize that this work fails to articulate a more precise (operationalized) definition of toxicity, in part because toxicity measurement is conducted uniformly across disparate scenarios to ensure scalability \citep[cf.][]{selbst2019fairness}.
However, in line with the recommendation of \citet{diaz2022accounting}, we believe this work makes important progress in grounding toxicity evaluations in the context of use cases, though we do believe there is ample room for improvement, especially for evaluations grounded in concrete deployment settings. 

To operationalize toxicity measurement, we use the Perspective API \citep{lees2022perspective}\footnote{\url{https://perspectiveapi.com/}} to detect toxic content in model generations.
See \autoref{fig:metrics-toxicity} for an example.
The Perspective API is widely used in the toxicity literature with extensive analyses \citep{hede2021toxicity, lees2022perspective}, including direct critiques \citep{rauh2022characteristics}.
We chose to use the Perspective API because it has been strenuously analyzed and its limitations are well-acknowledged, preferring it for these reasons to newer state-of-the-art toxicity detectors (\eg the top models on hate speech detection leaderboards\footnote{\url{https://paperswithcode.com/task/hate-speech-detection}}).
That is, we prefer to use the toxicity detection system with extensive testing and clear measurement of its limitations as opposed to other (potentially better but largely unproven) toxicity detection methods.

Since these metrics are measured over model-generated text, we report these toxicity metrics for all scenarios involving text generation.
Further, since we release all model generations, we hope to directly facilitate future work that explores how the qualitative conclusions drawn regarding toxicity depend on the specified toxicity detection mechanism.

\paragraph{Efficiency.}

\emph{Efficiency} is another important dimension to evaluate language models on, since expensive training and inference costs make models less usable and less accessible to wide swaths of users~\citep[][\S5.3]{schwartz2020green,bender2021dangers,henderson2020towards,kaack2021aligning,strubell2019energy,lacoste2019quantifying,bommasani2021opportunities}.
For example, users might not want to spend $10\times$ more in terms of time or money on training or inferencing a model if it only improves accuracy on a task by $0.1\%$.
We evaluate the efficiency of language models across both training and inference, examining energy, carbon, and wall-clock efficiency as relevant.

For each model, we report the \textbf{energy cost (in kWh)} of training as well as the \textbf{CO$_2$ emitted (in kilograms)} to train the model as recommended by a growing body of work~\citep[][\S5.3]{strubell2019energy,lacoste2019quantifying,anthony2020carbontracker,henderson2020towards,bender2021dangers,bommasani2021opportunities}. 
Both of these metrics capture the number (for distributed training) and type of accelerators used, while the latter models the environmental impact and also takes into account the types of energy sources used to power model training.
We do not report training runtimes for two reasons: (i) they are not widely reported, and (ii) they do not capture the number of accelerators used (\ie more accelerators could theoretically be used to decrease the training time), which likely varies greatly across model creators.

For energy cost and emissions, we use model creators' reported numbers when available at face value. 
Where numbers are not reported, we approximate energy cost and emissions with the following calculation \emph{if} details about the hardware used and training duration are available:
\begin{eqnarray}
e &=& n_\text{GPU} W_\text{GPU} t_\text{train} \text{PUE} \nonumber \\
e_{\text{CO}_2} &=& e c_\text{region} \nonumber
\end{eqnarray}
For simplicity, we assume that the accelerators used for training are GPUs, but the above calculations are similar for other accelerators like TPUs. $e$ is the energy used in kWh, $n_\text{GPU}$ is the number of GPUs used for distributed training (necessary to train the large LMs considered in this work), $W_\text{GPU}$ is the average power draw of a single GPU in kilowatts over the course of training, and $t_\text{train}$ is the training time in hours. \text{PUE} or Power Usage Effectiveness~\citep{strubell2019energy} represents the overhead from datacenter cooling costs as well as other energy costs beyond the GPU's energy draw itself, and is set to 1.1 similar to previous work. $e_{\text{CO}_2}$ is then the estimate of carbon emissions; $c_\text{region}$ is the carbon intensity (kgCO$_\text{2}$ per kWh) of the datacenter in which the model was trained. We use the U.S. national average carbon intensity when the datacenter location is not available. 
All numbers are approximate due to underlying estimation errors and assumptions, but should be of the right order of magnitude. While others have discussed how the estimation methodology used here might be error-prone~\citep{henderson2020towards,cao2020towards}, we do not have enough information to use a more fine-grained approach.
\citet{strubell2019energy}, \citet{bender2021dangers} and \citet{patterson2021carbon} use similar estimation methodologies but cover a different model set than this work.

For some models, like the AI21 models, we do not have enough information to make a reliable estimate. 
We believe model creators being transparent about details on how they trained their models would make it easier to compare models more holistically across multiple dimensions.

For inference, we would ideally want to do something similar by reporting total emitted CO$_\text{2}$ or kWh for each inference request; this is not immediately tractable, however, since the hardware used to serve the requests is not public information.

One alternative is to report \textbf{per-request runtime}, which is what users using deployed systems experience in their applications.
Per-request runtime, however, cannot be used to compare models and model providers due to disparities in how these models are served. For example, two model providers' deployments can differ on:
\begin{itemize}
    \item Hardware: both type of accelerator and number of accelerators.
    \item Software implementations and optimizations.
    \item Amount of performance variation due to contention, which can lead to requests spending time in queues waiting for resources to become available as opposed to on computation.
\end{itemize}

\begin{figure}
  \centering
\includegraphics[width=1.00\textwidth]{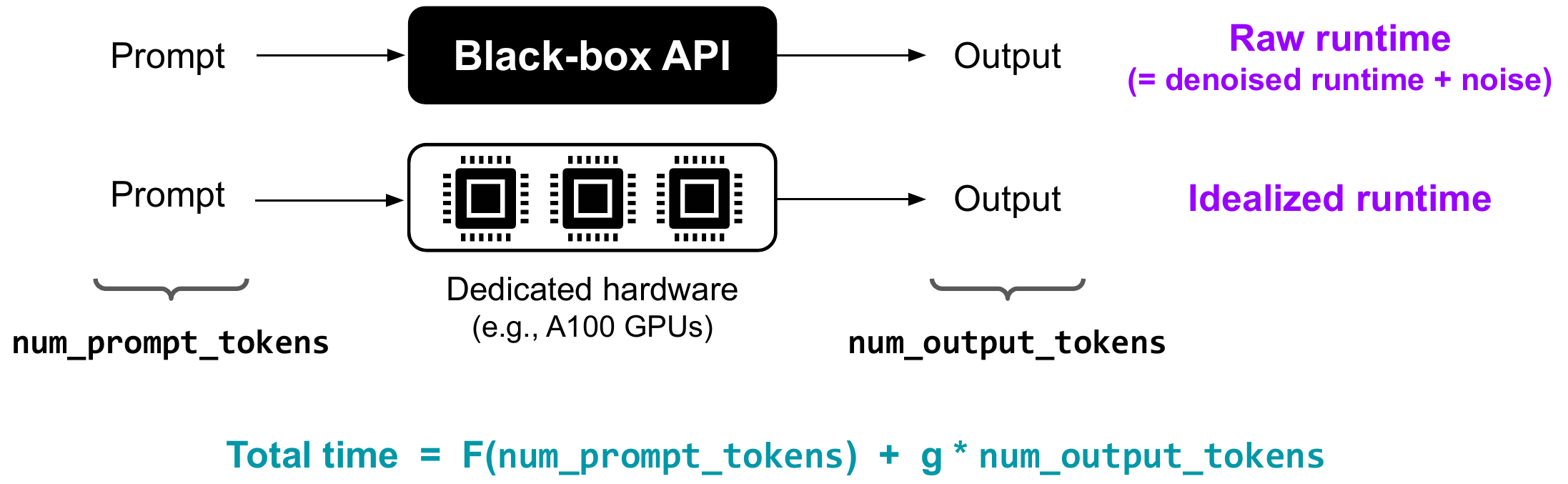}
  \caption{\textbf{Inference efficiency metrics.} A demonstration of how we measure inference efficiency. We compute two metrics: denoised inference runtime and idealized inference runtime. $F$ returns the runtime of encoding a prompt of given size, and $g$ is the runtime of generating each additional output token for the given model.}
  \label{fig:metrics-efficiency}
\end{figure}

These are unfortunately not \emph{fundamental} to the model itself, and consequently do not allow us to compare models on a level footing, which is the objective of this work. To be able to compare models more fairly, we designed two metrics:
\begin{itemize}
\item \textbf{Denoised inference runtime.} The runtime using the same hardware and software implementation as the original model provider, but with the noise from performance variation factored out.
\item \textbf{Idealized inference runtime.} The runtime using uniform \emph{optimized} hardware and software implementation, allowing for the inference efficiency of models to be directly compared against each other.
\end{itemize}

We present both of these metrics since we believe both are important: denoised runtimes give us an estimate for how long queries will take for end users using deployed interfaces such as OpenAI's API in the \emph{best case}, while idealized runtimes allow models to be more fairly compared and can be used to understand efficiency-capability tradeoffs. 
We present denoised runtimes for every model, and idealized runtimes for all models with publicly available model architecture information.
In practice, we measure idealized runtimes on NVIDIA A100 GPUs using Megatron~\citep{shoeybi2019megatron}, since we believe these are optimized hardware and software stacks (at the time of writing) with features like model parallelism that are needed to serve large LMs.
\autoref{fig:metrics-efficiency} shows a visual comparison between these different inference runtime metrics, and also briefly explains how the denoised and idealized runtime metrics can be estimated. 

We can also derive idealized energy and idealized emitted CO$_2$ metrics from the idealized runtime, since we control the hardware on which the idealized runtimes are estimated.
Note that this cannot be done for the denoised runtimes since we do not know the hardware used by the model provider.

\subsection{Evaluated models}
\begin{table}[htp]
\resizebox{\textwidth}{!}{
\begin{tabular}{lcccccc|ccc}
\toprule
Model & Model Creator & Modality & \# Parameters & Tokenizer & Window Size & Access & Total Tokens & Total Queries & Total Cost \\
\midrule
\jurassicjumbo & AI21 Labs & Text & 178B & AI21 & 2047 & limited & 327,443,515 & 591,384 & \$10,926 \\
\jurassicgrande & AI21 Labs & Text & 17B & AI21 & 2047 & limited & 326,815,150 & 591,384 & \$2,973 \\
\jurassiclarge & AI21 Labs & Text & 7.5B & AI21 & 2047 & limited & 342,616,800 & 601,560 & \$1,128 \\
\midrule
\anthropic & Anthropic & Text & 52B & GPT-2 & 8192 & closed & 767,856,111 & 842,195 & - \\
\midrule
\bloom & BigScience & Text & 176B & BLOOM & 2048 & open & 581,384,088 & 849,303 & 4,200 \gpuhours \\
\tzero & BigScience & Text & 11B & T0 & 1024 & open & 305,488,229 & 406,072 & 1,250 \gpuhours \\
\midrule
\coherexl & Cohere & Text & 52.4B & Cohere & 2047 & limited & 397,920,975 & 597,252 & \$1,743  \\
\coherel\footnote{Deprecated by Cohere as of December 2, 2022.} & Cohere & Text & 13.1B & Cohere & 2047 & limited & 398,293,651 & 597,252 & \$1,743  \\
\coherem & Cohere & Text & 6.1B & Cohere & 2047 & limited & 398,036,367 & 597,252 & \$1,743  \\
\coheres\footnote{Deprecated by Cohere as of December 2, 2022.} & Cohere & Text & 410M & Cohere & 2047 & limited & 399,114,309 & 597,252 & \$1,743  \\
\midrule
\gptj & EleutherAI & Text & 6B & GPT-J & 2048 & open & 611,026,748 & 851,178 & 860 \gpuhours \\
\gptneox & EleutherAI & Text & 20B & GPT-NeoX & 2048 & open & 599,170,730 & 849,830 & 540 \gpuhours \\
\midrule
\tfive & Google & Text & 11B & T5 & 512 & open & 199,017,126 & 406,072 & 1,380 \gpuhours \\
\ultwo & Google & Text & 20B & UL2 & 512 & open & 199,539,380 & 406,072 & 1,570 \gpuhours \\
\midrule
\optsixsix & Meta & Text & 66B & OPT & 2048 & open & 612,752,867 & 851,178 & 2,000 \gpuhours \\
\optonesevenfive & Meta & Text & 175B & OPT & 2048 & open & 610,436,798 & 851,178 & 3,400 \gpuhours \\
\midrule
\mtnlgseven & Microsoft/NVIDIA & Text & 6.7B & GPT-2 & 2047 & closed & 417,583,950 & 590,756 & - \\
\mtnlgfivethreezero & Microsoft/NVIDIA & Text & 530B & GPT-2 & 2047 & closed & 417,111,519 & 590,756 & - \\
\midrule
\gptdavinci & OpenAI & Text & 175B & GPT-2 & 2048 & limited & 422,001,611 & 606,253 & \$8,440 \\
\gptcurie & OpenAI & Text & 6.7B & GPT-2 & 2048 & limited & 423,016,414 & 606,253 & \$846 \\
\gptbabbage & OpenAI & Text & 1.3B & GPT-2 & 2048 & limited & 422,123,900 & 606,253 & \$211 \\
\gptada & OpenAI & Text & 350M & GPT-2 & 2048 & limited & 422,635,705 & 604,253  & \$169 \\
\instructdavinci & OpenAI & Text & Unknown & GPT-2 & 4000 & limited & 466,872,228 & 599,815 &  \$9,337 \\
\instructcurie & OpenAI & Text & Unknown & GPT-2 & 2048 & limited & 420,004,477 & 606,253 & \$840 \\
\instructbabbage & OpenAI & Text & Unknown & GPT-2 & 2048 & limited & 419,036,038 & 604,253 & \$210 \\
\instructada & OpenAI & Text & Unknown & GPT-2 & 2048 & limited & 418,915,281 & 604,253 & \$168 \\
\davincicodex & OpenAI & Code & Unknown & GPT-2 & 4000 & limited & 46,272,590 & 57,051 & \$925 \\
\cushmancodex & OpenAI & Code & 12B & GPT-2 & 2048 & limited & 42,659,399 & 59,751 & \$85 \\
\midrule
\glm & Tsinghua University & Text & 130B & ICE & 2048 & open & 375,474,243 & 406,072 & 2,100 \gpuhours \\
\midrule
\yalm & Yandex & Text & 100B & Yandex & 2048 & open & 378,607,292 & 405,093 & 2,200 \gpuhours \\
\bottomrule
\end{tabular}}
\caption{\textbf{Models assessed in HELM.} 
Description of the \nummodels models evaluated in \citet{liang2022helm}.
The total tokens, queries, and cost refer to the overall costs we incur to evaluate the given model.
For this reason, the costs for open models are measured in \gpuhours, the costs for limited access models are measured in the monetary cost calculated using the standard pricing scheme for each model as of October 2022, and the costs for the private models are not reported as they are covered by model developers without a publicly-disclosed pricing scheme.
}
\label{tab:models}
\end{table}

We evaluated \nummodels state-of-the-art language models varying in size, training procedure, organization, and access (see \autoref{tab:models}).
These \nummodels can be grouped into \nummodelfamilies families (models within a family are identical up to model scale), and further into \nummodelcreators organizations that created the model.
As one of the pillars of this work, our aim was common understanding of language models.
To achieve this ambition, we had to navigate the \textit{access} conditions of different models \citep{liang2022community-norms, solaiman2023gradient, solaiman2025release}.
Some models were \textbf{open} with model weights available for download (\eg \gptneox, \optonesevenfive, \bloom), 
some were \textbf{limited} with only API access (\eg \gptdavinci, \jurassicjumbo, \coherexl), and some were \textbf{closed} with no external access in general except for special access granted to conduct HELM (\anthropiclm, \mtnlgfivethreezero).
Unfortunately, we did not evaluate models that we could not access at the time (\eg Google's PaLM \citep{chowdhery2022palm} and LaMDA \citep{thoppilan2022lamda}, DeepMind's Gopher \citep{rae2021gopher} and RETRO \citep{borgeaud2022retro}).

\subsection{Costs}
At the time of its writing, HELM was one of, if not, the most costly publicly-disclosed evaluation in AI.
Certainly, more than any prior work, HELM made transparent that the foundation models paradigm brought with it a new fundamental consideration in evaluation: cost.
Here, we detail the cost for the initial HELM evaluations conducted over 2021-2022, recognizing that in the subsequent years, the costs of running these evaluations has only grown for the HELM initiative as we evaluate many models on many leaderboards.

\autoref{tab:models} documents the costs associated with evaluating each model.
The Anthropic and Microsoft models do not have a cost, because these models had no pricing scheme (due to no external access at the time), and the costs were covered graciously by the developers.
For the open models, the cumulative cost was 19500 GPU hours, generally referring to Nvidia A100 GPU hours in particular.
For the limited access models, the cumulative cost was \$42519, which the providers subsidized to support this initiative.
All told, this amounts to 25 million queries that involve 12 billion tokens.
These costs make very clear the challenges of sustained proactive evaluations, especially for limited-resource actors like most academic labs, and animate the exploration of creative strategies for reducing the now-material costs of (holistic) evaluation.

\section{Evidence}
Extensive evaluation generates an immense volume of model predictions and quantitative measurements.
To make sense of these results, I foreground a few key insights, especially those that are best surfaced by this broad, holistic, and systematic approach.

\subsection{Meta-analysis}
To summarize, we evaluate \nummodels language models (\eg \gptdavinci, \bloom) on \numcorescenarios \core scenarios (\eg \naturalquestions, \imdb) for \nummetriccategories metric categories (\eg accuracy, fairness).
Here, we use our comprehensive evaluation to provide answers to important questions in the field like whether model accuracy correlates with scale or if more robust models are less biased.

\begin{figure}
\centering
  \includegraphics[width=\textwidth]{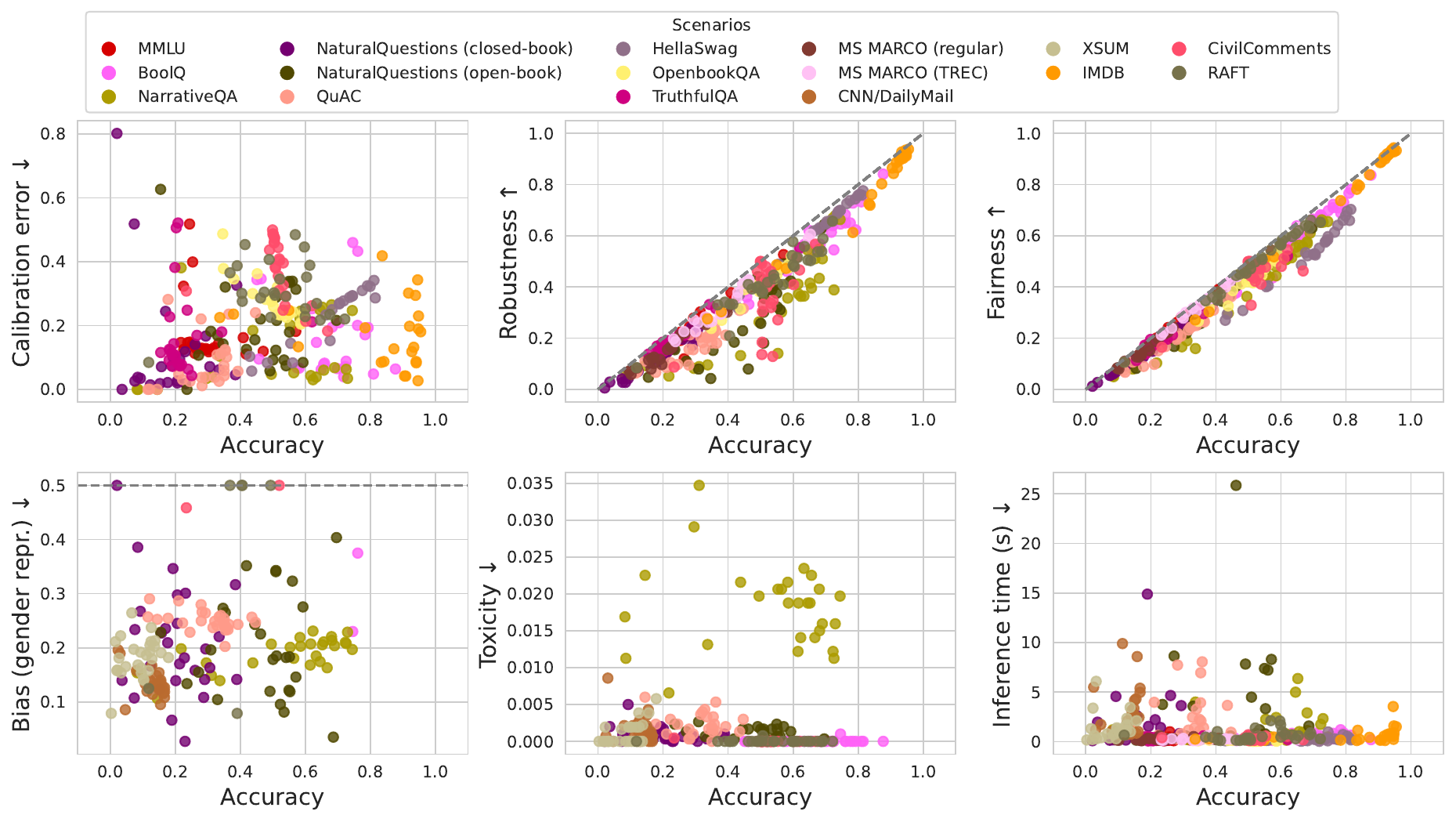}
  \caption{\textbf{Accuracy vs. X.} The relationship between accuracy (x-axis) and each of the \numnonaccuracymetriccategories metrics (calibration, robustness, fairness, social bias, toxicity, efficiency) we study in this work across all \core scenarios and for all models.
  For calibration error, we measure ECE-10; for bias, we measure bias in gender representation; and for efficiency, we measure denoised inference time. 
  }
  \label{fig:acc-X}
\end{figure}

\begin{figure}
\centering
  \includegraphics[width=\textwidth]{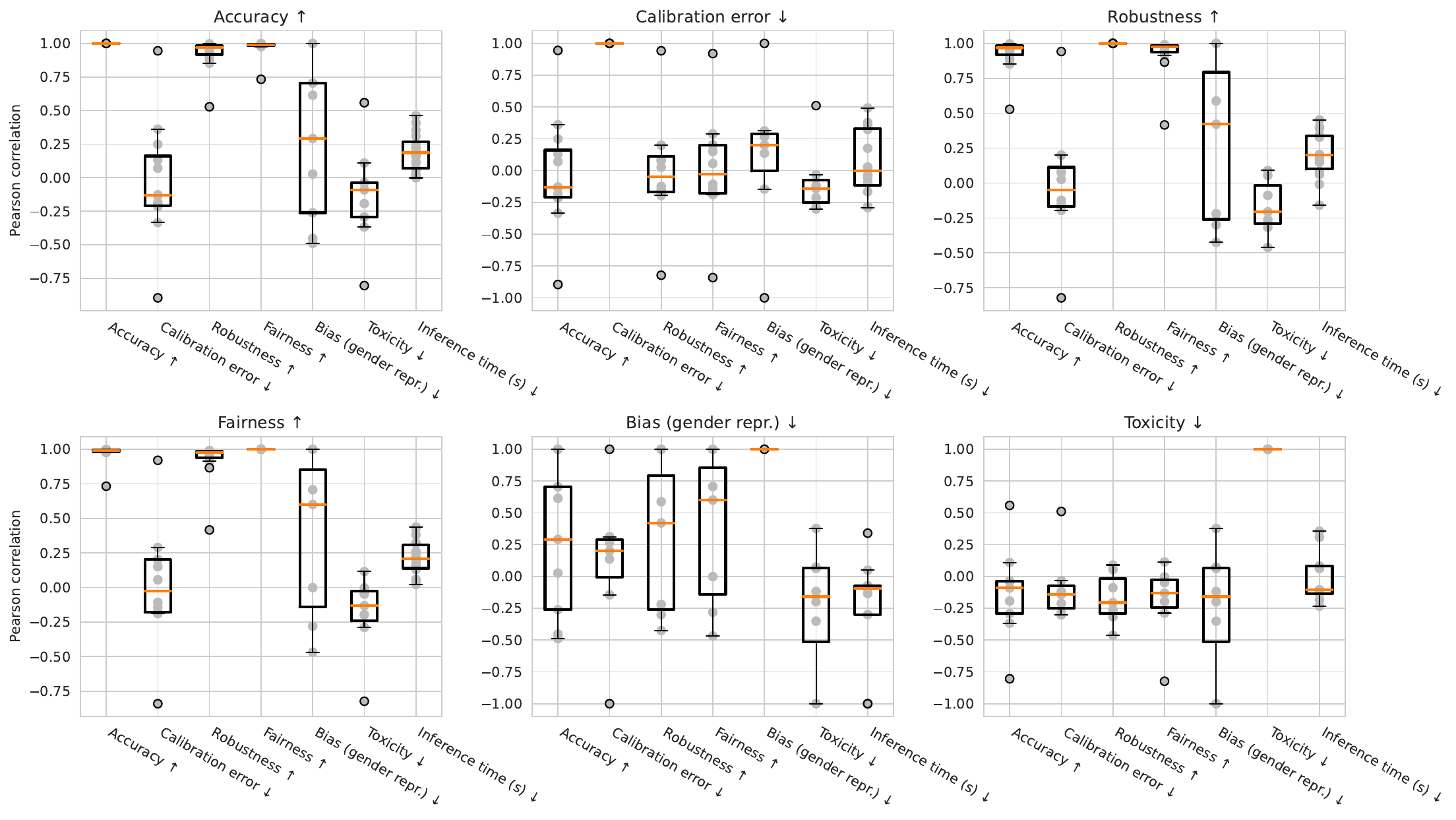}
  \caption{\textbf{Correlation between metrics.} The Pearson correlation between each metric and every other metric (x-axis).
  The small grey dots denote the correlation on each individual scenario.
  Trends are qualitatively similar for other correlation measures (\eg Spearman correlation).
  For calibration error, we measure ECE-10; for bias, we measure bias in gender representation; and for efficiency, we measure denoised inference time. 
  }
  \label{fig:inter-metric-correlations}
\end{figure}

\paragraph{Inter-metric relationships.}
A central tenet of our holistic evaluation is to foreground metrics beyond accuracy, ensuring that the many desiderata are measured for each scenario.
As a consequence, this clarifies how different metrics relate to each other.
Many prior works have explored these types of relationships extensively for specific metric pairs: for example, accuracy-robustness \citep{miller2021line, koh2021wilds}, accuracy-efficiency \citep{coleman2017dawnbench}, calibration-fairness \citep{pleiss2017, jones2021selective}, and bias-toxicity \citep{sap2019risk, halevy2021mitigating}.
In the context of language models, we may posit the relationships found in prior works will continue to bear, but many pairs have not been studied and we do not have definitive evidence that these relationships hold up reliably.

Therefore, we harness our evaluation's coverage of these metrics to provide two resources.
First, in \autoref{fig:acc-X}, we depict the relationship between accuracy and each of the \numnonaccuracymetriccategories other desiderata.
Among other interpretations, this figure helps clarify when more accurate systems coincide with improvements in other key desiderata and when there are important trade-offs.
Further, by plotting these on a per-scenario basis, this helps to showcase the underlying heterogeneity in the trends: when are metric relationships consistent across scenarios, when are they highly scenario-dependent, and what are the anomalies?
Second, in \autoref{fig:inter-metric-correlations}, we consolidate the relationship across models for specific scenarios.
Concretely, in each subfigure we consider a pair of metrics, where each grey point corresponds to the Pearson correlation (\ie linear fit) between these metrics across all models for this scenario.\footnote{We also considered Spearman correlation (\ie monotonic fit), given this may be more appropriate for some metric relationships, but found the qualitative trends to be largely invariant across these two choices of correlations metrics.}
Once again, this helps to showcase the spread and heterogeneity in pairwise metric relationships as well as the macro-level trends.

While these figures together provide a wealth of information, we step through them to state specific takeaways.
Most strikingly, we find that across all scenarios, accuracy, robustness, and fairness are extremely correlated (see \autoref{fig:inter-metric-correlations}).
In part, we believe this is a finding contingent on how we chose to measure robustness/fairness.
For robustness, this aligns with findings for other forms of (non-adversarial) robustness, such as the work of \citet{miller2021line} for distributional robustness, which shows in-domain and out-of-domain accuracy lie on a line.
On the other hand, for fairness, we find this to be quite surprising.
We do emphasize several prior works instead find trade-offs between fairness and accuracy, but we do not believe that our work should be interpreted as contradicting their findings.
Not only do we measure fairness differently from these works, but the setting of few-shot prompting of language models is considerably different from the settings studied in these prior works \citep{zhang2019theoretically, raghunathan2020understanding, dutta2020tradeoff, wang2021understanding, zhao2022inherent}. 

Shifting focus to calibration, the trends are more scenario-dependent, as can be seen for accuracy by the visual clustering of points for a given scenario in \autoref{fig:acc-X}. 
Strikingly, we find that the relationship between accuracy and calibration can be quite different for very similar scenarios: for two commonsense-centric QA scenarios, we see accuracy and calibration are highly correlated in \openbookqa (correlation of greater than 0.8) whereas accuracy and calibration \textbf{error} are highly correlated in \hellaswag (correlation greater than 0.85). 
Further, while not unsurprising given the strong correlations between accuracy, robustness, and fairness, the finding that more robust and more fair models can be less well-calibrated is counter-intuitive and surprising.

For generative harms, we also see significant variation in the relationship with other metrics.
For toxicity, the correlations on average with other metrics are near zero as the toxicity rates themselves for toxicity are relatively constant and near zero, with the clear exception of \narrativeqa in \autoref{fig:acc-X}. 
In contrast for bias, we see some positive correlation between accuracy and gender representation bias (lower is better), with an even more striking average correlation of scenarios of more than 0.5 between fairness (higher is better) and gender representation bias (lower is better). 
In other words, we see a clear and surprising trade-off: models that tend to have better fairness performance, which depends on task-specific performance, tend to have worse gender bias, which depends on model generations but not task-specific performance.
As to the two generative harms, we see there are some scenarios where they are fairly anti-correlated, which indicates the less gender biased models tend to be more toxic for these scenarios.
These findings indicate it may not suffice to only measure one of these types of harms, and that efforts to reduce one may have side-effects for the other \citep[\cf][]{xu2021detoxifying}, which is especially relevant given the harms of toxicity are more straightforward/immediate (\eg more likely to draw media/journalistic attention) whereas the harms of bias may be more subtle/gradual. 

Finally, for efficiency, we see fairly weak correlations with other metrics.
In particular, we do not see a strong overarching trade-off between accuracy and efficiency, which we may have anticipated \textit{a priori}.
This is made clear by \autoref{fig:acc-X}, which instead showcases that the trends for efficiency tend to be quite scenario-dependent.
This is especially true for the denoised metric, given a lot of what drives the measurements relates to the fundamental distributional properties of the scenario (\ie the length of inputs and outputs).

\paragraph{Direct model comparisons.}
A central objective for our evaluation is to achieve a common and unified understanding of all the models available.
In particular, the \nummodels models we evaluate in this work have been previously evaluated under different conditions for different evaluation suites.
In particular, the degree of overlap is surprisingly low and uneven across different models.
Consequently, we lack clarity as a field on the relationship between different models.

\begin{figure}
\centering
\includegraphics[width=\textwidth]{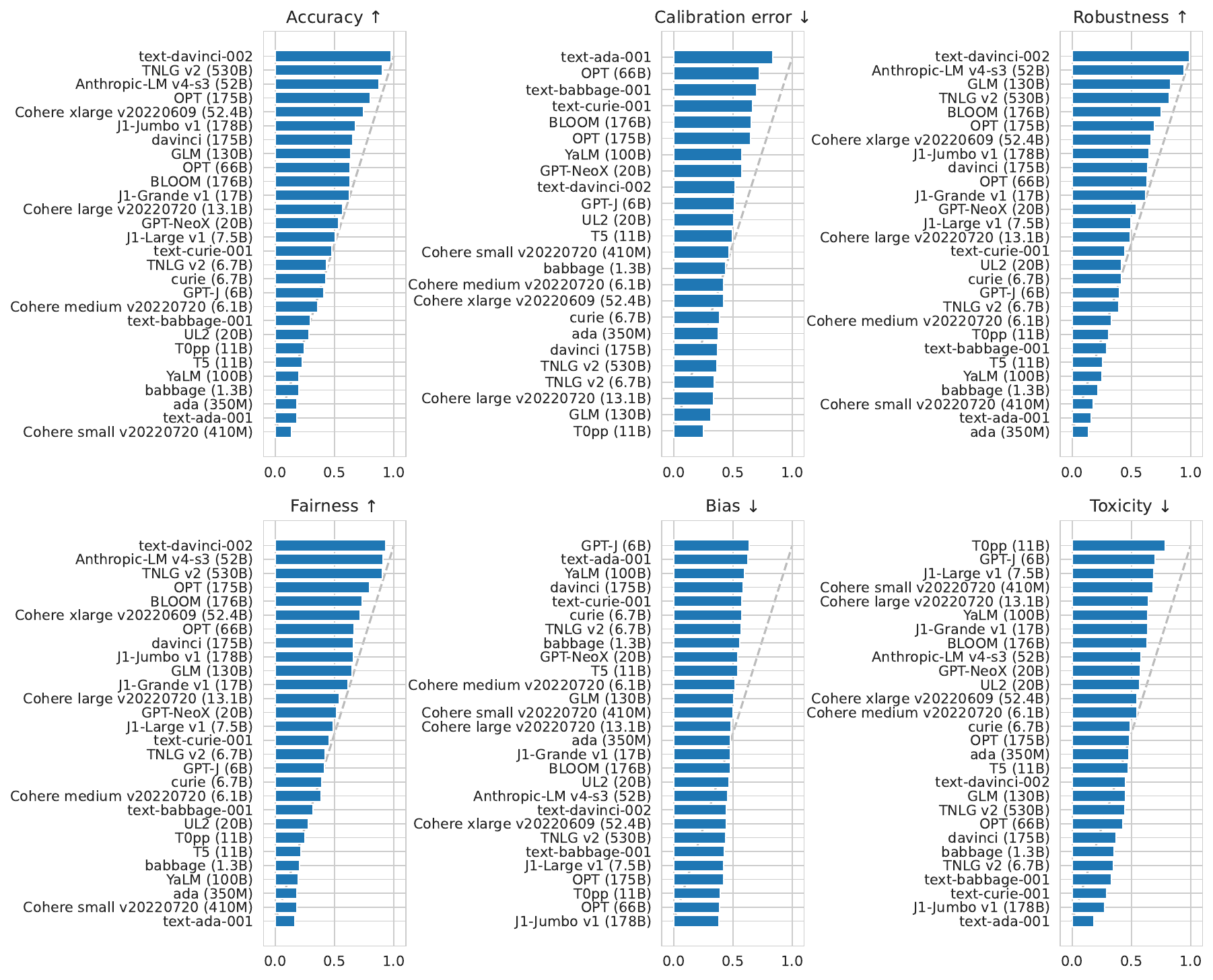}
  \caption{\textbf{Head-to-head win rate per each model.} 
  We report the fraction of head-to-head comparisons between the given model and all other models, across all scenarios, where the given model is higher along the metric (\eg more accurate in the accuracy subfigure).
  If a model was the highest for the given metric for every scenario, it would receive a score of 1.0; if a model received a score of 0.5, then if a scenario and second model were chosen at random, the outcome of the comparison would be a coin flip.
  For metrics where higher is better (Accuracy, Robustness, Fairness), the better models are towards the top, and vice versa.
  For calibration error, we measure ECE-10; for bias, we measure bias in gender representation; and for efficiency, we believe these comparisons should not be made without consideration of accuracy. 
  }
  \label{fig:head-to-head}
\end{figure}

In \autoref{fig:head-to-head}, we report how different models would fare in a head-to-head comparison for each metric across all the \core scenarios.\footnote{We exclude efficiency as we believe head-to-head comparisons based solely on efficiency are not meaningful; instead efficiency needs to be considered in conjunction with other metrics like accuracy.
}
We find that \instructdavinci is clearly the most accurate model, winning in these comparisons more than 90\% of the time.
Of the remaining models, we see that the largest model \mtnlgfivethreezero is the second most accurate, followed by \anthropiclm.
In light of the significant difference in model scale of a factor of more than 10 between \mtnlgfivethreezero and \anthropiclm, given \instructdavinci and \anthropiclm share instruction-tuning in common (which other models, beyond smaller OpenAI text variants, do not), this suggests that instruction-tuning and the use of human feedback is an efficient and effective  means for improving model accuracy.

Further, while we will later see the overall relationship between model scale and accuracy can be complicated (\autoref{fig:scale-accuracy}), here we see a clear thresholding effect in terms of model scale.
All of the models with a win rate for accuracy clearly above chance (\eg 55\% or more) have more than 50B parameters, whereas the sole exception of a large model outside this group of the top 10 is \yalm.
\yalm has a surprisingly poor accuracy win rate below 25\%, perhaps because of significant training on Russian instead of English). 
Within the top 10, we see model scale appears to be less predictive of rank, as two of the largest models (\bloom and \jurassicjumbo; both 100B+ parameters) are at the bottom of this tier, whereas \anthropiclm and \coherexl (the two smallest models in the tier) are in the top half of this tier.
Similarly, within a model family, we see model scale is perfectly monotonically correlated with model accuracy win rate (\eg see the OpenAI and Cohere model variants). 
We emphasize this certainly does not imply the models that fare poorly, or the small models, are inaccurate under all conditions: we expect that the prompting approaches we used are poorly suited for these smaller models and, perhaps, it is better to fine-tune these models \citep[\cf][]{liu2022fewshot}.

Consistent with previous findings of accuracy being correlated with robustness and fairness, we see similar rankings of models for those scenarios.
However, we note that \bloom notably performs considerably better (in a relative sense) for robustness/fairness than its accuracy would suggest, with \optonesevenfive and \glm roughly interchanging positions for robustness and accuracy (\ie \glm is more robust and less accurate in a relative sense compared to \optonesevenfive). 
In contrast, perhaps also as expected given the weak correlations of generative harms with other metrics, we see the rankings for toxicity and bias are notably quite different.
We highlight how \tzero and \gptdavinci in particular demonstrate the opposite trends when comparing their bias win rates to their toxicity win rates.
That is, \tzero is the most toxic when compared to all other models, but one of the three least (gender) biases, whereas \gptdavinci is one of the four most biased head-to-head, but one of the less toxic models.  

\begin{figure}
\centering
  \includegraphics[width=\textwidth]{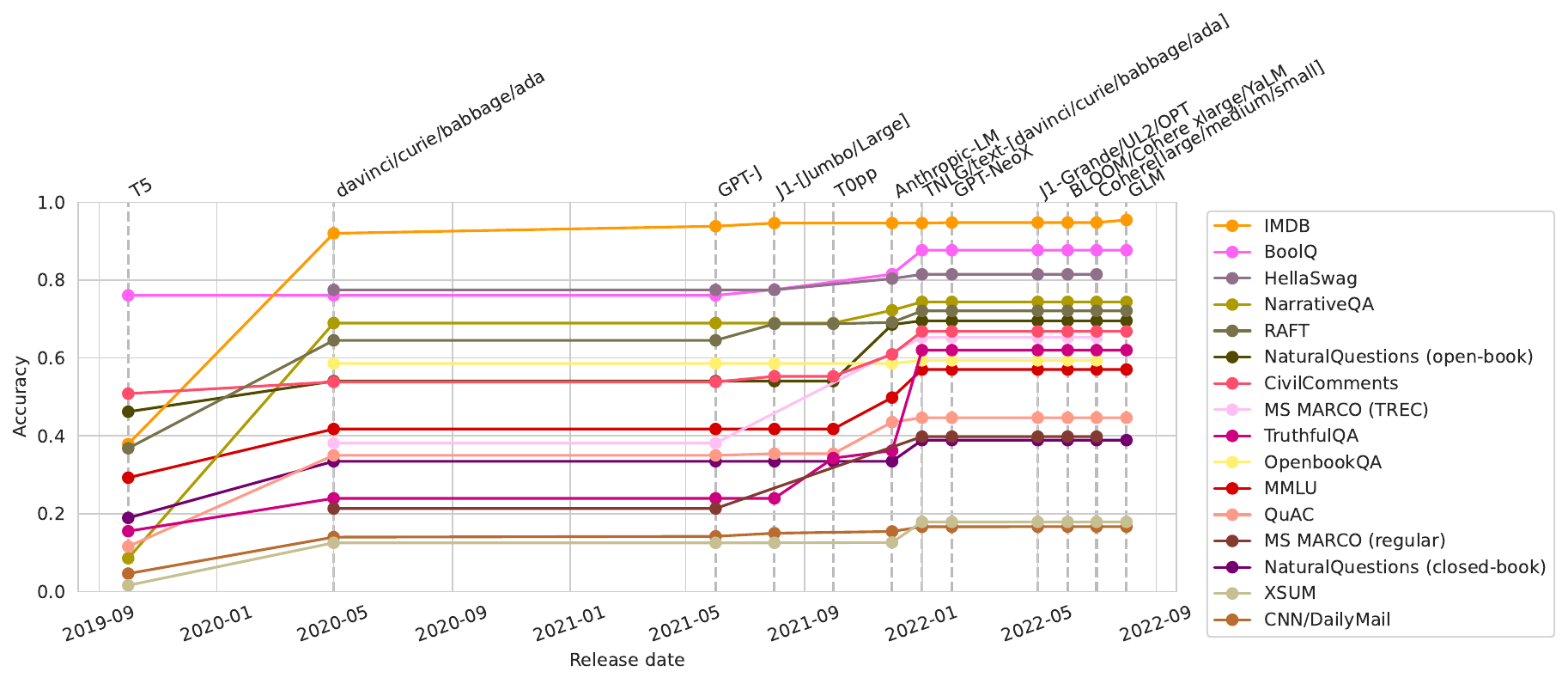}
  \caption{\textbf{Cumulative accuracy over time.} The relationship between time (x-axis) and the accuracy of the most accurate model released up to that point (y-axis) across \numcorescenarios \core scenarios. 
  That is, the graph tracks the progress in the state-of-the-art (SOTA) accuracy over time for each scenario.
  }
  \label{fig:accuracy-over-time}
\end{figure}

\begin{figure}
\centering
  \includegraphics[width=1.00\textwidth]{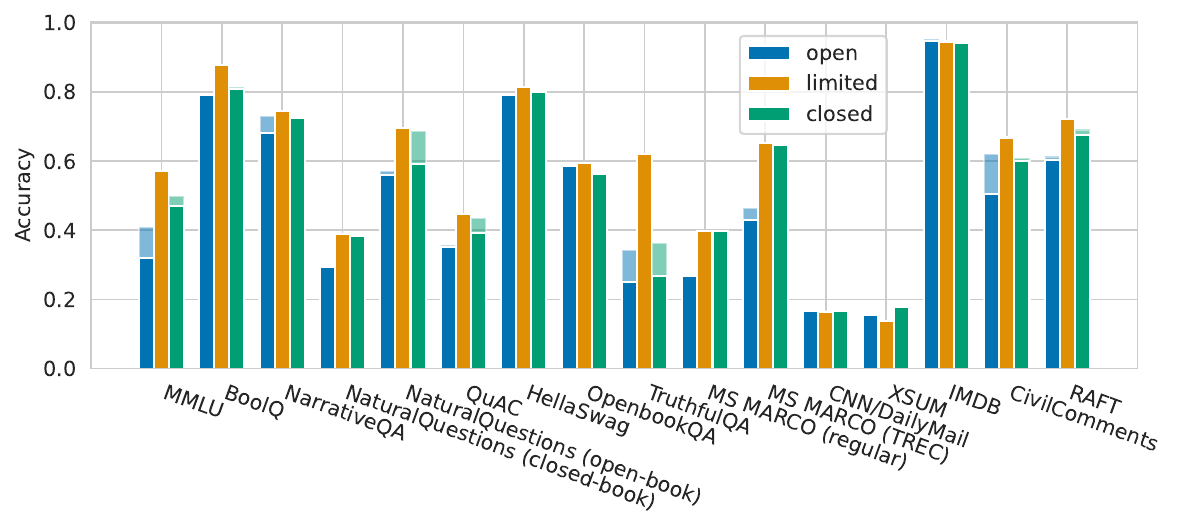}
  \caption{\textbf{Accuracy as a function of model access.} The relationship between access (\textbf{open} vs. \textbf{limited} vs. \textbf{closed}) and model accuracy for each of the \numcorescenarios \core scenarios. Shaded bars indicate the performance of the best model for that scenario, whereas the solid bars indicate the performance of the overall most accurate model across all \core scenarios based on \autoref{fig:head-to-head}.} 
  \label{fig:accuracy-vs-access}
\end{figure}

\paragraph{Accuracy as a function of other variables.}
To build on these model comparisons, we now consider trends in model accuracies as a function of other relevant variables. 
In \autoref{fig:accuracy-over-time}, we report model accuracies as a function of time (\ie model release date). 
As a function of time, we see that the release of GPT-3 \citep{brown2020gpt3} clearly establishes a strong baseline for future models across all scenarios, with a distinctive improvement over \tfive.
To the extent there is upward movement in accuracy since then, we often see it coming with the introduction of \anthropiclm roughly 18 months later in December, 2021 (the first model to use reinforcement learning with human feedback of those we evaluate), though we see a surprisingly clear monotonic improvement over the intermediary period for \naturalquestions. 

In \autoref{fig:accuracy-vs-access}, we stratify the results for accuracy based on model accessibility \citep[see][]{liang2022community-norms}.
For each access category, we report the per-scenario accuracy of the model in that category that is the most accurate (full bar) and that is generally the most accurate (smaller dark sub-bar; based on \autoref{fig:head-to-head}).
This means the sub-bar is always the accuracy of \optonesevenfive for \textbf{open} models, \instructdavinci for \textbf{limited access} models, and \mtnlgfivethreezero for \textbf{closed} models.
We see that these models for \textbf{limited} access are consistently the best in their categories, but \optonesevenfive is sometimes improved over (\eg 10 point improvement for \mmlu and \civilcomments) as is \mtnlgfivethreezero (\eg more than 5 point improvement for \naturalquestions (open-book) and \truthfulqa).

Overall, we see the accuracies generally fall in the direction \textbf{limited access} > \textbf{closed} > \textbf{open}, which aligns with \autoref{fig:head-to-head}. 
With that said, we see some reason for optimism regarding open science through open-sourced models, in that the best public model is usually somewhat competitive in accuracy (\ie within 5 points of the best non-open model), with the notable exceptions of more knowledge-intensive scenarios of \mmlu, \naturalquestions (closed-book) and \truthfulqa as well as both information retrieval scenarios. 
This is especially noteworthy given we have yet to see models being open-sourced with significant use of human preferences and reinforcement learning from human feedback, which may further bridge this gap.
However, we do highlight there are other confounds: some models may not be disclosed to the public (let alone released), some models may be disclosed after significant delay, and some highly accurate models are not yet evaluated in \benchmarkname (\eg Gopher, Chinchilla, PaLM).
We hope our findings on the relationship between model accessibility and model performance (\eg accuracy) will inform the development of community norms and standards around model access/release \citep{liang2022community-norms}. 

\begin{figure}
\centering
  \includegraphics[width=\textwidth]{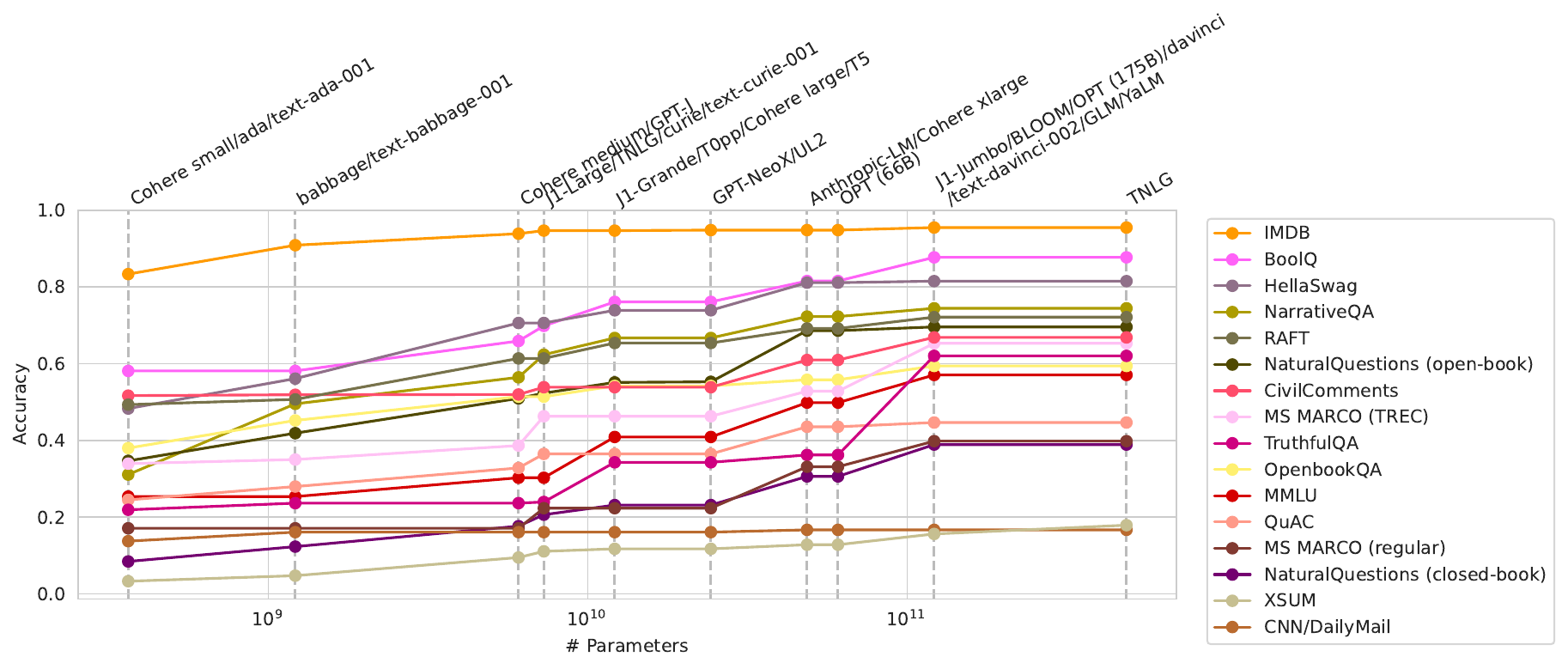}
  \caption{\textbf{Model size vs. accuracy.} 
    The relationship between model parameter size (x-axis) and the accuracy of the most accurate model released up to that scale on each \core scenario.
    That is, the graph tracks the progress in the state-of-the-art (SOTA) accuracy as a function of scale for each scenario.
  }
  \label{fig:scale-accuracy}
\end{figure}

\begin{figure}
\centering
  \includegraphics[width=\textwidth]{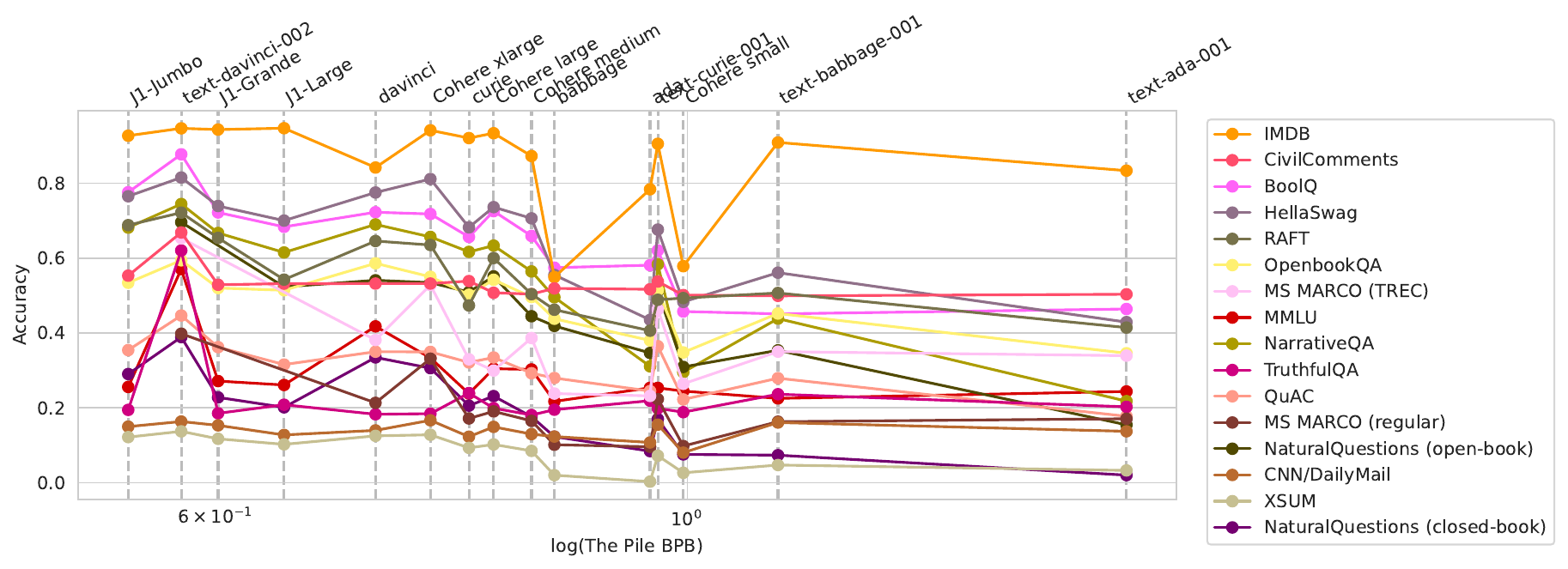}
  \caption{\textbf{\pile loss vs. accuracy.} The relationship between log bits-per-byte (BPB) on \pile and the accuracy on each \core scenario.}
  \label{fig:perplexity-accuracy}
\end{figure}

\paragraph{Predicting accuracy.}
Given the resource-intensive nature of producing language models, especially the most capable models (for which we take accuracy for different scenarios as a proxy), the ability to predict model capabilities\footnote{It would be similarly interesting to predict our properties (\eg those related to harms), though here we focus on accuracy. We do note that our evaluation deliberately surfaces all the necessary metrics required to project other properties.} is valuable for orienting decision-making.
Prior work demonstrates model scale (whether measured in model size, data size, or compute/training FLOPs) reliably predicts model loss for language models, establishing \textit{scaling laws} \citep{kaplan2020scaling, hoffmann2022chinchilla}.
On the other hand, work has also shown that certain qualitative capabilities \citep[\eg in-context learning;][]{brown2020gpt3} \textit{emerges} unpredictably beyond a critical scale threshold \citep[see][]{wei2022emergent}.
And many works have implicitly assumed and demonstrated that improvements in language modeling loss (\ie perplexity) translate to improvements on downstream use cases. 
In \autoref{fig:scale-accuracy}, we report on the relationship between model scale and accuracy, while in \autoref{fig:perplexity-accuracy}, we report perplexity (measured on \pile) vs. accuracy.

In light of the very consistent picture that model scale within a family led to reliable improvements in accuracy (\autoref{fig:head-to-head}), we see that this does not generalize across model families.
If we are to plot model scale vs. accuracy, we find the relationship is very chaotic; model scale (agnostic to model family) as measured in parameters is a poor predictor of trends in accuracy.
Unfortunately, the data on different models was too sparse to attempt a similar analysis in terms of training FLOPs, but we expect such an analysis may be more helpful in demonstrating a crisper relationship between training compute and downstream accuracy.
When we accumulate the performance of models in \autoref{fig:scale-accuracy}, so as to understand if scale appears more necessary to improve accuracy, we do see some notable jumps for many scenarios both in the range of 12B parameters and 50B parameters.
These jumps are specifically attributable to \tzero and \anthropiclm, which likely instead indicates that attributing these jumps to scale is confounded by other changes (\ie the training procedure of \tzero to deliberately improve few-shot prompting performance and the RLHF involved in \anthropiclm).
And we similarly see the relationship between perplexity (or, more accurately, log bits-per-byte) on \pile and downstream accuracy is similarly messy (\autoref{fig:perplexity-accuracy}), though we expect the confound of contamination (some models are trained on \pile and others are not) exacerbates the messiness of these findings.

\subsection{Prompting analysis}
While the benchmark we design is general, we evaluate \nummodels models by adapting them through few-shot prompting.
Consequently, here we study variations across several of these models with respect to different design decisions involved in prompting.\footnote{We do not conduct all of these analyses on every model due to cost. We exclude all commercial models due to monetary cost, as well as TNLGv2 models due to long wait times given the scale of the model.}
Together, these analyses help clarify the potential brittleness of model behavior to prompt design, which bears on the legitimacy and reliability of using prompting to build practical systems.

\paragraph{Choice of in-context examples.}
Given that we adapt language models through few-shot prompting, the selection of the specific in-context examples could significantly influence model performance.
In particular, unlike prior work \citep[\eg][]{brown2020gpt3}, we use the same examples across all evaluation instances, rather than selecting different in-context examples for each evaluation instance, as this better reflects the realities of few-shot adaptation \citep{perez2021true}.
(We describe our specific selection algorithm which, to first-approximation, is random selection from the scenario's training set while ensuring coverage of different classes for classification problems.)
As a consequence, this exacerbates the variation observed for different choices of in-context examples (consistent with the high variability one might witness in practice if they truly only have a few examples). 

\begin{figure}
\centering
\includegraphics[width=\textwidth]{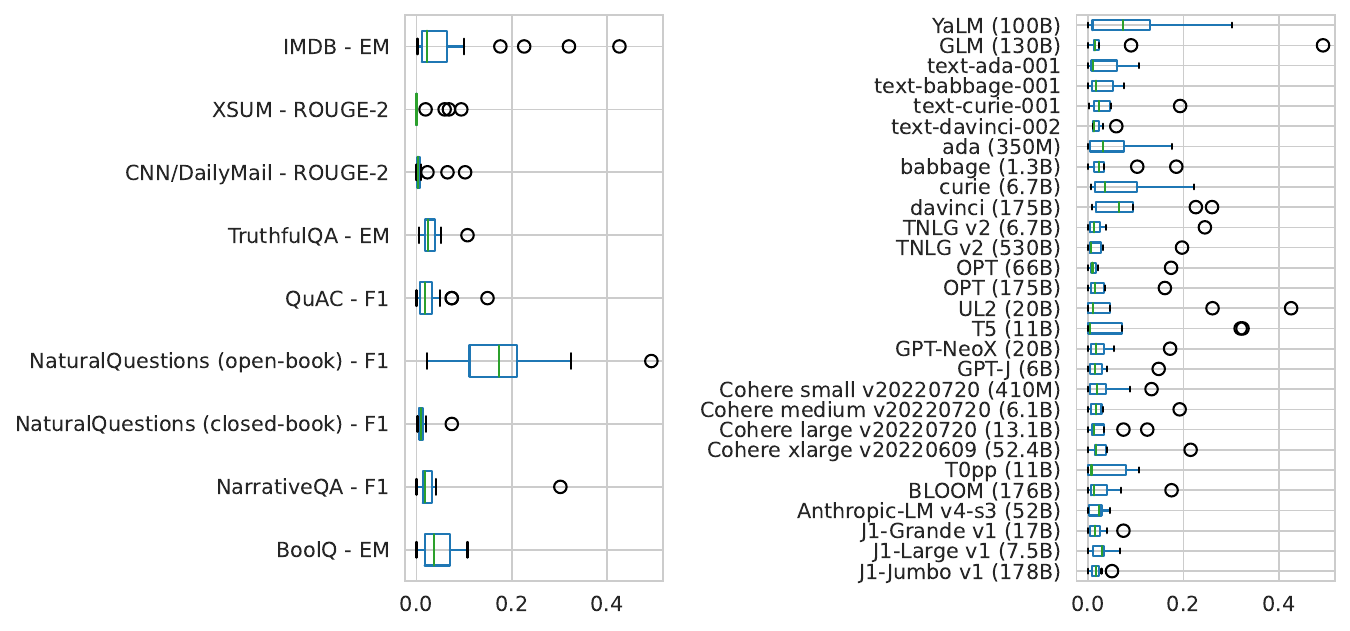}
  \caption{\textbf{Variance across seeds.} 
    For a subset of models and scenarios, we evaluate each scenario with three different random sets of in-context examples.
    We compute the range of the accuracy metric (maximum minus minimum value over the three random seeds) and visualize across models and scenarios.
  }
  \label{fig:seed-variance}
\end{figure}

To measure this sensitivity, we repeat all evaluations for \numruns runs, meaning \numruns choices of the in-context examples.
To explore the sensitivity of model results to this choice, hover over a specific entry to see the variance across runs on the website.
Further, the predictions across the \numruns runs can be found for a given evaluation instance.
Additionally, we visualize the accuracy range (best minus worst performance) across seeds in \autoref{fig:seed-variance}.
We observe that the performance of most models tends to be relatively consistent: the median range (across scenarios) is below 0.03 for 24 out of 28 models (exceptions are \yalm, \gptdavinci, \gptcurie, \gptada).
Interestingly, we find that there are scenarios where all models tend to exhibit higher variance.
A prominent example is \naturalquestions (open-book), where the median range is 0.1730 (e.g., \gptdavinci obtains F1 scores of 0.376, 0.611, and 0.636 across the three sets of in-context examples).

\begin{figure}
\centering
\includegraphics[width=\textwidth]{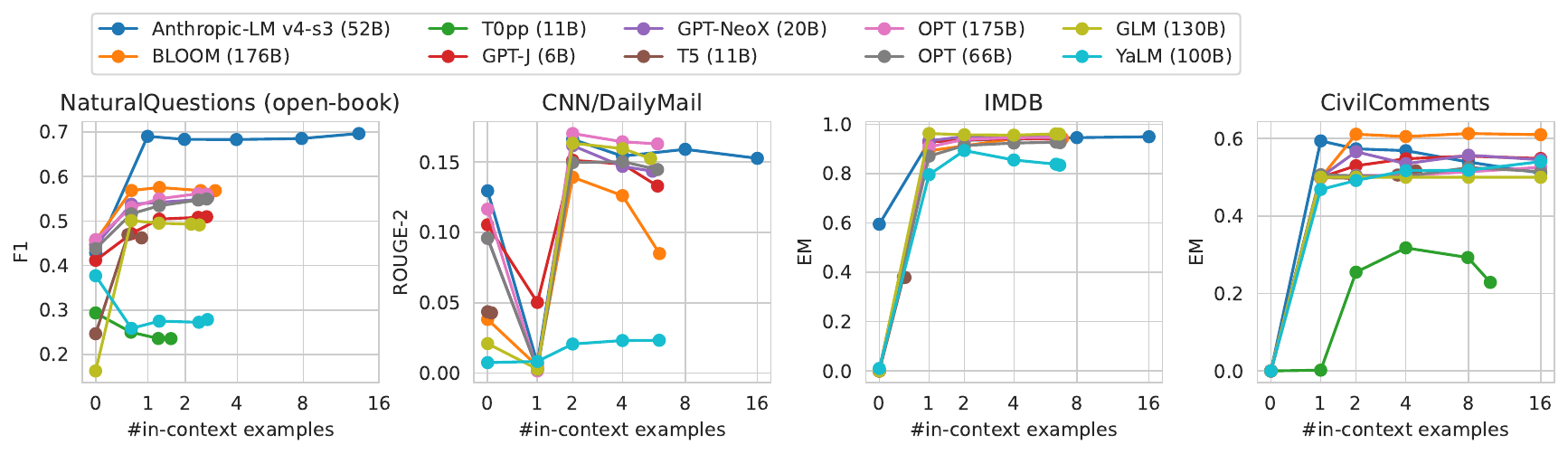}
  \caption{\textbf{Number of in-context examples.} 
    For each model, we set the maximum number of in-context examples to [0, 1, 2, 4, 8, 16] and fit as many in-context examples as possible within the context window.
    We plot performance as a function of the average number of in-context examples actually used.
  }
  \label{fig:in-context-ablations}
\end{figure}

\paragraph{Number of in-context examples.}
By default, we either use \numincontext in-context examples, or fewer examples for scenarios where \numincontext examples do not fit within the context window. 
To test how the number of examples (\ie the sample efficiency of adaptation) influences performance, we vary the maximum number of examples across $n \in \{0, 1, 2, 4, 8, 16\}$.
In \autoref{fig:in-context-ablations}, we plot model performance as a fraction of the average number of in-context examples provided (which may be fewer than the maximum stated above if they do not fit inside the context window).

We find that all models show clear improvement from $n = 0$ to $n = 1$, sometimes having 0\% accuracy in the zero-shot setting, with the consistent exception of \cnndm where zero-shot accuracy is better for almost all models.
We posit
that models may not effectively understand the appropriate length distribution and the poor reference summaries may comparatively mislead the model in the one-shot setting compared to the zero-shot setting.
However, for larger numbers of in-context examples, we do not see consistent benefits across all models and all scenarios.
The sole exception is \optonesevenfive which, besides \cnndm, shows a perfectly monotonically increasing relationship between number of shots and model accuracy for \naturalquestions (open-book), \imdb, and \civilcomments.

\paragraph{Formatting of prompt.}
Beyond the in-context examples and the evaluation instance, there are several other details required to fully specify a prompt (\eg instructions that describe what the model should do).
Since this formatting exists in the space of natural language, it is difficult to specify concrete axes to systematically vary across (\ie in contrast to how we can specify a range we consider for the number of in-context examples).
Consequently, we consider the following motivated but fairly \textit{ad hoc}/arbitrary changes to the prompt format involving instructions, input prefixes, output prefixes, and input suffixes.

\begin{figure}
\centering
\includegraphics[width=\textwidth]{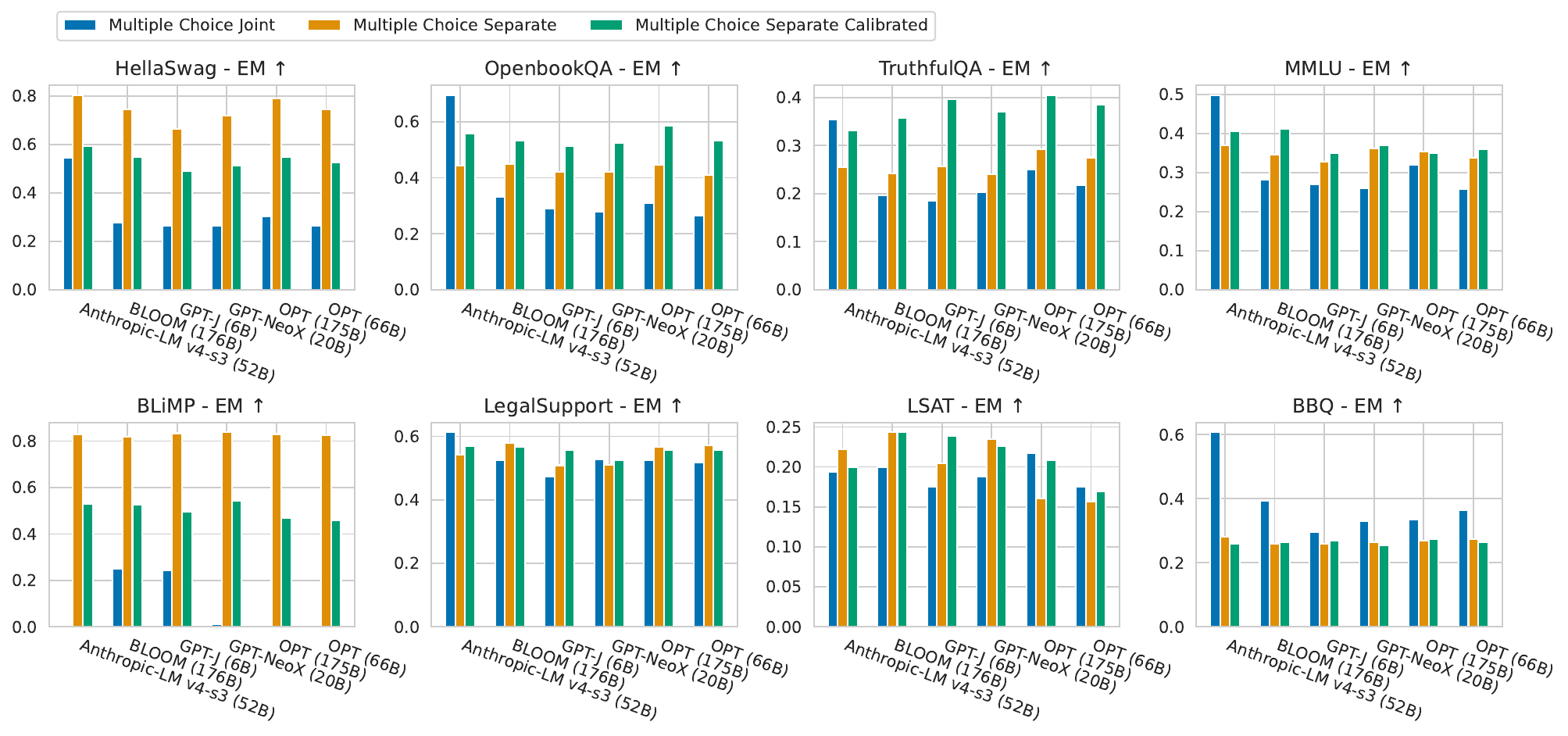}
  \caption{\textbf{Multiple-choice adaptation.} 
    For each adaptation method (joint, separate, and separate calibrated), we compare models across scenarios.
  }
  \label{fig:mc-ablations}
\end{figure}

The clear finding is that the best prompt formatting is not consistent across models (\ie models can stand to improve in their interoperability).
In particular, one variant leads to an accuracy of 67.3\% for \anthropiclm on \naturalquestions (open-book), whereas the prompt performs very poorly for \bloom, which drops from an accuracy around 60\% to 8.5\%. 
In some cases, we believe the prompting changes may lead to poor/undesirable interactions with the tokenizer, given the models use different tokenizers in several cases. 
For \glm, we are intrigued to see the prompt involving mentioning the model is an expert AI assistant performs best across all four scenarios (\naturalquestions (open-book), \cnndm, \imdb, and \civilcomments) we look at in terms of accuracy.
Specifically, the prompt includes \texttt{instructions} of  ``I am an expert AI assistant who is here to help you with the following.'', along with an \texttt{input\_prefix} of ``Passage: '',  \texttt{input\_suffix} of `` "'', and an \texttt{output\_prefix} of ``Answer:''

\paragraph{Formulation of multiple choice scenarios.}
Beyond the details of the prompt, we can conceptually imagine different ways to make use of the language interface to perform the same underlying scenario.
In the case of multiple choice scenarios, we could provide the model with all of the answer choices and task it with selecting the correct choice (\textit{joint}).
Or we could provide the model with each answer choice as a separate query of the model, and see which answer choice is assigned higher probability (\textit{separate}).
And we could further calibrate the model-assigned probabilities by the probability the model assigns to the corresponding answer choice when presented standalone (\textit{separate-calibrated}).

The results for these multiple choice scenarios are visualized in \autoref{fig:mc-ablations}.
We observe that the method that maximizes accuracy is largely determined by the scenario, whereas it is generally consistent across models for a given scenario.
Taking \hellaswag as a case study, for all six models, the methods follow \textit{separate} $>$ \textit{separate-calibrated} $>$ \textit{joint}.
This choice can be very consequential for the model accuracy: for \optonesevenfive, the accuracies are 79.1\%, 54.8\%, and 30.2\% respectively. 
Further, the trends are entirely inverted for calibration error, as may be expected.
So, overall for \hellaswag, across both desiderata and for all six models, there is clear and strict Pareto dominance in adaptation methods of \textit{separate} $>$ \textit{separate-calibrated} $>$ \textit{joint}.
In the case of \hellaswag, given the dataset is designed as completions of an incomplete textual sequence, the preference for the \textit{separate} adaption method over the \textit{joint} adaptation method is fairly intuitive and precisely aligns with the empirical findings. 
However, we do note the methods pattern differently for specific models in terms of efficiency.
In particular, the \textit{separate} methods require a query for each answer choice, whereas the \textit{joint} method requires a single, much longer (especially due to the inclusion of in-context examples) query.
This manifests in lower denoised inference times for \textit{separate} over \textit{joint} for \bloom (0.075s vs. 0.271s), but the opposite trend for \optonesevenfive (0.71s vs. 0.187s), despite the fact the models are evaluated on the same infrastructure and are near-identical in overall parameter count.

For the other scenarios, we do not observe model-agnostic preferences over the adaptation methods.
In general, for accuracy, we find \textit{separate-calibrated} $>$ \textit{separate} $>$ \textit{joint} for \openbookqa, \truthfulqa, and \mmlu for five of the six models.
The consistent exception is \anthropiclm where the preference is consistently \textit{joint} $>$ \textit{separate-calibrated} $>$ \textit{separate}.
We find this to be quite striking: the adaptation method that elicits the most accurate behavior from \anthropiclm elicits the least accurate behavior from the other five models.
These gaps are significant as well: on \openbookqa, presenting accuracies in the order (\textit{separate-calibrated}, \textit{separate}, \textit{joint}), we see (58.6\%, 44.6\%, 30.9\%) for \optonesevenfive, whereas for \anthropiclm we see (55.8\%, 44.4\%, 69.6\%).
That is, if \optonesevenfive and \anthropiclm were compared using the \textit{separate-calibrated} adaptation method (or the \textit{separate} method), they would achieve near-equal accuracies (within 3\%), but they are almost 40\% apart when using the \textit{joint} method, and 12\% apart when comparing the per-model accuracy-maximizing method.
This raises fundamental questions about the appropriate way to compare across models and the sense in which uniform standardization is fair; we encourage future work to more extensively look into this and develop clearer best practices.

\subsection{Task-level analysis}

\begin{figure}[!ht]
    \centering
    \includegraphics[width=\textwidth]{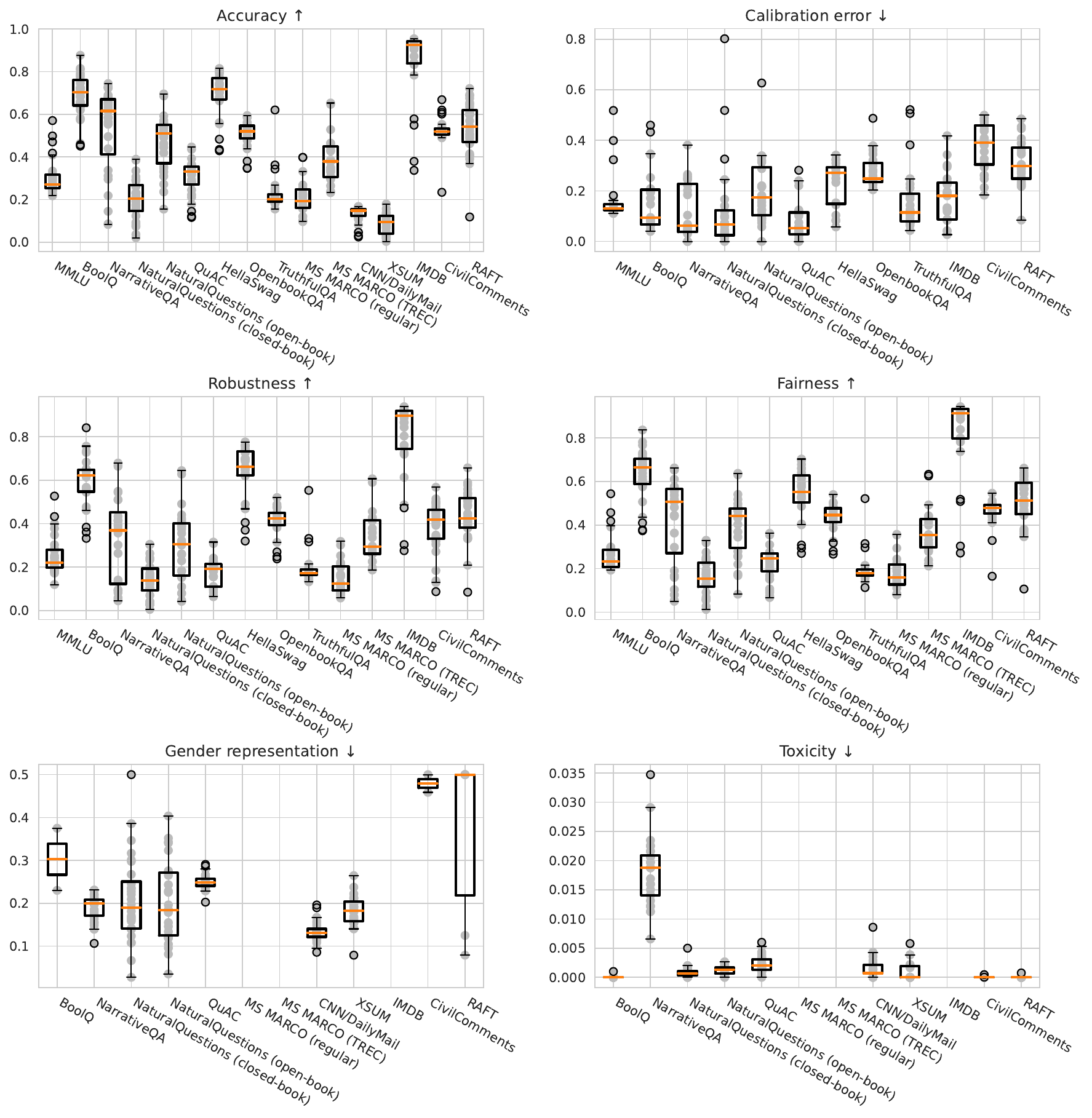}
    \caption{\textbf{Metric spread for \core scenarios.}
    Metrics for every model on every \core scenario as a means for indicating the spread on a per-metric basis.}
    \label{fig:metric-spread}
\end{figure}

Since we organize the \numcorescenarios \core scenarios by the broader task, we highlight findings at the task level.
To provide a sense of the spread in accuracies for each of these scenarios, we provide \autoref{fig:metric-spread}.
The results we provide here are highlighting specific trends/phenomena, though we encourage interactively looking at the quantitative results and underlying model behaviors.

\paragraph{Question answering.}
\begin{table*}[htp]
\centering
\resizebox{\textwidth}{!}{
\begin{tabular}{lrrrrrrrrr}
\toprule
Model & MMLU (EM) & BoolQ (EM) & NarrativeQA (F1) & NaturalQ (closed-book) (F1) & NaturalQ (open-book) (F1) & QuAC (F1) & HellaSwag (EM) & OpenbookQA (EM) & TruthfulQA (EM) \\
\midrule
Jurassic Jumbo (178B) & 0.256 & 0.791 & 0.688 & 0.288 &  & 0.365 & 0.765 & 0.534 & 0.177 \\
Jurassic Grande (17B) & 0.276 & 0.727 & 0.691 & 0.226 &  & 0.372 & 0.739 & 0.52 & 0.173 \\
Jurassic Large (7.5B) & 0.265 & 0.712 & 0.634 & 0.197 &  & 0.304 & 0.7 & 0.514 & 0.19 \\
Anthropic-LM (52B) & 0.502 & 0.829 & 0.718 & 0.28 & 0.672 & 0.449 &  &  & 0.375 \\
BLOOM (176B) & 0.281 & 0.745 & 0.671 & 0.227 & 0.456 & 0.358 & 0.744 & 0.532 & 0.199 \\
T0++ (11B) & 0.413 & 0.592 & 0.355 & 0.096 & 0.604 & 0.154 &  &  & 0.336 \\
Cohere xlarge & 0.341 & 0.784 & 0.647 & 0.299 & 0.394 & 0.368 & 0.81 & 0.55 & 0.196 \\
Cohere large & 0.309 & 0.744 & 0.641 & 0.227 & 0.472 & 0.343 & 0.736 & 0.542 & 0.211 \\
Cohere medium & 0.307 & 0.643 & 0.578 & 0.176 & 0.331 & 0.284 & 0.706 & 0.496 & 0.185 \\
Cohere small & 0.247 & 0.481 & 0.31 & 0.074 & 0.234 & 0.213 & 0.483 & 0.348 & 0.185 \\
GPT-J (6B) & 0.267 & 0.544 & 0.547 & 0.165 & 0.412 & 0.319 & 0.663 & 0.512 & 0.191 \\
GPT-NeoX (20B) & 0.273 & 0.72 & 0.605 & 0.2 & 0.446 & 0.342 & 0.718 & 0.53 & 0.193 \\
T5 (11B) & 0.295 & 0.831 & 0.086 & 0.189 & 0.29 & 0.119 &  &  & 0.179 \\
UL2 (20B) & 0.241 & 0.806 & 0.092 & 0.17 & 0.191 & 0.154 &  &  & 0.199 \\
OPT (66B) & 0.258 & 0.757 & 0.652 & 0.26 & 0.438 & 0.35 & 0.745 & 0.534 & 0.219 \\
OPT (175B) & 0.32 & 0.816 & 0.671 & 0.291 & 0.458 & 0.365 & 0.791 & 0.586 & 0.248 \\
GPT-3 (175B) & 0.42 & 0.734 & 0.697 & 0.334 & 0.376 & 0.355 & 0.775 & 0.586 & 0.2 \\
GPT-3 (6.7B) & 0.242 & 0.642 & 0.594 & 0.207 & 0.376 & 0.319 & 0.682 & 0.502 & 0.237 \\
GPT-3 (1.3B) & 0.219 & 0.502 & 0.516 & 0.117 & 0.307 & 0.273 & 0.555 & 0.438 & 0.187 \\
GPT-3 (350M) & 0.256 & 0.61 & 0.333 & 0.086 & 0.304 & 0.226 & 0.435 & 0.38 & 0.22 \\
Davinci Instruct v2 & 0.574 & 0.863 & 0.748 & 0.388 & 0.68 & 0.445 & 0.815 & 0.594 & 0.633 \\
Curie Instruct & 0.259 & 0.616 & 0.569 & 0.169 & 0.403 & 0.376 & 0.676 & 0.514 & 0.202 \\
Babbage Instruct & 0.229 & 0.394 & 0.429 & 0.076 & 0.33 & 0.274 & 0.561 & 0.452 & 0.203 \\
Ada Instruct & 0.25 & 0.509 & 0.218 & 0.016 & 0.205 & 0.177 & 0.429 & 0.346 & 0.196 \\
GLM (130B) & 0.321 & 0.839 & 0.739 & 0.153 & 0.163 & 0.292 &  &  & 0.209 \\
YaLM (100B) & 0.267 & 0.648 & 0.536 & 0.14 & 0.375 & 0.307 &  &  & 0.229 \\
\bottomrule
\end{tabular}}
\caption{\textbf{Question answering accuracy.} These accuracy of all 30 models assessed for question answering in \citet{liang2022helm}.}
\label{fig:accuracy (question_answering)}
\end{table*}
For the 9 question answering scenarios, we see significant heterogeneity across the results.
However, in terms of accuracy, what is very consistent is that \instructdavinci is the most accurate model for all 9 scenarios.
The margin however varies greatly across different scenarios: the largest margin is on \truthfulqa where \instructdavinci achieves accuracy of 62.0\% compared to second place from \anthropiclm at 35.4\%, whereas the smallest margins are for \naturalquestions (closed-book) (38.9\% for \instructdavinci vs. 38.5\% for \mtnlgfivethreezero) and \hellaswag (81.5\% for \instructdavinci vs. 81.1\% for \coherexl). 
As this suggests, the most accurate models across the question answering beyond \instructdavinci is much more variable: all of \anthropiclm, \tzero, \coherexl, \optonesevenfive, \mtnlgfivethreezero, \gptdavinci, and \glm are a top-3 model in terms of accuracy for some QA scenario, though for 5 scenarios \anthropiclm is the second-most accurate and for 6 \mtnlgfivethreezero is the third-most accurate. 

For calibration, we see several instances of models being very poorly calibrated. 
For example, on \hellaswag, which is one of the scenarios where models have the highest absolute accuracies, the two most accurate models (\instructdavinci, \coherexl) both have calibration errors of at least 0.28 (0.286, 0.341 respectively). 
In fact, we tend to see the J1 models are poorly calibrated on these scenarios, with all 3 models incurring an ECE-10 of roughly 0.5 or more for both \naturalquestions variants, \quac, \narrativeqa, and \truthfulqa. 
In contrast, some models are quite calibrated for some scenarios (\eg \coherexl has a calibration error of 0.06 on \narrativeqa). 

\begin{figure}
\centering
\includegraphics[width=\textwidth]{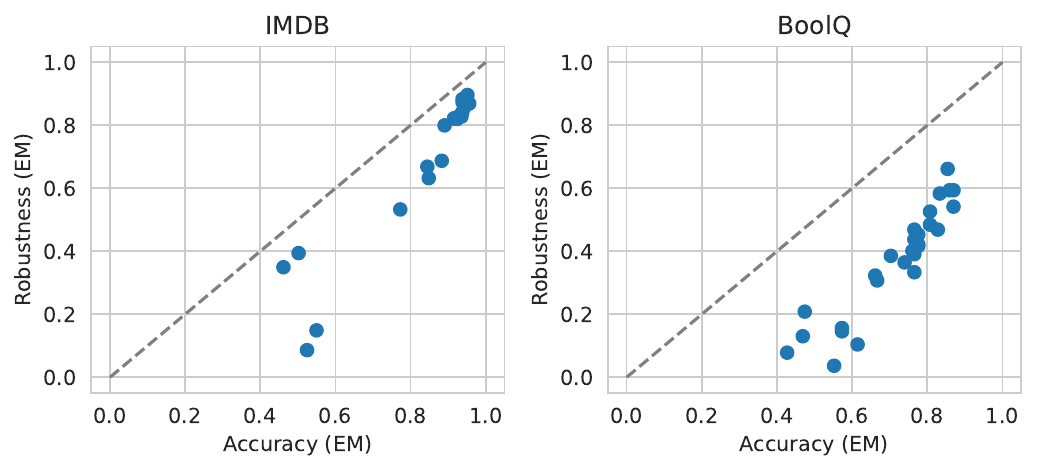}
  \caption{\textbf{Robustness--equivariance via contrast sets.} For the two scenarios where we have access to hand-crafted contrast sets, for each model, we plot the robustness of the model on that scenario (worst-case performance across perturbations of each instance) as a function of its standard accuracy. 
  }
  \label{fig:contrast-sets}
\end{figure}

For robustness and fairness, we see all models tend to show consistent drops of 5 to 10 points for all scenarios.
Consistent with the broader trends, we find robustness and fairness are strongly correlated with accuracy, with no observed cases of the most accurate models suffering especially large drops for robustness/fairness.
One exception is \hellaswag, where the three most accurate models are the only models with standard accuracies above 80\% (\instructdavinci = 81.5\%, \coherexl = 81.1\%, \anthropiclm = 80.4\%), and only \instructdavinci remains about 70\% in the presence of fairness perturbations  at 70.3\%.
\coherexl in particular experiences the largest drop of more than 15 points to 66\%, which is a worse robust accuracy than \mtnlgfivethreezero at 67.8\% despite \mtnlgfivethreezero being 1.2\% worse in standard accuracy.
In other words, the model with higher standard accuracy has lower accuracy in the presence of fairness perturbations, with a 3\% larger drop due these perturbations.
In terms of robustness to semantics-altering perturbations (equivariance), we only have access to Contrast Sets~\citep{gardner2020contrast} for the \boolq scenario.
In \autoref{fig:contrast-sets}, we find that the performance of all models drops significantly when evaluated for equivariance.
Interestingly, there is a cluster of models with moderate accuracy, but really low robustness: e.g.,\ \gptj achieves an accuracy of 55\% but a robustness of just 3.6\%.

For generative harms, we find the toxicity rates are quite low across all QA scenarios (\eg less than 1\% for all models on both \naturalquestions variant as well as \quac). 
The one exception is \narrativeqa, where several models have toxicity rates between 2--3\%, which may be attributable to the story domain (and/or false positives from the Perspective API being more common for this domain). 
Looking at biases for \narrativeqa, we see models generally show similar biases for all four of \{gender, race\} $\times$ \{demographic representation, stereotypical associations\}. 
The one exception is for racial representation, where all but three models achieve bias scores of 0.667 (lower is better/less biased), whereas \ultwo and \tfive have scores around 0.35 and \yalm around 0.45. 
Looking at the model generations, we do see some evidence for discrepancies in model generations, though we also expect the effects referenced are still somewhat weak given models infrequently mention racial words that we track for this scenario.
Beyond \narrativeqa, models tend to demonstrate similar biases for a given scenario.

\paragraph{Information retrieval.}
Unless otherwise stated, our results report the scores from the \textit{boosted} setting, which aim in principle to establish an intuitive upper-bound on model quality. Whereas in the \textit{vanilla} settings models re-rank the top-30 passages from BM25, in the boosted setting models re-rank the top-30 passages from BM25 as well as every passage that has an explicit relevance assessment, even when these passages aren't retrieved by BM25.

For information retrieval on \msmarcoregular and \msmarcotrec, we begin by observing that the best-performing models show competitive accuracy scores, especially under the boosted setting. The most effective models on \msmarcoregular are
\instructdavinci, which scores 39.8\% RR@10 in the boosted setting and 22.4\% RR@10 in the vanilla BM25 top-30 setting, and \mtnlgfivethreezero, which scores 39.7\% RR@10 (boosted) setting and 22.5\% RR@10 (vanilla), with \coherexl being the next most accurate model at 33.1\% RR@10 (boosted) and 17.6\% (vanilla). To put these scores in perspective, the RR@10 of our underlying classical retriever (i.e., BM25) is 19.0\%. That is, in the vanilla top-30 setting, both \instructdavinci and \mtnlgfivethreezero obtain a better ranking than the underlying retriever on average, whereas other models like \coherexl perform worse. 

We see a similar trend but with larger gains on the more densely-annotated \msmarcotrec. In particular, \instructdavinci is the most accurate model at 65.3\% NDCG@10 (boosted) and 61.0\% NDCG@10 (vanilla), then \mtnlgfivethreezero at 64.7\% (boosted) and 60.7\% (vanilla), and \coherexl at 52.8 (boosted) and 53.4\% (vanilla). On \msmarcotrec, BM25's NDCG@10 is 50.6\%. Given the much more dense annotations in the \msmarcotrec setting, we see that these models deliver considerably larger gains over BM25, under both the boosted and the vanilla setting, compared with their gains on \msmarcoregular. For \msmarcotrec, the boosted setting includes all of the assessed passages, including many relevant and many non-relevant passages. Naturally, the gains for the best-performing models are larger in the boosted setting. Perhaps surprisingly, for \coherexl, quality is diminished in the boosted setting: effectively, this model is distracted by the non-relevant passages that are in the assessed pool more than it benefits from the relevant documents that are not in BM25's top-30 results.

Overall, these few-shot effectiveness scores are relatively competitive but they nonetheless trail the current state-of-the-art retrieval systems by a considerable margin. At the time of writing, in \msmarcoregular, the top-scoring system on the public leaderboard scores 45\% RR@10 on the test set. Compared to this, the models we test score up to 39.8\% RR@10 in the boosted setting, which is comparable to recent state-of-the-art standalone retrievers like ColBERTv2~\citep{santhanam2021colbertv2}, and 22.5\% in the vanilla setting that re-ranks BM25's top-30, which is comparable to simple neural ranker submissions from late 2018 and early 2019 like Conv-KNRM~\citep{dai2018convolutional} and DuetV2~\citep{mitra2019updated}. In \msmarcotrec, the best-scoring system that participated in the original NIST competition scored 76.5\% NDCG@10. Compared to this, the models we test score up to 65.3\% NDCG@10 in the boosted setting, which is within 2--3 points of the ANCE~\citep{xiong2020approximate} system from mid 2020, and 61.0\% in the vanilla setting, which is more effective than the Siamese BERT embeddings baseline of \citet{gao2021complement}.

These promising accuracy scores for few-shot adaptations of language models are promising, especially since it is not inherently obvious that such models are well-suited for the nuances of these passage ranking scenarios. In particular, these scenarios (at least as realized through our adaptations) necessitate a strong degree of calibration, in which the probabilities assigned to each of the ``Yes''/``No'' outputs accurately reflects the continuous degree of relevance of a passage to a query. Out of the MS MARCO corpus with 9M passages, numerous passages can carry varying degrees of relatedness to a given query~\citep{craswell2020overview}.

Beyond accuracy, we hone in on the robustness and fairness of the systems we study. Robustness and fairness have been considered in many ranking problems \citep[\eg][]{zehlike2017fair, singh2019policy}, but such concerns have only become mainstream in the past few years.\footnote{For example, see the TREC Fair Ranking track: \url{https://fair-trec.github.io/}.}
We find that \instructdavinci only presents minor drops in accuracy in the presence of both robustness and fairness perturbations for \msmarcotrec, but otherwise most models show some drops, though no model shows especially large drops. To highlight an interesting case, \coherexl shows the largest drops for \msmarcoregular, going from 33.1 standard RR@10 to 29.4 fairness RR@10 and 25.3 RR@10 in the presence of robustness perturbations.
We highlight the potential for future work to further explore cross-lingual retrieval settings for both language and dialects, in that we currently perturb both the passages and the query for our dialect perturbations but it may be more realistic to assume the passages are, for example, in SAE but the queries are in AAE to understand performance for AAE speakers. 

Finally, in spite of the promising results in terms of model accuracy, we note that scale is critical to many information retrieval applications, necessitating models be efficient.
For example, to deploy information retrieval systems on the web in search engines, systems must perform very efficient inference to be useful, which is why amortized indexing strategies have been widely studied and adopted. 
In contrast, we see these language models when adapted via prompting are drastically less efficient, perhaps also because of the large model sizes (\eg both of the most accurate models are more than 100B parameters).
In particular, the idealized denoised inference time for \instructdavinci is 0.21s per request (i.e., a query--passage pair to be scored). If these requests have to be issued sequentially per query, this is likely too slow, even though we are evaluating models on a variant where only ~30 passages are being ranked for each query. However, it is in principle possible to parallelize such scoring across passages (for a given query), and thus a latency of 0.21s could be acceptable if such parallelism is seen as cost-effective.

\begin{table}[htp]
\centering
\resizebox{0.55\textwidth}{!}{
\begin{tabular}{lrr}
\toprule
Model & CNN/DailyMail (ROUGE-2) & XSUM (ROUGE-2) \\
\midrule
Jurassic Jumbo (178B) & 0.149 & 0.122 \\
Jurassic Grande (17B) & 0.152 & 0.117 \\
Jurassic Large (7.5B) & 0.13 & 0.103 \\
Anthropic-LM (52B) & 0.155 & 0.126 \\
BLOOM (176B) & 0.113 & 0.032 \\
T0++ (11B) & 0.121 & 0.106 \\
Cohere xlarge & 0.166 & 0.128 \\
Cohere large & 0.15 & 0.102 \\
Cohere medium & 0.129 & 0.085 \\
Cohere small & 0.08 & 0.027 \\
GPT-J (6B) & 0.143 & 0.095 \\
GPT-NeoX (20B) & 0.145 & 0.095 \\
T5 (11B) & 0.054 & 0.017 \\
UL2 (20B) & 0.053 & 0.05 \\
OPT (66B) & 0.147 & 0.122 \\
OPT (175B) & 0.167 & 0.156 \\
GPT-3 (175B) & 0.148 & 0.139 \\
GPT-3 (6.7B) & 0.137 & 0.103 \\
GPT-3 (1.3B) & 0.125 & 0.02 \\
GPT-3 (350M) & 0.107 & 0.003 \\
Davinci Instruct v2 & 0.163 & 0.146 \\
Curie Instruct & 0.157 & 0.075 \\
Babbage Instruct & 0.161 & 0.047 \\
Ada Instruct & 0.134 & 0.033 \\
GLM (130B) & 0.153 & 0.136 \\
YaLM (100B) & 0.022 & 0.022 \\
\bottomrule
\end{tabular}}
\caption{\textbf{Summarization accuracy.} These accuracy of all 30 models assessed for summarization in \citet{liang2022helm}.}
\label{fig:accuracy (summarization)}
\end{table}
\paragraph{Summarization.}
For summarization on \cnndm and \xsum, we begin by observing that the \texttt{ROUGE} scores tend to be much lower than our qualitative judgments of summary quality, consistent with \citet{goyal2022news}.
With that said, broadly, we did find that the \texttt{ROUGE-2} scores did correlate with more accurate models across-the-board (\eg the top models on both datasets based on \texttt{ROUGE-2} largely overlapped with those at the top of the accuracy subfigure of \autoref{fig:head-to-head}).
In general, we saw a strong correlation with model size, as the largest models tended to be those with high \texttt{ROUGE-2} scores for both scenarios, notably with \mtnlgfivethreezero having the highest score for both scenarios with a reasonable margin for \xsum at 17.9 points when compared with second-place \optonesevenfive at 15.6 points and no other model above 14 points. 

Beyond model accuracy, we found that getting models to produce summaries of appropriate length (so as to match the distribution of reference summaries in the dataset) was a key challenge, especially given models only observe a few in-context examples and, therefore, may be poorly specialized to the distribution (when compared to models explicitly trained/fine-tuned on the distribution, for which automated \texttt{ROUGE} scores may be more useful). 
In particular, comparing the compression scores across models, we see considerable variation, and the trends were not consistent across the two datasets.
As a very striking example, \gptada was one of the most compressive on \cnndm but the least compressive on \xsum by a wide margin (1.65 vs the next least compressive in \gptbabbage at 6.12 where higher scores means more compression), suggesting the \gptada model especially did not ``understand'' the specification of length requirements in the instructions in the prompt. 
In terms of the degree of abstraction in model generations, we found that the relationship between model quality and abstraction (measured in terms of both coverage and density from \citet{grusky2018newsroom}) was very variable, with few consistent trends.

Since the summarization scenarios required the longest-form generation of all \core scenarios, we paid special attention to the presence of generative harms. 
For stereotypical associations (for both race and gender) on both datasets, we found that all models exhibited very similar biases, especially on \cnndm.  
In part, we think alternative forms of measurement that propose means for controlling for bias in the source documents that are being summarized (\ie attributing bias to the dataset vs. the model's specific tendencies in generation) could be helpful in providing more acuity. 
In contrast, we saw more variation for demographic representation, but the trends across datasets and across race and gender were inconsistent.
Interestingly, with respect to demographic representation, \yalm demonstrated the greatest racial bias on both datasets and the greatest gender bias on \cnndm (\eg racial bias of 0.651 on \cnndm for race compared to the next highest of 0.399 from \gptj), but was one of the least gender biased models on \xsum. 
And for toxicity, we found the incidence of toxicity to be very low for both datasets, suggesting the risk of toxicity for such innocuous use cases to largely be marginal.
With that said, we emphasize this is summarization of news documents, where models are inherently less likely to generate toxic content given the domain of the documents.
And we do note, while a very low rate of 0.6\%, \mtnlgfivethreezero achieves the highest toxicity rate in addition to the highest \texttt{ROUGE-2} accuracy on \xsum. 

\begin{table}[htp]
\centering
\resizebox{0.3\textwidth}{!}{
\begin{tabular}{lr}
\toprule
Model & IMDB (EM) \\
\midrule
Jurassic Jumbo (178B) & 0.927 \\
Jurassic Grande (17B) & 0.949 \\
Jurassic Large (7.5B) & 0.952 \\
Anthropic-LM (52B) & 0.93 \\
BLOOM (176B) & 0.944 \\
T0++ (11B) & 0.57 \\
Cohere xlarge & 0.938 \\
Cohere large & 0.925 \\
Cohere medium & 0.886 \\
Cohere small & 0.521 \\
GPT-J (6B) & 0.953 \\
GPT-NeoX (20B) & 0.945 \\
T5 (11B) & 0.382 \\
UL2 (20B) & 0.403 \\
OPT (66B) & 0.921 \\
OPT (175B) & 0.948 \\
GPT-3 (175B) & 0.913 \\
GPT-3 (6.7B) & 0.902 \\
GPT-3 (1.3B) & 0.52 \\
GPT-3 (350M) & 0.898 \\
Davinci Instruct v2 & 0.947 \\
Curie Instruct & 0.915 \\
Babbage Instruct & 0.923 \\
Ada Instruct & 0.839 \\
GLM (130B) & 0.955 \\
YaLM (100B) & 0.8 \\
\bottomrule
\end{tabular}}
\caption{\textbf{Sentiment analysis accuracy.} These accuracy of all 30 models assessed for sentiment analysis in \citet{liang2022helm}.}
\label{fig:accuracy (sentiment_analysis)}
\end{table}
\paragraph{Sentiment analysis.}
For sentiment analysis on \imdb, we see model accuracies are often quite high (\ie greater than 90\%), with the top models hitting an accuracy around 95\% (\jurassicgrande, \jurassiclarge, \bloom, \coherexl, \gptneox, \optonesevenfive, \mtnlgfivethreezero, \instructdavinci, \glm) and \glm reporting the top accuracy by a thin margin at 95.5\%.
The strong performance in terms of accuracy of the much smaller \gptj and \jurassiclarge models is noteworthy for this scenario, as well the performance of \bloom, which generally underperforms on other scenarios given its multilingual training contrasted with our English-specific evaluation.
Further, we are surprised to see that \jurassicjumbo lags a few points behind the other J-1 models and the generally high-accuracy \anthropiclm model is also more middle-of-the-pack with an accuracy of  around 93.4\%.
A few models perform worse than chance/majority baseline, namely \tfive and \ultwo, with \tzero, \coheres, and \gptbabbage slightly above 50\% (though \gptada is well above chance at 78.4\%). 

The most accurate models are fairly well-calibrated, with \glm being somewhat miscalibrated with a calibration error of ECE-10 = 0.18 and \bloom being quite miscalibrated with ECE-10 = 0.353.
\yalm is especially poorly calibrated with ECE-10 = 0.418 alongside a standard accuracy of 83.6\%. 
For robustness and fairness, we see the size of the drops in accuracy are generally consistent: the gaps are notably quite small for \instructdavinci (less than a 1\% drop for either fairness or robustness) and quite large for \gptneox (94.8\% standard accuracy drops to 91.2\% robust accuracy). 
In particular, \gptneox is the second most accurate model but it is outside the top 10 models in terms of robust accuracy. 
Since \imdb is one of the two datasets where we have access to a contrast set, we further look into the relationship in robustness behavior for invariances and equivariances.
In contrast to the invariances that use automatic perturbations, the human-authored contrast set examples that are semantics-altering show significantly larger drops in accuracy (see \autoref{fig:contrast-sets}). 
Notably, the most accurate model in \glm experiences one of the largest drops, with its accuracy dropping from 95.6\% standard accuracy for the contrast set examples to 86.9\% in the presence of these perturbations.
This comes in clear contrast to the synthetic perturbations that were semantics-preserving, which yielded a drop of only 1.7\% for \glm. 
Similar to the case of \boolq, we find a cluster of models with moderate accuracy, but really low robustness: e.g., \gptbabbage achieves an accuracy of 52\% but robustness of only 8.6\%.
In this sense, approaches like the contrast sets and counterfactual techniques explored by \citet{kaushik2019learning} and \citet{gardner2020contrast} may be necessary to more completely surface robustness issues in language models (and automated/scalable techniques may underestimate such concerns). 
After all, a model that does not correctly change its prediction when the input changes in a semantics-altering way, is likely latching on irrelevant features of the input.

\begin{table}[htp]
\centering
\resizebox{0.45\textwidth}{!}{
\begin{tabular}{lr}
\toprule
Model & CivilComments (EM) \\
\midrule
Jurassic Jumbo (178B) & 0.407 \\
Jurassic Grande (17B) & 0.788 \\
Jurassic Large (7.5B) & 0.777 \\
Anthropic-LM (52B) & 0.589 \\
BLOOM (176B) & 0.614 \\
T0++ (11B) & 0.672 \\
Cohere xlarge & 0.389 \\
Cohere large & 0.277 \\
Cohere medium & 0.212 \\
Cohere small & 0.805 \\
GPT-J (6B) & 0.284 \\
GPT-NeoX (20B) & 0.224 \\
T5 (11B) & 0.604 \\
UL2 (20B) & 0.62 \\
OPT (66B) & 0.198 \\
OPT (175B) & 0.215 \\
GPT-3 (175B) & 0.294 \\
GPT-3 (6.7B) & 0.337 \\
GPT-3 (1.3B) & 0.646 \\
GPT-3 (350M) & 0.662 \\
Davinci Instruct v2 & 0.688 \\
Curie Instruct & 0.679 \\
Babbage Instruct & 0.809 \\
Ada Instruct & 0.799 \\
GLM (130B) & 0.81 \\
YaLM (100B) & 0.753 \\
\bottomrule
\end{tabular}}
\caption{\textbf{Toxicity detection accuracy.} These accuracy of all 30 models assessed for toxicity detection in \citet{liang2022helm}.}
\label{fig:accuracy (toxicity_detection)}
\end{table}
\paragraph{Toxicity detection.}
For toxicity detection on \civilcomments, the most striking finding is that models do not do particularly well, with many achieving accuracies marginally above 50\% (\ie chance accuracy for this binary classification task). 
\anthropiclm, \bloom, \mtnlgfivethreezero, and \instructdavinci are the only models above 60\% accuracy, with all other models being at or below 55\% accuracy.
Of these, \instructdavinci is clearly the most accurate at 66.8\%, whereas the other three are around 61.0\% accuracy. 
In fact, some models that are generally quite accurate (\eg \coherexl, \optonesevenfive, \gptdavinci) all get at most 53.2\% accuracy, with \optonesevenfive basically achieving chance accuracy at 50.2\%. 

For calibration, all models are poorly calibrated (\eg ECE-10 of at least 0.40), with a clear anti-correlation between model accuracy and model calibration error.
Even more strikingly, we see models fare quite poorly in the presence of perturbations, with most models having accuracies in the presence of either robustness or fairness perturbations belong 50\%. 
For example, \instructdavinci drops precipitously from well above 50\% to below it in the presence of fairness perturbations (66.8\% to 46.3\%), and \mtnlgfivethreezero shows a similar-sized drop in the presence of robustness perturbations (60.1\% to 40.9\%). 

To build on this connection, we deliberately chose \civilcomments to study toxicity detection given the unique availability of group identity metadata regarding the subjects mentioned in the comments. 
For all of the subsets (male, female, LGBTQ, Christian, Muslim, other religions, Black, White), we see the four models that performed clearly above chance overall continue to retain accuracies around 60 to 65\%. 
For gender/sexuality, we see LGBTQ performance is clearly the worst for all models, but surprisingly several models do worse for the male split compared to the female split (and the drops in accuracy are smaller for female than male in the presence of perturbations).
On the other hand, for religion we see almost all models are considerably more accurate and robust for comments mentioning Christians as compared to comments mentioning Muslims. 
For example, for the Christian split, the \anthropiclm accuracy improves to 62.8\% from the overall accuracy of 61.0\%, whereas it declines to 54.7\% for the Muslim split. 
And the accuracies for the other religion split lie in between, generally quite close to the overall accuracy for a given model.
Finally, for race we find the accuracies on the Black and White splits are similar, but models are generally significantly less robust on the Black split, most notably with \optonesevenfive dropping precipitously from a standard accuracy of 51.3\% on the Black split to 8.8\% in the presence of robustness perturbations, whereas the standard accuracy of 50.8\% on the White split drops considerably less to 24.3\%.
We note this is an especially critical relationship between fairness, robustness, and toxicity detection performance, given the potential disparate impact of censoring the voices of the already-marginalized on Internet platforms (\ie the primary site of automated toxicity detection and content moderation), building on the findings of \citet{sap2019risk}. 

\begin{table}[htp]
\centering
\resizebox{0.38\textwidth}{!}{
\begin{tabular}{lr}
\toprule
Model & RAFT (EM) \\
\midrule
Jurassic Jumbo (178B) & 0.702 \\
Jurassic Grande (17B) & 0.664 \\
Jurassic Large (7.5B) & 0.536 \\
Anthropic-LM (52B) & 0.68 \\
BLOOM (176B) & 0.584 \\
T0++ (11B) & 0.411 \\
Cohere xlarge & 0.602 \\
Cohere large & 0.609 \\
Cohere medium & 0.523 \\
Cohere small & 0.509 \\
GPT-J (6B) & 0.616 \\
GPT-NeoX (20B) & 0.511 \\
T5 (11B) & 0.336 \\
UL2 (20B) & 0.293 \\
OPT (66B) & 0.575 \\
OPT (175B) & 0.6 \\
GPT-3 (175B) & 0.655 \\
GPT-3 (6.7B) & 0.484 \\
GPT-3 (1.3B) & 0.491 \\
GPT-3 (350M) & 0.386 \\
Davinci Instruct v2 & 0.691 \\
Curie Instruct & 0.495 \\
Babbage Instruct & 0.53 \\
Ada Instruct & 0.423 \\
GLM (130B) & 0.593 \\
YaLM (100B) & 0.555 \\
\bottomrule
\end{tabular}}
\caption{\textbf{Text classification accuracy.} These accuracy of all 30 models assessed for text classification in \citet{liang2022helm}.}
\label{fig:accuracy (miscellaneous_text_classification)}
\end{table}
\paragraph{Miscellaneous text classification.}
For miscellaneous text classification on \raft, we see that the models display the widest spread in accuracies.
\glm is clearly the most accurate, with an overall accuracy of 85.8\%, though surprisingly \jurassicgrande is the only other model with at least 80\% accuracy at exactly 80\% accuracy.
\jurassicjumbo, \coherexl, and another surprise in \gptj all achieve accuracies of at least 78\%, whereas the overall most accurate models in \instructdavinci (66.7\%) and, especially, \anthropiclm (50.8\%) are considerably less accurate on \raft. 
In fact, this is a very rare instance where \gptdavinci outperforms \instructdavinci in terms of accuracy, with an accuracy of 67.5\%. 

In terms of calibration, the accurate models all have calibration errors around ECE-10 = 0.2, with the clear exception of \gptj, which is very miscalibrated while being quite accurate with an ECE-10 of 0.457. 
The models also pattern quite differently in terms of robustness, with \jurassicgrande dropping nearly 10 points from its overall standard accuracy to its robust accuracy, whereas \glm only drops 3.3\%. 
In contrast, all of the models do not show significant drops in the presence of fairness perturbations with the clear exception of \instructdavinci, which goes from 66.7\% overall standard accuracy to 40.0\% in the presence of fairness perturbations. 

Since \raft itself is composed from 11 relatively diverse sub-scenarios, we further consider how model performance breaks down across these sub-scenarios. 
In general, we see that the difficulty of different sub-scenarios can be very variable: some models achieve accuracies of 97.5\% on the Systematic Review Inclusion task, whereas the best accuracy on the One Stop English task is 62.5\% from \gptj, which itself is a notable outlier as most models are belong 30\%.  
Across the subsets, we see that \instructdavinci consistently performs accurately, whereas the performance of other accurate models across splits is more variable (\eg \glm).
However, \instructdavinci's overall accuracy is brought down by a very poor accuracy of 40.8\% on Systematic Review Inclusion (the second worst accuracy for this subset) compared to \glm, which gets an accuracy of 97.5\%.
The heterogeneity in the sub-scenarios in \raft proves to be quite useful in discriminating different models that are generally performant for all of these classification problems in the long-tail vs. those that may be highly accurate for some problems but not fare as well for other problems.
In this sense, \raft helps speak to the generalizability of models across the broad canon of text classification that may be encountered in practical deployments, given the ubiquity of natural language and the demand for classification/categorization. 

\section{Conclusion}
Evaluations have been foundational to many disciplines, including artificial intelligence: determining the right objectives to measure progress is necessary for that progress to have meaning.
The relative importance of evaluations in artificial intelligence has waxed and waned over its history \citep{bommasani2022evaluation}.
Today, many more pay attention to AI evaluations than ten years ago, yet no current evaluation eclipses ImageNet's field-defining impact.
However, this may reflect the reality that evaluation today must be done ``by committee'' given the broad range of interests: while we pursued this holistic vision with our work on HELM, the right trade-off between expansiveness and simplicity still remains unclear to me.
What is even more important is not the specific evaluations that exist, but the growing recognition of the broader ambition required to meet the moment.
Evaluation as a whole can no longer just serve as a guide for the AI research community, or even the entire AI industry.
Evaluations will be expected to disambiguate not just the relative strengths of different models, but the absolute societal needs of which models should warrant trust.\footnote{At the start of my PhD, I briefly considered how measurement theory can provide specific criteria for trust in measures \citep{bommasani2022trustworthy}, which is a prerequisite for trust in models based on those measures.}
I look forward to the field adopting a more expansive conceptualization of evaluation in AI to achieve these more consequential mandates.
\chapter{Empirical Understanding: Part II}\label{chapter:empirics-index}
\chaptermark{\small Composite Index}
Model-level approaches, such as the evaluation platforms we design, are the most common approach for understanding AI.
However, these approaches do not reflect the broader sociotechnical context we introduce in our discussion of supply chains, thereby leaving a deficit in empirical understanding.
Therefore, we continue to develop proactive methods to improve transparency into the broader foundation model ecosystem.
While the previous chapter describes the first branch of methods to generate new information, this chapter describes the second branch of methods to disclose existing information.
This chapter introduces a novel family of methods, known as composite indexes, which have not been previously applied in computer science but have a rich history in economics, statistics, and social science.
We built the first third-party composite index to measure foundation model developers' transparency \citep{bommasani2023fmti, bommasani2024fmti}.
When viewed together, my work on empirical understanding has amounted to two large-scale multi-year initiatives, at both the technological and organizational level, that have yielded comprehensive insight into AI in practice.
\newpage
\noindent Transparency is a core ethical principle undergirding responsible AI \citep{fjeld2020principled, Hagendorff2020}.\footnote{See UNESCO's Recommendation on the Ethics of Artificial Intelligence, which was adopted by its 193 member states and constitutes the first global normative instrument on AI ethics. Our conceptualization of transparency covers several of UNESCO's 10 principles, namely Transparency and Explainability. See \url{https://www.unesco.org/en/artificial-intelligence/recommendation-ethics}}  
Evaluations are the standard method for increasing transparency in AI \citep{bommasani2023transparency}.
Alongside evaluations, documentation is a growing practice for specifying broader information about the broader context that situates model and system development.
Formative works like data sheets \citep{gebru2021datasheets} and model cards \citep{mitchell2018modelcards} brought this approach to the fore, complementing evaluations by articulating design decisions and developer positions involved in creating assets.
Documentation efforts are most common at the level of data \citep{gebru2021datasheets, bender2018data, pushkarna2022data} and models \citep{mitchell2018modelcards, crisan2022interactive}, with evaluations providing further insight into models \citep{deng2009imagenet, ribeiro2020beyond, perez2022red, liang2022helm, bommasani2023transparency}.
More recently, several efforts have studied the broader ecosystem-wide transparency of AI and its supply chains \citep{bommasani2023ecosystem, cen2023supplychain}, though transparency on the downstream impacts of AI is comparatively understudied \citep{narayanan2023transparencyreports}.

The barrier to transparency is incentives, not methods.
As the societal impact of foundation models increased, the transparency of their developers decreased. 
For example, \openai states plainly its intention to be nontransparent in the \gptfour technical report, which “contains no further details about the architecture (including model size), hardware, training compute, dataset construction, training method, or similar” \citep{openai2023gpt4}. 
Companies often claim that such information is proprietary or that sharing it would undermine their market position and pose a danger to society as a whole, but this does not negate the enormous risks stemming from foundation models these same companies openly acknowledge, as well as the value of greater transparency. 

Transparency is an essential precondition for public accountability, scientific innovation, and effective governance of digital technologies.
To advance these goals, transparency must be understood broadly: the prior conceptualization of transparency in AI is woefully inadequate for a major public technology that may very well restructure the entirety of human society.
Without adequate transparency, stakeholders cannot understand foundation models, who they affect, and the impact they have on society.
Historically, digital technologies often follow a familiar pattern: a new technology provides opportunities and benefits, but companies are not transparent in how they develop and deploy the technology, and this opacity eventually leads to harm. 
In the case of social media, companies have not been transparent about the ways in which they moderate content and share user data, contributing to massacres like the Rohingya genocide in Myanmar\footnote{\url{/https://about.fb.com/wp-content/uploads/2018/11/bsr-facebook-myanmar-hria_final.pdf}} and gross violations of privacy like the Cambridge Analytica scandal.\footnote{\url{https://www.nytimes.com/2018/04/04/us/politics/cambridge-analytica-scandal-fallout.html}}
Consequently, a chorus of academics, civil society organizations, firms, and governments have called for foundation model developers to improve transparency.
Groups such as the Partnership on AI, Mozilla, and Freedom House have noted that increased transparency is a crucial intervention.
UN Secretary-General António Guterres has proposed that the international community should “make transparency, fairness and accountability the core of AI governance ... [and] Consider the adoption of a declaration on data rights that enshrines transparency.”

Here, I describe our work on the Foundation Model Transparency Index \citep[FMTI;][]{bommasani2023fmti, bommasani2024fmti}, which pioneered the use of composite indices in AI and computer science.
A composite \textit{index} measures a complex construct (\eg transparency) as the basis for scoring/ranking entities (\eg foundation model developers) by aggregating  many low-level quantifiable \textit{indicators} of transparency.
Indexes are not common in AI\footnote{We note that the AI Index from the Stanford Institute for Human-Centered AI \citep{zhang2022ai, maslej2023ai} is a related effort, but the AI Index tracks broader trends in AI, rather than scoring specific entities or aggregating to a single value.} but are a standard methodology in the social sciences: iconic examples include the United Nations Development Programme's Human Development Index \citep{undp2022hdr}, which ranks countries, and Ranking Digital Rights' Corporate Accountability Index, which ranks companies \citep{rdr2020index}.
We score the transparency of foundation model developers in an effort to promote responsible business practices and greater public accountability. 
I primarily describe the inaugural 2023 Foundation Model Transparency Index, describing the 2024 Foundation Model Transparency to illustrate changes to the ecosystem that reflect improvements to overall transparency.

\section{Constructs}
To develop FMTI, we first realized the woeful insufficiency of the AI community's conceptualization of transparency.
Simply put, transparency into specific technical details (\eg model architecture) is, at best, marginally relevant for the study of societal impact and, at worst, outright distracting.
Therefore, we began our efforts by strengthening the conceptualization before developing concrete indicators that operationalized the abstract construct.
Transparency is broadly understood as the property of being visible and easily understood \citep{aristotle350deanima, kalderon2015transparency}, and is often a fundamental prerequisite of social responsibility and accountability \citep{florini2007right,robinson2012nations}.
In general, transparency is desirable from a variety of standpoints.
For example, transparently disclosing information makes that information available, shareable, legible, and verifiable. 
Transparency when conducting a complex process can make clear the processes' scope, stakes, and pitfalls \citep{lathrop2010open}. 
Similarly, transparency in decision-making can help those who are not involved in the decision assess the motivations behind the decision, the evidence used to justify it, as well as its costs and benefits.
Various philosophers, political theorists, scientists, and journalists have emphasized the importance of transparency across these and other domains \citep{johnston2006good,florini2007right, benkler2013practical, schudson2015rise}. 
Civil society, grassroots organizations, and consumers also regularly call for transparency as a mechanism for fact finding, accountability, and holding organizations responsible for harm \citep{heikkila2023high,diresta2022openblackbox}.

Transparency in digital technologies is particularly relevant for three reasons.
First, new digital technologies, such as AI, are not well understood by society, often appearing as a black box \citep{castelvecchi2016openblackbox}. 
Second, digital technologies are easily rendered invisible, meaning it is difficult for non-experts to understand when processes like algorithmic decision-making are taking place \citep{ng_conceptualizing_2021}.
Third, these technologies can have a profound influence on billions of users across society.
And yet these technologies are built by a small cadre of industry actors who do not represent society as a whole.
Under these conditions, transparency functions as a prerequisite for public accountability and responsible innovation \citep{klyman2023open}.
Shared visibility engenders public trust and facilitates interventions in the public interest \citep{hardin2002trust}. 
Without sufficient understanding of industry practices, researchers cannot characterize the societal impact of digital technologies, let alone propose concrete actions to improve business practices \citep{pasquale2015black}.
While the effects of transparency are often difficult to measure as they are diffuse and indirect, transparency helps to expose malpractice and enables the public to respond to such malpractice.

\paragraph{Calls for transparency.}
In recent years, transparency has been a rallying cry for activists, a boon to researchers, and a tractable first step for governments interested in regulating foundation models. 
Here we outline some of the salient calls for transparency to illustrate the different stakeholders with an interest in a more transparent foundation model ecosystem.

A wide variety of governments have made transparency in the development of foundation models a top priority in their wider agenda for AI regulation. 
In the U.S., the White House has secured voluntary commitments from 16 companies that include a commitment "to publicly reporting their AI systems’ capabilities, limitations, and areas of appropriate and inappropriate use" in the form of "transparency reports."\footnote{See \url{https://www.whitehouse.gov/briefing-room/statements-releases/2023/07/21/fact-sheet-biden-harris-administration-secures-voluntary-commitments-from-leading-artificial-intelligence-companies-to-manage-the-risks-posed-by-ai/} and \url{https://www.whitehouse.gov/wp-content/uploads/2023/07/Ensuring-Safe-Secure-and-Trustworthy-AI.pdf} and \url{https://www.whitehouse.gov/briefing-room/statements-releases/2023/09/12/fact-sheet-biden-harris-administration-secures-voluntary-commitments-from-eight-additional-artificial-intelligence-companies-to-manage-the-risks-posed-by-ai/} and \url{https://www.whitehouse.gov/wp-content/uploads/2023/09/Voluntary-AI-Commitments-September-2023.pdf}} 
The AI Risk Management Framework from the U.S. National Institute for Standards and Technology outlines the U.S. federal government’s current approach to transparency for foundation models and other AI systems.\footnote{\url{https://nvlpubs.nist.gov/nistpubs/ai/NIST.AI.100-1.pdf}}
The AI Risk Management Framework states "Trustworthy AI depends upon accountability. Accountability presupposes transparency. Transparency reflects the extent to which information about an AI system and its outputs is available to individuals interacting with such a system ... Meaningful transparency provides access to appropriate levels of information based on the stage of the AI lifecycle and tailored to the role or knowledge of AI actors or individuals interacting with or using the AI system." 

The SAFE framework for regulating AI proposed by Senate Majority Leader Schumer aims to ensure that "AI is developed and deployed in a responsible and transparent manner" and to "support US-led innovation in AI technologies–-including innovation in security, transparency and accountability."\footnote{\url{https://www.democrats.senate.gov/imo/media/doc/schumer_ai_framework.pdf}}
Transparency is also one of the five pillars of the bipartisan framework for a U.S. AI Act proposed by Senators Hawley and Blumenthal; their framework specifically suggests "requiring transparency from the companies developing and deploying A.I. systems" as it relates to training data, limitations, accuracy, safety, and user interaction with an AI system.\footnote{\url{https://www.blumenthal.senate.gov/imo/media/doc/09072023bipartisanaiframework.pdf}}
A variety of other draft legislation in the U.S. would require a higher level of transparency for foundation model developers, such as the Algorithmic Accountability Act\footnote{See \url{https://www.congress.gov/bill/118th-congress/house-bill/5628/all-info?s=2&r=1} and \url{https://docs.google.com/document/d/1A1bJ1mkIfE3eZuSbDmz3HGVtOvQDegHl53q3ArO7m44/}} at the federal level and California's Safety in Artificial Intelligence Act.\footnote{\url{https://leginfo.legislature.ca.gov/faces/billTextClient.xhtml?bill_id=202320240SB294}} 

In the EU,  transparency and information sharing have become a central focus of the draft EU AI Act. 
For instance, Article 52 of the Act imposes "transparency obligations" for some types of AI systems.
The European Parliament's draft of the AI Act included specific obligations for foundation model developers: "foundation models should have information obligations and prepare all necessary technical  documentation for potential downstream providers to be able to comply with their obligations under this Regulation. Generative foundation models should ensure transparency about the fact the content is generated by an AI system, not by humans."\footnote{\url{https://www.europarl.europa.eu/doceo/document/TA-9-2023-0236_EN.pdf}}
Developers of high-risk AI systems may also be required to provide additional transparency about their systems such that deployers have adequate information about risks and how to mitigate them.

China has gone a step further, with the central government adopting regulations that impose transparency requirements on foundation model deployers. 
China's "Interim Measures for the Management of Generative Artificial Intelligence Services" state that organizations deploying foundation models, including via an API, must "employ effective measures to increase transparency in generative AI services."\footnote{\url{http://www.cac.gov.cn/2023-07/13/c_1690898327029107.htm}} 
The law further specifies that "providers shall formulate clear, specific, and feasible tagging rules" for data and that "providers shall establish and complete mechanisms for making complaints and reports, setting up easy complaint and reporting portals, disclosing the process for handling them and the time limits for giving responses."

Many other governments have also highlighted the importance of transparency in the development and use of foundation models. 
Canada has released a "Voluntary Code of Conduct on the Responsible Development and Management of Advanced Generative AI Systems," which has been signed by Cohere, the Montreal Institute for Learning Algorithms, and the Vector Institute among other organizations.\footnote{\url{https://ised-isde.canada.ca/site/ised/en/voluntary-code-conduct-responsible-development-and-management-advanced-generative-ai-systems}} 
Canada's Voluntary Code of Conduct states that signatories commit to achieve transparency such that "sufficient information is published to allow consumers to make informed decisions and for experts to evaluate whether risks have been adequately addressed."
It further specifies that "developers of advanced generative systems available for public use" are required to "Publish information on capabilities and limitations of the system... Develop and implement a reliable and freely available method to detect content generated by the system, with a near-term focus on audio-visual content (e.g., watermarking). ... Publish a description of the types of training data used to develop the system, as well as measures taken to identify and mitigate risks." Japan is reportedly in the process of adopting its own code of conduct, which may go beyond voluntary commitments.\footnote{\url{https://english.kyodonews.net/news/2023/10/3b83adf1e28d-japans-ai-draft-guidelines-ask-for-measures-to-address-overreliance.html}}

India's report on "Impact, Opportunity, and Challenges of Generative AI," coauthored by India's Ministry of Electronics and Information Technology, states that transparency should be a central feature of India's regulatory framework for ensuring responsible use of generative AI.\footnote{\url{https://indiaai.s3.ap-south-1.amazonaws.com/docs/generative-ai-report.pdf}} The United Arab Emirates' generative AI guide, published by the Office of the Minister for Artificial Intelligence, Digital Economy, and Remote Work Applications,  highlights the importance of transparency for generative AI in terms of data protection: "Transparency is crucial to data privacy because it enables individuals to know how their data is collected, processed, and used by organizations. By being transparent, organizations can provide clear and concise information about their data privacy practices, policies, and procedures."\footnote{\url{https://ai.gov.ae/wp-content/uploads/2023/04/406.-Generative-AI-Guide_ver1-EN.pdf}} Data protection authorities around the world are "de facto regulating generative AI" by using their existing authorities, including those related to information sharing; for example, data protection authorities in Brazil, Japan, and South Korea launched investigations into OpenAI's ChatGPT in 2023.\footnote{\url{https://fpf.org/blog/how-data-protection-authorities-are-de-facto-regulating-generative-ai/}}

Some governments have highlighted the fact that existing transparency requirements already apply to foundation model developers and ought to be enforced as such.
The UK Competition and Markets Authority notes that transparency requirements are already in place under consumer protection law, and that foundation model developers must comply with the transparency provisions of the UK Consumer Rights Act.\footnote{\url{https://www.gov.uk/government/publications/ai-foundation-models-initial-report}} The U.S. Federal Trade Commission has stated that "we take note–and can take action–if companies aren’t upfront about what consumers are buying, who made it, how it was made, or what rights people have in their own creations. ... When offering a generative AI product, [companies] may need to tell customers whether and the extent to which the training data includes copyrighted or otherwise protected material."\footnote{https://www.ftc.gov/business-guidance/blog/2023/08/cant-lose-what-you-never-had-claims-about-digital-ownership-creation-age-generative-ai}

It is also worth noting that many governments have emphasized the importance of transparency in the development and use of AI systems outside of the context of foundation models. The national AI strategies of Colombia,\footnote{\url{https://colaboracion.dnp.gov.co/CDT/Conpes/Económicos/3975.pdf}}, Egypt,\footnote{\url{https://mcit.gov.eg/Upcont/Documents/Publications_672021000_Egypt-National-AI-Strategy-English.pdf}} Indonesia,\footnote{\url{https://ai-innovation.id/images/gallery/ebook/stranas-ka.pdf}}, and India\footnote{\url{https://www.niti.gov.in/sites/default/files/2019-01/NationalStrategy-for-AI-Discussion-Paper.pdf}} highlight the importance of transparency as do the national AI strategies of other countries.\footnote{\url{https://oecd.ai/en/dashboards/overview}} 

The UN High Commissioner for Human Rights, Volker Türk, has argued that existing rules for businesses squarely apply to foundation model developers. 
In a speech in July 2023, Türk stated that generative AI "companies must live up to their responsibilities to respect human rights in line with the Guiding Principles on Business and Human Rights."\footnote{\url{https://www.ohchr.org/en/statements/2023/07/artificial-intelligence-must-be-grounded-human-rights-says-high-commissioner}}
In addition to requiring human rights due diligence, the UN Guiding Principles on Business and Human Rights explicitly refer to transparency as it relates to a company's obligation to (i) transparently communicate the human rights impact of its products and (ii) be transparent in administering grievance processes.\footnote{For instance, the UN Guiding Principles on Business and Human Rights state, "The responsibility to respect human rights requires that business enterprises have in place policies and processes through which they can both know and show that they respect human rights in practice. Showing involves communication, providing a measure of transparency and accountability to individuals or groups who may be impacted and to other relevant stakeholders, including investors." See \url{https://www.ohchr.org/sites/default/files/documents/publications/guidingprinciplesbusinesshr_en.pdf}}

Türk further argued that without adequate guarantees of transparency, generative AI and other types of AI systems should be banned or suspended.
He said "regulations need to require assessment of the human rights risks and impacts of AI systems before, during, and after their use. Transparency guarantees, independent oversight, and access to effective remedies are needed, particularly  when the State itself is using AI technologies. AI technologies that cannot be operated in compliance with international human rights law must be  banned or suspended until such adequate safeguards are in place."

UN Secretary-General António Guterres has foregrounded transparency as well. 
The UN's digital agenda, summarized in Guterres' Global Digital Compact, makes three key proposals related to transparency: (i) the international community should "make transparency, fairness and accountability the core of AI governance," (ii) governments should "consider the adoption of a declaration on data rights that enshrines transparency," and (iii) researchers and companies should be responsible for transparently communicating the risks of AI systems.\footnote{\url{https://indonesia.un.org/sites/default/files/2023-07/our-common-agenda-policy-brief-gobal-digi-compact-en.pdf}}

The G7 Hiroshima AI Process, which was launched in May 2023 and focuses on generative AI, makes "promotion of transparency" one of its core aims.\footnote{\url{https://www.whitehouse.gov/briefing-room/statements-releases/2023/05/20/g7-hiroshima-leaders-communique/}} A September 2023 joint statement on the Hiroshima AI Process by G7 Digital and Technology Ministers committed the G7 to "develop guiding principles for organizations developing, deploying, and using advanced AI systems, in particular foundation models and generative AI," and stated that one such guiding principle could be "publicly report models’ capabilities, limitations and domains of appropriate and inappropriate use, ensuring sufficient transparency."\footnote{\url{https://www.politico.eu/wp-content/uploads/2023/09/07/3e39b82d-464d-403a-b6cb-dc0e1bdec642-230906_Ministerial-clean-Draft-Hiroshima-Ministers-Statement68.pdf}}

More broadly, international organizations have long noted that transparency is essential for responsible development of AI systems.
The OECD AI Principles, adopted in 2019, include transparency as one of five principles for trustworthy AI. 
The principle on “transparency and explainability” reads: “AI Actors should commit to transparency and responsible disclosure regarding AI systems. To this end, they should provide meaningful information, appropriate to the context, and consistent with the state of art: (i) to foster a general understanding of AI systems; (ii) to make stakeholders aware of their interactions with AI systems, including in the workplace; (iii) to enable those affected by an AI system to understand the outcome; and, (iv.) to enable those adversely affected by an AI system to challenge its outcome based on plain and easy-to-understand information on the factors, and the logic that served as the basis for the prediction, recommendation or decision.”\footnote{\url{ https://legalinstruments.oecd.org/en/instruments/OECD-LEGAL-0449}} 
The G20 AI Principles, also adopted in 2019, include this OECD principle on transparency verbatim. \footnote{\url{https://wp.oecd.ai/app/uploads/2021/06/G20-AI-Principles.pdf}} 
A number of other countries have committed to the OECD AI Principles, including Argentina, Brazil, Egypt, and Singapore.\footnote{\url{https://oecd.ai/en/ai-principles}}   

Foundation model developers have also called for greater transparency and touted the benefits of transparency in their own business practices. 
For example, in June 2022 AI21 Labs, Cohere, and OpenAI published "Joint Recommendation for Language Model Deployment" that advocated for increased transparency \citep{cohere2022}. 
Their recommendations stated that developers should “Publish usage guidelines and terms of use of LLMs ... Document known weaknesses and vulnerabilities, such as bias or ability to produce insecure code ... Documentation should also include model and use-case-specific safety best practices."

Individual developers have highlighted the importance of transparency as well. 
Anthropic ties the importance of transparency to interpretability in its paper on Constitutional AI and in describing the company's "Core Views on AI Safety" \citep{bai2022constitutional}.\footnote{As the blog post summarizing the paper states, "Constitutional AI is also helpful for transparency: we can easily specify, inspect, and understand the principles the AI system is following." See \url{https://www.anthropic.com/index/claudes-constitution and https://www.anthropic.com/index/core-views-on-ai-safety}} Inflection prioritizes transparency in its decision-making about the choices it makes with regard to safety. Inflection's Safety Policy states "Safety at its heart is a question of values. Companies choose what risks to prioritize, and how to address them. We believe the best principle is to be deliberate about these choices, and transparent with our users about the specific values we build into our AIs. We may prioritize values that you disagree with. That’s OK. We think that there is room for many perspectives ... We commit to sharing publicly what positions we aim to take in our AIs."\footnote{\url{https://inflection.ai/safety}}

OpenAI has argued that transparency can help companies work together to mitigate safety concerns regarding foundation models.\footnote{\url{https://openai.com/research/cooperation-on-safety}}
\citet{askell2019role} argue "information that companies provide about their intentions and actions—how transparent they are—can play an important role in whether other companies will cooperate with them."
OpenAI also requires transparency from its suppliers: OpenAI's Supplier Code of Conduct states that "OpenAI expects all Suppliers to adhere to the highest standards of integrity, transparency, honesty, and ethical conduct in all their business dealings."\footnote{\url{https://openai.com/policies/supplier-code}} 

Cohere states that transparency is important for its responsible development of large language models, noting that it has "invested in technical and non-technical measures to mitigate potential harm and make our development processes transparent."\footnote{\url{https://cohere.com/responsibility}} 
Cohere's Usage Guidelines prohibit users from using Cohere's platform for applications with "no transparency," meaning those that  "do not disclose that the content is generated through automated means."\footnote{\url{https://docs.cohere.com/docs/usage-guidelines}} 

Stability AI has called for transparency in connection with its advocacy for open foundation models. In a May 2023 report submitted to the U.S. Senate Judiciary Subcommittee on Privacy, Technology, and the Law, Stability AI wrote "Models like Stable Diffusion and StableLM demonstrate our commitment to AI technology that is transparent, accessible, and human-centric: ... We develop open models for transparency. Researchers can `look under the hood' to verify performance, identify potential risks, and help develop safeguards. Organizations across the public and private sector can customize these models for their own needs without exposing sensitive data or ceding control of their AI capabilities."\footnote{\url{https://stability.ai/blog/stability-ai-letter-us-senate-ai-oversight}} The report further argues "These principles can help to advance important policy objectives. Transparent models promote safety and security. ... open models enable the transparent identification, assessment, and management of risks consistent with the National Institute of Standards and Technology AI Risk Management Framework."

Hugging Face has also called for transparency as part of its push for open foundation models. In written testimony before the U.S. House Committee on Science, Space, and Technology, Hugging Face CEO Clement Delangue stated "Rigorous documentation practices for AI systems, with transparent reporting that follows well-defined protocols, serves three main goals: incentivizing responsible development; ensuring researchers and developers consider values and priorities that may otherwise be overlooked; and creating a paper trail for review. ... transparency from entities about how and where they deploy AI systems to understand what evaluations are most urgently needed."\footnote{\url{https://republicans-science.house.gov/_cache/files/5/5/551f066b-4483-4efd-b960-b36bc02d4b66/B82DBAFFA56F31799E058FB2755C2348.2023-06-22-mr.-delangue-testimony.pdf}} Hugging Face has, along with various partners, released a number of artifacts that advance transparency such as tools for exploring datasets \citep{piktus2023roots}. 

In articulating Meta's position with respect to Llama 2, \citet{touvron2023llama} state that "It is important to understand what is in the pretraining data both to increase transparency and to shed light on root causes of potential downstream issues, such as potential biases. ... open releases promote transparency and allow more people to access AI tools, democratizing the technology and decentralizing AI expertise." Meta's Responsible Use Guide for Llama 2 encourages downstream developers to "build transparency and reporting mechanisms in user interactions ... consider ways to provide transparency to end users regarding potential risks and limitations of the system prior to or at the time of user interaction."
\footnote{\url{https://ai.meta.com/static-resource/responsible-use-guide/}} 

Amazon makes clear that transparency is important with respect to the way in which it communicates its policies to users.
Amazon Web Services' Data Privacy Center states that "Our contracts are written in plain, straightforward language to be transparent and help you understand the data privacy protections that we offer. We also provide ongoing data transparency reporting."\footnote{\url{https://aws.amazon.com/compliance/data-privacy/Privacy_at_AWS_}} 

Google highlights transparency in its AI principles, writing "For datasets and models, the consistent outcome is to create and publish detailed documentation of datasets and models in the form of structured transparency artifacts known as data and model cards (see the following section for details), which function like nutrition labels, providing information such as the provenance of the data (if a data card) and model performance when tested for fairness (if a model card)."\footnote{\url{https://ai.google/static/documents/ai-principles-2022-progress-update.pdf}} 
Google's AI principles also detail the "Transparency Artifacts" that Google researchers have built, such as Healthsheets and a Data Cards Playbook.

Microsoft has also produced such artifacts, namely in the form of "Transparency Notes," which "are intended to help you understand how our AI technology works, the choices system owners can make that influence system performance and behavior, and the importance of thinking about the whole system, including the technology, the people, and the environment."\footnote{\url{https://learn.microsoft.com/en-us/legal/cognitive-services/language-service/transparency-note}} 

A large number of developers and deployers that we do not assess have also expressed the importance of transparency \citep{jobin2019global,fjeld2020principled,wef2023presidio}. Notable among them is EleutherAI, a non-profit research group that is a leading developer of open foundation models \citep{skowron2023euaiact}. 
\citet{phang2022eleutherai} write that "EleutherAI’s approach to research goes beyond transparency: by doing research entirely in public, anyone in the world can observe and contribute at every stage," adding that such public-facing research fosters a highly collaborative, diverse, and innovative research community. 

While governments and companies have consistently underscored the value of transparency, less powerful actors have banded together to push public and private entities to meaningfully improve transparency along with the business practices that transparency uncovers. 

Researchers have driven much of the improvement in transparency for foundation model developers, with innovations like model cards, datasheets, and data statements leading to substantial gains \citep{mitchell2018modelcards,gebru2021datasheets, bender2018data}.
Some have sought to solidify these improvements in transparency by strengthening the field of algorithmic auditing \citep{costanzachock2022audit}. 
Mozilla's Open Source Audit Tooling project calls for better infrastructure to evaluate and audit AI systems ~\citep{raji2022mozilla}.
Another proposal to bolster the auditing ecosystem is for governments to conduct third-party audits of AI systems under their existing authority to protect consumers and data subjects ~\citep{miller2021radical}.

Recently, coalitions of researchers led by organizations like LAION have come together to call for greater transparency in the foundation model ecosystem \citep{laion2023transparentai}.
In recent congressional hearings, expert testimony has expressed "The Need for Transparency in Artificial Intelligence" ~\citep{gregory2023testimony}.
\citet{belli2023igf} detail the central importance of transparent foundation models from the perspective of experts across Asia, Africa, and Latin America. 
Other researchers still have argued that transparency, while necessary, is far from sufficient to regulate AI ~\citep{hartzog2023oversight}.

Data workers employed as contractors by foundation model developers have also mobilized for increased transparency \citep{gray2019ghost}.\footnote{Some policymakers have focused on the importance of transparency with respect to data labor. For example, in a letter to the CEOs of major foundation model developers, eight members of the U.S. Congress wrote "Tech companies also must be more transparent about the role data workers play in their AI, so that consumers can make informed choices about the products they use. Unfortunately, many companies have sidestepped these duties, and that must change. ... Please share any plans your company has to be more transparent about the role its data workers play and their working conditions." See \url{https://www.markey.senate.gov/imo/media/doc/letter_to_artificial_intelligence_companies_on_data_worker_labor_conditions_-_091323pdf1.pdf}}
For example, in July 2023 members of the African Content Moderators Union filed a petition with Kenya's parliament requesting an investigation into OpenAI, Meta, Google, and other multinational technology companies that employ content moderators in Kenya.\footnote{The African Content Moderators Union has also sued Meta, alleging that it unlawfully fired workers for their union organizing. See \url{https://techcrunch.com/2023/08/23/meta-and-moderators-agree-to-mediation/}} 
The petition states that OpenAI used a vendor, Sama, to hire the petitioners as contractors who "trained the ChatGPT algorithm," and alleges that "the contracts did not sufficiently describe the nature of the job ... we were not properly informed of the nature of the work we would be undertaking." The petition further alleges that although this data labor included "reading and viewing material that depicted sexual and graphic violence and categorizing it accordingly so that ChatGPT's artificial intelligence could learn it for the purposes of its future interactions with people ... throughout the contract of training ChatGPT we were not afforded psychosocial support."\footnote{\url{https://x.com/mercymutemi/status/1678984336996028416?s=46}}

The Partnership on AI has advocated for transparency with respect to the employment of data enrichment workers, writing "While shifting how the broader field approaches data enrichment is not a trivial task, increasing transparency regarding current practices and developing more practical guidance can move the field towards improved conditions for data enrichment workers. Greater transparency can help emphasize the central role of data enrichment workers, create the basis for a rich public dialogue of how to improve conditions for workers, and increase confidence in AI models themselves."\footnote{In addition to conducting a case study in partnership with Google DeepMind exploring how to increase transparency regarding data labor, the Partnership on AI has separately published a white paper recommending that developers increase transparency in wages and pay structure for data enrichment workers. See \url{https://partnershiponai.org/wp-content/uploads/2022/11/case-study_deepmind.pdf} and \url{http://partnershiponai.org/wp-content/uploads/2021/08/PAI-Responsible-Sourcing-of-Data-Enrichment-Services.pdf}}

Civil society groups with a range of different focus areas agree that transparency is a pressing priority for policymakers and foundation model developers.
For instance, 123 civil society organizations, including AccessNow, Algorithm Watch, and the European Center for Not-for-Profit Law, released a statement advocating for the prioritization of more serious transparency requirements in the EU AI Act.\footnote{\url{https://www.fairtrials.org/app/uploads/2022/05/Civil-society-reacts-to-EP-AI-Act-draft-report_FINAL.pdf}}
The statement advocates the inclusion of a "mandatory impact assessments are a
crucial measure to ensure foresight and accountability for potential AI-related harms," and that "information on all uses of AI systems by public authorities, regardless of
the systems’ risk level should be made public in the EU database." 
Additionally, they call for "an obligation for providers and/or users to include
information regarding the environmental impact of AI systems," which is not a provision in the EU AI Act.
Freedom House has also warned that "AI has allowed governments to refine their online censorship" and threatens to exacerbate the decline in global internet freedom.
AI has allowed governments to enhance and refine their online censorship, and foundation models may exacerbate this trend.\footnote{\url{https://freedomhouse.org/report/freedom-net/2023/repressive-power-artificial-intelligence}}
Freedom House points to transparency requirements as a mechanism to identify and combat evolving and subtle censorship pressures.


In October 2023, the U.S. Federal Trade Commission convened a workshop on the "Creative Economy and Generative AI," where creators from across different industries demanded increased transparency. In the words of one participant, "The creative economy only works when the basic tenets of consent, credit, compensation, and transparency are followed. ... Without transparency, we can't even know the extent of how much of these companies have taken. They took our work and data to train for-profit technologies that then directly compete against us in our own markets using generative media that is meant to mimic us."\footnote{\url{https://www.ftc.gov/system/files/ftc_gov/pdf/creative-economy-and-generative-ai-transcript-october-4-2023.pdf}} 

Despite its limits, transparency is a necessary and broadly popular first step towards accountability for harm caused by AI systems \citep{kaminski_2020, bates2023socially}. In the context of the rapid rollout of extremely powerful AI systems such as foundation models, transparency is all the more urgent. Companies developing and deploying foundation models should heed the call.

\paragraph{Limitations of transparency.} 
Transparency is far from sufficient on its own and it may not always bring about the desired change \citep{corbett2023interrogating}. 
Salient critiques of transparency include:
\begin{itemize}\itemsep0em
    \item Transparency does not equate to responsibility. Without broad based grassroots movements to exert public pressure or concerted government scrutiny, organizations often do not change bad practices \citep{boyd2016algorithmic,ananny2016limits}.
    \item Transparency-washing provides the illusion of progress. 
    Some organizations may misappropriate transparency as a means for subverting further scrutiny. 
    For instance, major technology companies that vocally support transparency have been accused of \emph{transparency-washing}, whereby "a focus on transparency acts as an obfuscation and redirection from more substantive and fundamental questions about the concentration of power, substantial policies and actions of technology behemoths" \citep{zalnieriute2021transparency}.
    \item Transparency can be gamified. Digital platforms have been accused of performative transparency, offering less insightful information in the place of useful and actionable visibility ~\citep{doi:10.1177/20539517231164119, Mittelstadt2019}. 
    As with other metrics, improving transparency can be turned into a game, the object of which is not necessarily to share valuable information.\footnote{According to Goodhart's Law, "when a measure becomes a target, it ceases to be a good measure" \citep{goodhart1984problems}.}
    \item Transparency can inhibit privacy and promote surveillance. 
    Transparency is not an apolitical concept and is often instrumentalized to increase surveillance and diminish privacy \citep{han2015transparency,Mohamed2020,birchall2021radical}. 
    For foundation models, this critique underscores a potential tension between adequate transparency with respect to the data used to build foundation models and robust data privacy.
    \item Transparency may compromise competitive advantage or intellectual property rights.
    Protections of competitive advantage play a central role in providing companies with the incentives to innovate, thereby yielding competition in the marketplace that benefits consumers.
    Consequently, work in economics and management studies have studied the interplay and potential trade-off between competitive advantage and transparency \citep{bloomfield1999market, granados2013transparency, liu2023competitive}, especially in the discourse on corporate social responsibility.
\end{itemize}

Transparency is not a panacea. 
In isolation, more information about foundation models will not necessarily produce a more just or equitable digital world. 
But if transparency is implemented through engagement with third-party experts, independent auditors, and communities who are directly affected by digital technologies, it can help ensure that foundation models benefit society.

\paragraph{Transparency in practice for prior digital technologies.}
Digital technologies are marked by a long track record of poor transparency.
While each major new technology has dramatically restructured society, the powerful corporations that build these technologies have wielded outsized influence and maintained opacity to advance their commercial interests.
Consider the following examples of digital technologies that suffer from a lack of transparency as well as associated interventions/studies to reduce opacity:
the fight for net neutrality for internet service providers like Comcast \citep{crs2021netneutrality}, web cookies for online advertising like Google Ads \citep{englehardt2015cookies,englehardt2016online,narayanan2017princeton}, labor practices for crowd-sourcing platforms like Amazon Mechanical Turk \citep{gray2019ghost, crawford2021atlas}, wage schemes for ride sharing platforms like Uber \citep{rosenblat2016algorithmic}, and dark patterns for game companies like Epic Games.

Stepping through these examples, efforts like the Princeton Web Transparency Project \citep{englehardt2015cookies,englehardt2016online,narayanan2017princeton} have unveiled the ecosystem of online third-party tracking using cookies, which ``led to greater public awareness, the cessation of some privacy-infringing
practices, and the creation of new consumer privacy tools.''
Similarly, \citet{rosenblat2016algorithmic} empirically demonstrated that Uber drivers were the subject of a severely asymmetric power dynamic given the control exerted by Uber over their drivers, to the detriment of the ride sharing market.
In the context of crowd-sourcing, \citet{gray2019ghost} and \citet{crawford2021atlas} demonstrated exploitation of the ``ghost" workers powering AI, such as on Amazon Mechanical Turk, that was made invisible on these platforms.
More recently, these efforts have prompted the scrutiny of lawmakers to improve transparency and, thereby, labor conditions.
As a final example, dark patterns have a pervasive practice for myriad technologies, leading to mismanaged consumer expectations and overall opacity. 
To this end, the FTC's recent inquiry into Epic Games for dark patterns used to deceive gamers, and particularly children, amounted to a \$245M fine on Epic Games.

Building on these prior examples, we consider social media more specifically.
Social media platforms provide a vivid example of transparency challenges in recent years, and the increasing level of acknowledgement among some technology companies that a baseline level of transparency is a necessity. 
Given the profound impact of social media in mediating how humans form relationships, communicate with each other, buy goods and services, and access information, a broad body of work argues for greater transparency \citep[see][]{keller2022platform}. 
Social media platforms have slowly begun to adopt transparency reporting practices.
For example, Facebook now hosts its own Ad Library\footnote{\url{https://www.facebook.com/ads/library/}}, Content Library\footnote{\url{https://transparency.fb.com/researchtools/meta-content-library}}, and a transparency center\footnote{\url{https://transparency.fb.com/}} that reports on content enforcement, widely viewed content, regulatory transparency, government data requests, and intellectual property, among other pieces of mostly voluntary transparency.
In parallel, transparency requirements have been enshrined in laws like the EU Digital Services Act \citep{dsa2022} and legislative proposals like the U.S. Platform Accountability and Transparency Act \citep{pata2021}.

\section{Methods}
A (composite) index is a standard methodology \citep{oecd2008handbook, greco2019methodological} for assessing entities (\eg companies, countries) in relation to a specific construct (\eg transparency, responsibility).
Methodologically, the score on an index for a specific entity is the aggregate of multiple low-level indicators that can be more directly quantified. 
Composite indexes as a methodology has seen broad adoption across the social sciences, including to directly address major political, economic, and societal concerns such as public corruption \citep[\eg Transparency International’s Corruption Perceptions Index;][]{ti2022corrupt}, environmental welfare \citep[\eg the World Economic Forum’s Environmental Sustainability Index;][]{whitford2009political} and living standards \citep[\eg the United Nations Development Programme’s Human Development Index;][]{hopkins1991human}. 
However, indexes have not played a major role in mainstream AI discourse.
In contrast to the composite indexes here, the AI Index neither directly scores specific entities nor does it aggregate individual indicators into a singular aggregate.

Indexes are designed to further several objectives and have certain characteristic strengths \citep{joint2008handbook, saisana2002state}. 
Most fundamentally, indexes can transform complex and amorphous constructs into straightforward and concrete scores.
Indexes and the aggregate quantitative metrics they provide can therefore allow for broad engagement on certain topics, furthering public understanding as well as providing a strong basis for various forms of decision-making  such as regulatory intervention. 
In addition, when indexes are maintained over time, they encourage a long-term focus and can be vital in fostering improvement over time \citep{kogen2022rdr}.
In this way, while operating at a different level of abstraction and involving a different set of design decisions, indexes are analogous to model benchmarks that are commonplace in AI \citep{deng2009imagenet, wang2019superglue, liang2023holistic} and appeal to a similar theory of change \citep{donoho2017fifty, ethayarajh2020utility, raji2021benchmark, bommasani2022evaluation}.
Indexes also have shortcomings: namely, they can be reductive and overly subjective \citep{saisana2002state, oecd2008handbook, greco2019methodological}.
To design and score an index, researchers must make simplifying decisions about which indicators to include, how to weigh those indicators, and how to grade indicators.
Beyond these methodological issues, indexes are subject to a broader conceptual critique that they may oversimplify concepts that are intrinsically complex, discarding valuable nuances.\footnote{The literature and theory on composite indexes is much too extensive to be easily summarized here.
We recommend the Handbook on Constructing Composite Indicators: Methodology and User Guide \citep{oecd2008handbook} as a proper introduction to the subject: \url{https://doi.org/10.1787/9789264043466-en}.} 

To design the Foundation Model Transparency Index, we describe (i) the indicators/scoring criteria, (ii) the developers/scored entities, and (iii) the scoring procedure.

\subsection{Indicators}
We define \numindicators indicators that comprehensively characterize transparency for foundation model developers.  
To select these indicators, we compiled relevant concepts raised across past scientific literature as well as concerns animated by public discourse on foundation models and other digital technologies.
These indicators cover each dimension of the foundation model supply chain, from the data, compute, and labor required to build foundation models to model evaluations and developers' policies to restrict their use. 
We divide our indicators into three broad domains as described in \autoref{fig:supply-chain}: indicators that are \textit{upstream} of the model, indicators that relate to the \textit{model} itself, and indicators that are \textit{downstream} of the model.

\paragraph{Upstream indicators.}
The upstream indicators identify the \emph{ingredients and processes} involved in building a foundation model. 
There are \numupstreamindicators upstream indicators, which we further taxonomize into the following \numupstreamsubdomains subdomains:

\begin{itemize}
    \item \textbf{\data (10 indicators).} 
    Assesses transparency regarding the size and composition of the data used to build the model; the creators whose content is present in the data; and any steps to curate or augment the data. 
    These indicators also address transparency regarding the inclusion of personal, copyrighted, or licensed data. 
    \item \textbf{\labor (7 indicators).} 
    Assesses transparency regarding the use of human labor in producing the data used to build the model, including the wages, labor protections, employer, and geographic distribution of workers who contributed to data annotation and curation.
    These indicators also address transparency regarding the third parties that foundation model developers partnered with to construct their models. 
    \item \textbf{\dataaccess (2 indicators).} 
    Assesses the scope of data access given to external parties.
    \item \textbf{\compute (7 indicators).} 
    Assesses transparency regarding the hardware and computation used to build the model, as well as the resulting energy use and environmental impacts.
    \item \textbf{\methods (4 indicators).} 
    Assesses basic technical specifications for the model's training stages and objectives, as well as the software frameworks and dependencies used.
    \item \textbf{\datamitigations (2 indicators).} 
    Assesses transparency regarding steps taken to mitigate data privacy and copyright concerns.
\end{itemize}

\begin{figure}
\centering
\includegraphics[keepaspectratio, height=0.7\textheight]{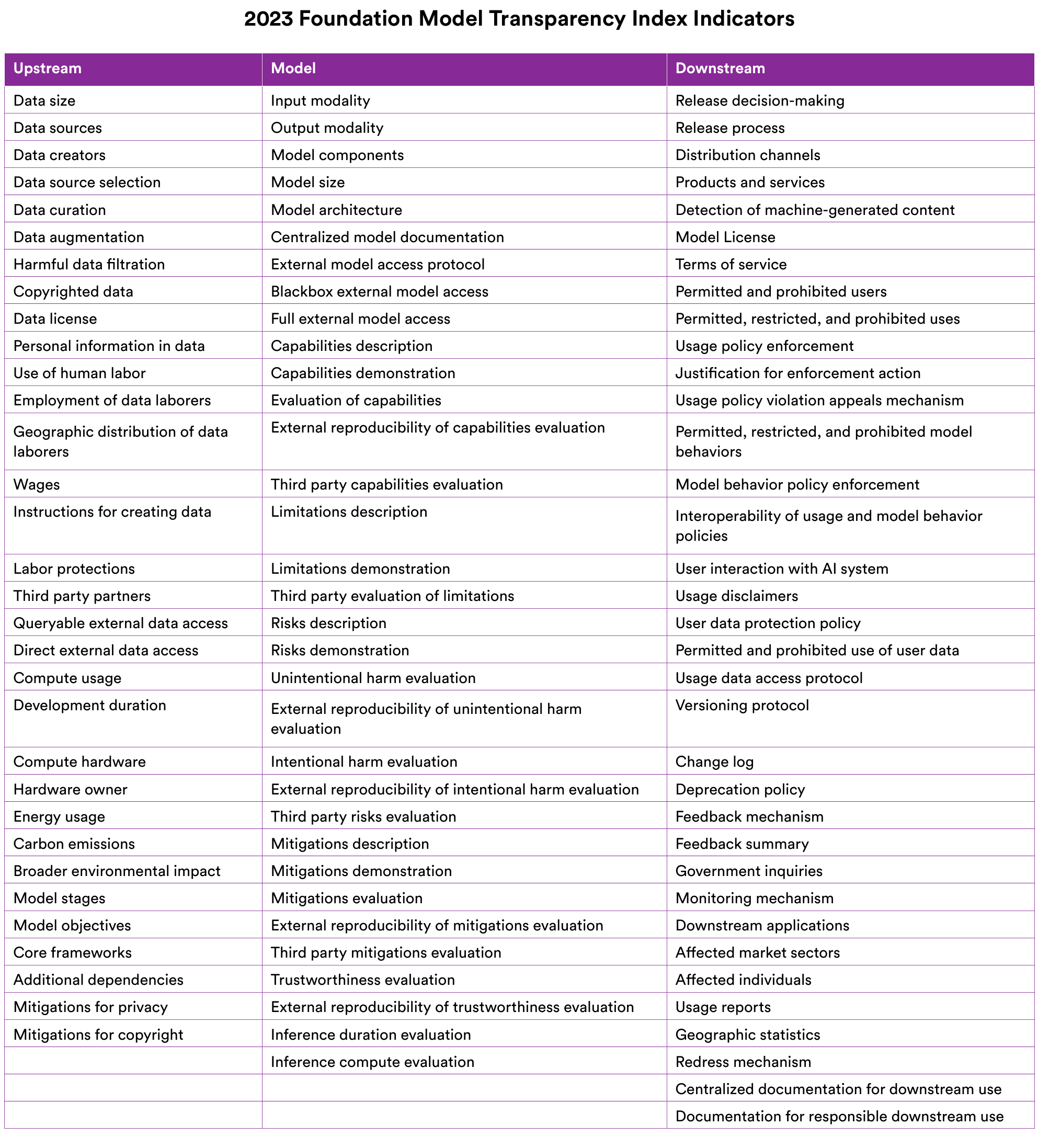}
\caption{\textbf{Indicators.} The \numindicators indicators of the \projectname spanning the \numdomains domains: upstream, model, and downstream.
}.
\label{fig:indicators}
\end{figure}

\begin{figure}
\centering
\includegraphics[keepaspectratio, height=0.9\textheight]{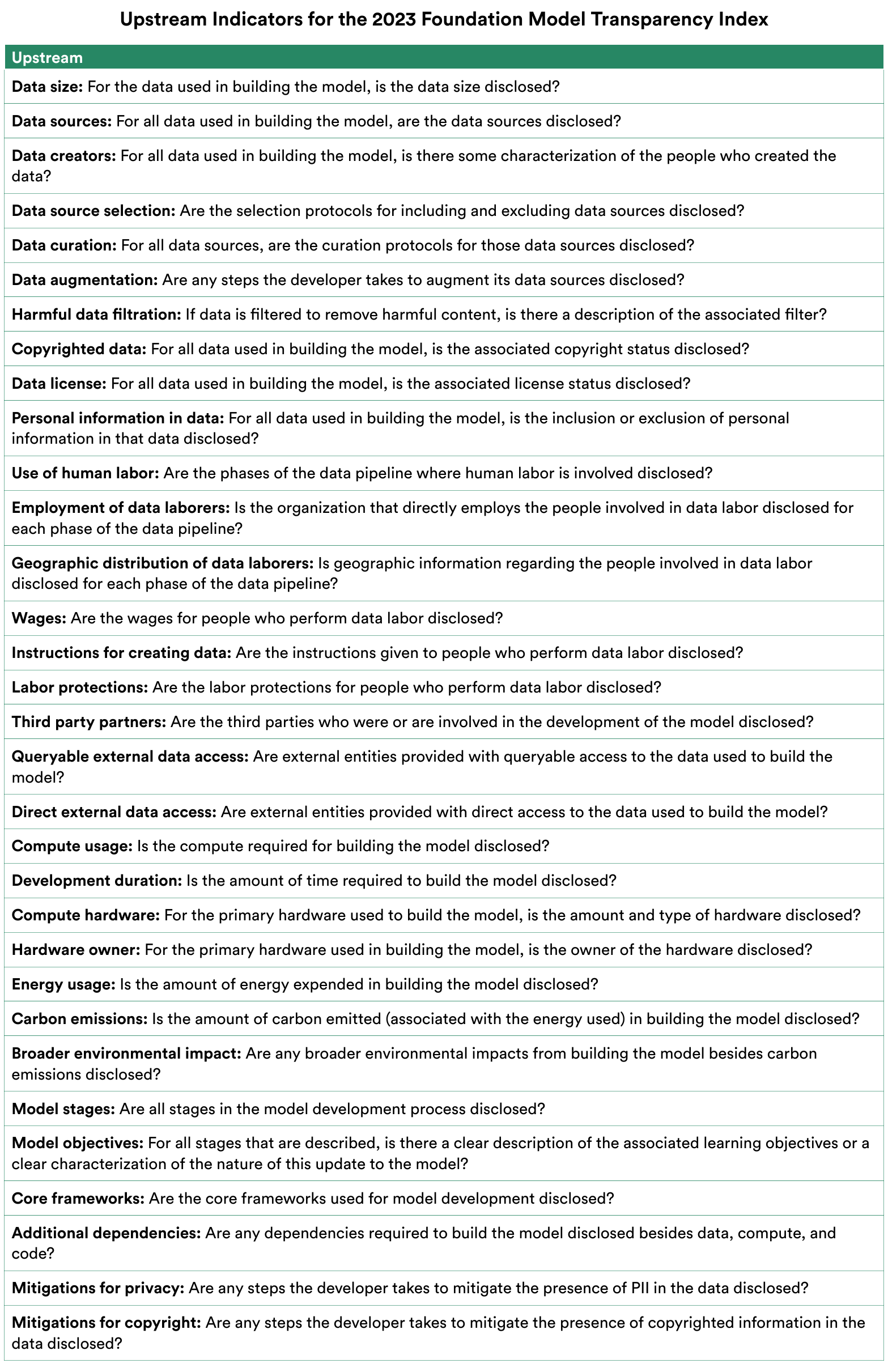}
\caption{\textbf{Upstream indicators.} The \numupstreamindicators upstream indicators that span \data, \labor, \dataaccess, \compute, \methods, and \datamitigations.}
\label{fig:upstream-indicators}
\end{figure}

\noindent We depict the upstream indicators in \autoref{fig:upstream-indicators}.
Researchers have widely advocated for greater transparency in relation to \data and \dataaccess \citep{bender2018data, gebru2021datasheets, hutchinson2021towards, dodge2021c4, bandy2021addressing} as a means for contextualizing model capabilities ~\citep{sambasivan2021everyone, longpre2023pretrainer} and risks related to privacy, bias, and copyright ~\citep{buolamwini2018gender,bender2021dangers, kandpal2022deduplicating,sobel2017artificial}.
\labor indicators uplift concerns related to labor practices, including irresponsible or exploitative use of human labor ~\citep{gray2019ghost, crawford2021atlas, hao2023cleaning, kittur2013future, dzieza2023ai, west2019data}.
\compute indicators relate to concerns around the high computational cost and energy expenditure associated with building foundation models, which can result in environmental harm \citep{lacoste2019quantifying,strubell2019energy,schwartz2020green,patterson2021carbon,bender2021dangers,henderson2020towards,luccioni2023counting,vipra2023comments}.
\datamitigations indicators also relate to the growing legal and sociotechnical concerns over data privacy, copyright, and licensing \citep{henderson2023foundation, brown2022does, lee2023talkin,genlaw2023,tremblay2023openai}.

\paragraph{Model indicators.}
The model indicators identify the \emph{properties and function} of the foundation model. 
There are \nummodelindicators model indicators, which we further taxonomize into the following \nummodelsubdomains subdomains:
\begin{itemize}
    \item \textbf{\modelbasics (6 indicators).} 
    Assesses transparency regarding fundamental information about the model such as modalities, size, and architecture as well as the presence of centralized model documentation.
    \item \textbf{\modelaccess (3 indicators).} 
    Assesses the scope of model access given to external entities.
    \item \textbf{\capabilities (5 indicators).} 
    Assesses transparency regarding the capabilities of the model, including evaluations.
    \item \textbf{\limitations (3 indicators).} 
    Assesses transparency regarding the limitations of the model, including evaluations.
    \item \textbf{\risks (7 indicators).} 
    Assesses transparency regarding the risks of the model, including evaluations, with specific focus on both unintentional harm (\eg bias) and intentional harm (\eg fraud).    
    \item \textbf{\modelmitigations (5 indicators).} 
    Assesses transparency regarding model-level mitigations, including evaluations of their efficacy.
    \item \textbf{\trustworthiness (2 indicators).} 
    Assesses transparency regarding the trustworthiness of the model, including evaluations.
    \item \textbf{\inference (2 indicators).} 
    Assesses transparency regarding standardized inference with the model.
\end{itemize}

\begin{figure}
\centering
\includegraphics[keepaspectratio, height=0.9\textheight]{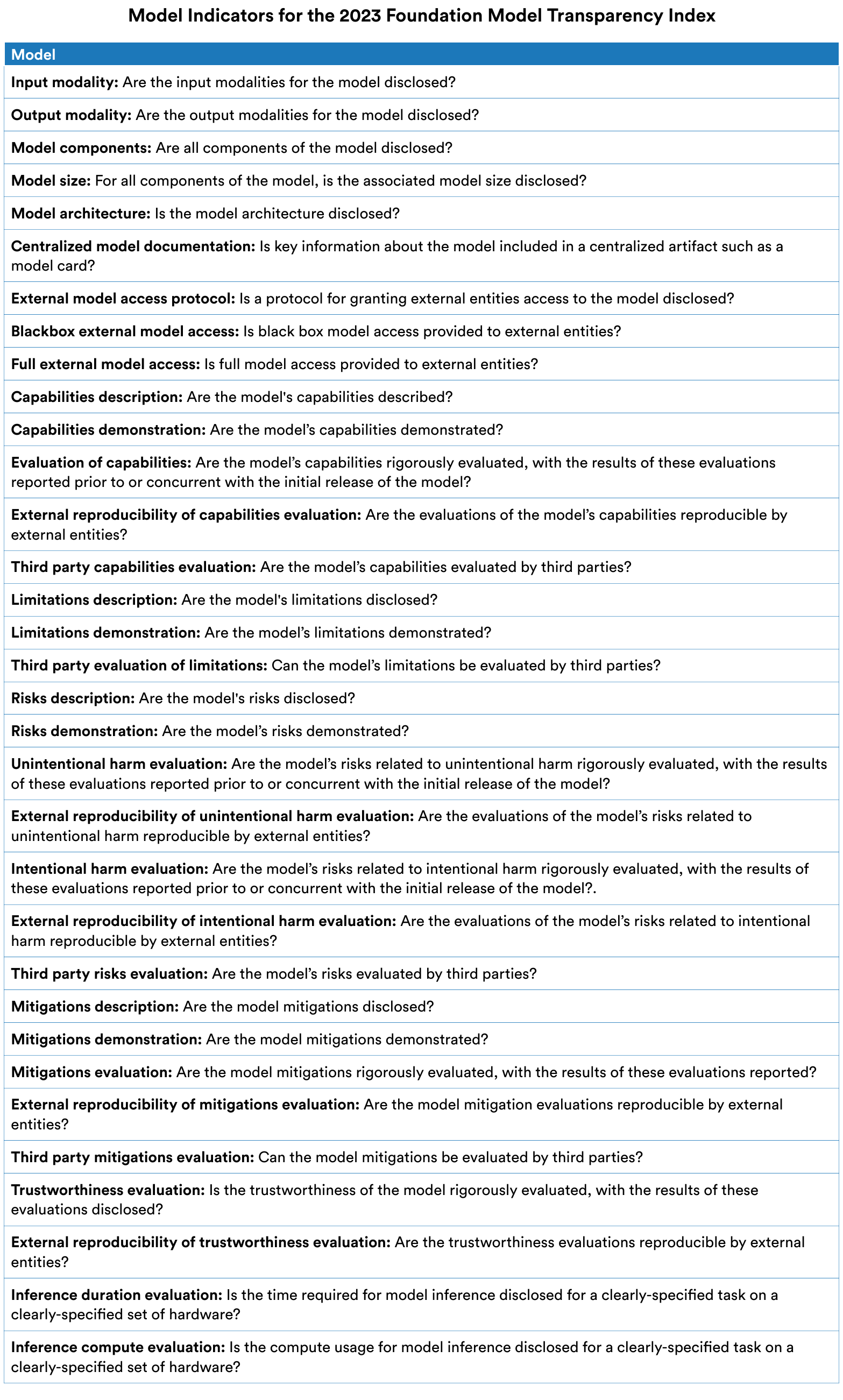}
\caption{\textbf{Model indicators.} 
The \nummodelindicators model indicators that span \modelbasics, \modelaccess, \capabilities, \limitations, \risks, \modelmitigations, \trustworthiness, and \inference.}
\label{fig:model-indicators}
\end{figure}

\noindent 
We depict the model indicators in \autoref{fig:model-indicators}.
\modelbasics indicators refer to fundamental information that is expected by model documentation standards \citep{mitchell2019model, crisan2022interactive, bommasani2023ecosystem} and, historically, have been reliably reported in the release of machine learning models. 
\modelaccess indicators reflect literature tied to the spectrum of model release and the associated differences in external access \citep{solaiman2019release, sastry2021release, shevlane2022structured, liang2022norms, solaiman2023gradient}. 
The indicators on \capabilities, \limitations, \risks and \modelmitigations are motivated by a common understanding that these factors jointly influence the societal impact of machine learning models and AI systems \citep{tabassi2023airmf, weidinger2023sociotechnical}. 
For these subdomains, the description and demonstration indicators gauge whether there is some non-technical articulation and legibility of these concepts, primed by concerns surrounding public understanding of foundation models.\footnote{See \url{https://www.gov.uk/government/publications/public-perceptions-towards-the-use-of-foundation-models-in-the-public-sector}.} 
To make these assessments more rigorous, the evaluation indicators build on the extensive tradition of evaluation in AI spanning iconic benchmarks like ImageNet \citep{deng2009imagenet}, broader benchmarks like SuperGLUE \citep{wang2019superglue}, and extensive meta-benchmarks like LM-Harness, BIG-bench, HELM and BEHAVIOR \citep{gao2021harness, srivastava2022bigbench, liang2023holistic, srivastava2021behavior}. 
Indicators assessing evaluations also highlight the importance of reproducibility \citep{lipton2019troubling, kapoor2023reforms, kapoor2023leakage}\footnote{See the ML Reproducibility challenge: \url{https://paperswithcode.com/rc2022}, CodaLab worksheets for reproducible ML: \url{https://worksheets.codalab.org/}, and Joelle Pineau's reproducibility checklist: \url{https://www.cs.mcgill.ca/~jpineau/ReproducibilityChecklist.pdf}.} and independent assessment \citep{sandvig2014auditing, raji2019actionable, metaxa2021audit, costanzachock2022audit, raji2022audit, raji2022mozilla, lam2022user, weidinger2023sociotechnical}, which enable open science and external verification of developers' claims about their models.
In the case of risks, finer distinctions between unintentional harms (\eg biases, toxicity) and intentional harms (\eg disinformation, fraud) build on harm taxonomies \citep{bender2021dangers, bommasani2021opportunities, weidinger2021ethical, nist2023airmf, weidinger2023sociotechnical}.
Indicators on trustworthiness and inference are especially motivated by the Trustworthy ML Initiative\footnote{\url{https://www.trustworthyml.org/}} and MLPerf \citep{reddi2020mlperf} respectively, among other works \citep{brundage2020toward, cammarota2020trustworthy, kumar2020trustworthy, liu2022trustworthy, shneiderman2020bridging, patterson2021carbon, narayanan2023cheaply}.

\paragraph{Downstream indicators.}
The downstream indicators identify the \emph{use} of the foundation model, including details about its \textit{release}.
There are \numdownstreamindicators downstream indicators, which we further taxonomize into the following \numdownstreamsubdomains subdomains:

\begin{itemize}
    \item \textbf{\distribution (7 indicators).} 
    Assesses transparency regarding the release process, the distribution channels for the model, and the products and services that arise through internal use.
    Additionally, this subdomain assesses the presence of model licenses, terms of service, and mechanisms for detecting model-generated content.
    \item \textbf{\usagepolicy (5 indicators).} 
    Assesses transparency regarding the developer's acceptable use policy such as restrictions on specific uses or users, as well as transparency regarding how it enforces such policies.
    \item \textbf{\modelbehaviorpolicy (3 indicators).} 
    Assesses transparency regarding the developer's policy on acceptable and unacceptable model behavior as well as transparency regarding enforcement of this policy and expectations in the event of usage policy violations.
    \item \textbf{\interface (2 indicators).} 
    Assesses transparency in the user interface for the developer's flagship distribution channel, if the channel includes a user interface. 
    \item \textbf{\dataprotection (3 indicators).} 
    Assesses transparency regarding the developer's policies with respect to user data protection, such as how data is stored, shared, and accessed. 
    \item \textbf{\updates(3 indicators).} 
    Assesses transparency regarding the developer's versioning protocol, change log, and deprecation policy.
    \item \textbf{\feedback (3 indicators).} 
    Assesses transparency regarding mechanisms for reporting feedback on the model, summaries of feedback received, and related government inquiries.
    \item \textbf{\impact (7 indicators).} 
    Assesses transparency regarding the downstream impact of the model on society, such as affected market sectors, individuals, and geographies. 
    Additionally, this subdomain assesses transparency regarding downstream applications, usage statistics, and mechanisms for monitoring usage as well as providing redress in the event of harm to users.
    \item \textbf{\documentation (2 indicators).} 
    Assesses the presence of centralized documentation for downstream use and documentation for responsible downstream use.     
\end{itemize}

\begin{figure}
\centering
\includegraphics[keepaspectratio, height=0.9\textheight]{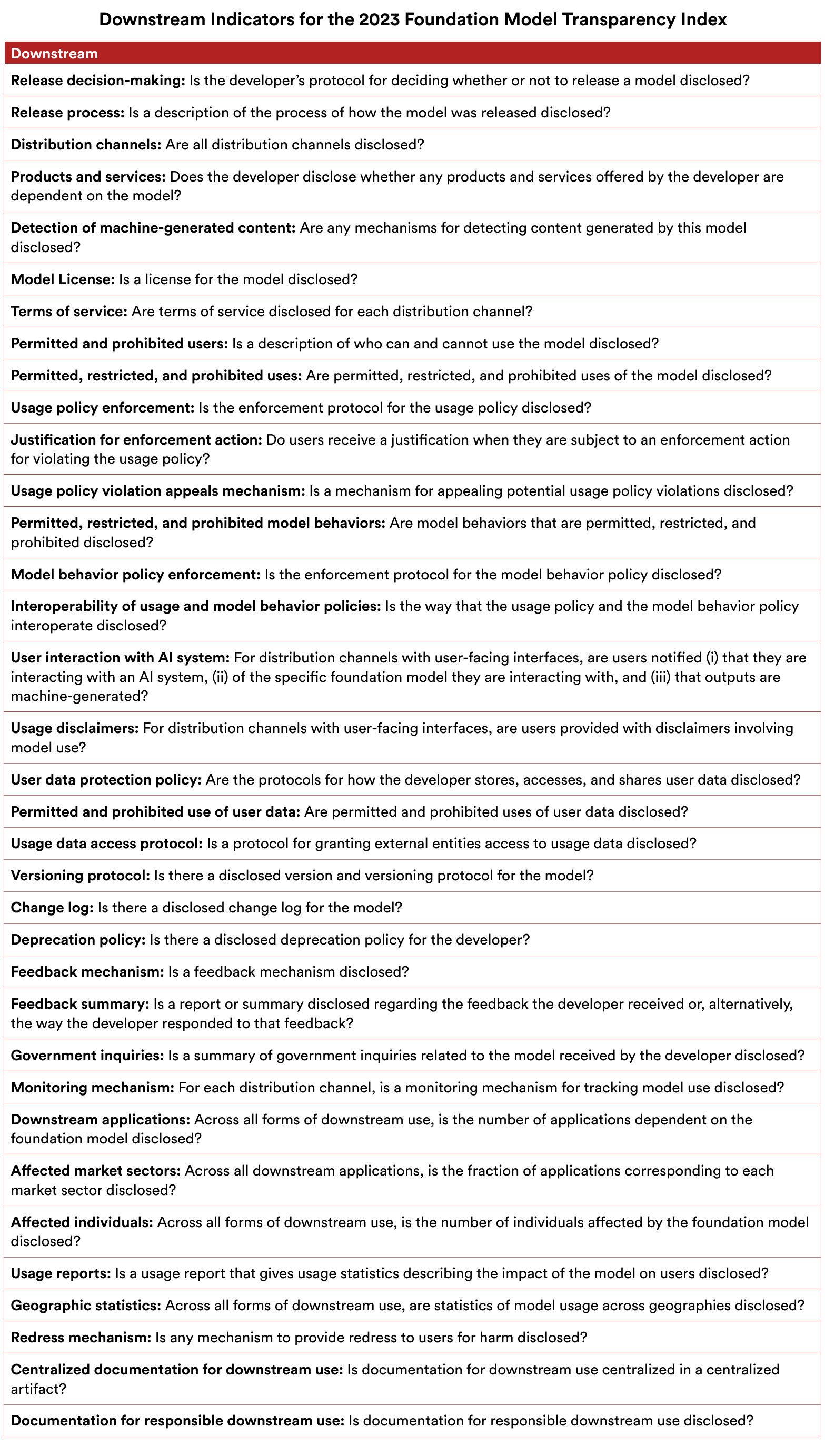}
\caption{\textbf{Downstream indicators.} 
The \numdownstreamindicators downstream indicators that span \distribution, \usagepolicy, \modelbehaviorpolicy, \interface, \dataprotection, \updates, \feedback, \impact, and \documentation.}
\label{fig:downstream-indicators}
\end{figure}

\noindent We depict the downstream indicators in \autoref{fig:downstream-indicators}.
Given that foundation models are the basis for a downstream supply chain \citep{bommasani2021opportunities}, the distribution indicators are informed by the literature on AI supply chains \citep{bommasani2023ecosystem, vipra2023concentration, cen2023supplychain, cobbe2023supply, widder2023thinking, brown2023allocating} and release practices \citep{liang2022community-norms, solaiman2023gradient, henderson2023foundation, pmlr-v202-kirchenbauer23a, Kuditipudi2023RobustDW, liesenfeld2023opening}. 
Usage policy indicators draw from company publications on responsible model deployment \citep{cohere2022} as well as precedents from social media. 
Model behavior policy indicators are rooted in literature that discusses AI behavior and trustworthiness, risks, mitigation and refusal \citep{kumar2022language, weidinger2021ethical, brundage2020toward, cammarota2020trustworthy, kumar2020trustworthy, liu2022trustworthy, reuter2023im}. 
User interface indicators are derived from research on safety by design and human-centered user interfaces \citep{qiaosi2023ux, nakao2022responsible}. 
User data protection indicators are inspired by policy recommendations on user data minimization, privacy, preservation, protection and contextual integrity \citep{eu2016, brown2022does,vipra2023, winograd2023privacy, nissenbaum2004contextual, king2020privacy, mulligan2017privacy}.
Model updates indicators stem from work focused on adequately updating systems and version control of AI systems \citep{sathyavageesran2022privacy, chen2023chatgpts}. 
For feedback, impact and downstream documentation, the indicators were motivated by the literature on algorithmic auditing \citep{liang2022community-norms, solaiman2023gradient, raji2022audit} as well as transparency reporting practices for social media.\footnote{See \url{https://www.tspa.org/curriculum/ts-fundamentals/transparency-report/}, \url{https://transparencyreport.google.com/} and \url{https://transparency.fb.com/reports/}.}

Our view of transparency is expansive, considering the broader supply chain beyond just foundation models.
Existing conceptualizations of transparency in AI, at best, often consider upstream resources (especially data) in addition to machine learning models.
But these works and broader public discourse usually do not foreground the downstream use and impact of AI, even though this is the most direct way in which AI affects society.
To this end, we include the entire downstream domain to bring greater attention to this vital topic.

In particular, while we are assessing foundation model developers, we assess them in relation to distribution channels and other factors that determine their downstream impact.
At present, we recognize that characterizing the downstream impact of foundation models may be challenging, especially for open model developers.
By releasing a model openly, developers may cede the ability to easily monitor the model's downstream use and impact.
In addition, we believe in the potential for greater coordination between foundation model developers and distribution channels to increase transparency; for example, distribution channels could supply information about how the model is used to the foundation model developer.
Partnerships with distribution channels that promote transparency provide a promising means for all foundation model developers to share more information about the impact their models have on society.

\subsection{Developers}
\label{sec:developers}
\begin{table*}[htp]
\resizebox{\textwidth}{!}{
\begin{tabular}{ccccccccccc}
\toprule
Name & Flagship & Release & Input & Output & Status & Headquarters & WH1 & WH2 & WH3 & FMF \\
\midrule
\aitwentyone & \jurassic & API & Text & Text & Startup & Tel Aviv, Israel & \xmark & \xmark & \xmark & \xmark \\
\amazon & \titan & API & Text & Text & Big Tech & Seattle, USA & \xmark & \cmark & \xmark & \xmark \\
\anthropic & \claude & API & Text & Text & Startup & San Francisco, USA & \cmark & \cmark & \xmark & \cmark \\
\cohere & \command & API & Text & Text & Startup & Toronto, Canada & \cmark & \xmark & \cmark & \xmark \\
\google & \palm & API & Text & Text & Big Tech & Mountain View, USA & \cmark & \cmark & \xmark & \cmark \\
\huggingface & \bloomz & Open weights, open data & Text & Text & Startup & Brooklyn, USA & \cmark & \xmark & \xmark & \xmark \\
\inflection & \inflectionone & No access (API forthcoming) & Text & Text & Startup & Palo Alto, USA & \xmark & \cmark & \xmark & \xmark \\
\meta & \llama & Open weights & Text & Text & Big Tech & Menlo Park, USA & \xmark & \cmark & \xmark & \xmark \\
\openai & \gptfour & API & Text, Images & Text & Startup & San Francisco, USA & \cmark & \cmark & \xmark & \cmark \\
\stability & \stablediffusion & Open weights, open data & Text & Images & Startup & London, UK & \cmark & \xmark & \cmark & \xmark \\
\bottomrule 
\end{tabular}}

\caption{\textbf{Foundation model developers assessed in the 2023 FMTI.} 
Information on the 10 foundation model developers in the 2023 FMTI: the developer name, their flagship model, the release strategy for the model , the input and output modalities for the model, the developer's status as either Big Tech or Startup, and the developer's headquarters.
We note which of the developers were involved in the White House's initiative for public evaluation of AI systems announced in May 2023 (WH1), voluntary commitments for the management of risks posed by AI announced in July 2023 (WH2), and commitments by additional organizations on the same matters of risks by AI announced in September 2023 (WH3). 
Additionally, we note which of the developers are founding members of the Frontier Model Forum, announced in July 2023.
}
\label{tab:developer-info}
\end{table*}
\noindent 
Transparency initiatives in AI (\eg datasheets and model cards) often introduce frameworks that support machine learning developers in achieving greater transparency in their own work.
In contrast, we proactively assess foundation model developers for their transparency using the \numindicators indicators we specify.
By conducting the assessment ourselves, we sidestep concerns of uneven uptake that have arisen with past transparency initiatives \citep[\eg][]{gebru2021datasheets, mitchell2018modelcards} and provide greater consistency in the scoring of each indicator across developers.
Most importantly, scoring many developers allows for the comparison of their scores, which provides a rich context for how to improve transparency in the foundation model ecosystem.

Efforts like Ecosystem Graphs \citep{bommasani2023ecosystem} and the UK Competition and Markets Authority (CMA) report on the foundation model market\footnote{\url{https://www.gov.uk/government/publications/ai-foundation-models-initial-report}} track the organizations that develop foundation models.
At the time of writing in September 2023, the CMA report documented 160 foundation models (based on data drawn from Ecosystem Graphs) built by more than 50 organizations.\footnote{\url{https://assets.publishing.service.gov.uk/government/uploads/system/uploads/attachment_data/file/1185508/Full_report\_.pdf\#page=22}} 
However, as the CMA report states, a small number of developers control the majority of the market at present \citep{vipra2023concentration}.
Due to this intense level of market concentration, we decided to assess \numcompanies major foundation model developers.

\paragraph{Selecting developers.}
We considered a variety of selection criteria in choosing the \numcompanies developers to assess, arriving at the following three principles:
\begin{enumerate}
    \item \textbf{Impact.} We selected developers that have built the most influential foundation models.
    \item \textbf{Diversity.} We selected developers that, when considered collectively, represent many axes of variation in the foundation model ecosystem. For example, developers that release models along different points on the release gradient \cite[\eg open vs. closed,][]{solaiman2023gradient}, build models with different modalities (\eg text-to-text vs. text-to-image), and occupy different positions in the market (\eg startups vs. Big Tech). 
    \item \textbf{Companies.} 
    We selected developers that are established companies as enduring targets for longitudinal improvement. 
    This to some extent parallels current regulatory initiatives that explicitly focus on companies as the target of policy for foundation models.\footnote{See \url{https://www.blumenthal.senate.gov/imo/media/doc/09072023bipartisanaiframework.pdf}.} 
\end{enumerate}
On this basis, we chose 10 companies that all are influential foundation model developers: 
\aitwentyone, \amazon, \anthropic, \cohere, \google, \huggingface, \inflection, \meta, \openai, and \stability.
These \numcompanies provide significant diversity in terms of release strategy (\eg \anthropic, \meta, and \huggingface all release flagship models with different levels of openness, modality (\eg \cohere, \openai, and \stability all provide different input-output modalities), and market position (\eg \google, \inflection, and \openai occupy different market positions).

Additionally, in parallel to our research, the White House made three announcements involving companies that develop foundation models: a red-teaming exercise announced in May 2023,\footnote{\url{https://www.whitehouse.gov/briefing-room/statements-releases/2023/05/04/fact-sheet-biden-harris-administration-announces-new-actions-to-promote-responsible-ai-innovation-that-protects-americans-rights-and-safety/}} a set of voluntary commitments announced in July 2023,\footnote{\url{https://www.whitehouse.gov/briefing-room/statements-releases/2023/07/21/fact-sheet-biden-harris-administration-secures-voluntary-commitments-from-leading-artificial-intelligence-companies-to-manage-the-risks-posed-by-ai/}} and another set of voluntary commitments announced in September 2023.\footnote{\url{https://www.whitehouse.gov/briefing-room/statements-releases/2023/09/12/fact-sheet-biden-harris-administration-secures-voluntary-commitments-from-eight-additional-artificial-intelligence-companies-to-manage-the-risks-posed-by-ai/}}
Separately, three of the companies we assess jointly announced the formation of the Frontier Model Forum in July 2023.\footnote{\url{https://blogs.microsoft.com/on-the-issues/2023/07/26/anthropic-google-microsoft-openai-launch-frontier-model-forum/}} 
When taken together, these announcements name 16 companies: Adobe, Amazon, Anthropic, Cohere, Google, Hugging Face, IBM, Inflection, Meta, Microsoft, NVIDIA, OpenAI, Palantir, Salesforce, Scale AI, and Stability AI.
We note that 9 of the 10 companies we selected are within this set of 16 (all but \aitwentyone).

\paragraph{Flagship models.}
Almost all major foundation model developers release multiple foundation models over time and, even at the time of writing, many have multiple salient foundation models (often across different modalities).
For example, OpenAI has developed GPT, GPT-2, GPT-3, GPT-4, InstructGPT, WebGPT, Codex, CLIP, DALL-E, DALL-E 2, DALL-E 3, Jukebox, and Whisper among other models.
Given that developers are not guaranteed to provide uniform transparency for each foundation model (\eg OpenAI releases the weights openly for some of these models but not others), we decide to assess developers in relation to their \textit{flagship} foundation model.
By flagship foundation model, we mean the foundation model that is most salient and/or capable from the developer based on our judgment, which is directly informed by the company's public description of the model.
We provide basic information about each of the developers and their flagship model in \autoref{tab:developer-info}.\footnote{For OpenAI, we evaluate GPT-4, which was released in March 2023, not GPT-4V, a model OpenAI released in September 2023 after we completed the 2023 FMTI. With respect to input and output modality, \citet{openai2023gpt4} states that GPT-4 is "a large multimodal model capable of processing image and text inputs and producing text outputs."}

\subsection{Scoring}
By selecting the indicators and companies, we abstractly specify the form of the index.
By defining each indicator and designating the flagship foundation model to be assessed for each developer, we move to a more precise operationalization. 
To make the index fully precise, we describe how we sourced the information that was used to assess each developer on each indicator, resulting in the final scores.

\paragraph{Search protocol.}
To source information that we use to score developers, we exclusively use publicly available information provided by developers themselves.
We recognize that this information may be incomplete (\eg clients or governments may have greater access to information from the developer), but given that our focus includes public accountability, and we are academic researchers, we choose to consider only publicly available information.
Given that public information may change, we use information available as of \informationfreezedate.

For each developer, we initially compile a basic set of resources disclosed by the developer about their model development practices and their flagship foundation model.
To gather information for a specific indicator, we perform a structured search to identify all relevant information that is public.

\paragraph{Initial scoring.}
Having identified the information basis for scoring an indicator, \numgraders researchers on the team independently scored the developer on the indicator.
This entails specifying a \textit{score} (\ie 0 or 1), \textit{source} used in arriving at that score (\eg one or more web pages), and a textual \textit{justification} for how the evidence from sources is weighed against the criteria for the indicator in determining the score. 
Given these initial score assignments, the researchers reviewed their scores to identify any errors. 

Binary scoring provided several advantages. First, it simplified the scoring process by allowing researchers to focus on the sharp distinction between 0 and 1 point for each indicator. 
Second, a narrow criterion for making a binary scoring decision for each indicator reduced subjectivity in the initial scoring. 
Third, by reducing the level of complexity of each indicator we were able to reduce overlap between indicators, ensuring that we assess distinct dimensions of transparency.
At the same time, binary scoring limits the level of complexity of each indicator, potentially leaving out valuable information that can be captured by more complex scoring schemes \citep[\cf][]{bommasani2023eu-ai-act}.

In some instances, the researchers responsible for the same (indicator, developer) pair arrived at different scores, indicating disagreement. 
Given the systematic information gathering process, the iterative refinement of indicator definitions, and the binary scoring scheme, we found that disagreements were fairly infrequent.
Disagreements generally related to relevant information being erroneously neglected by one researcher or differences in the fine-grained interpretation of how to score an indicator.
Overall, across all \numindicators $\times$ \numcompanies (indicator, developer) pairs, the agreement rate was 85.2\% \citep[Cohen's $\kappa = 0.67$, indicating substantial agreement;][]{landis1977agreement}. 
To resolve disagreements, the researchers discussed and jointly came to a resolution.
Following the disagreement resolution, the scores were finalized and sources and justifications were merged to yield an initial set of \numcells (score, source, justification) triples for all \numcells (indicator, developer) pairs. \clearpage

\paragraph{Company feedback.}
Given that these scores constitute a direct assessment of specific companies, we engaged these companies to provide them with the opportunity to review, respond, and potentially rebut or contest the scores we assigned. 
Concretely, we contacted leaders at each of the companies with (i) a description of the \projectname, 
(ii) the \numindicators indicators and their definitions, and (iii) their \numindicators (score, source, justification) triples. 
We encouraged each company to review our scores, provide any general feedback and, especially, to directly contest any scores the company viewed as incorrect (by referencing public information available as of \informationfreezedate).
Companies were provided two business weeks to respond with clear assurance that all correspondence would be strictly private. 

Of the \numcompanies companies, all 10 responded.
Of these, 8 companies (\amazon, \anthropic, \cohere, \huggingface, \inflection, \meta, \openai, \stability) provided rebuttals for specific scores, which we extensively reviewed. 
In most cases, we did not change scores, though some rebuttals led to improvements in the scores (an average increase of 1.25 points across the 8 developers that contested on average 8.75 scores).
Rather than improving developers' scores, these rebuttals often revealed misunderstandings regarding definitions of indicators or our justifications for scores, leading to more robust definitions and justifications.
Beyond the scores, several companies scheduled calls with us or provided broader forms of feedback, which provided insight regarding how they conceptualize best practices for transparency and responsible AI.
Following company feedback, we again verified all scores, sources, and justifications that constitute the finalized materials used throughout this paper and made publicly available. 

We also notified the companies prior to the release of this paper, responding to their feedback. 
In addition, we encouraged companies to provide a public written response regarding their perspective on this initiative, their specific scores, and their broader approach as an organization to transparency and responsible AI as it relates to foundation models.

\section{Results}

The finalized results of the 2023 Foundation Model Transparency Index are the scores for each of the \numindicators indicators across all \numcompanies companies.
Here, we specifically consider overarching trends in the results, along with more specific trends based on the structure of the index.
Namely, we analyze along the rows/indicators (\eg domains), the columns/companies (\eg release strategy), as well as data-driven trends (\eg correlations).

\subsection{Overarching results}

\begin{figure}
\centering
\includegraphics[width=\textwidth]{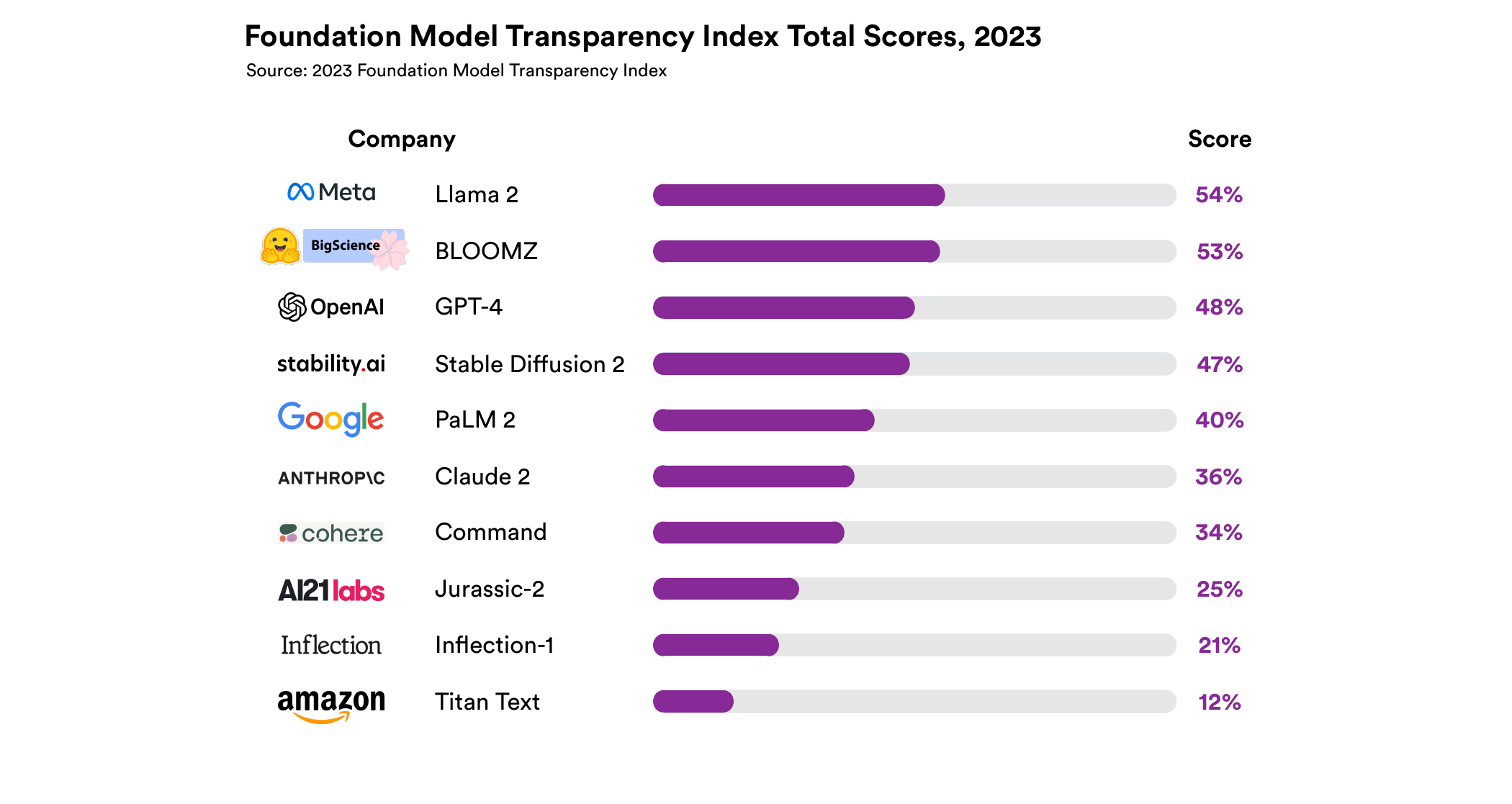}
\caption{\textbf{Overall 2023 FMTI scores.} The overall 2023 Foundation Model Transparency Index score and ranking across all \numindicators indicators.}
\label{fig:overall-scores}
\end{figure}

\begin{figure}
\centering
\includegraphics[keepaspectratio, height=\textheight, width=\textwidth]{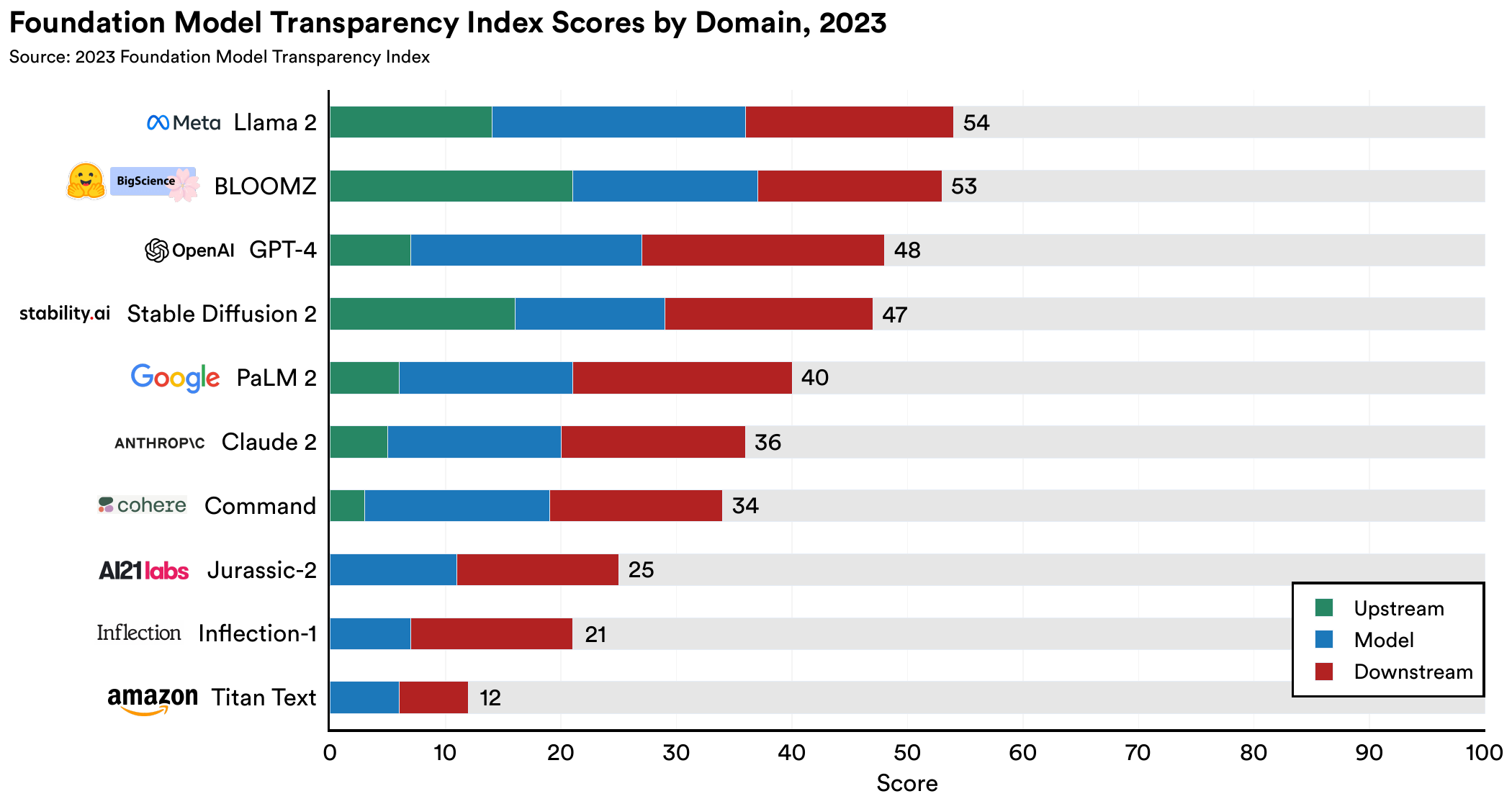}
\caption{\textbf{Scores by domain.} The aggregate 2023 FMTI score of each developer broken down by the three domains: upstream, model, and downstream.
}
\label{fig:domain-scores}
\end{figure}

\begin{figure}
\centering
\includegraphics[width=\textwidth]{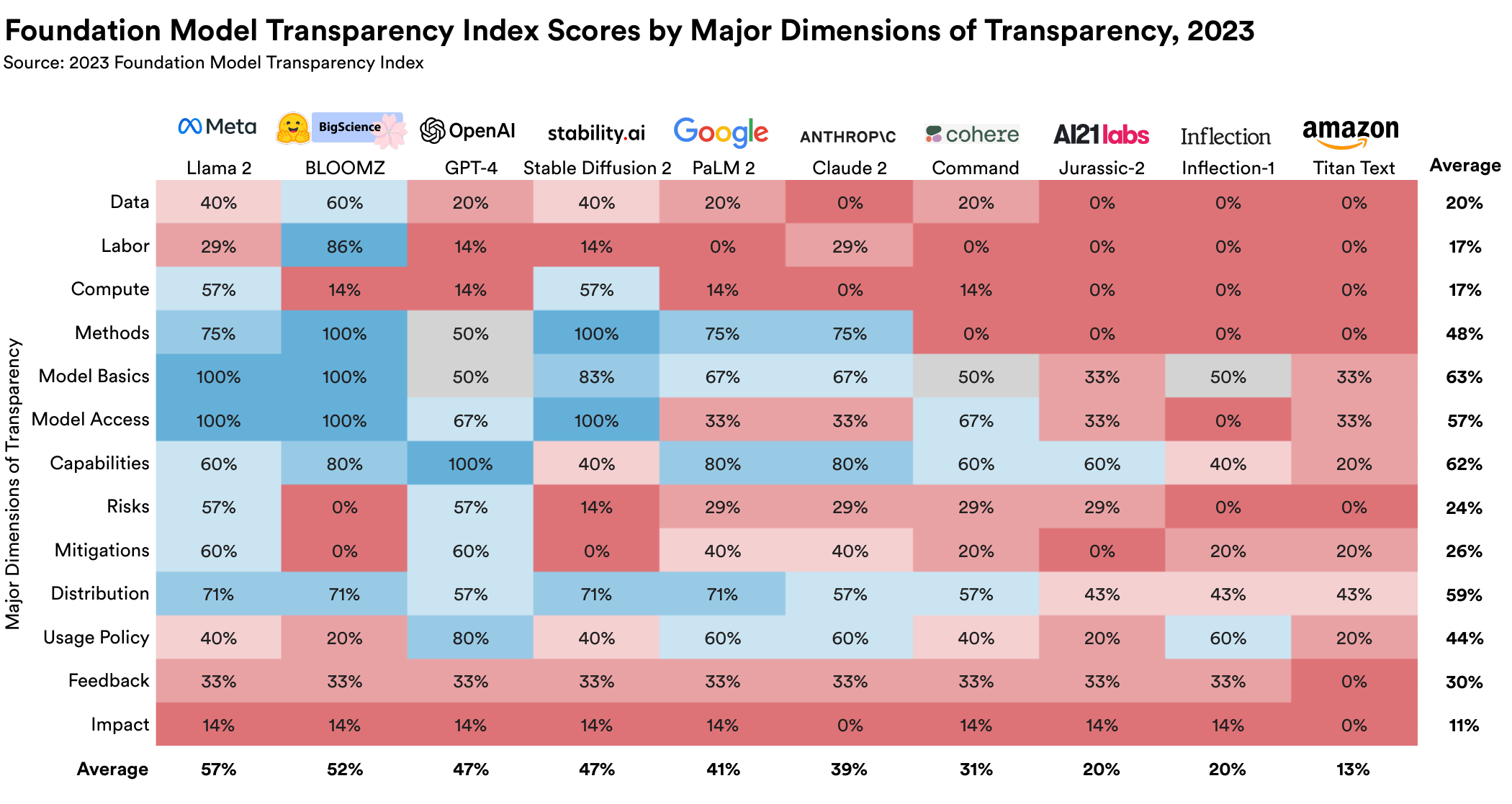}
\caption{\textbf{Scores by major dimensions of transparency.} 
The fraction of achieved indicators in each of the \nummajorsubdomains major dimension of transparency in the 2023 FMTI. 
Major dimension of transparency are large subdomains within the \numsubdomains subdomains.}
\label{fig:major-subdomain-scores}
\end{figure}

We begin our analysis by first establishing the broad trends when viewing the index as a whole.
We consider results aggregated at the level of a single overall score per company (\autoref{fig:overall-scores}) as well as the scores broken down into the \numdomains domains (upstream, model, downstream; \autoref{fig:domain-scores}).
We supplement our findings on these overarching trends with a more granular consideration of the \textit{major dimensions of transparency} in the index in \autoref{fig:major-subdomain-scores}.\footnote{
The major dimensions of transparency we highlight are \nummajorsubdomains large subdomains among the \numsubdomains subdomains. }

\paragraph{All developers have significant room for improvement. But most transparency indicators are very obtainable, having been implemented by at least one developer.}
Based on \autoref{fig:overall-scores}, the highest-scoring developer scores points for \maxscore of the \numindicators indicators, and the average score across all developers is \meanscore.
This establishes a pervasive lack of transparency across major foundation model developers.
With that said, for \numfeasible of the \numindicators indicators, there exists some developer that scores points, and of these there are \numfeasiblemultiple where multiple developers score points. 
Consequently, there is clear reason to believe that across all developers, the necessary change to become more transparent is feasible.
That companies' competitors are more transparent in certain issue areas suggests that such transparency, even if not fully costless, is unlikely to cause serious damage to their business.
Companies can emulate the higher level of transparency their competitors exhibit on certain indicators, providing a precedent and a starting point for improving transparency in the foundation model ecosystem. 

\paragraph{Developers show significant variance in overall transparency scores.}
While all developers have significant room for improvement, the current transparency of developers is strikingly uneven. 
Namely, the range in overall scores is \scorerange between the highest-scoring \meta at \maxscore and the lowest-scoring \amazon at \minscore.
Even excluding \amazon's score as especially low, we still see an effective range of 30 points between \meta and the next lowest \inflection.
Overall, with respect to the mean of \meanscore, the standard deviation is \stdev, which is quite substantial.
The four top-scoring developers (\meta, \huggingface, \openai, \stability) all cluster well above the mean, the next three are very close to the mean (\google, \anthropic, \cohere), and the three lowest-scoring developers (\aitwentyone, \inflection, \amazon) are well below the mean.
In many cases, the lowest-scoring developers have clear opportunities for improvement through straightforward changes related to some of the least challenging indicators. 
Examples include improved documentation  (\eg change logs, versioning protocols, model cards, centralized documentation for downstream use), clearer language in corporate policies  (\eg usage policies, model behavior policies, deprecation policies), and disclosing additional information that is unlikely to have implications for business competitiveness or safety (\eg basic details on methods, dependencies, feedback).

\paragraph{The upstream domain sees the worst transparency scores.}
To gain additional insight beyond developers' basic overall scores, we consider scores broken down by the \numdomains top-level domains in \autoref{fig:domain-scores}.
On this basis, we see clear evidence that developers are, on average, least transparent with respect to the upstream resources required to build their models, such as data, labor, and compute. 
Concretely, the mean score on upstream indicators is 7.2 out of 32 (22.5\%), compared to 14.1 out of 33 (42.7\%) for model indicators and 15.7 out of 35 (44.9\%) for downstream indicators.
To confirm this is not overly biased by outliers, we note that the medians show the same trend: the median score on upstream indicators is 3.5, compared to 12.5 for model indicators and 16 for downstream indicators.
We specifically highlight that the four lowest-scoring developers overall (\autoref{fig:overall-scores}) also fare the worst on the upstream domain (\autoref{fig:domain-scores}), with \cohere receiving 3 points and all of \aitwentyone, \inflection, and \amazon receiving 0 points.
In contrast, for both the model and downstream domains, all \numcompanies companies receive at least 6 points. 

\paragraph{Domain-level discrepancies explain some of the differences between companies with similar overall scores.}
We partition the \numcompanies companies into three groups based on whether their overall score (\autoref{fig:overall-scores}) is well-above (\meta, \huggingface, \openai, \stability), around (\google, \anthropic, \cohere), or well-below (\aitwentyone, \inflection, \amazon) the mean. 
Within these groups, while companies receive somewhat similar scores, we find that their domain-level scores clarify discrepancies between them. 
Among the highest scorers, \openai is considerably less transparent on upstream matters (7) as compared to the other three high-scoring companies (\meta with 14, \huggingface with 21, \stability with 16).
In particular, \openai and \stability receive the nearly the same overall score, with \openai making up the deficit to \stability on upstream transparency mostly through better model-level transparency (and, specifically, many of the indicators on evaluations and risks).
For the middle category of \google, \anthropic, and \cohere, the discrepancies are less stark, but we do see that \cohere is at 3 in the upstream category compared to \google with 6 and \anthropic with 5.
Among the three lowest-scoring developers, we see that \aitwentyone and \inflection are differentiated by the model domain, with both scoring a zero on the upstream domain and similarly on the downstream domain.

\paragraph{\data, \labor, and \compute are pervasive blind spots across developers.} 
While the overall and domain-level results provide a basic lay of the land, we find that the major dimensions of transparency provide the Goldilocks region for clear and incisive analysis as shown in \autoref{fig:major-subdomain-scores}.
In particular, these dimensions of transparency are subdomains with several indicators (so the subdomain scores are more reliable) that are tied to broadly-understandable concepts like labor and capabilities. 
We hone in on the following major dimensions of transparency: \data, \labor, \compute, \methods, \modelbasics, \modelaccess, \capabilities, \risks, \modelmitigations, \distribution, \usagepolicy, \modelbehaviorpolicy, \updates, \dataprotection, \feedback, and \impact. 
Analysis at this level reveals actionable insight into what types of transparency or opacity lead to many of our top findings. 
For example, we find that the poor upstream transparency stems from low performance on the \data, \labor, and \compute subdomains; developers average just 20\%, 17\%, and 17\% for \data, \labor, and \compute respectively. 
In terms of smaller subdomains, developers on average score 25\% of the available points on \datamitigations.

\paragraph{\modelbasics, \capabilities, \limitations, and \dataprotection are the most transparent subdomains at present, but still short of the ideal.}
Developers score the highest proportion of points on indicators related to the following subdomains: \interface (85\%), \documentation (70\%), \dataprotection (67\%), \modelbasics (63\%), and \updates (63\%).
This reflects some baseline level of transparency across developers with respect to notifying users they are interacting with AI systems, providing centralized documentation for downstream use, publishing data protection policies, and disclosing the modalities associated with their model. 
Still, there are gaps even for these subdomains. 
No developer provides a protocol for accessing usage data.
Most developers (8 of 10) do not disclose the size of their model.
And only half of the developers provide any form of deprecation policy.

\subsection{Upstream results}

\begin{figure}
\centering
\includegraphics[keepaspectratio, width=0.7\pdfpagewidth]{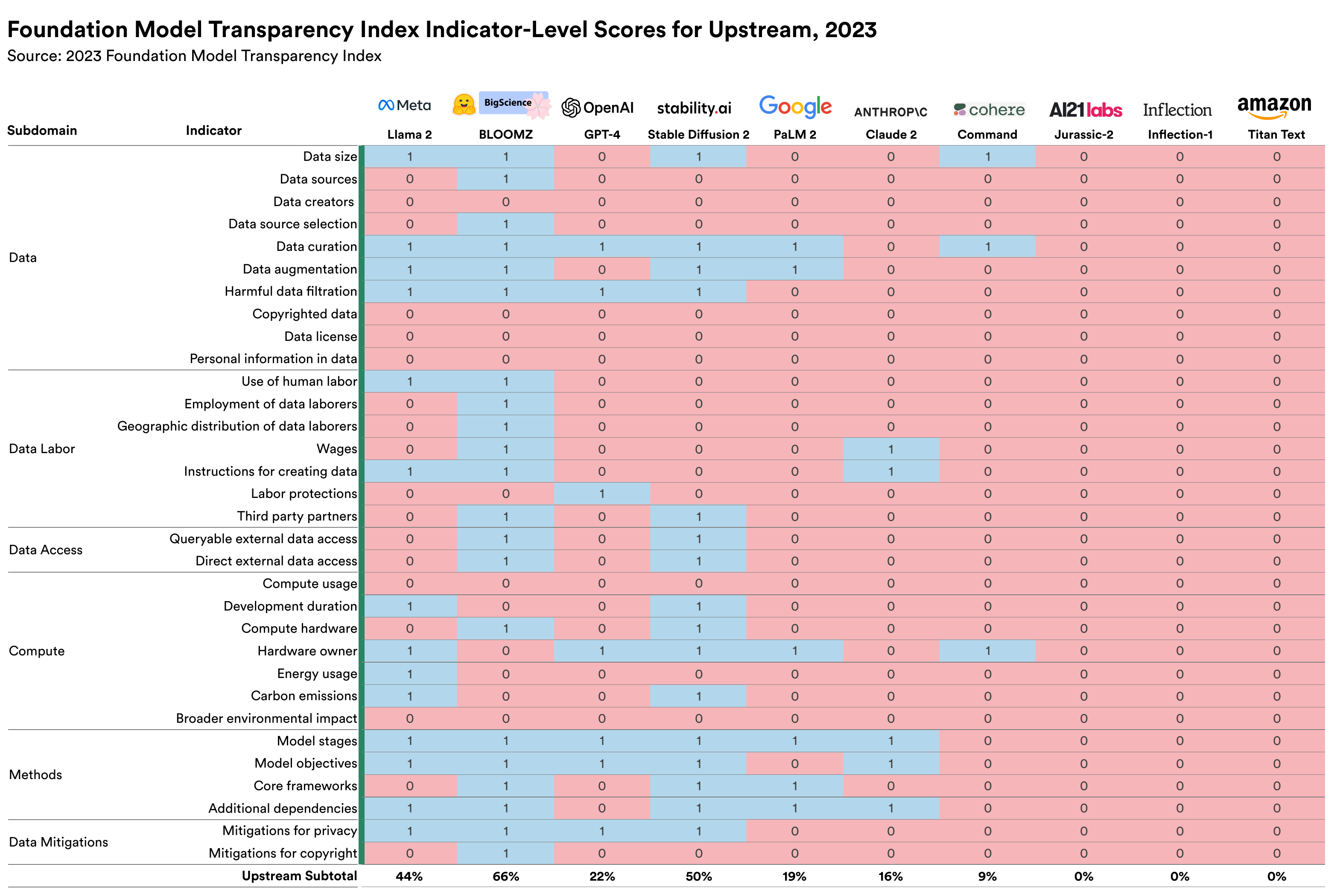}
\caption{\textbf{Upstream scores by indicator.} The 2023 FMTI scores for each of the \numupstreamindicators upstream indicators.
}
\label{fig:upstream-scores}
\end{figure}

Upstream indicators assess transparency regarding the ingredients that go into the foundation model including data, labor, compute, methods, and code. 
These ingredients are important predictors of the capabilities and risks of the foundation model they produce, as well as externalities of the model development process (\eg impacts on human laborers and the environment). 
As we show in \autoref{fig:domain-scores}, the upstream indicators are the most sparsely awarded (22.5\% coverage on average).
Here, we analyze at the level of subdomains and indicators based on \autoref{fig:upstream-scores}. 

\paragraph{The upstream domain shows the greatest spread.}
Building on the fact that developers score worst on the upstream domain--with several developers scoring exactly or nearly 0 points--we find the range in scores is the greatest for this domain. Namely, only one developer (\huggingface) scores more than half of the indicators (21 of the available \numupstreamindicators indicators; 65.6\%), yielding a range of 21 when compared to the lowest-scoring developers: \aitwentyone, \inflection, and \amazon (0 of the available \numupstreamindicators indicators; 0\%).
We emphasize this striking disparity given that many of the fundamental societal issues in connection with foundation models relate to upstream resources: bias, copyright, and privacy in relation to data, worker protections and fair compensation in relation to labor, environmental impact and energy expenditure in relation to compute, reproducibility in relation to methods, and cybersecurity in relation to code. 

\paragraph{The \methods subdomain is the most transparent in aggregate, while \labor is the least transparent.} 
Among the upstream subdomains, only \methods shows some degree of coverage, with six of the developers giving some description of training stages, training objectives, and dependencies.
On the other end of the spectrum, \labor sees little to no coverage with the exception of \bloomz, which involved volunteers providing data.
Developers generally share no information about the use of human labor in their data pipeline, the employer, wages, and geographical distribution of these workers, instructions they give to data annotators, or any labor protections they implement.
This industry norm of being non transparent with respect to data labor is in tension with the fact that such information is critical to reinforcement learning with human feedback \citep{ziegler2019finetuning, ouyang2022instructions, casper2023open}.
That data labor is one of the two least transparent subdomains is consistent with prior work documenting widespread ethical challenges with data labor \citep{gray2019ghost, crawford2021atlas, hao2023cleaning}. 

\paragraph{The \compute subdomain shows major discrepancies among developers.} 
\meta and \stability document some aspects of compute, energy, and hardware usage, as well as the carbon footprint of model development, whereas many developers do not. 
Given the significant compute expenditure required to build many foundation models, the practice of documenting energy use and environmental impact is well-established along with associated tooling to measure these quantities \citep{lacoste2019quantifying,strubell2019energy,schwartz2020green,luccioni2023counting}. 
In spite of this, most developers do not disclose minimal, or sometimes any, details related to compute usage, particularly with respect to energy usage, carbon footprint, and environmental impact.

The broader environmental impact of building foundation models is also essential to consider; although there has been significant public attention concerning energy expenditure, other matters such as water usage may be of similar consequence environmentally \citep{luccioni2023counting}.
\citet{luccioni2022estimating} provides an excellent example, documenting the embodied emissions, dynamic consumption, and idle consumption associated with BLOOM \citep{scao2022bloom}.
Given that \bloomz is derived from BLOOM, we note the potential for \textit{documentation transience}, where prior documentation is not updated to reflect substantial changes and, therefore, does not correctly persist to the new asset. 
In particular, the additional broader environmental impact of deriving \bloomz from BLOOM is not disclosed.

\paragraph{Widespread lack of upstream transparency on data creators, data license, copyrighted data and associated mitigations, and broader environmental impact.} 
Of the \numupstreamindicators indicators, no company scores points on six of them.
These are the indicators for data creators, data license status, copyrighted data, copyright mitigations, compute usage and broader environmental impact.
For data creators, in part we believe this reflects the nascent status of methods for providing web-scale understanding of who created the data (\eg text, images) scraped from the Internet. 
However, we recognize that \huggingface in particular has taken important steps to characterize aspects of who created the data, along with associated metadata for copyright, license, and personal information, for the ROOTS corpus used to build BLOOM (though not the additional data involved in building \bloomz). 
With respect to the copyrighted data and data license status indicators, we emphasize that information related to these indicators  is at issue in ongoing litigation. 
In particular, \stability has explicitly argued that training foundation models on copyrighted data is protected by fair use doctrine in the U.S.\footnote{See \url{https://www.judiciary.senate.gov/imo/media/doc/2023-07-12_pm_-_testimony_-_brooks.pdf} and \url{https://www.documentcloud.org/documents/23589439-openai-motion-to-dismiss} as well as \citet{lemley2020fair}.} 
Closed developers may also view information related to their data as a key competitive advantage, or be disincentivized to share this information due to a perception of legal risk.
Additionally, we note that we are surprised no developer directly discloses the compute usage in FLOPs to sufficient precision, though several disclose information that could be used to compute an estimate or upper bound. 



\begin{figure}
\centering
\includegraphics[keepaspectratio, width=0.7\pdfpagewidth]{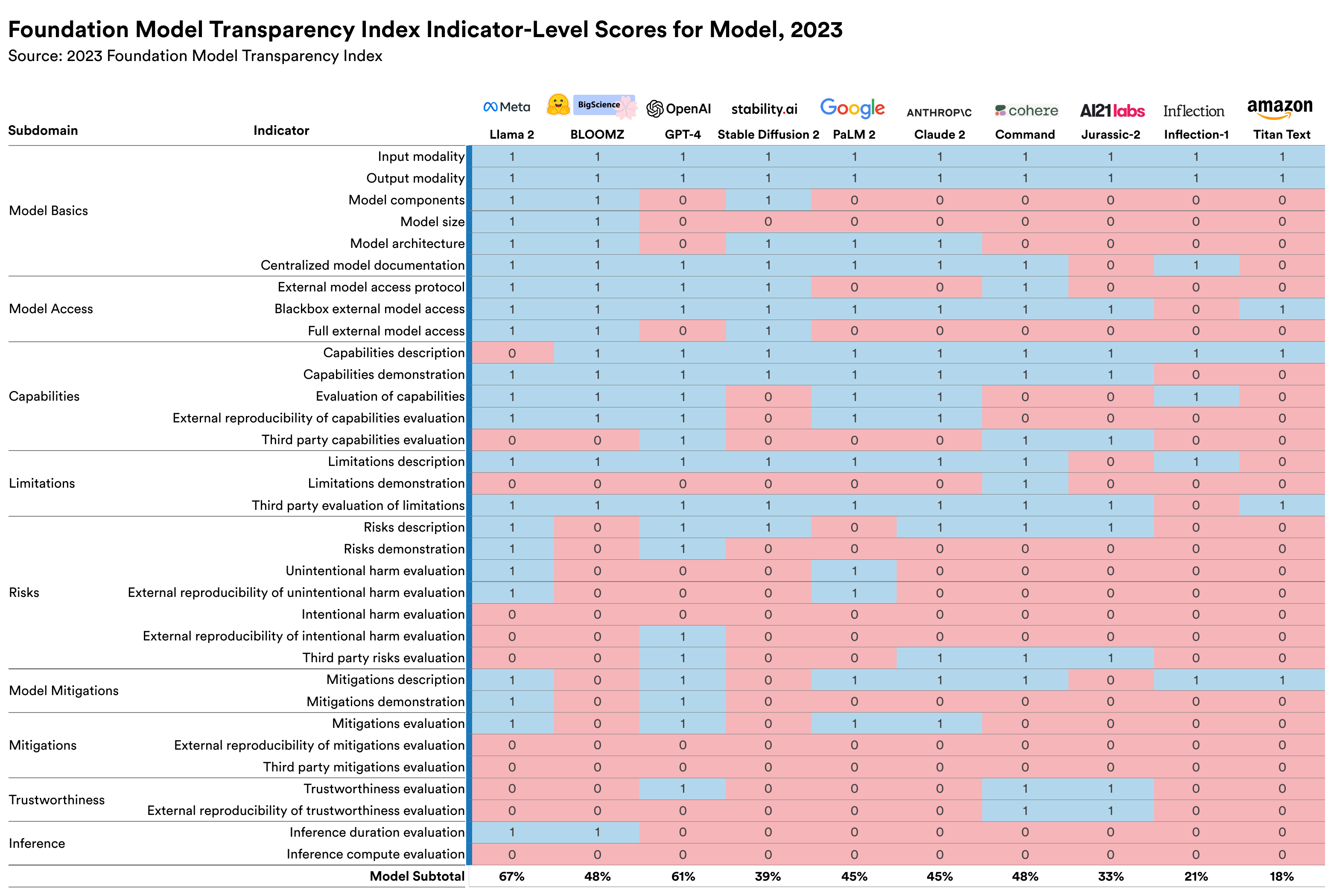}
\caption{\textbf{Model scores by indicator.} The 2023 FMTI scores for each of the \nummodelindicators model indicators.
}
\label{fig:model-scores}
\end{figure}

\paragraph{No upstream indicators are satisfied by all developers.}
At the indicator level, there is no upstream indicator for which every developer receives points.
Of course, this is guaranteed by the presence of (multiple) developers that score 0 points on the entire upstream domain.
Even putting these 3 developers aside, there is no indicator that is satisfied by all of the remaining 7.
The indicators where the greatest number of developers score points are data curation (all but \anthropic) and model stages (all but \cohere), which both suggest that developers are generally willing to describe the basics of the overall pipeline of model development.
With that said, we take the absence of any upstream indicator where all companies score points, and the fact that 5 or more developers score no points on 30 of \numupstreamindicators upstream indicators, as strong evidence that upstream transparency is the domain with the broadest room for improvement.

\subsection{Model results}

Model indicators assess transparency regarding the function of foundation models, spanning model access, capabilities, risks, limitations, mitigations, trustworthiness and inference efficiency, as well as basic information about the model. 
The indicators in this domain comprehensively characterize the foundation model as a standalone artifact: what tasks the model can and cannot perform, what is the model's basic structure, who has access to the model, and more. 
Here, we analyze developers at the level of subdomains and indicators based on \autoref{fig:model-scores}.

\paragraph{Model subdomains are some of the highest-scoring across the index.}
Overall, the mean score on model indicators is 14.1 out of 33 (42.7\%) and the median developer receives 12.5 points (37.9\%). 
With this in mind, several of the highest-scoring subdomains belong to the model domain.
Developers score best on \modelbasics (63\%), \capabilities (62\%), \limitations (60\%), and \modelaccess (57\%) within the domain.
These scores arise partially because of very generous indicators within these subdomains (\eg input modality, output modality, description of capabilities, description of limitations).

\paragraph{Transparency on capabilities does not translate to transparency on limitations, risks, or mitigations.}
Of the 33 model indicators, 20 are in the \capabilities, \limitations, \risks, and \modelmitigations subdomains.
Within these subdomains, \capabilities is clearly the most transparent subdomain: nearly all developers provide descriptions (9 of 10) and demonstrations (8 of 10) of multiple model capabilities, with the majority reporting evaluations (6 of 10), half reporting reproducible evaluations (5 of 10), and few providing third party evaluations (3 of 10).
In general, we see a decline in the number of developers who score the point from the most rudimentary (\ie description) to the most substantive (\ie third party evaluations) across these four subdomains.
With respect to \capabilities, while we assume most or all developers conduct internal evaluations, they may not score points on evaluations indicators because (i) they do not disclose sufficient details about internal evaluations for these evaluations to be externally reproducible, (ii) they do not assess multiple capabilities, or (iii) they do not report the results of the evaluations, perhaps due to a concern that a model may underperform competitors' models. 

With this in mind, developers consistently score worse on \limitations, \risks, and \modelmitigations indicators than on \capabilities.
For example, only \cohere receive points for demonstrating limitations, while 8 developers score points for demonstrating capabilities. 
These asymmetries where companies are more willing to share information about capabilities than limitations, risks, and mitigations are concerning, as they may lead to an inflated sense of trust in companies' foundation models. 
In fact, these asymmetries are especially pronounced for \risks (average score of 24\%) and \modelmitigations (average score of 26\%), given that these scores are considerably worse than the average scores for \capabilities (62\%) and \limitations (60\%). 

\paragraph{Developers score poorly on \trustworthiness, largely in line with \risks and \modelmitigations.}
With respect to the \trustworthiness subdomain, only \openai, \cohere, and \aitwentyone provide information about rigorous evaluations of their flagship model related to robustness, reliability, hallucinations, calibration, or explainability.
Of those developers, only \cohere and \aitwentyone provide sufficient detail for their evaluations to be deemed externally reproducible due to their use of the HELM benchmark \citep{liang2023holistic}, compared to \openai's unclear description of their evaluations of model calibration.
Given the previous asymmetry we establish around greater disclosure of capabilities as compared to limitations, risks, and mitigations, the absence of trustworthiness evaluations exacerbates these concerns.
Put together, the lack of sufficient public information on limitations, risks, mitigations, and trustworthiness makes it more likely that consumers will not have well-calibrated expectations.
In turn, this could lead to undesirable overreliance on foundation models because not enough is done to calibrate consumers on the appropriate levels of trust \citep{parasuraman2010complacency}.\footnote{See \url{https://www.theverge.com/2023/5/30/23741996/openai-chatgpt-false-information-misinformation-responsibility} as an example.}
With this said, we do acknowledge that developers may take other routes towards improving trustworthiness including methods like reinforcement learning from human feedback \citep{ziegler2019finetuning, ouyang2022instructions} and constitutional AI \citep{bai2022constitutional}, though transparency is lacking on these approaches \citep{casper2023open}.

\paragraph{\modelaccess reveals slight differences beyond just release strategy.}
In aggregate, companies score 17 of the 30 points (57\%) in the \modelaccess subdomain across the 3 indicators and \numcompanies companies.
On the external model access protocol indicator, \meta, \huggingface, \openai, and \stability are the only developers to score points.
We find this particularly interesting given \meta, \huggingface and \stability release their models openly in terms of both model weights and data, whereas \openai is considerably more closed, providing only API access.
However, in particular, \openai has a clear researcher access program with a form to request access, criteria it discloses for granting access, and a period of 4--6 weeks disclosed as the expected turnaround for a decision.
This demonstrates that developers across the release spectrum \citep{solaiman2023gradient} may achieve transparency on some indicators while taking substantively different approaches.
In practice, we find that several closed developers have access forms that allow external entities greater access to the model, but these forms often lack key components of transparency that clarify the specific steps the developer will take to assess and grant applications (\eg in comparison to \openai's process).
With that said, the indicator for full external model access is exclusively achieved by the three open developers, though every developer other than \inflection provides black box access to its model.

\paragraph{\modelmitigations are a weak point for most developers.} 
Developers on average scored just 26\% of the total available points on the five \modelmitigations indicators. 
\huggingface, \stability, and \aitwentyone score 0 points, while \cohere, \inflection, and \amazon score only the point on mitigations description, which is the most lightweight of these indicators. 
In general, we highlight an important mismatch between the many risks that are enumerated and the relatively few mitigations that are described, implemented, and/or evaluated.
Even when mitigations are described, in scoring we find the mapping between stated risks and stated mitigations is often vague or nonexistent. 
Moving forward, we hope developers will directly aim mitigations at addressing specific risks, with appropriate evaluations to confirm the efficacy of mitigations in achieving the stated goals. 

\begin{figure}
\centering
\includegraphics[keepaspectratio, width=0.7\pdfpagewidth]{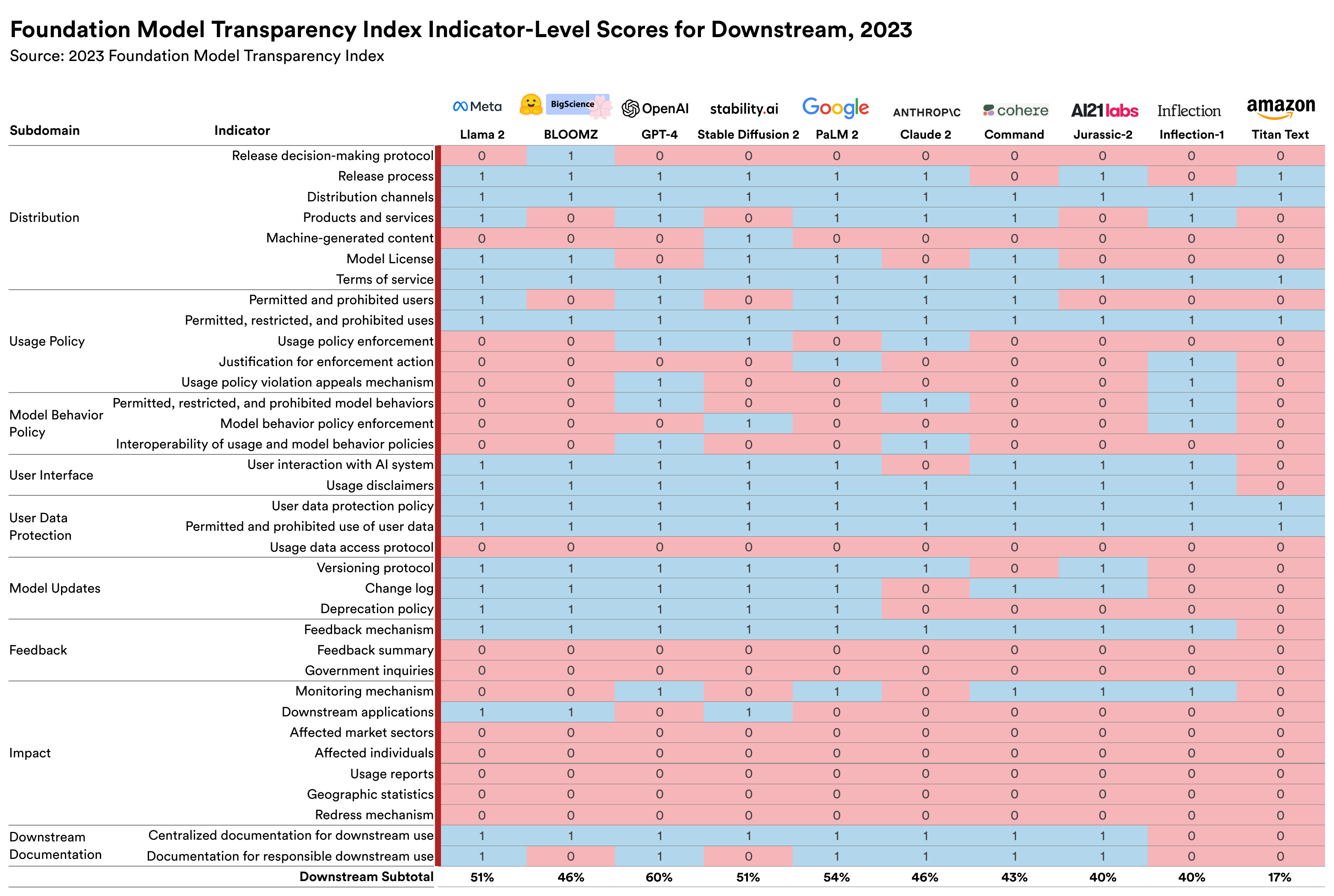}
\caption{\textbf{Downstream scores by indicator.} 
The 2023 FMTI scores for each of the \numdownstreamindicators downstream indicators.
}
\label{fig:downstream-scores}
\end{figure}

\paragraph{Most model indicators are scored by some developer, though most developers score poorly on indicators related to evaluating intentional harms, mitigations, and inference efficiency.}
Of the 33 indicators in the model domain, at least one developer scores a point on 29 of them. 
Further, multiple developers score points on 27 model indicators.
The 4 indicators for which no developer scores points are (i) intentional harm evaluation, (ii) external reproducibility of mitigations evaluations, (iii) third party mitigations evaluations, and (iv) inference compute evaluation.
The 2 additional indicators for which only one developer scores points are limitations demonstration (\cohere) and external reproducibility of internal harm evaluation (\openai).
While many companies describe risks (including the risk of intentional harms), they do not share sufficient information related to evaluations of intentional harm or the reproducibility of evaluations of mitigations.
In the case of inference, we believe standards are needed akin to MLPerf \citep{reddi2020mlperf} to rigorously benchmark the inference of foundation models \citep{narayanan2023cheaply} given the key role of efficient inference and low latency in the usability of models \citep{lee2023evaluating}.
We see that \bloomz in particular provides a potential benchmark for language models by tracking the time spent for a fixed task (generating 100 tokens given a 7 token prefix) on fixed hardware (a NVIDIA A100-80GB GPU), though compute is not measured.\footnote{See \url{https://huggingface.co/blog/habana-gaudi-2-bloom}.}

\subsection{Downstream results}
Downstream indicators assess transparency regarding the use of foundation models, spanning subdomains related to distribution, policies constraining the use and behavior of the model, user interfaces, user data protection, model updates, feedback, impact, and documentation. 
Indicators in these subdomains characterize transparency related to how the foundation model is deployed and its downstream effects on the ecosystem and society. 
Our analysis is based on publicly available information about how the foundation model is distributed, how it can and cannot be used, how users can give feedback and seek redress, broader societal impacts, and how the model affects actors downstream of the developer in the supply chain.
Here, we conduct a fine-grained analysis at the level of subdomains and indicators based on \autoref{fig:downstream-scores}.

\paragraph{Downstream scores show less spread across developers.}
Total scores on downstream indicators are tightly clustered around the mean of 15.7 out of 35, which corresponds to 44.9\% of the \numdownstreamindicators downstream indicators. 
With the exception of \amazon (6 out of \numdownstreamindicators; 17.1\%), the other nine developers all score between 14 and 21 points.
The highest-scoring on the downstream domain is \openai at 21 points and the lowest-scoring (barring \amazon) are \aitwentyone and \inflection at 14 points.

\paragraph{\impact is the least transparent subdomain in the entire index.}
To clarify the downstream impact of a given foundation model, the \impact subdomain includes indicators on monitoring mechanisms, affected market sectors, affected individuals, usage reports, geographic statistics, and redress mechanisms.
Strikingly, the mean score across all developers on this subdomain is just 11\%, with 8 developers scoring points on just 1 of the possible 7 indicators and the remaining 2 scoring none of the indicators. 
No developer scores points on affected market sectors, affected individuals, usage reports, geographic statistics, or redress mechanism.
This means that there is essentially no information about how many people, sectors, and regions foundation models are impacting. 
\openai, \google, \cohere, \aitwentyone, and \inflection are the only developers to disclose a potential monitoring mechanism for tracking model use. 
And only open foundation model developers share limited information about downstream applications, whereas the rest provide no information.\footnote{We score the downstream applications indicator quite generously: all of the open developers score points because they discloses which Hugging Face "Spaces" are also using the model via Hugging Face's platform.
However, we emphasize that this is still a poor proxy for the number of applications dependent on the foundation model.} 

\paragraph{Developers are significantly more transparent about \distribution than other major dimensions of (downstream) transparency.}
Across the four major dimensions of transparency in the downstream domain (\distribution, \usagepolicy, \feedback, \impact), mean scores are on the higher end only for \distribution at 59\%, with the other three all below 50\%. 
Every developer shares information about distribution channels, or the pathways by which the model is made available to entities beyond the model developer organization.
Every developer provides terms of service that cover the distribution of its foundation model.\footnote{As with several downstream indicators, we assessed the terms of service of the primary distribution channel. For example, this meant that we assessed Microsoft Azure's terms of service for Meta.}
Most developers share information about their process for releasing their flagship model (8 of 10) as well as the developer's products and services that use the foundation model (6 of 10).
Half of developers share information about the license under which the model is distributed.

\paragraph{In spite of broad transparency on the \distribution subdomain, developers are highly opaque around release decisions.}
Within the \distribution subdomain, developers score poorly on the release decision-making protocol indicator; \huggingface is the only developer that shares information about its decision-making protocol for release.
Although there has been an extensive focus on release strategies in the literature on foundation models \citep{solaiman2019release, sastry2021release, shevlane2022structured, liang2022community-norms, liang2022condemning, solaiman2023gradient, widder2023open, seger2023open}, developers across the release spectrum share very little information about how and why they release their flagship models. 
In particular, we highlight that many of companies we assess have written about the broader topic of release, but not in a way that is precise to their specific decisions for their flagship models.\footnote{We note that following \informationfreezedate, \anthropic released information about its approach to responsible scaling: \url{https://www.anthropic.com/index/anthropics-responsible-scaling-policy}.}

\paragraph{\usagepolicy and \modelbehaviorpolicy subdomain scores are uneven across developers.}
Scores on the \usagepolicy subdomain are uneven, with all developers scoring points on the indicator for permitted, restricted, and prohibited uses, but only two (\openai and \inflection) scoring points on the usage policy violation appeals indicator.
This reflects the lack of industry standards regarding precisely how foundation model developers should restrict the use of their models. 
We found that different developers provide this information in different types of documentation, ranging from standalone Acceptable Use Policies to Content Policies to terms in the model license, and that many developers share some of this information in several different documents. 

While developers did provide some transparency on usage policies related to a user's obligations, they did not provide a similar level of transparency on the restrictions they place on their model's behavior. 
Scores on indicators in the \modelbehaviorpolicy subdomain were relatively weaker, with a mean across the 3 indicators of 23\% compared to 44\% for the 5 usage policy indicators.
\openai, \anthropic, and \inflection are the only developers who provide information about permitted, restricted, and prohibited model behaviors, while only \inflection and \stability provide information about how they might enforce such restrictions.
\openai and \anthropic are the only developers who make clear how their models are expected to behave in the event that a user violates the usage policy. 
In part, we believe the norms and standards around model behavior are rather immature, meaning that developers do not provide a clear conceptualization of if/how they impose a model behavior policy.
For example, the role of modeling decisions (\eg the use of reinforcement learning from human feedback or constitutional AI) on behaviors (\eg model refusals to specific requests) are not made clear.

\paragraph{Identical scores on the \dataprotection subdomain across all developers.}
For the \dataprotection subdomain, scores are uniform across developers, with every developer scoring points on user data protection policy, as well as permitted and prohibited uses of user data.
However, no developer scores points on usage data access protocol. 
This may reflect that few, if any, companies actually share usage data externally, meaning companies may perceive that the need to develop protocols for sharing such data is limited. 
However, developers' data protection policies include many provisions that would allow them to share such usage data, and specific protocols for how and when they do so are not transparent.

\paragraph{Developers lack transparency on the \feedback subdomain.} 
Developers score relatively poorly on \feedback indicators, scoring only 30\% of the available points. 
While every developer but \amazon has a public mechanism for collecting feedback on its model, none provide information such as a feedback summary or details on government inquiries, such as requests for user data (which social media companies disclose).
This is likely a function of how nascent the foundation model ecosystem is: companies have only been collecting feedback for a few years, and it took social media companies several years to respond to public calls for transparency around the feedback they receive from users and governments.
Moving forward, more robust transparency reporting practices that provide the public with more information regarding these forms of feedback will likely be necessary.\footnote{For example, consider the EU's DSA Transparency Database, implemented on the basis of the Digital Services Act to provide transparency on content moderation decisions: \url{https://transparency.dsa.ec.europa.eu/}.}

\paragraph{Developers are fairly transparent on the \updates subdomain.}
5 of 10 developers provide clear information about their versioning protocol, change log, and deprecation policy.
\inflection and \amazon, however, score zero points on these indicators, which may be due in part due to the fact that \inflectionone and \titan are at an earlier stage of release than some other flagship models. 
While there is a wide variation in the type, specificity, and quality of documentation provided related to \updates, as with other indicators, we assess these metrics generously and allocate points on the basis of transparency alone. 

\paragraph{Developers score well on the \interface subdomain, though this may change due to deployments on mobile phones.}
Developers scored highly on \interface indicators (average score of 85\%), with more than half of developers scoring points on both indicators, which assess if users are told they are interacting with an AI system and if users are provided appropriate disclaimers. 
Developers frequently disclose to users that they are interacting with a specific foundation model by including the name of the foundation model somewhere in the user interface, while they give usage disclaimers upon sign-up for the user interface via a link to the terms of service or usage policy. 
Unlike all other indicators, we generally had to make use of step 7 in the and directly interact with developers' models via a user interface to assess these indicators. 
However, \amazon did not have a publicly available user interface in advance of \informationfreezedate, meaning that it could not receive these points. 
We initially assessed transparency of deployments on mobile devices in some cases, though we ultimately did not consider these deployments for scoring.
With that said, we highlight that the same standard for transparency of user interfaces does not currently appear to be met by mobile deployments from \openai and \inflection.
Overall, we believe in the importance of providing transparency through user interfaces as it can help foundation models avoid the formation of the "dark patterns" we have seen develop with other digital technologies \citep{10.1145/3359183}.
For example, we highlight that \anthropic does not make clear that a user is interacting with an AI system, except for the textual description "Message Claude."

\subsection{Results for open and closed developers}
\begin{figure}
\centering
\includegraphics[keepaspectratio, height=\textheight, width=\textwidth]{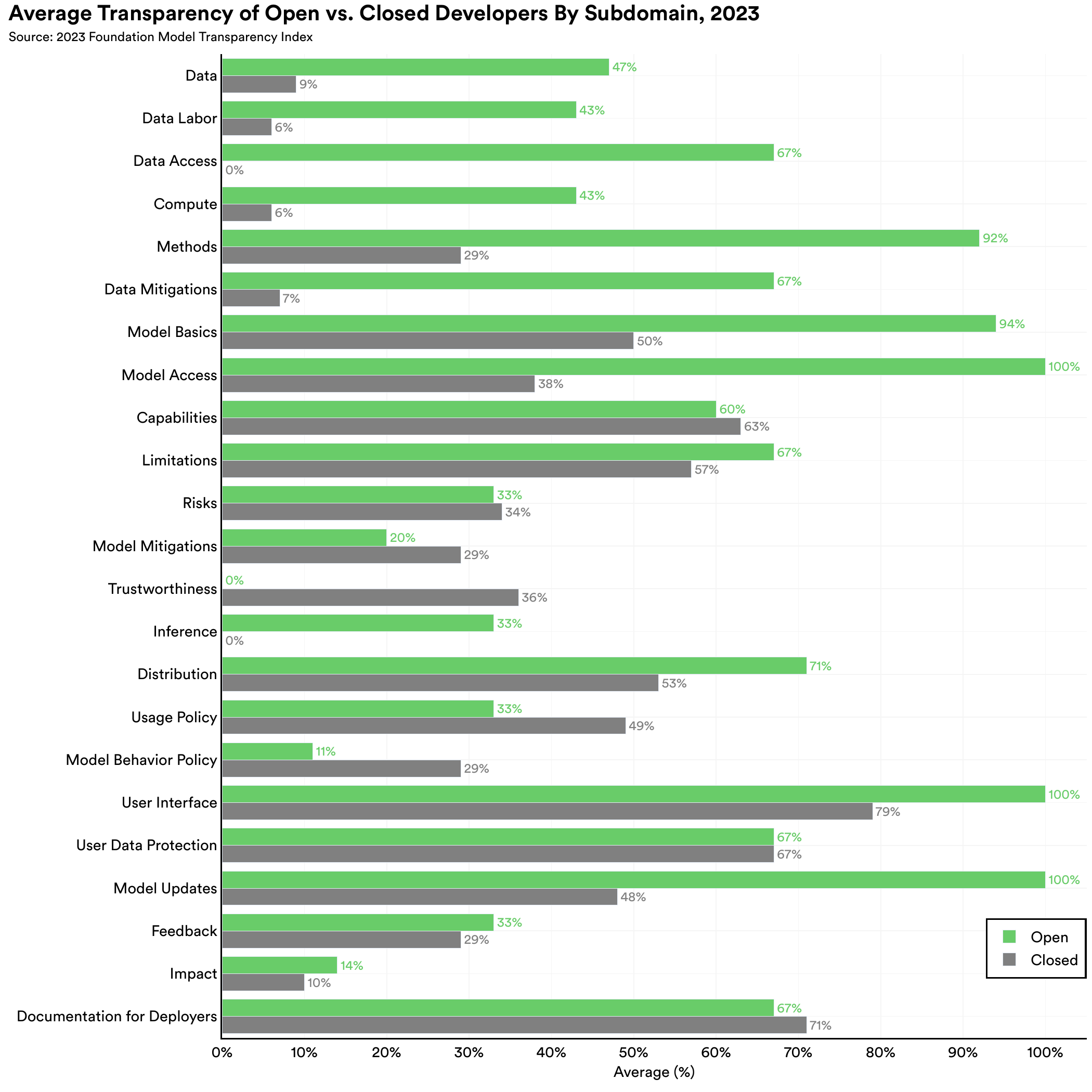}
\caption{\textbf{Open vs. closed by subdomain.} 
The 2023 FMTI average  score for the 3 open developers (\meta, \huggingface, \stability) and the 7 closed developers (\openai, \anthropic, \google, \cohere, \aitwentyone, \inflection, \amazon) across each of the 23 subdomains. Note: the number of indicators per subdomain varies widely.
}
\label{fig:open-closed}
\end{figure}

Foundation models are released by different developers using a variety of release strategies \citep{liang2022community-norms, solaiman2023gradient}.
In particular, we deliberately chose several developers that are more \textit{open} (\eg release the weights of their model, perhaps along with the data used to build the model)  and others that are more \textit{closed} (\eg only provide access via an API).
The topic of release and the (reductive) dichotomy of open vs. closed has emerged as a primary topic of technical and policy research on foundation models \citep{solaiman2019release, sastry2021release, shevlane2022structured, liang2022community-norms, liang2022condemning, solaiman2023gradient, widder2023open, seger2023open}.
To clarify how transparency differs between the open developers we assess (\ie \meta, \huggingface, \stability) and the closed developers (\ie \openai, \google, \anthropic, \cohere, \aitwentyone, \inflection, \amazon), we emphasize the distinction in \autoref{fig:open-closed}. 

\paragraph{Open developers score higher in aggregate and on every domain.}
We establish a clear trend that the open developers score higher overall, with all three being among the four highest-scoring developers (see \autoref{fig:overall-scores}).
In particular, every open developer is nearly at least as transparent in terms of aggregate score as the highest-scoring closed developer (\openai): \meta and \huggingface are at least 5 points higher, and \stability is within a point of \openai.
Further, this trend is established more strongly through domain-level analysis, where open developers score higher on average than closed developers across all domains (\ie upstream, model, downstream).
The mean score of open developers on upstream indicators is 53\% compared to 9\% for closed developers, 51\% for open developers on model indicators compared to 39\% for closed developers, and 49\% on downstream indicators compared to 43\% for closed developers.
To ensure these trends are robust to outliers, we highlight that the trends hold even when considering medians instead of means (upstream: 50\% to 9\%, model: 48\% to 45\%, downstream: 51\% to 43\%).

We emphasize that our findings confirm common hypotheses that open developers will in general be more transparent with respect to the upstream resources required to build their models (which also aligns with some making the data they use publicly available), but our findings dispute hypotheses that open developers will be less transparent on downstream matters due to their weaker control over downstream use.
While we believe that closed developers providing APIs are better positioned to collect information on the downstream use of their models, in practice these developers do not disclose this information to provide greater public transparency.

\paragraph{Open developers score higher on most subdomains.}
Open developers score higher than closed developers on 15 of the \numsubdomains subdomains, which account for 68 of the 100 indicators. 
The mean score of closed developers is higher than that of open developers on indicators in the subdomains of \capabilities, \risks, \modelmitigations, \trustworthiness, \usagepolicy, \modelbehaviorpolicy, and \documentation.
We highlight that these seven subdomains point to two broader themes: closed developers in some cases may be higher-resourced or face stronger incentives to proactively address certain matters around responsible AI (\eg \risks, \modelmitigations, \trustworthiness).
In addition, closed developers often have a closer coupling between the foundation model we assessed and downstream services, meaning that certain user-related aspects of transparency are potentially of higher priority (namely the \usagepolicy). 
For example, many closed developers provide products built on top of their flagship foundation model, providing users of their platforms and clients who license their proprietary foundation models with an opportunity to push for transparency.

The mean score of open developers is higher than closed developers on every upstream subdomain, with major score differentials especially for the \data, \compute, and \methods subdomains.
Looking at the difference in average scores by release strategy, we see large disparities in favor of open models in each domain, with the largest gaps for \dataaccess (67\% to 0\%), \methods (92\% to 29\%), and \datamitigations(67\% to 7\%).
We also observe similar large differentials (40\%+) for \modelbasics, \modelaccess, and \updates.
While less stark, we highlight the superior transparency on average for the \distribution subdomain as especially surprising given that closed developers maintain greater control over distribution by virtue of being closed.

\paragraph{Indicator-level analysis further demonstrates the disparity between open and closed developers.}
At the indicator level, the median open developer outscores the median closed developer on 28 indicators (18 upstream, 7 model, 3 downstream), while the median closed developer scores higher on just 6 indicators (0 upstream, 2 model, 4 downstream). 
The median open developer and the median closed developer both score points on 22 indicators and neither scores points on 44 indicators. 

\paragraph{The open developers we assessed provide greater transparency than their closed counterparts.}
Overall, each level of analysis points in the same direction: open developers are reliably more transparent.
In particular, we highlight that the release of assets (\eg model weights, data, code) may be significantly underweighted in terms of its broader transparency effects.
Our findings dispel the belief that closed developers are more likely to be transparent about downstream matters due to their greater control over deployment, while emphasizing that both open and closed developers continue to be extremely opaque in terms of the downstream impact of their foundation models.
With this in mind, we caution that our assessment is necessarily based on the practices of some of the highest-resourced open and closed developers, so these trends should not be taken as sufficient evidence to claim that all open developers are more transparent than closed developers.

\subsection{Correlations between companies}
\begin{figure}
\makebox[\textwidth][c]{\includegraphics[keepaspectratio, width=0.7\textwidth]{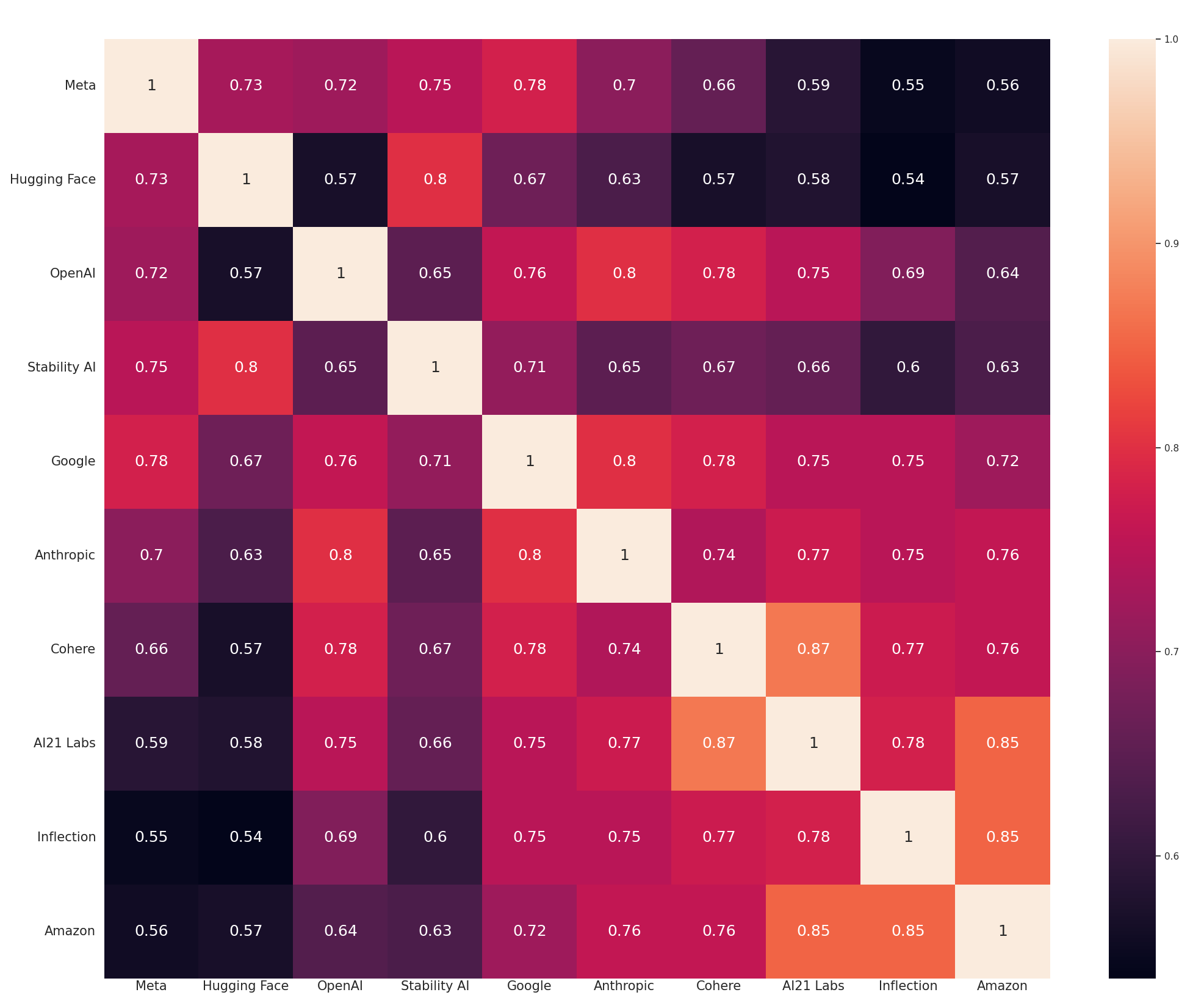}}
\caption{\textbf{Correlations between Companies.} The correlation between the 2023 FMTI scores for pairs of companies across all indicators. Correlation is measured using the simple matching coefficient (\ie agreement rate), which is the fraction of all indicators for which both companies receive the same score (\ie both receive the point or both do not receive the point).
}
\label{fig:overall-correlations}
\end{figure}
\paragraph{Measuring correlations.}
The \numindicators $\times$ \numcompanies scores introduces data-driven structure.
In particular, it clarifies relationships that arise in practice between different regions of the index.
Here, we consider the \textit{correlations}, in scores, focusing on company-to-company similarity for simplicity.
For example, if two companies receive similar aggregate scores, is this because they satisfy all the same indicators or do they score points on two very different sets of indicators?

In \autoref{fig:overall-correlations}, we plot the correlation between every pair of companies.
To measure correlation, we report the simple matching coefficient (SMC) or the agreement rate.
The SMC is the fraction of the \numindicators indicators for which both companies receive the same score (\ie both receive a zero or both receive a 1). 
As a result, a SMC of 0 indicates there is no indicator such that both companies receive the same score and a SMC of 1 indicates that for all indicators both companies receive the same score. 
For this reason, the correlation matrix is symmetric and guaranteed to be 1 on the diagonal. 

To systematically analyze the results, we consider three patterns in the correlation matrix:
(i) individual cells with very small or very large values (\ie highly similar or highly dissimilar company pairs),
(ii) individual rows with consistently small, consistently large, or highly varied values (\ie unusual companies),
and
(iii) structural patterns across the correlation matrix.

\paragraph{Strongly correlated company practices.}
In terms of the most correlated company pairs, we identify a few regions of the correlation matrix.
First, we identify the three most correlated pairs: 
(\cohere, \aitwentyone; SMC = 0.87),
(\aitwentyone, \amazon; SMC = 0.85),
and
(\inflection, \amazon; SMC = 0.85).
These pairs are all among the four lowest-scoring companies, though we note the inclusion of \cohere is interesting given \cohere's overall score (34) is closer to the average (37) and the middle-scoring group of companies (\ie including \google and \anthropic).
In addition to these pairs, if we consider the other highly-correlated pairs (SMC $\geq$ 0.8), we identify:
(\huggingface, \stability; SMC = 0.80),
(\openai, \anthropic; SMC = 0.80),
and
(\google, \anthropic; SMC = 0.80).
In particular, we observe that the company correlations identify clear structure: \huggingface and \stability are the only two developers to release both data and models openly, and the trio of \openai, \google, and \anthropic are the three members of the Frontier Model Forum that we assess.

\paragraph{Weakly correlated company practices.}
In contrast, we see that the least correlated pairs (SMC < 0.6) are pairs involving \meta and the three lowest-scoring developers as well as pairs involving \huggingface and five of the seven closed developers (\openai, \cohere, \aitwentyone, \inflection, \amazon).
These are all pairings between an open and a closed developer.
More broadly, we highlight that \meta is the sole developer that is not correlated with SMC at least 0.80 with any other developer, with the most similar other developer being Google at 0.78 (see below for further analysis). 
This means \meta is rather unique in terms of the indicators where it scores points; it is the sole developer that is not strongly correlated with any other company, even including the two other open developers.
Nevertheless, the least correlated pair of companies still agrees in over half the indicators (SMC = 0.54), which is not surprising given that all the companies are opaque (\eg if all the companies all scored 0 on every indicator, they would necessarily be perfectly correlated with SMC = 1).

\begin{figure}
\makebox[\textwidth][c]{\includegraphics[keepaspectratio, width=0.7\textwidth]{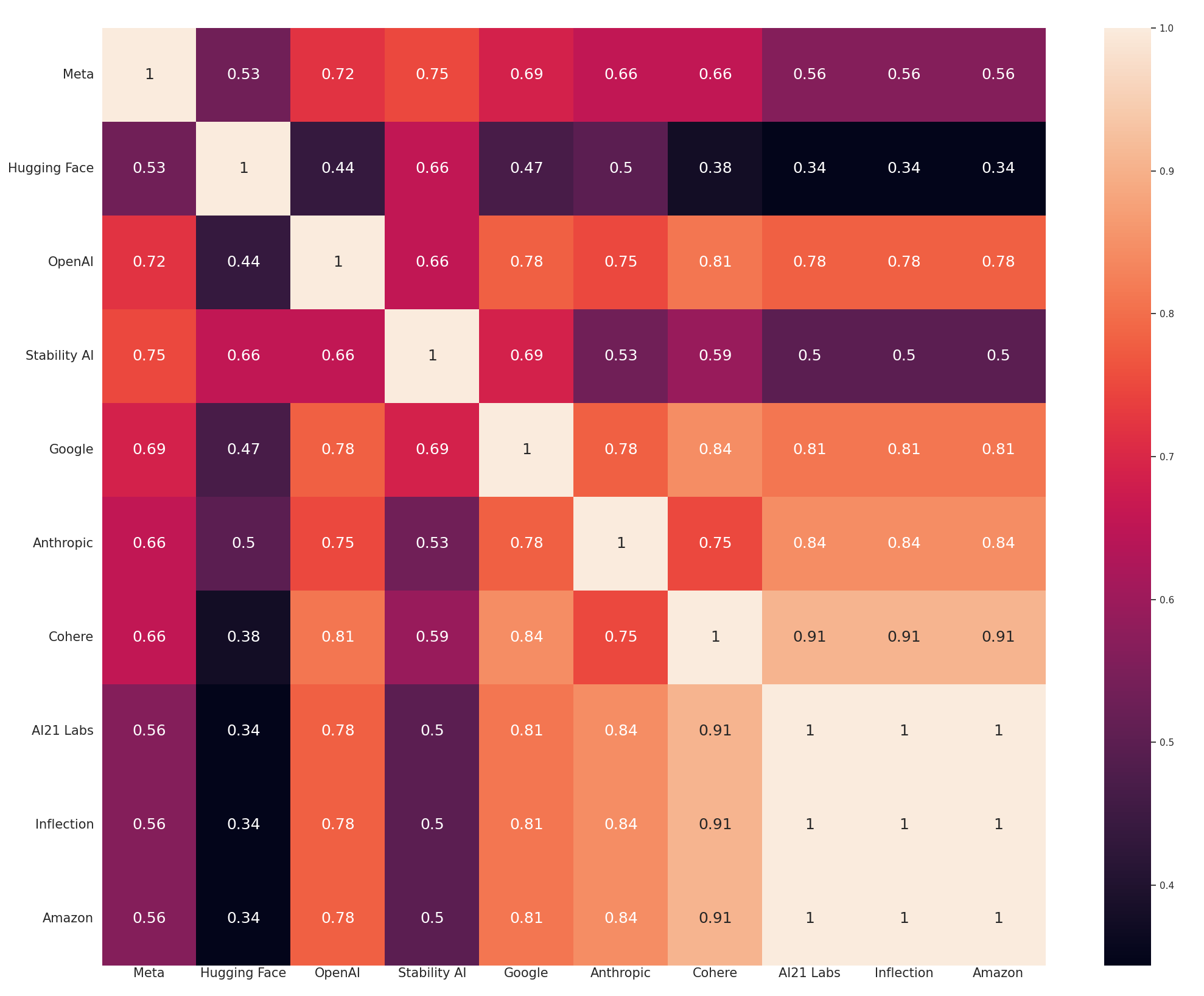}}
\caption{\textbf{Correlations between companies (upstream).} The correlation between the 2023 FMTI scores for pairs of companies across all indicators when only considering upstream indicators. Correlation is measured using the simple matching coefficient (\ie agreement rate), which is the fraction of all indicators for which both companies receive the same score (\ie both receive the point or both do not receive the point).
}
\label{fig:upstream-correlations}
\end{figure}

\paragraph{Upstream correlations.}
In \autoref{fig:upstream-correlations}, we plot the correlation between every pair of companies when considering only indicators from the upstream domain.
Since the four lowest-scoring companies overall also score zero (or near-zero in the case of \cohere) points on the upstream indicators, they are necessarily extremely correlated.
For the same reason, the extent to which the remaining six companies are correlated with the three lowest-scoring companies is precisely proportional to their own opacity on the upstream domain.
Looking at the three companies that score in the middle overall (\google, \anthropic, \cohere), we see their indicator-level transparency is reasonably correlated.
We also see a similar trend where \openai, \google, and \anthropic are correlated, though in this case \openai and \cohere are even more correlated with an SMC of 0.81.
Interestingly, while the three open developers score much higher overall than any of the seven closed developers for the upstream domain, the correlations between them are somewhat different than in the other domains: there is a weaker correlation between \huggingface and \stability, and \meta's correlation with \openai and \stability is stronger than its correlation with \huggingface.
Despite the fact that \meta and \huggingface are the two highest-scoring companies on upstream, they are not especially correlated (SMC = 0.53) in that domain.
These discrepancies coincide with the indicators where \huggingface scores points and the other two open developers (\meta, \stability) do not, namely those relating to data sources and data labor. 
Given the large spread in scores across developers in the upstream domain, we see the related effect that the correlations can be quite variable with some at or near 1 and others well below 0.5 (minimum upstream SMC = 0.34).

\begin{figure}
\makebox[\textwidth][c]{\includegraphics[keepaspectratio, width=0.7\textwidth]{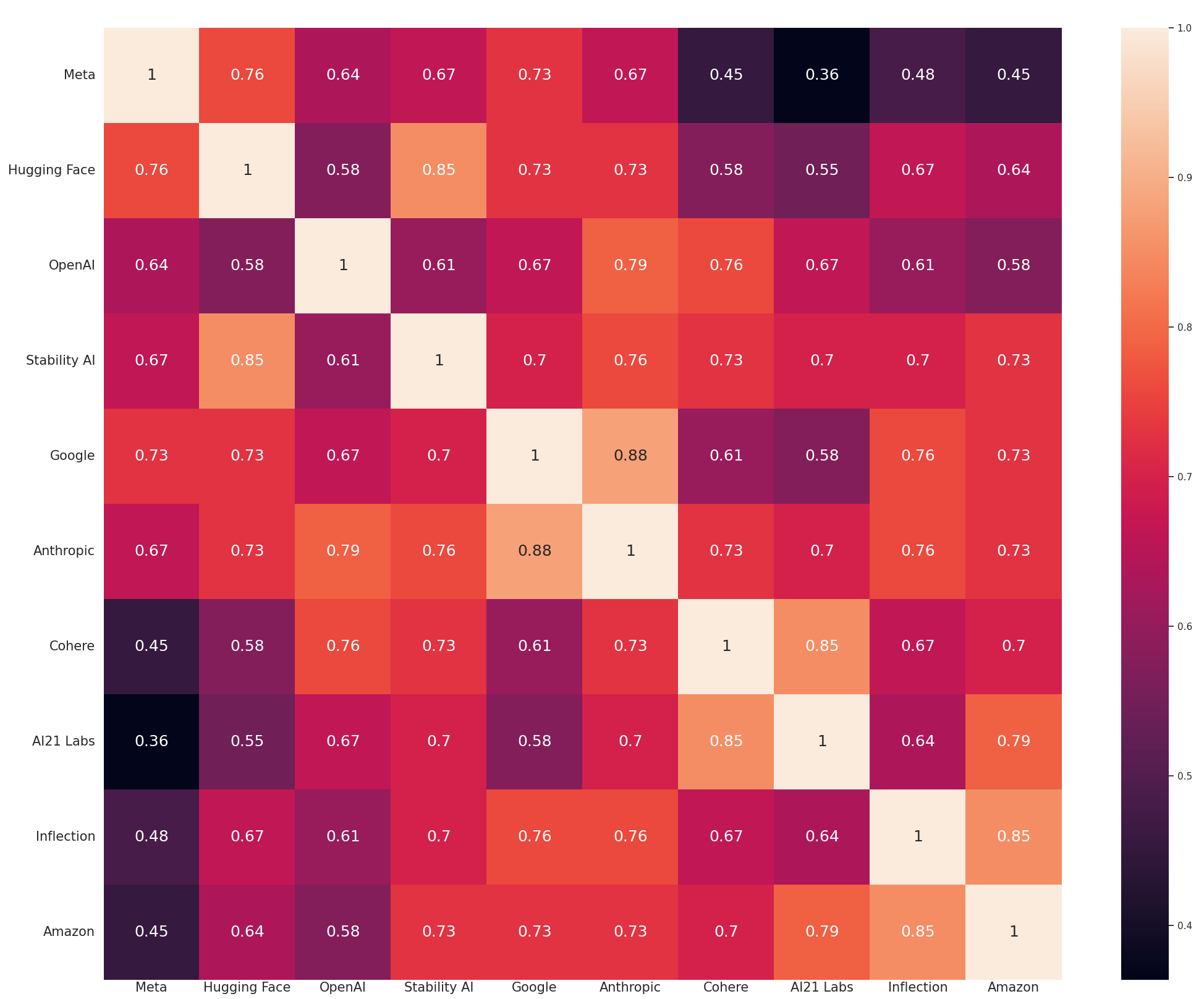}}
\caption{\textbf{Correlations between companies (model).} The correlation between the 2023 FMTI scores for pairs of companies across all indicators when only considering model indicators. Correlation is measured using the simple matching coefficient (\ie agreement rate), which is the fraction of all indicators for which both companies receive the same score (\ie both receive the point or both do not receive the point).
}
\label{fig:model-correlations}
\end{figure}

\paragraph{Model correlations.}
In \autoref{fig:model-correlations}, we plot the correlation between every pair of companies when considering only indicators from the model domain.
In contrast to the upstream correlations, we see a much more varied picture.
First, much like the overall correlations, we see strong correlations for (\cohere, \aitwentyone; SMC = 0.85) and (\inflection, \amazon; SMC = 0.85) but not necessarily for the other pairs between these four companies. 
Among the three Frontier Model Forum companies, we see a very strong correlation of 0.88 between \google and \anthropic, a fairly high correlation of 0.79 between \openai and \anthropic, but a considerably lower correlation for the third pair of \openai and \google at 0.67.
These trends, where \anthropic is highly correlated with both, but \openai and \google are not necessarily correlated, mirror what we observe for the overall correlations.
Similar to what we observed for the overall correlations, \huggingface and \stability are quite correlated as well with a correlation of 0.85, and \meta is not particularly correlated with any company (the highest is \huggingface at 0.76).

\begin{figure}
\makebox[\textwidth][c]{\includegraphics[keepaspectratio, width=0.7\textwidth]{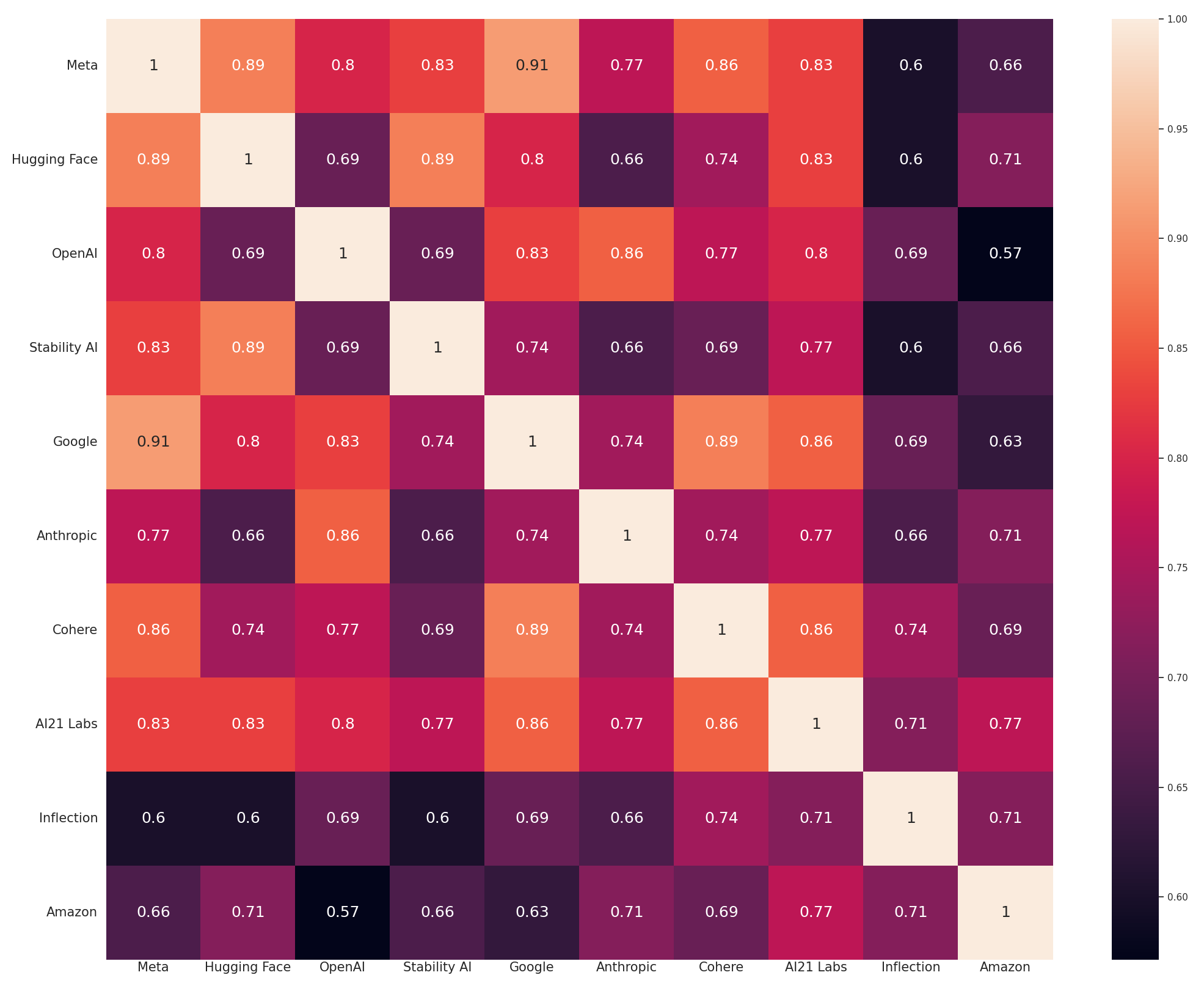}}
\caption{\textbf{Correlations between companies (downstream).} The correlation between the 2023 FMTI scores for pairs of companies across all indicators when considering only downstream indicators. Correlation is measured using the simple matching coefficient (\ie agreement rate), which is the fraction of all indicators for which both companies receive the same score (\ie both receive the point or both do not receive the point).
}
\label{fig:downstream-correlations}
\end{figure}

\paragraph{Downstream correlations.}
In \autoref{fig:downstream-correlations}, we plot the correlation between every pair of companies when considering only indicators from the downstream domain.
The downstream correlations surface considerably different trends from the overall correlations or those for the other two domains.
In particular, we first highlight that \meta is strongly correlated with \google in their scores on downstream indicators.
Given that several of the downstream indicators related to broader corporate practices, the similarities between these companies may contribute to this result, though both companies are not strongly correlated with \amazon, the other Big Tech company we assess.
Relatedly, we see fairly strong correlations between \openai and \anthropic, which again may relate to their fairly similar business practices mapping onto specific downstream indicators (\eg indicators in the \modelbehaviorpolicy subdomain).
On the other hand, akin to the upstream subdomain, we see that \inflection is especially dissimilar from all of the open model developers (\meta, \huggingface, \stability). 
And, unlike the other correlation matrices, \openai and \amazon are more dissimilar than usual. 
Overall, while we do not observe it as clearly in the other correlation analyses, here we see all three pairs of open developers are highly correlated:
(\meta, \huggingface; SMC = 0.89),
(\meta, \stability; SMC = 0.83),
(\huggingface, \stability; SMC = 0.89).
This may reflect that all open developers have shared transparency challenges on specific indicators within the downstream domain (\eg monitoring mechanism and model behavior policy enforcement), perhaps stemming from the weaker control they have over downstream use. 
Overall, we find the complex structure and heterogeneity in the correlation for the downstream domain especially intriguing, given the aggregate scores for this domain are the most tightly clustered.
That is to say, disaggregated indicator-level analysis is especially revealing for this domain compared to domain-level analysis.

\section{Improvements}
A central objective of an index is to track a concept over time to characterize changes.
In doing so, many notable indices have evolved over time to reflect changing circumstances and priorities.
For instance, the Human Development Index (HDI) was changed multiple times in the 1990s and 2000s, largely in response to academic criticism \citep{klasen2018human, stanton2007}.
Despite these changes, which complicate direct comparisons across index iterations, the HDI remains one of the most trustworthy and popular indices for human development. 

As an index is conducted repeatedly, and the world changes as measured by the index,
a natural question is how the index contributes to this change.
Attributing why corporate behavior (such as disclosure practices) changes is notoriously difficult.
Companies generally do not reveal why they make changes and changes generally reflect a confluence of multiple factors.
\citet{kogen2022rdr} provides a unique demonstration of an index's impact, analyzing the 2018 Ranking Digital Rights Index (RDR), which ranked the freedom of expression and privacy policies of 26 of the world's largest ICT companies.
By reviewing internal RDR documents and interviewing relevant stakeholders (\eg representatives from 11 companies and 14 civil society groups), Kogen concluded that RDR had clear influence and that indexes can be useful resources for social movements.
In the case of FMTI, we expect that the Index brings attention to the disclosure practices of companies, making it easier for media, policymakers, investors, customers and the public to apply additional pressure that engenders greater transparency.
Additionally, the Index provides clarity to companies by setting concrete targets and empowers employees within companies to push for greater transparency.

To reason about an index's impact over time, we also draw inspiration from \citet{raji2019actionable}.
In 2017, \citep{buolamwini2018gender} demonstrated significant performance disparities across demographic groups in 3 face recognition systems (from IBM, Microsoft, and Megvii).
A year later, \citet{raji2019actionable} audited five systems (IBM, Microsoft, Megvii, Amazon, Kairos): they found that the original 3 systems had reduced the performance disparities considerably, whereas the 2 new systems in 2018 showed large disparities comparable to those seen in the 3 systems from 2017.

\subsection{2024 FMTI}
The 2024 FMTI involves four steps: indicator selection, developer selection, information gathering, and scoring.
We describe these steps and how they relate to their implementation in the 2023 FMTI below, noting that we deliberately selected the same indicators to facilitate comparison.

\paragraph{Developer selection.}
In the 2023 FMTI, we selected 10 foundation model developers: all 10 were companies developing salient foundation models with consideration given for diversity (\eg type of company, type of foundation model).
Further, for each foundation model developer, we designated a \textit{flagship} foundation model that was used as the basis for scoring the developer.
In the 2024 FMTI, we require companies to submit \textit{transparency reports}\footnote{As we describe later, companies submitted an initial report that was modified through the 2024 FMTI process. The final report that we publish is validated by the company but, therefore, different from this initial report. For brevity, we refer to both as transparency reports.
}: we reached out to leadership at 19 companies: 01.AI, Adept, AI21 Labs, Aleph Alpha, Amazon, Anthropic, BigCode/Hugging Face/ ServiceNow, Cohere, Databricks, Google, IBM, Inflection, Meta, Microsoft, Mistral, OpenAI, Stability AI, Writer, and xAI.
14 developers agreed to prepare reports and designated their flagship foundation model.\footnote{We provided guidance that the flagship foundation model should be ``based on a combination of the following factors: greatest resource expenditure, most advanced capabilities, and greatest societal impact.''}

8 of the 14 developers are hold-overs from the 2023 FMTI.\footnote{Cohere and Inflection declined to participate in the 2024 FMTI.} 
Three developers are assessed for the same models as 2023 (\jurassic for \aitwentyone, \llama for Meta, \gptfour for \openai), whereas five are assessed for new models (\titan for \amazon, Claude 3 for Anthropic, StarCoder for BigCode/HuggingFace/ServiceNow, Gemini 1.0 Ultra API for Google, and Stable Video Diffusion for Stability AI).
The six new developers and their models are Fuyu-8B for Adept, Luminous Supreme for Aleph Alpha, Granite for IBM, Phi-2 for Microsoft, Mistral 7B for Mistral, and Palmyra-X for Writer.
In aggregate, the 2024 FMTI composition has broader geographic coverage (\eg from 0 to 2 companies headquartered in the European Union), broader modality coverage (\eg Gemini takes audio and video as input), and a more even balance of open and closed foundation models (\ie from 3 of 10 open models in 2023 to 6 of 14 open models in 2024).

\paragraph{Information gathering.}

In the 2023 FMTI, we identified publicly-available sources of information for each developer through a systematic protocol for searching the Internet, which provided the information for all scoring decisions.
This approach has four potentially undesirable properties.
First, given that information is decentralized across the Internet, the researchers may have missed information.\footnote{However, this does beg the question of whether the information being public is truly constitutive of transparency if it is not discovered through a systematic and high-effort search.}
Second, the relationship between a piece of public information and an indicator may be indirect and oblique, leading to greater subjectivity in scoring.
Third, focusing on public information aligns with a developer's current level of transparency but does not provide developers with an opportunity to disclose further information.
Finally, and most fundamentally, this search significantly adds to the cost of executing the index.

In the 2024 FMTI, we request transparency reports from each developer that directly address each of the 100 indicators.
This change in the information gathering process alters the dynamics for the four aforementioned considerations.
First, if we assume developers are strongly incentivized to be their own best advocates and are certainly the most knowledgeable entities about their models, then the information they compile should be complete.
Second, by having developers directly clarify information on indicators affirmatively, uncertainties that contributed to more subjective scoring are addressed.
Third, by allowing developers to include information that was not-previously public, which is made public through this process, opportunities arise for greater transparency.
Finally, by having developers gather information, the cost we bear is reduced. 

\paragraph{Scoring procedure.}
In the 2023 FMTI, once information was identified, two researchers independently scored each of the 1000 (indicator, developer) pairs.
The agreement rate was 85.2\% (148 disagreements): in the event of disagreement, the researchers discussed and came to agreement.
These initial scores were sent to developers to permit rebuttal: following a two-week rebuttal process, final scores were published in October 2023.
For the 2024 FMTI, using the information identified through the developer-submitted transparency reports, two researchers independently scored each of the 1400 (indicator, developer) pairs.
The agreement rate was 85.3\% (206 disagreements): in the event of disagreement, the researchers discussed and came to agreement.

These initial scores were sent to developers to permit rebuttal: in contrast to the 2023 FMTI, the rebuttal process was a more iterative multi-week process that involved email exchanges and video meetings.
Through this correspondence, companies clarified existing information and disclosed new information, which is reflected in the final form of the transparency reports we publish.
Ultimately, companies validated their transparency reports and approved their release.
In doing so, unlike the 2023 FMTI, these reports make explicit instances where (i) developers disclose useful information yet (ii) we felt the disclosure was insufficient to award a point.

\subsection{Results}
We focus on major changes relative to 2023 here, deferring many of the 2024 results to \citet{bommasani2024fmti}.

\begin{figure}
\centering
\includegraphics[keepaspectratio, height=\textheight, width=\textwidth]{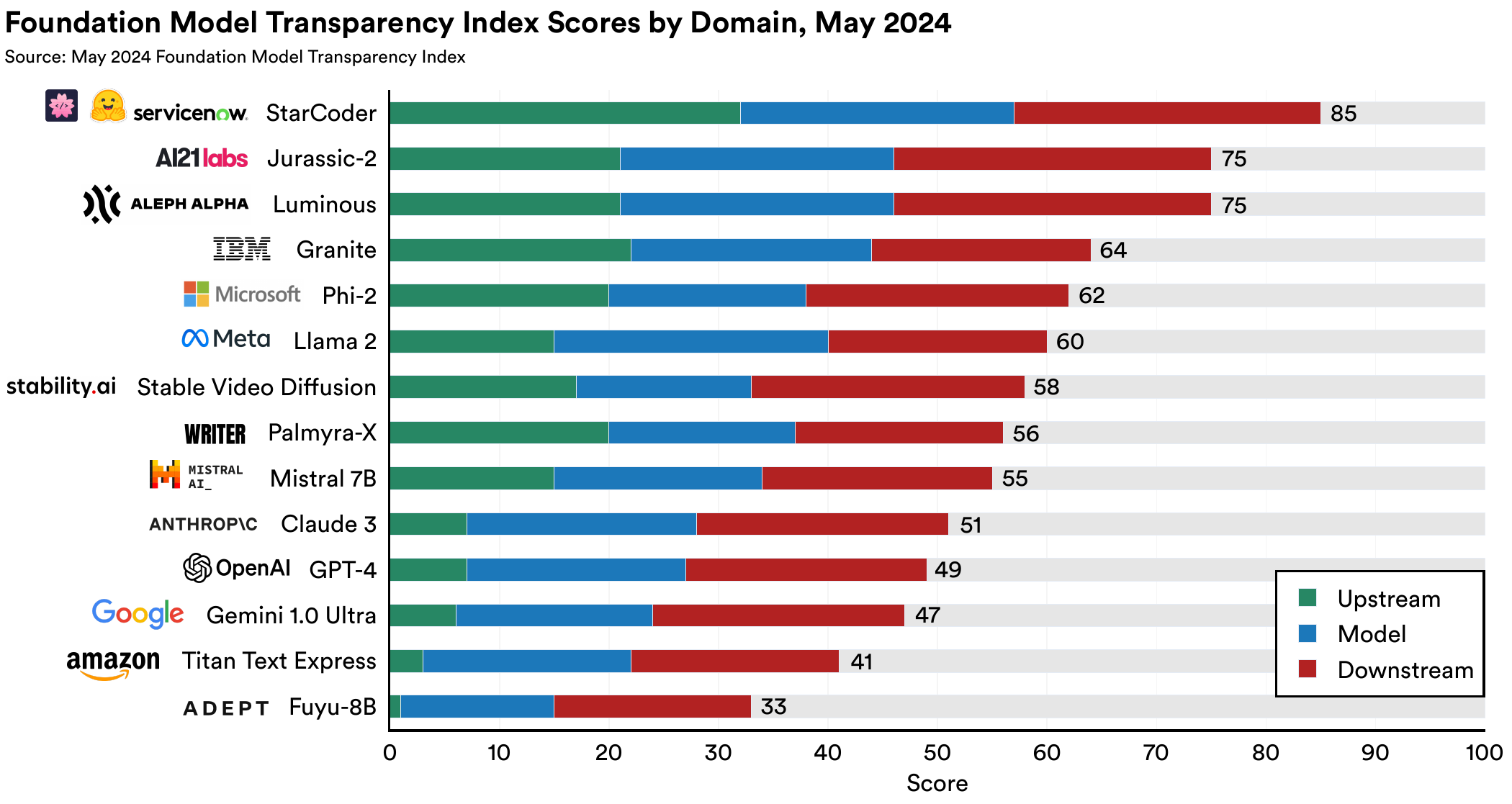}
\caption{\textbf{Scores by domain.} The overall 2024 FMTI scores disaggregated into the three domains: upstream, model, and downstream.}
\label{fig:total-domain}
\end{figure}

\paragraph{While the average score on the Index has significant room for improvement, there is high variability among developers.} 
Based on the overall scores (right of Figure 1), 11 of the 14 developers score below 65, showing that there is a significant lack of transparency in the foundation model ecosystem and substantial room for improvement across developers. 
The mean and median are 57.93 and 57 respectively, with a standard deviation of 13.98. 
The highest-scoring developer scores points for 85 of the 100 indicators, while the lowest-scoring developer scores 33. 
The 3 top-scoring developers (BigCode/Hugging Face/ServiceNow, Aleph Alpha, and AI21 Labs) are more than one standard deviation above the mean, the next 9 are near the mean (IBM, Microsoft, Meta, Stability AI, Writer, Anthropic, OpenAI, Mistral, Google), and the 2 lowest-scoring developers are more than one standard deviation below the mean (Amazon, Adept). 

\paragraph{Improvement is feasible for each developer.} 
In spite of significant opacity, for 96 of the 100 indicators there exists some developer that scores points, and of these there are 89 where multiple developers score points. 
The disclosures that developers make to satisfy these indicators provide a concrete example of how all developers can be more transparent. 
If developers emulate the most-transparent developer for each indicator, overall transparency would improve sharply. 

\paragraph{Developers disclose significant new information, which contributes to their scores.} 
A developer’s total score on the Foundation Model Transparency Index reflects the information that it discloses about its flagship foundation model in relation to each of the 100 indicators in the Index. 
In this version of the Index, we note where this information is disclosed as part of the process of conducting the Index (i.e. where developers disclosed the information for the first time via their report, or where developers updated their documentation in response to the Index process). 
This new information contributes to developer’s overall scores: developers on average release new information related to 16.6 indicators, which improves the average score by 14.2 points. 
For example, AI21 Labs and Writer newly disclosed the carbon emissions associated with building their flagship foundation models (200-300 tCO2eq and 207 tCO2eq respectively).

\paragraph{The upstream domain is the most opaque and the region where the most transparent developers distinguish themselves.}
Breaking the results down by domain (\autoref{fig:total-domain}), developers performed best on indicators in the downstream domain, scoring 65\% of available points overall in comparison to 61\% on the model domain and 46\% in the upstream domain.
Developers scored worse across upstream indicators: of the 20 indicators where developers score highest, just one indicator (model objectives) is in the upstream domain. 
High-scoring developers often differentiate themselves in terms of their transparency in the upstream domain; the top scorer overall, BigCode/Hugging Face/ServiceNow, scored all 32 points, whereas the lowest scorer overall, Adept, scored just one point. 
The standard deviation of scores on the upstream domain (8.8) is more than double that of the other two domains (3.6 each), reflecting the much greater spread across the domain.
On the whole, developers are less transparent about the data, labor, and compute used to build their models than how they evaluate or distribute their models.
Prior work has emphasized the importance of these particularly opaque domains \citep{crawford2021atlas, gebru2021datasheets, luccioni2023counting}.

\begin{figure}
\centering
\includegraphics[keepaspectratio, height=\textheight, width=\textwidth]{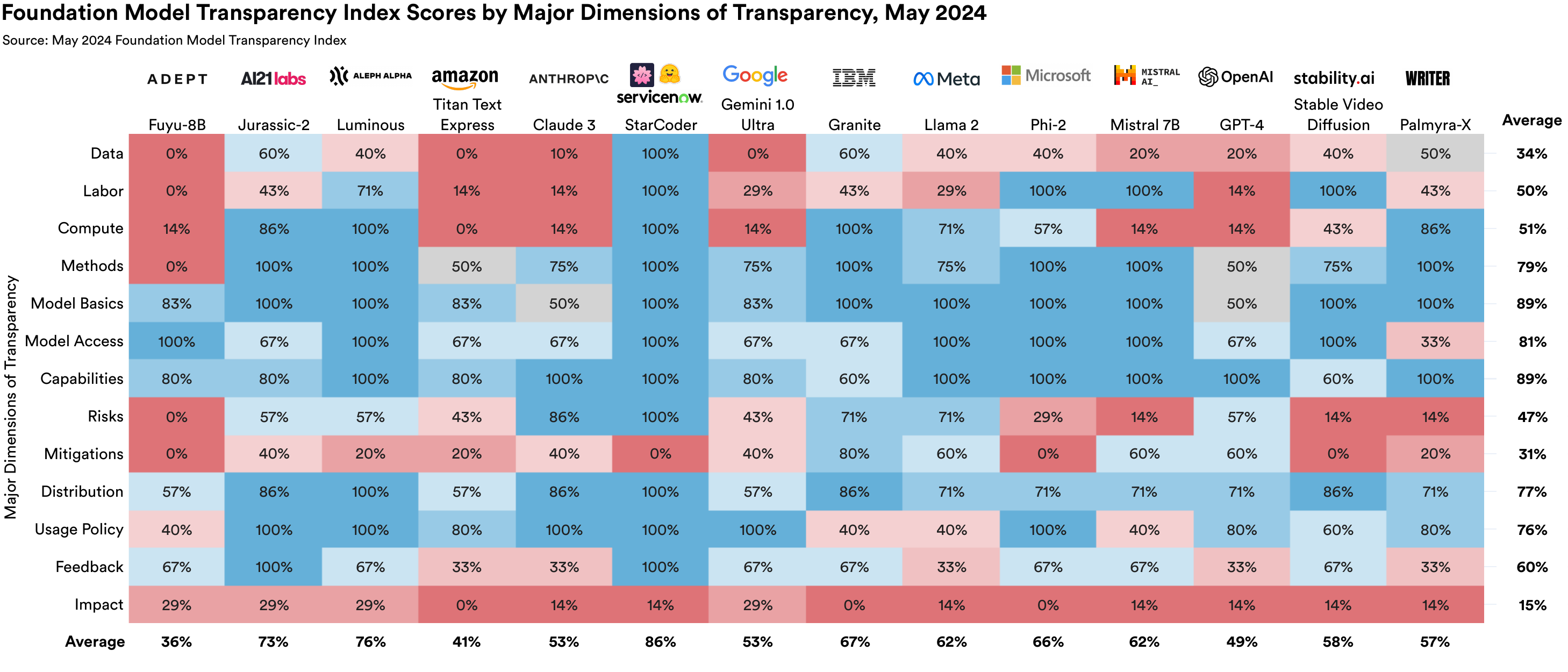}
\caption{\textbf{Scores by major dimensions of transparency.}  The fraction of achieved indicators in each of the 13 major dimensions of transparency in the 2024 FMTI. Major dimensions of transparency are large subdomains within the 23 subdomains.}
\label{fig:overall-subdomains}
\end{figure}

\paragraph{Scores varied across subdomains, with developers scoring best on user interface, capabilities, and model basics.}
Disaggregating each domain, we consider the 23 subdomains with \autoref{fig:overall-subdomains} showing results for 13 major subdomains.
Scores varied greatly across subdomains: 86 percentage points separate the average scores on the most transparent and least transparent subdomains.
The subdomains with the highest scores are user interface (93\%) capabilities (89\%), model basics (89\%), documentation for downstream deployers (89\%), and user data protection (88\%).
Each of these subdomains is in the downstream domain, where developers scored near or above 70\% on 6 of the 9 subdomains. 
These high scores were achieved in part through the release of new information. 
For instance, AI21 Labs released its first model card for Jurassic-2 during the 2024 FMTI process, and Aleph Alpha and Stability AI made significant changes to their existing model cards.
High scores on these subdomains reflects that disclosure in these areas is relatively less onerous for developers---disclosing documentation for downstream deployers as well as information about capabilities and model basics makes it easier and more appealing for customers to make use of a companies' flagship foundation model, meaning it is in developers' interest to do so.

\paragraph{Data access, impact, and trustworthiness are the least transparent subdomains.}
The subdomains with the lowest total scores are data access (7\%), impact (15\%), trustworthiness (29\%), and model mitigations (31\%). 
Developers score 50\% or less on 10 of 23 subdomains in the index, including 3 of the 5 largest subdomains---impact (15\%), data (34\%), and data labor (50\%). 
The lack of transparency in these subdomains shows that the foundation model ecosystem is still quite opaque---there is little information about how people use foundation models, what data is used to build foundation models, and whether foundation models are trustworthy.

\begin{figure}
\centering
\includegraphics[keepaspectratio, height=\textheight, width=\textwidth]{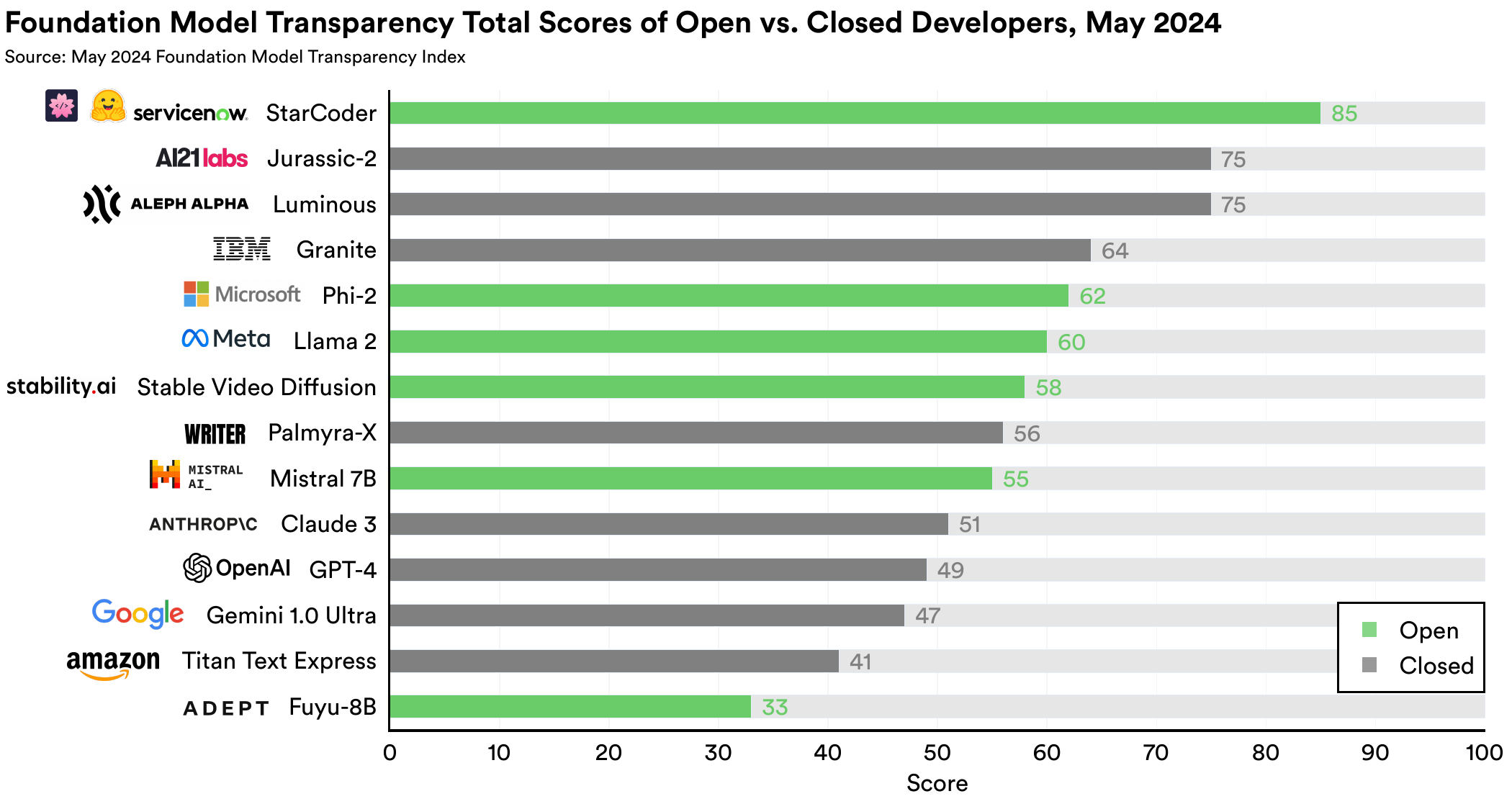}
\caption{\textbf{Overall scores by release strategy.} The overall 2024 FMTI scores for the 6 open developers (Adept, BigCode/Hugging Face/ServiceNow, Meta, Microsoft, Mistral, Stability AI) and the 8 closed developers (AI21 Labs, Aleph Alpha, Amazon, Anthropic, Google, IBM, OpenAI, Writer).}
\label{fig:open-total}
\end{figure}

\paragraph{Open 
developers match closed developers on downstream indicators, and exceed them on upstream indicators.}
Developers adopt different release strategies \citep{solaiman2023gradient} for their flagship foundation models: six developers release open foundation models \citep{kapoor2024societal}, meaning the model weights are widely available, whereas the other eight employ a more closed release strategy.\footnote{Aleph Alpha shares model weights for its flagship foundation model with some customers on premises, but this does not mean that the model weights are widely available and so it is considered a closed foundation model developer per the definition of \citet{kapoor2024societal}.}
Open developers generally outperform closed developers (\autoref{fig:open-total}: the median open developer scores 5.5 points higher than the median closed developer.
While making the weights of a model openly available is correlated with greater overall transparency, it does not necessarily imply greater transparency about matters like data, compute, or usage.

The difference in transparency between open and closed developers is attributable to the substantial gap in upstream transparency: within the upstream domain, the median open developer scores 3 additional points on indicators in the upstream subdomain over the median closed developer. 
Within each subdomain, the median open developer scores as many or more points than the median closed developer on all but 5 of the 23 subdomains (i.e., risks, model mitigations, trustworthiness, distribution, usage policy, and model behavior policy). 
Open developers outscore closed developers by the widest margin on the following subdomains: data labor, data, and model access. 
Though open developers may struggle to gauge the downstream use of their models \citep{klyman2024aups}, which closed developers may be able to directly monitor, the median open developer scores the same number of points on the downstream domain as the median closed developer.

Developers that are part of the AI Alliance (Hugging Face,
IBM, Meta, ServiceNow, Stability AI), which often advocates for open releases of model weights, outscore developers that are founding members of the Frontier Model Forum (Anthropic, Google, Microsoft, OpenAI) by a median of 12 points, with higher average scores across upstream, model, and downstream indicators.\footnote{The AI Alliance is a coalition of developers, universities, and researchers ``who collaborate to advance safe, responsible AI rooted in open innovation.'' The Frontier Model Forum is an industry group whose founding members are Anthropic, Google, Microsoft, and OpenAI; the Frontier Model Forum committed to ``advancing AI safety research,'' ``identifying best practices,'' ``collaborating across sectors,'' and ``help[ing] AI meet society's greatest challenges.'' See \url{https://thealliance.ai/} and \url{https://www.frontiermodelforum.org/}.}

Still, closed developers outperform open developers in several areas related to policies and enforcement.
Closed developers generally share more information about if and how they enforce their policies regarding user and model behavior, outperforming open developers on these subdomains by 2 and 1 points respectively.\footnote{As with some other subdomains, the difference in scores here is driven by just one or two of the 14 developers.}
Closed developers also score higher on the risks and model mitigations subdomains, as several open developers do not describe or demonstrate risks associated with their flagship foundation model and closed developers are more likely to describe and demonstrate any risk mitigations that are taken at the model level.\footnote{Open developers, however, outscore closed developers on data mitigation indicators. They often do not score points on model mitigation indicators because they do not explicitly state that no such mitigations were applied at the model level.}
The discrepancy in transparency between open and closed foundation model developers is a reflection of the current state of the ecosystem, not a fundamental reality about the transparency of developers that do or do not make the weights of their foundation models widely available.

\begin{figure}
\centering
\includegraphics[keepaspectratio, height=\textheight, width=\textwidth]{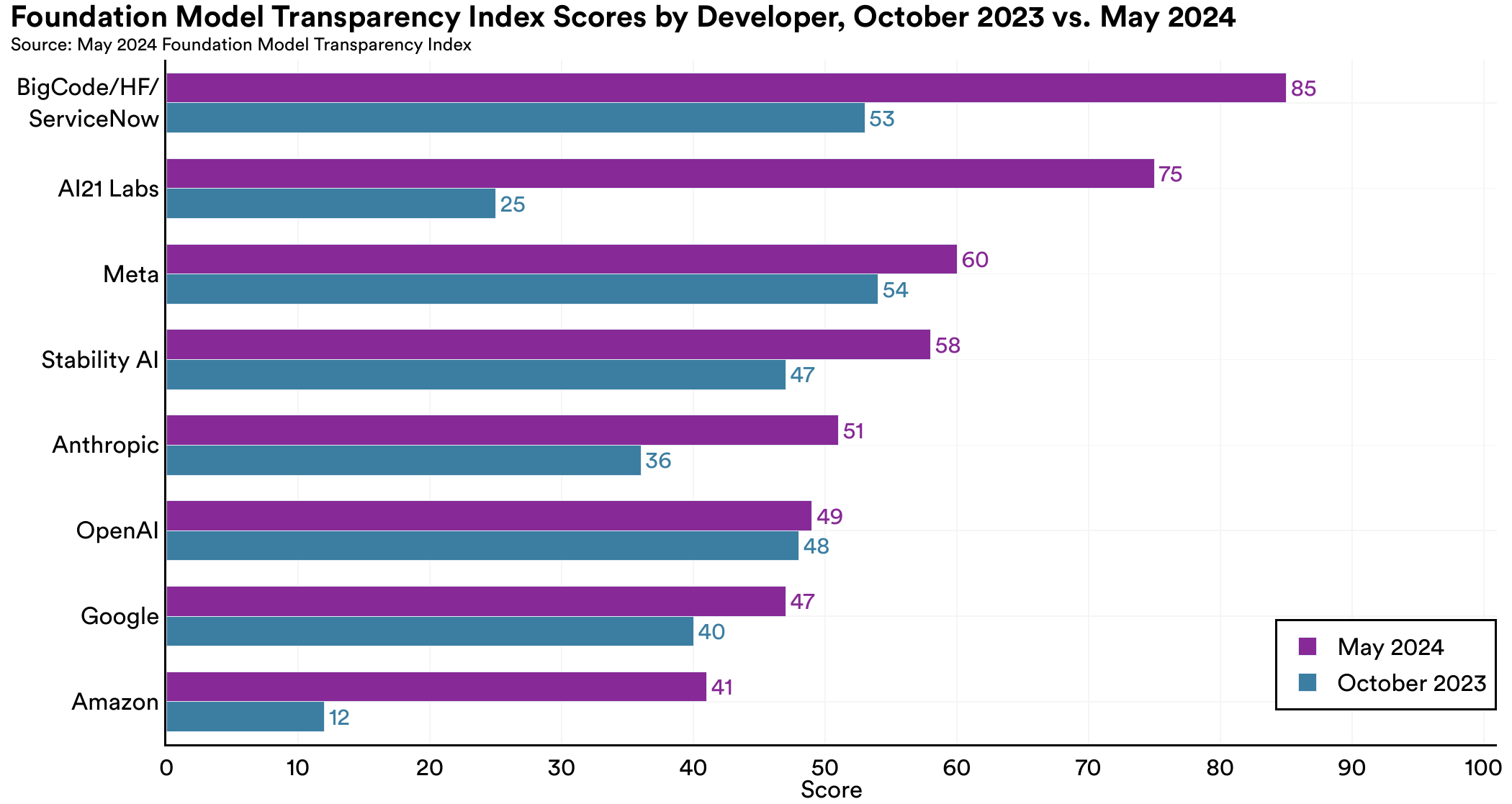}
\caption{\textbf{Change in overall scores.} The 2023 FMTI and 2024 overall scores for the eight developers assessed in both versions.
}

\label{fig:total-change}
\end{figure}

\paragraph{Transparency increased across the board from 2023 to 2024.} Foundation model developers significantly improved their scores between October 2023 and May 2024, with the average score rising from 37 to 58 out of 100. 
Scores improved on every domain, with average upstream scores improving by the greatest margin (+7.6 points) followed by downstream (+7.2) and model (+6.1). 
As a result, there is significantly more information publicly available about the upstream resources developers use to build foundation models, the models themselves and how they are evaluated, and their downstream distribution and use.

\paragraph{Transparency increased in nearly all subdomains from 2023 to 2024.} Developers improved their scores on every subdomain with the exception of data access. The largest improvements in subdomain scores were in compute (average increase of +2.4 indicators per developer), data labor (+2.3), and risks (+1.6). This broad improvement demonstrates that the overall trend in recent years toward reduced transparency is more nuanced than is commonly understood, though transparency is still lacking with respect to the data used to build foundation models and the impact they have once deployed. 

\paragraph{Transparency increased in subdomains that were especially opaque in 2023.} Several of the areas of the index that were least transparent in 2023 show significant improvement in 2024, including subdomains such as compute, methods, risks, and usage policy. For example, the total score across companies for the compute subdomain rose from 17\% in 2023 to 51\% in 2024. Compute is potentially one of the most intractable areas for disclosure as it relates to the environmental impact of building foundation models—and therefore could be associated with additional legal liability for developers and deployers—yet we see improvement across compute indicators. This improvement is driven to a significant degree by new information that companies have disclosed, with companies disclosing new information on compute usage (6 companies disclosed new information), development duration (4), energy usage (4), compute hardware (3), hardware owner (3), and carbon emissions (2). In this way, transparency about compute expenditure has spillover effects for transparency about environmental impact, providing a potential model for translating information disclosure about one area into details about the impact of the foundation model supply chain.

\begin{figure}
\centering
\includegraphics[keepaspectratio, height=\textheight, width=\textwidth]{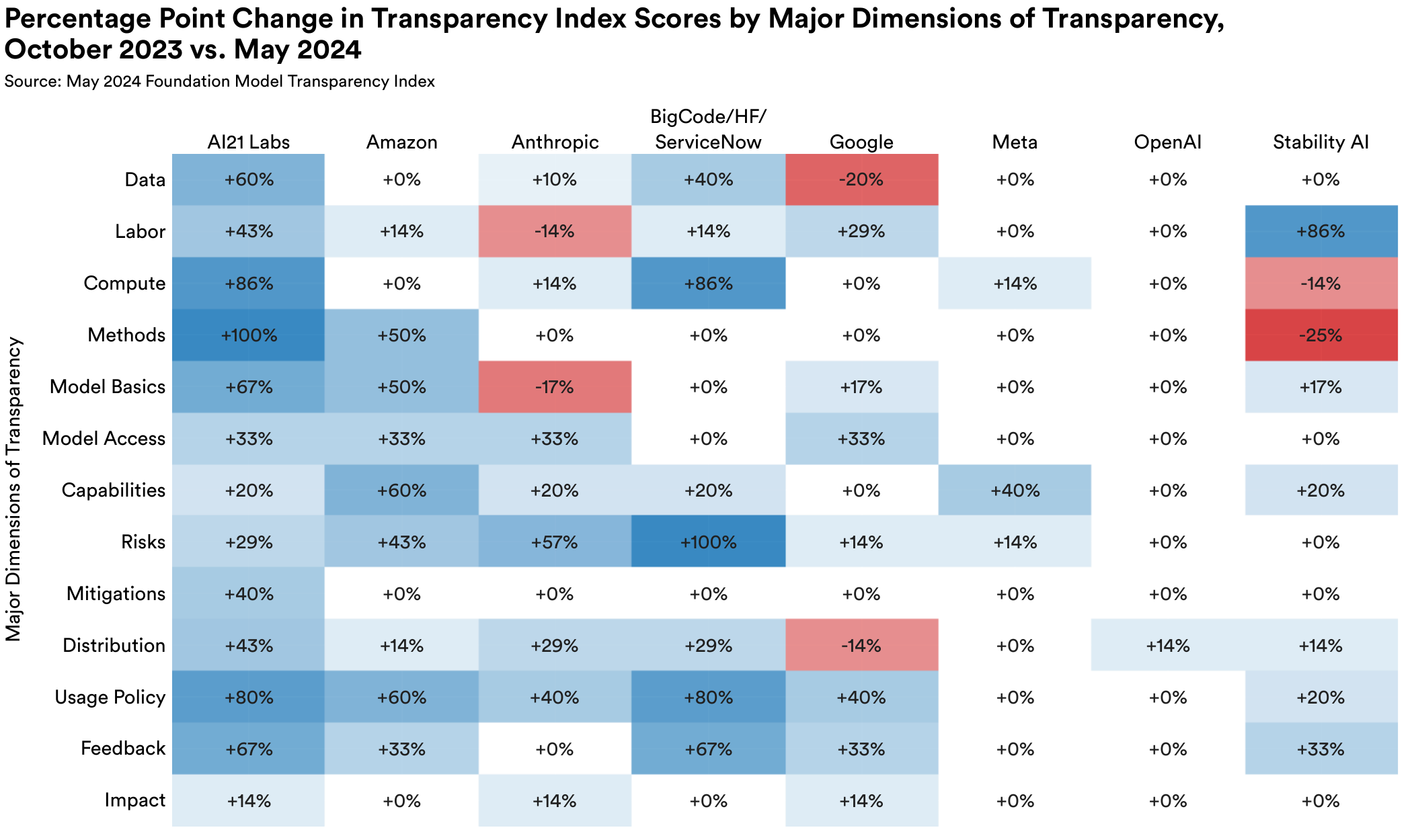}
\caption{\textbf{Change in subdomain scores from 2023 FMTI to 2024 FMTI.} This figure shows the percentage point change in scores for major subdomains for the eight developers that are included in both the October 2023 and May 2024 versions of the Foundation Model Transparency Index.}
\label{fig:heatmap-change}
\end{figure}

Transparency has improved substantially with respect to companies’ policies regarding acceptable use and behavior of their models. The total score across companies for the usage policy subdomain rose from to 44\% to 57\%, which was driven by increased transparency related to how these policies are enforced. Similarly, the total score across companies for the model behavior policy subdomain rose from 23\% to 69\%. While increased transparency in these domains is a relatively light lift—as companies merely need to state whether they have such policies and, if so, describe how they are enforced—this form of transparency is not costless for companies as it highlights potential failure modes for their models \citep{klyman2024aups}. Transparency about how companies restrict the use and behavior of their models can facilitate independent evaluation and red teaming due to reduced legal uncertainty, promoting research on safety and trustworthiness \citep{longpre2024safe}. 

\paragraph{Data remains a key area of opacity.} Several areas of the Index exhibit sustained and systemic opacity: almost all developers remain opaque on these matters. Developers display a fundamental lack of transparency with respect to data, building on frustrations of data documentation debt \citep{bandy2021addressing, sambasivan2021everyone}. Transparency on the data subdomain rose from 20\% to 34\% from 2023 to 2024, but BigCode/Hugging Face/ServiceNow is the only developer that scores points on indicators relating to data creators, data copyright status, data license status, and personal information in data. Only 3 developers (Aleph Alpha, BigCode/Hugging Face/ServiceNow, IBM) score points on 6 or more of the 10 data indicators, while 6 developers score 2 or fewer points. These low scores reflect the ongoing crisis in data provenance \citep{longpre2024data, longpre2023data}, wherein companies share no information about the license status of their datasets, preventing downstream developers from ensuring they are complying with such licenses. 

While scores on data labor improved from 17\% to 50\% from 2023 to 2024, this was driven in large part by an increase in the number of companies that do not use data labor outside of their own organization (BigCode/Hugging Face/ServiceNow, Microsoft, Mistral). Considering only the 11 developers who do not disclose that they do not use external data labor, scores on the data labor subdomain fall to 36\%, which would make it the sixth lowest scoring subdomain. Only one developer (Stability AI) that discloses its use of external data labor discloses the wages that it pays data laborers, highlighting how a lack of transparency may limit accountability for worker exploitation \citep{gray2019ghost}. 

Transparency in data access, one of the lowest scoring subdomains across developers in 2023, declined in 2024 from 20\% to 7\%. 
This reflects the significant legal risks that companies face associated with disclosure of the data they use to build foundation models.
In particular, these companies may face liability if the data contains copyrighted, private, or illegal content such as child sexual abuse material \citep{lee2024talkin, Solove2024, thorn2024csam}. 

\paragraph{Developers disclose little information about the real-world impact of their foundation models.} Developers still lack transparency about the real-world impact of their models, and any steps they take to mitigate negative impacts pre-deployment. 
Of the major dimensions of transparency in \autoref{fig:overall-subdomains}, developers score worst on the impact subdomain (as they did in 2023). Each of the four indicators where no developer scores points (affected market sectors, affected individuals, usage reports, and geographic statistics) are in the impact subdomain. This means that the public has little to no information about who uses foundation models, where foundation models are used, and for what purpose. The lack of transparency regarding these matters inhibits effective governance of foundation models, as it is difficult for governments or civil society organizations to pressure companies to responsibly deploy their models if there is no information about the impact of deployment. In many cases this opacity stems from a lack of information sharing between developers, deployers, and customers; developers generally do not know how their foundation model is being used unless a deployer monitors use or receives and shares information about use from its customer.

\paragraph{The 8 developers evaluated in both October 2023 and May 2024 showed marked improvement, or 19 points on average.}
For 3 of these developers we evaluate the same flagship foundation model (Jurassic-2 for AI21 Labs, Llama 2 for Meta, and GPT-4 for OpenAI) as 2023 FMTI while for the other 5 we evaluate a different flagship foundation model (Titan Text Express for Amazon, Claude 3 for Anthropic, StarCoder for BigCode/Hugging Face/ServiceNow, Gemini 1.0 Ultra API for Google, and Stable Video Diffusion for Stability AI). 
All 8 companies' scores increased, as shown in (\autoref{fig:total-change}): AI21 Labs (+50), BigCode/Hugging Face/ServiceNow (+32) and Amazon (+29) made the largest improvements.

\paragraph{Developers that were assessed only in 2024 performed slightly worse than those assessed in both 2023 and 2024.}
The 6 developers that were assessed only in 2024 (Adept, Aleph Alpha, IBM, Microsoft, Mistral, Writer) have slightly lower scores than those assessed in both 2023 and 2024.
Their mean total score is 57.5, which is 1 point lower than that of the other 8 developers.\footnote{This is despite the fact that 4 of the 6 developers that are assessed only in 2024 FMTI are open developers, which score higher on average.}   
Developers that were included in 2023 FMTI had the benefit of having already evaluated their own transparency practices in relation to this initiative (\ie in 2023 FMTI they had the opportunity to rebut scores provided by we, meaning it may have been relatively easier for them to compile transparency reports. 
The 6 developers assessed only in 2024 FMTI include the lowest scoring developer and the two lowest scoring open model developers.

A key feature of the 2024 FMTI methodology is that companies were able to disclose \textit{new information}, meaning information that was not public at the onset of the 2024 FMTI process.
In some cases, this information is directly made public for the first time via the 2024 FMTI transparency reports.
In other cases, this information was incorporated into preexisting or new publicly available documentation from companies.
For example, as we noted previously, AI21 Labs released the first model card for Jurassic-2 and Stability AI significantly updated the model card for Stable Video Diffusion.

\begin{figure}
\centering
\includegraphics[keepaspectratio, height=\textheight, width=\textwidth]{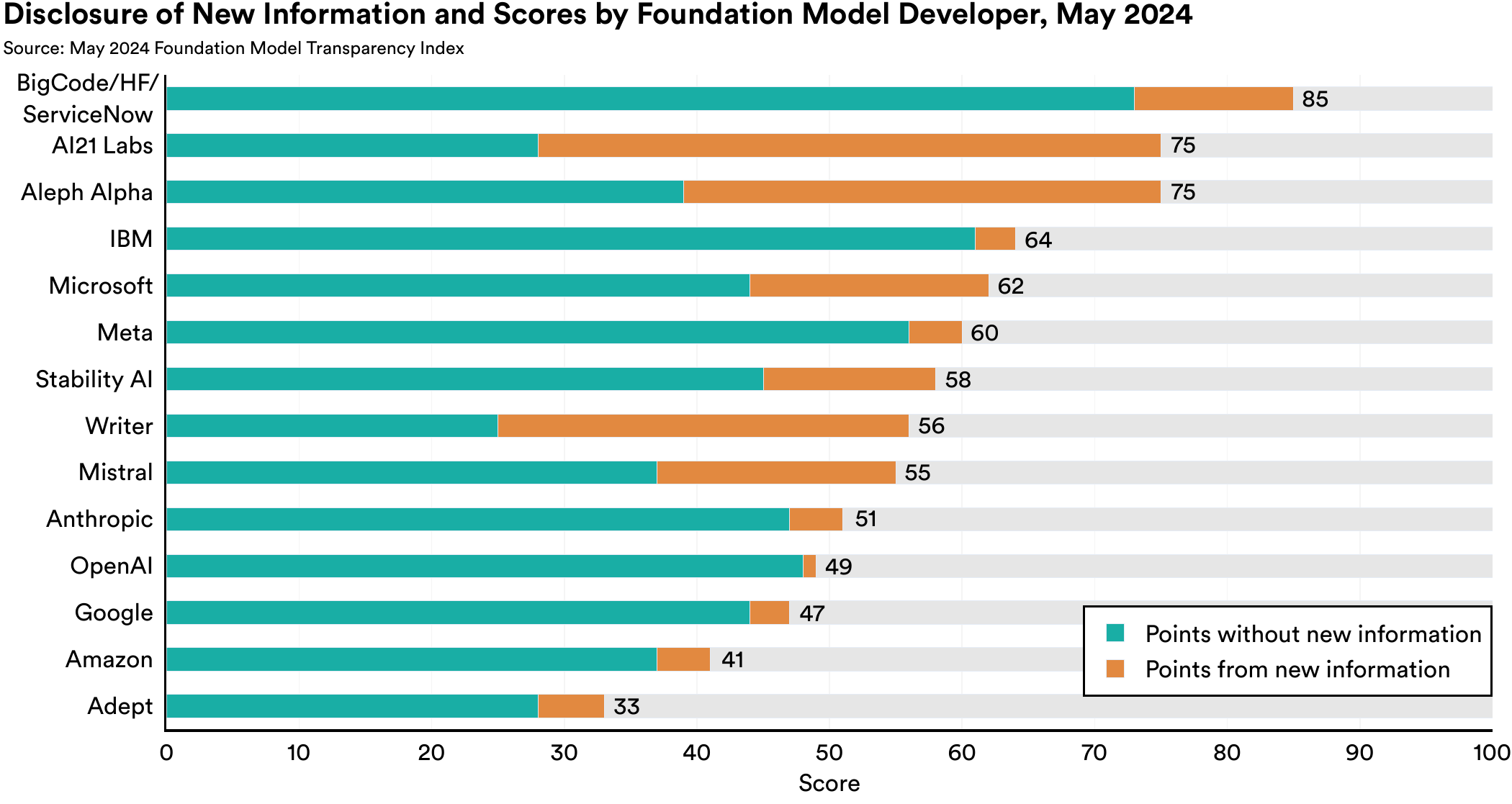}
\caption{\textbf{Scores by new information status.} 
The overall 2024 FMTI scores disaggregated based on whether the information was newly disclosed.}
\label{fig:new-company}
\end{figure}

\paragraph{New information constitutes a large fraction of the score for several companies.}
\autoref{fig:new-company} breaks down each developer's overall 2024 FMTI score based on which indicators were awarded for new information vs. information that was previously publicly available.
For three developers (AI21 Labs, Aleph Alpha, Writer), new information constitutes roughly half the points awarded.
In the case of Aleph Alpha, several new disclosures are made about data labor: all laborers are employed by Aleph Alpha, work in Germany, and are afforded labor protections as stipulated by German law.
For Writer, new information is provided on compute: models are trained on 1024 NVIDIA A100 80GB GPUs for 74 days (910k GPU hours) on the Writer cluster, amounting to $8.2 \times 10^{23}$ FLOPs in compute, 812 MWh in energy, and 207tCO2eq in emissions.

\paragraph{All developers disclose some new information.}
In the case of OpenAI, the sole change to its disclosures from 2023 FMTI is in relation to detecting machine-generated content.
Specifically, OpenAI clarified that it ``originally released a classifier that was taken down due to lack of accuracy. Our commitments are around audio / visual content for now, so this implies lack of ability to detect GPT-4 generated content''. 
In other cases, new information is disclosed to clarify existing information that was difficult to interpret based on publicly available documentation.
For example, Amazon clarified how its model behavior policy and usage policy interoperate: ``In the Bedrock user guide, we stated that AWS is committed to the responsible use of AI, and we use an automated abuse detection mechanisms to identify potential violations, we may request information about customers’ use of Amazon Bedrock and compliance with our terms of service or a third-party provider’s AUP. In the event that a customer is unwilling or unable to comply with these terms or policies, AWS may suspend access to Amazon Bedrock.''  

\begin{figure}
\centering
\includegraphics[keepaspectratio, height=\textheight, width=\textwidth]{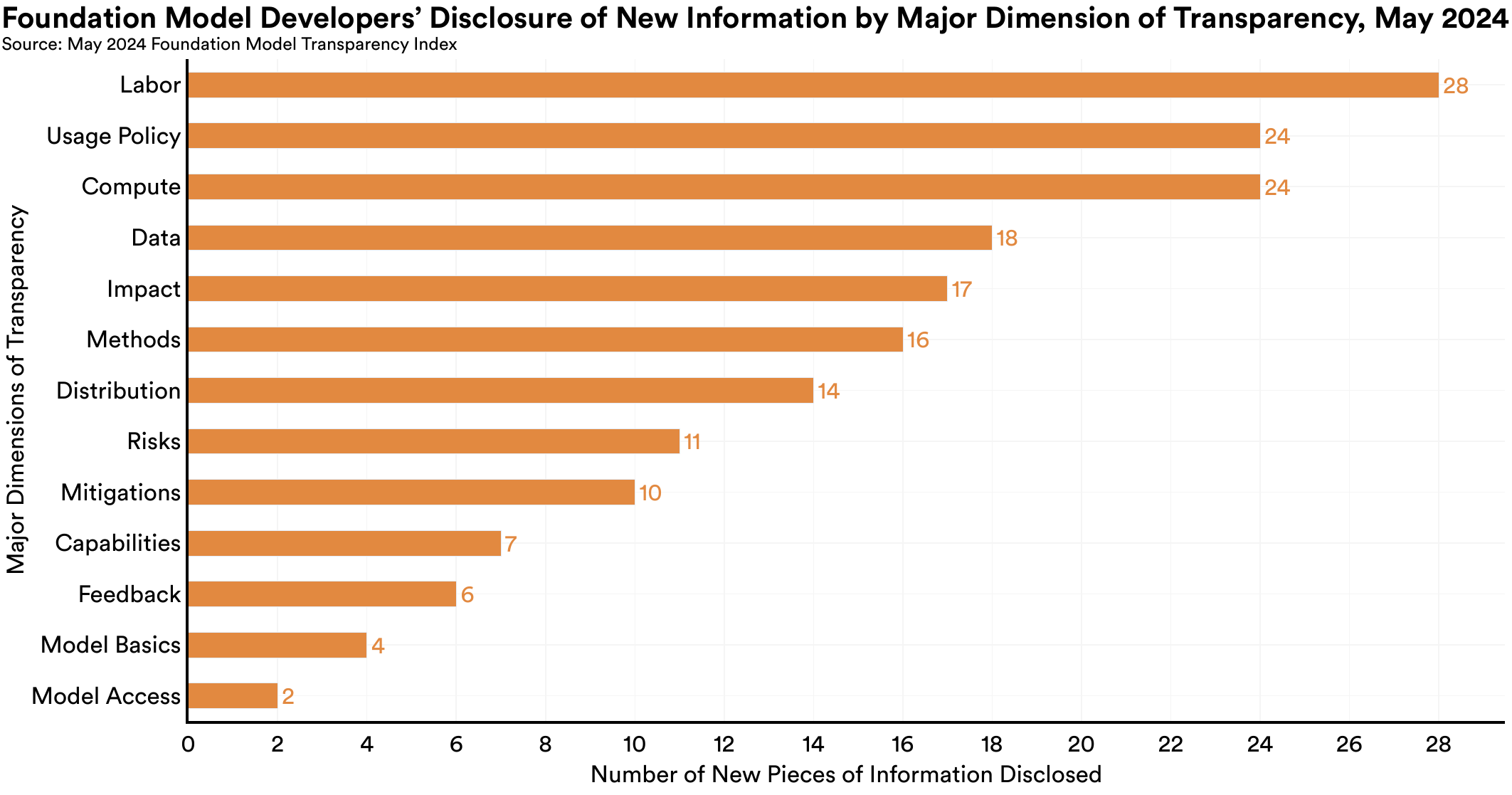}
\caption{\textbf{Aggregate new information by major dimensions of transparency.}
The number of pieces of new information, aggregated across all developers, for each of the 13 major domains of transparency in the 2024 FMTI. 
Note: major dimensions of transparency each have a different numbers of indicators---the largest is data (10 indicators), followed by compute (7), distribution (7), impact (7), labor (7), risks (7),  model basics (6), capabilities (5), mitigations (5), usage policy (5), feedback (3), methods (3), and model access (3).}
\label{fig:new-subdomain}
\end{figure}

\paragraph{New information drives much of the transparency for areas of large improvement.}
\autoref{fig:new-subdomain} totals the amount of new information across all developers for each major dimension of transparency.
The most information is in the area of labor, which we noted previously is largely due to multiple companies clarifying they do not use data labor in building their flagship foundation models.
Beyond this, other areas of large improvement from 2023 FMTI to 2024 FMTI are precisely those with large amounts of new information.
Namely, more than 10\% of the 233 pieces of new information are in the areas of compute and usage policy. 

\paragraph{Companies disclose new information for which they do not score points.}
For example, several companies provide additional details about their data labor practices like general information about instructions to annotators and assurances that wages exceed local minimum wage.
And on the indicator about data creators, which is awarded to only one of the 14 developers (BigCode/Hugging Face/ServiceNow), AI21 Labs discloses that ``Most of the internet-connected population is from industrialized countries, wealthy, younger, and male, and is predominantly based in the United States.''
While these disclosures are insufficient for the stated standards for the associated indicators, we nonetheless emphasize that it may still be valuable for developers to make such information public.
In a few cases, companies provide public justifications for why they do not disclose certain information.
Most notably, Aleph Alpha states that ``In line with the German copyright act (`Urheberrechtsgesetz'), data used for training is to be deleted after use and therefore can not be made available or distributed to external parties'' in relation to the indicators about data access.

\section{Conclusion}
Along with the previous chapter, this chapter is the largest contribution of this dissertation, when measured in terms of the number of pages allocated to it, and of my PhD, when measured in the time I spent on these specific works. 
In part, this reflects not only that both works were sustained over a multi-year span, but also that these works were designed to be highly comprehensive.
In some ways, these strengths are weaknesses: maintaining long-term efforts is not in keeping with the prevailing incentives in academia for novelty, and breadth introduces overhead for understanding the takeaways of this work, which are all the more important in today's society where attention is scarce.
Put together, the works speak to the central theme of the middle of my PhD: to advance the transparency of the foundation model ecosystem to extract empirical insights that sharpen the baseline conceptual account.
And the works complement each other in clear ways along two fundamental axes: HELM produces new information at the model level whereas FMTI discloses existing information at the organization level. 
In turn, this invites future work to complete this grid: better reporting practices for companies when publishing results of their own model evaluations and better investigative mechanisms for externals to understand AI companies.\footnote{In part, I have worked to some extent on the former problem, most notably in my work on train-test overlap that begin in HELM Appendix G \citep{liang2023holistic} and was made more complete in our position paper on the subject \citep{zhang2024languagemodeldevelopersreport}.}

\chapter{Evidence-based Policy}\label{chapter:policy}
\chaptermark{\small Evidence-based policy}
How do we capitalize on a superior understanding of the societal impact of foundation models?
While understanding the societal impact of foundation models has standalone value, this understanding for me is instrumental rather than terminal.
This chapter demonstrates how academic insight can guide society to better outcomes during a critical transitional period.
Specifically, I focus on public policy: policy of some form is necessary to govern a generational technology, but what form should it take and how should it use evidence?
While evidence-based policy has a storied history of striking successes and striking failures in more mature policy domains, here I put forward a vision for evidence-based AI policy.
In that frame, I show how research can be tightly coupled with policy across the policymaking process via evidence review, compliance pre-assessment, and policy design. 
This chapter, in turn, also expands the frontier of computer science, namely to improve public policy.
\newpage 
\noindent Policymakers around the world are grappling with how to govern this powerful technology. AI policy should advance AI innovation by ensuring its potential benefits are responsibly realized and widely shared. To achieve this goal, AI policymaking should place a premium on evidence: scientific understanding and systematic analysis should significantly inform policy, and policy should accelerate evidence generation.

While it may seem self-evident that evidence should inform policy, this is far from inevitable in the inherently messy policy process. Policy outcomes reflect institutional constraints, political dynamics, electoral pressures, stakeholder interests, media environment, economic considerations, cultural contexts, and leadership perspectives. 
Well-designed policy should integrate evidence that reflects scientific understanding and process \citep{cartwright2012evidence}. 
The harder question is how to optimize the relationship between evidence and policy to address the opportunities and challenges of increasingly powerful AI – the task this paper tackles.

\paragraph{What is evidence-based AI policy?}
Developing evidence-based policy is critical for AI, especially considering its critical role within the extant policy ecosystem. 
For example, the bipartisan Evidence-Based Policymaking Act of 2018 (Evidence Act), passed during the first Trump administration, requires federal agencies to provide evidence-building plans to justify funding, share non-sensitive data to enable research, and designate a Chief Evaluation Officer to oversee agency-wide evaluation of program effectiveness. 
Defining what counts as (credible) evidence is the first hurdle for evidence-based AI policy: more mature policy domains showcase differing norms for evidence. 
In health policy, evidence generally refers to randomized control trial results or observational data \citep{brownson2009understanding}. 
In economic policy, evidence tends to be more expansive, encompassing theoretical approaches (\eg macroeconomic forecasts) alongside data-driven approaches \citep{reiss2016error}. 

An evidence-based approach to AI policy should take advantage of empirical observations and leverage the scientific method. 
It also should recognize that many of the most important benefits and risks will require a broader range of analytical tools to ground policy in rigorously informed foresight. 
To that end, well-formulated evidence-based AI policy integrates multiple knowledge types across varying time horizons and levels of uncertainty. 
Such a framework includes (i) empirical observations of AI system behaviors and their impacts, (ii) historical comparisons, (iii) simulations or other analytical approaches to explore possible future outcomes, and (iv) evaluations. 
These predictive tools are critical to project nonlinear development in complex, dynamically evolving systems. Their credibility fundamentally hinges on systematicity in approach, transparency in methodology, and openness to refutation.

Crucial to this framework is recognizing that key sources of evidence in AI spans across a range of sectors and disciplines, encompassing diverse expert viewpoints and stakeholders with varying incentives.
Expertise in AI comes from a range of sectors and disciplines, encompasses a wide range of viewpoints, and includes stakeholders with a range of incentives. 
Academia has produced many of the defining papers on AI, but conventional incentives for promotion and tenure processes can slow evidence generation and result in the deprioritization of important research questions. 
The government houses deep domain expertise in agencies, national labs, and enforcement entities on biosecurity, cybersecurity, fraud, sexual imagery, discrimination and competition, but in some cases is constrained, for example, by national security imperatives. 
AI companies possess world-leading expertise on and evidence about AI capabilities and risks, but by their private sector nature are incentivized to prioritize profit generation.

By rigorously synthesizing multiple evidence streams from presently siloed areas of expertise, evidence-based policy is best positioned to produce desirable societal outcomes in accurate, information-rich evidence environments. By better aligning incentives, carefully tailored policy can help reduce the risk that evidence-based policy will be co-opted to justify inaction or promote negative societal outcomes \citep{casper2025evidence}. 
Historically, the tobacco industry relied on inconclusive studies to play up uncertainty, inhibiting policy to address documented tobacco-related harms \citep{proctorCancerWarsHow1994}. 
Similarly, fossil fuel companies misled the public about climate change despite company-internal reports that anticipated severe harms \citep{americanpetroleuminstituteTwoEnergyFutures1980}. 
These examples clarify not only the importance of the ambient evidence environment, but also the need for careful execution for evidence-based AI policy to advance positive public outcomes.

Evidence-based policy benefits from evidence that is not only credible, but also actionable. 
A focus on marginal risk \citep{kapoor2024societal}, meaning the additional risks posed by AI compared to existing technologies like internet search engines, identifies issues uniquely caused or exacerbated by AI as well as appropriate interventions to address these new risks. 
We present two forms of highly actionable evidence: (i) \textit{evidence reviews} that guide prioritization decisions by clarifying which threat vectors do or do not have resolution evidence of marginal risk and (ii) \textit{compliance pre-assessments} that give real time understanding of the status quo, absent a policy, to reason about the policy's marginal impact.

\section{Evidence Review}
I present an evidence review based on our work in \citet{kapoor2024societal}.
This evidence review addresses open foundation models , primed by high-stakes policy processes in the United States and worldwide.
At the time of this work (2023 -- 2024), and even still at the time of writing, policymakers were actively confronting how to govern open foundation models.
In the United States, Executive Order 14110 on Safe, Secure, and Trustworthy Development and Use of Artificial Intelligence mandated the Department of Commerce to prepare a report for the President on the benefits and risks of open foundation models \citep{EO14110}.
And the subject was equally relevant in other jurisdictions: for example, the European Union granted a partial exemption from obligations to open models under the EU AI Act, but the specifics of what qualified and why were unclear.

Given widespread disagreement within the AI community and uncertainty on how to design AI policy in light of differing releasing strategies, we conducted a multi-stage evidence review.
In short, I will present this in three steps: (i) the benefits of open foundation models at a high level, (ii) the risks of open foundation models with a more granular evidence review of specific works and (iii) the disparate impact of specific policies on open vs. closed foundation model developers.

\subsection{Benefits}
I identify 5 categories of benefits arising from open foundation models in particular.

\paragraph{Distributing who defines acceptable model behavior.}
\textit{Broader access and greater customizability expand who is able to specify the boundary of acceptable model behavior.
}

Developers of closed foundation models exercise unilateral control in determining what is and is not acceptable model behavior.
Given that foundation models increasingly intermediate critical societal processes \citep[\eg access to information, interpersonal communication;][]{lazar2023algorithmiccity}, much as social media platforms do today, the definition of what is acceptable model behavior is a consequential decision that should take into account the views of stakeholders and the context where the model is applied.
In contrast, while developers may initially specify and control how the model responds to user queries, downstream developers who use open foundation models can modify them to specify alternative behavior.
Open foundation models allow for greater diversity in defining what model behavior is acceptable, whereas closed foundation models implicitly impose a monolithic view that is determined unilaterally by the foundation model developer. 

\paragraph{Increasing innovation.}
\textit{Broader access, greater customizability, and local inference expand how foundation models are used to develop applications.
}

Since open foundation models can be more aggressively customized, they better support innovation across a range of applications.
In particular, since adaptation and inference can be performed locally, application developers can more easily adapt or fine-tune models on large proprietary datasets without data protection and privacy concerns. Similarly, the customizability of open models allows improvements such as furthering the state-of-the-art across different languages~\citep{pipatanakul_typhoon_2023}.
While some developers of closed foundation models provide mechanisms for users to opt out of data collection, the data storage, sharing, and usage practices of foundation model developers are not always transparent \citep{bommasani2023fmti}.

However, the benefits of open foundation models for innovation may have limits due to potential comparative disadvantages in improving open foundation models over time. For example, open foundation model developers generally do not have access to user feedback and interaction logs that closed model developers do for improving models over time.
Further, because open foundation models are generally more heavily customized, model usage becomes more fragmented and lessens the potential for strong economies of scale. However, new research directions such as merging models might allow open foundation model developers to reap some of these benefits ~\citep[akin to open source software;][]{raffel_building_2023}.
More generally, the usability of foundation models strongly influences innovation \citep{vipra2023concentration}: factors beyond whether a model is released openly such as the capabilities of the model and the quality of potential inference APIs shape usability. 

\paragraph{Accelerating science.}
\textit{Broader access and greater customizability facilitate scientific research. The availability of other key assets (especially training data) would further accelerate scientific research.}

Foundation models are critical to modern scientific research, within and beyond the field of artificial intelligence. 
Broader access to foundation models enables greater inclusion in scientific research, and model weights are essential for several forms of research across AI interpretability, security, and safety. 
Ensuring ongoing access to specific models is essential for the scientific reproducibility of research, something that has been undermined to date by the business practice of closed model developers to retire models regularly~\citep{kapoor_openais_2023}.
And since closed foundation models are often instrumented by safety measures by developers, these measures can complicate or render some research impossible.
For example, \citet{park_social_2022} use foundation models without safety filters because their research aims to simulate human behavior (including toxic speech). 
Most closed foundation models, which are generally available through content-moderated APIs, would suppress these outputs.

However, model weights alone are insufficient for several forms of scientific research.
Other assets, especially the data used to build the model, are necessary. 
For example, to understand how biases propagate, and are potentially amplified, requires comparisons of data biases to model biases, which in turn requires access to the training data \citep{wang_directional_2021}. 
Access to data and other assets, such as model checkpoints, has already enabled wide-ranging downstream research~\citep{tian2023joma,choi2023tools,longpre2023pretrainer}.
While some projects prioritize accessibility to such assets with the stated goal of advancing scientific research on foundation models~\citep{scao2022bloom,biderman2023pythia}, it is not common for open models in general.
In fact, even the basic validity of a model's evaluation depends on some transparency about the training data. For example, issues such as contamination might lead to overoptimistic results on benchmarks~\citep{kapoor_promises_2024,narayanan_gpt-4_2023}. Access to information about the data can allow us to assess the amount of overlap between the training data and the test set.

\paragraph{Enabling transparency.} 
\textit{Broad access to weights enables some forms of transparency. The availability of other key assets (such as documentation and training data) would further improve transparency.}

Transparency is a vital precondition for responsible innovation and public accountability.
Yet digital technologies are plagued by problematic opacity \citep[see][\S2.2]{bommasani2023fmti}.
Widely available model weights enable external researchers, auditors, and journalists to investigate and scrutinize foundation models more deeply.
In particular, such inclusion is especially valuable given that the foundation model developers often underrepresent marginalized communities that are likely to be subject to the harms of foundation models.
The history of digital technology demonstrates that broader scrutiny, including by those belonging to marginalized groups that experience harm most acutely, reveals concerns missed by developers \citep{sweeney2013discrimination, noble2018algorithms, buolamwini2018gender, raji2019actionable}. The Foundation Model Transparency Index as of May 2024 indicates that developers of major open foundation models tend to be more transparent than their closed counterparts \citep{bommasani2024fmti}.

Still, model weights only make some types of transparency (\eg evaluations of risk) possible, but they do not guarantee such transparency will manifest.
More generally, model weights do not guarantee transparency on the upstream resources used to build the foundation model (\eg data sources, labor practices, energy expenditure) nor transparency on the downstream impact of the foundation model (\eg affected markets, adverse events, usage policy enforcement).
Such transparency can help address prominent societal concerns surrounding bias \citep{birhane_into_2023,thiel2023identifying}, privacy \citep{ippolito_preventing_2023}, copyright \citep{henderson2023foundation, lee2023talkin, longpre2023data}, labor \citep{perrigo2023openai, hao2023cleaning}, usage practices \citep{narayanan2023transparencyreports}, and demonstrated harms \citep{guha2023ai}.

\paragraph{Mitigating monoculture and market concentration.} 
\textit{Greater customizability mitigates the harms of monoculture and broader access reduces market concentration.}

Foundation models function as infrastructure for building downstream applications, spanning market sectors \citep{bommasani2021opportunities, bommasani2023ecosystem, vipra2023concentration, cma2023ai}.
By design, they contribute to the rise of algorithmic monoculture \citep{kleinberg2021monoculture, bommasani2022homogenization}: many downstream applications depend on the same foundation model.
Monocultures often yield poor societal resilience and are susceptible to widespread systemic risk: consider the Meltdown and Spectre attacks, which led to massive security risks because of the widespread dependence on Intel and ARM-based  microprocessors~\citep{kocher_spectre_2018,lipp_meltdown_2018,staff_meltdown_2018}. 
Further, foundation model monocultures have been conjectured to lead to correlated failures \citep{bommasani2022homogenization} and cultural homogenization \citep{lee2022language, padmakumar2023does}.
Since open foundation models are more easily customized, they may yield more diverse downstream model behavior, thereby reducing the severity of homogeneous outcomes.

Broad access to model weights and greater customizability further enable greater competition in downstream markets, helping to reduce market concentration at the foundation model level from vertical cascading.
In the foundation model market, there are barriers to entry for low-resource actors in developing foundation models given their significant capital costs \citep{vipra2023concentration, cma2023ai}.
For example, training the Llama 2 series of models required 3.3 million GPU hours on NVIDIA A100-80GB GPUs \citep{touvron2023llama2}: at February 2024 cloud computing rates of \$1.8/GPU hour~\citep{lambda_gpu_2024},
training this model would cost around \$6 million.
Further, while open foundation models may increase competition in some regions of the AI supply chain, they are unlikely to reduce market concentration in the highly concentrated upstream markets of computing and specialized hardware providers~\citep{widder2023open}.

\subsection{Risks}
\begin{table*}[ht!]
  \centering
  \begin{adjustbox}{width=0.8\textwidth}
    \begin{tabular}{p{4.8cm}|p{3cm}|p{0.9cm}p{0.9cm}p{0.9cm}p{0.9cm}p{0.9cm}p{1.3cm}}
      \textbf{Misuse risk} & \textbf{Paper} & \rot{Threat identification} & \rot{Existing risk (absent open FMs)} & \rot{Existing defenses (absent open FMs)} & \rot{Evidence of marginal risk} & \rot{Ease of defense} & \rot{Uncertainty/assumptions}\\
      \midrule
      Spear-phishing scams & \citet{hazell_large_2023} & \bcircle & \wcircle & \ecircle & \ecircle & \wcircle & \ecircle\\
      \rowcolor[gray]{0.9} 
      Cybersecurity risk & \citet{seger2023open} & \wcircle & \ecircle & \wcircle & \ecircle & \wcircle & \ecircle\\
      Disinformation & \citet{musser_cost_2023} & \bcircle & \wcircle & \ecircle & \ecircle & \wcircle & \bcircle\\
      \rowcolor[gray]{0.9} 
      Biosecurity risk & \citet{gopal_will_2023} & \bcircle & \ecircle & \wcircle & \ecircle & \ecircle & \ecircle\\
      Voice-cloning scams & \hyperlink{ovadya_reducing_2019}{Ovadya et al.} \citeyearpar{ovadya_reducing_2019} & \bcircle & \wcircle & \wcircle & \wcircle & \wcircle & \bcircle\\
      \rowcolor[gray]{0.9} 
      Non-consensual intimate imagery & \citet{lakatos_revealing_2023} & \bcircle & \wcircle & \ecircle & \wcircle & \wcircle & \ecircle\\
      Child sexual abuse material & \citet{thiel_generative_2023} & \bcircle & \bcircle & \bcircle & \bcircle & \bcircle & \bcircle\\
      \bottomrule
    \end{tabular}
  \end{adjustbox}
\caption{\textbf{Evidence review of risks of open foundation models.}
\bcircle{} indicates the step of our framework is clearly addressed;  \wcircle{} indicates partial completion; \ecircle{} indicates the step is absent in the misuse analysis. Incomplete assessments do not indicate that the analysis in prior studies is flawed, only that these studies, on their own, do not show an increased marginal societal risk stemming from open foundation models.. 
}
  \label{table:risk_assessment}
\end{table*}

Technologists and policymakers have worried that open foundation models present risks, in particular, due to the inability to monitor, moderate, or revoke access.
We survey the literature on misuse vectors specifically associated with open foundation models, identifying biosecurity, cybersecurity, voice cloning scams, spear phishing, disinformation, non-consensual intimate imagery, and child sexual abuse material \citep{seger2023open,thiel_generative_2023,maiberg_civitai_2023}.\footnote{Some have also discussed that (open) foundation models may contribute to existential risk via speculative AI takeover scenarios, which we do not consider here.}
To understand the nature of these risks, we present a framework that centers \textit{marginal} risk: what additional risk is society subject to because of open foundation models relative to pre-existing technologies, closed models, or other relevant reference points? 

To assess the risk of open foundation models for a specific misuse vector, we present a six-point framework.
Underpinning this is an emphasis on communicating assumptions and uncertainty: misuse vectors often involve complex supply chains and the capabilities of foundation models are rapidly evolving, meaning the balance of power between attackers and defenders can be unstable~\citep{shevlane_offense-defense_2020}. 

The risk framework enables precision in discussing the misuse risk of open foundation models and is based on the threat modeling framework in computer security~\citep{drake_threat_2021,Shostack2014ThreatModeling, 10177704, Seaman2022CyberThreat, drake_threat_2021}.
For example, without clearly articulating the marginal risk of biosecurity concerns stemming from the use of open (natural) language models, researchers might come to completely different conclusions about whether they pose risks: open language models can generate accurate information about pandemic-causing pathogens \citep{gopal_will_2023}, yet such information is publicly available on the Internet, even without the use of open language models \citep{guha2023ai}.\footnote{In addition, two recent studies found that access to language models does not significantly increase access to information required to carry out biosecurity attacks compared to Internet access~\citep{mouton_operational_2024,patwardhan_building_2024}. More importantly, access to information might not be a major barrier for carrying out such attacks---stronger interventions might lie downstream~\citep{batalis_can_2023}.}

\paragraph{1. Threat identification.}
All misuse analyses should systematically identify and characterize the potential threats being analyzed~\citep{Shostack2014ThreatModeling, 10177704, Seaman2022CyberThreat, drake_threat_2021}. 
In the context of open foundation models, this would involve naming the misuse vector, such as spear-phishing scams or influence operations, as well as detailing the manner in which the misuse would be executed. 
To present clear assumptions, this step should clarify the potential malicious actors and their resources: individual hackers are likely to employ different methods and wield different resources relative to state-sponsored entities. 

\paragraph{2. Existing risk (absent open foundation models).}
Given a threat, misuse analyses should clarify the existing misuse risk in society. 
For example, \citet{seger2023open} outline the misuse potential for open foundation models via disinformation on social media, spear-phishing scams over email, and cyberattacks on critical infrastructure. 
Each of these misuse vectors already are subject to risk \textit{absent} open foundation models: understanding the pre-existing level of risk contextualizes and baselines any new risk introduced by open foundation models.

\paragraph{3. Existing defenses (absent open foundation models).}
Assuming that risks exist for the misuse vector in question, misuse analyses should clarify how society (or specific entities or jurisdictions) defends against these risks.
Defenses can include technical interventions (\eg spam filters to detect and remove spear-phishing emails) and regulatory interventions (\eg laws punishing the distribution of child sexual abuse material). 
Understanding the current defensive landscape informs the efficacy, and sufficiency, with which new risks introduced by open foundation models will be addressed.

\paragraph{4. Evidence of marginal risk.}
The threat identification, paired with an analysis of existing risks and defenses, provides the conceptual foundation for reasoning about the risks of open foundation models.
Namely, subject to the status quo, we can evaluate the \textit{marginal risk} of open foundation models.
Being aware of existing risk clarifies instances where open foundation models simply duplicate existing risk (\eg an open language model providing biological information available via Wikipedia).
Similarly, being aware of existing defenses clarifies instances where open foundation models introduce concerns that are well-addressed by existing measures~\citep[\eg, email and OS-based filters detecting spear-phishing emails, whether human or AI-generated;][]{craigmarcho_ie7_2007, apple_support_safely_2023, google_email_2023}.
Conversely, we can identify critical instances where new risks are introduced~\citep[\eg fine tuning models to create non-consensual intimate imagery of specific people;][]{maiberg_inside_2023} or where existing defenses will be inadequate~\citep[\eg AI-generated child sexual abuse material may overwhelm existing law enforcement resources;][]{harwell_ai-generated_2023}.

Further, the marginal risk analysis need not only be conducted relative to the status quo, but potentially relative to other (possibly hypothetical) baselines.
For example, understanding the marginal risk of open release relative to a more restricted release (\eg API release of a closed foundation model) requires reasoning about the relevant existing defenses for said restricted release.
This perspective ensures greater care is taken to not assume that closed releases are intrinsically more safe and, instead, to interrogate the quality of existing defenses \citep[\eg the fallibility of existing API safeguards;][]{qi2023finetuning}.

\paragraph{5. Ease of defending against new risks.}
While existing defenses provide a baseline for addressing new risks introduced by open foundation models, they do not fully clarify the marginal risk.
In particular, new defenses can be implemented or existing defenses can be modified to address the increase in overall risk.
Therefore, characterizations of the marginal risk should anticipate how defenses will evolve in reaction to risk: for example, (open) foundation models may also contribute to such defenses (\eg the creation of better disinformation detectors; \citet{zellers2019neuralfakenews} or code fuzzers; \citet{liu_ai-powered_2023}).

\paragraph{6. Uncertainty and assumptions.}
Finally, it is imperative to articulate the uncertainties and assumptions that underpin the risk assessment framework for any given misuse risk. 
This may encompass assumptions related to the trajectory of technological development, the agility of threat actors in adapting to new technologies, and the potential effectiveness of novel defense strategies. 
For example, forecasts of how model capabilities will improve or how the costs of model inference will decrease would influence assessments of misuse efficacy and scalability.

Using our risk assessment framework, we assess past studies that span different risk vectors in \autoref{table:risk_assessment}.
We find that the risk analysis is incomplete for six of the seven studies we analyze.
To be clear, incomplete assessments do not necessarily indicate that the analysis in prior studies is flawed, only that these studies, on their own, are insufficient evidence to demonstrate increased marginal societal risk from open foundation models. 

We instantiate the framework for two misuse risks, providing preliminary analyses of cybersecurity risks stemming from automated vulnerability detection and the risk of digitally altered NCII. 
For the former, we find that the current marginal risk of open foundation models is low and that there are several approaches to defending against the marginal risk, including using AI for defense. 
For the latter, open foundation models pose considerable marginal risk at present, and plausible defenses seem hard. 
Note that these are not the only risks from foundation models~\citep{barrett_identifying_2023}---for example, the creation of malware is another cybersecurity risk that requires separate analysis---yet when researchers talk about cybersecurity risks of open foundation models, they often club together different threats. 
This illustrates how the framework helps clarify the points of contention in debates on open foundation models. Critically, while many of the same properties of open foundation models are relevant for analyzing different misuse vectors (such as the inability to revoke access), the risk assessment framework helps introduce specifics that differentiate the misuse vector, for instance, by pointing out elements of the misuse supply chain where risk is better addressed. 

As the capabilities of foundation models (including open models) improve, the risk assessment framework can guide analyses of societal risks from increasing capability by providing a grounded analysis of whether model releases bring about increased marginal risk to society. Still, it is important to note the limitations on the scope of the framework's applicability. First, while the risk assessment framework can help clarify the societal risks of releasing a foundation model openly, note that it is not a complete framework for making release decisions since it does not provide a mechanism for trading the marginal benefits of openly releasing models against the marginal risk, nor does it look at the opportunity cost of \textit{not} releasing a model openly. Second, while the framework allows an evaluation of the risk of releasing models openly for known risks (such as cybersecurity, biosecurity etc.), it does not account for \textit{unknown unknowns}---risks that we have no prior understanding of. 
Third, there could be a number of coordination issues among actors for figuring out when to release models---for example, to reduce the risk of NCII, open model developers would need to coordinate with social media platforms as well as other downstream platforms like CivitAI. While the framework allows us to identify such opportunities, it does not automatically bring about the coordination of these actors.
Overall, while the framework improves the precision, rigor, and completeness of risk assessment, we expect other approaches to analyzing risk will be needed for addressing these limitations.  

\subsection{Impact}
As global policy efforts focus on foundation models, how do policy initiatives affect open foundation models?
Would policies impose greater compliance burdens on open foundation model developers compared with their closed counterparts, even though open developers are usually less-resourced?
Here I review analytical arguments for the relationship between proposed policies and foreseeable impact on open model developers based on \citet{bommasani2024open}.

\paragraph{Liability for downstream use.}
Because the distinction between open and
closed foundation models is predicated on
release, policies that impose penalties for
certain uses of foundation models are likely
to have differential impacts. 
The most famous such proposal is SB 1047 in California.
SB 1047 was a widely-discussed bill in 2024: the bill was proposed by Senator Scott Weiner, passed in both chambers of the California legislature, and vetoed by Governor Gavin Newsom.
SB 1047 would, potentially, impose liability for the downstream use of a foundation model, including for derivatives of the
foundation model that are the result of finetuning. 
Such policies aim to introduce
penalties for the release of unsafe models
that, potentially after modification, catalyze
misuse. 
Yet, liability for such downstream
harms could chill the open foundation model
ecosystem by exposing open foundation
model developers to severe liability risk. 
By
contrast, because closed foundation model
developers exercise greater control over
downstream use, some developers already
provide liability protections to downstream
users of their models (\eg Google offers users
of its generative AI products indemnification
for copyright claims). Although clarifying or
increasing liability for downstream use may
have benefits, these legislative proposals expose a broad and hard-to-control liability surface for open foundation model developers.
\\
\noindent \textit{Note on SB 1047 and this work.}
Much discussion of evidence-based AI policy in this chapter, and my related writings, were used extensively as the scientific basis to veto SB 1047.
As the most salient example, federal-level legislators released a public letter to California Governor Gavin Newsom, recommending he veto the bill as not being in the interest of California.
The legislators cite my work as discussed in this dissertation repeatedly, writing: ``There is little scientific evidence of harm of mass casualties or harmful weapons created \dots Understanding, measuring and monitoring the risks inherent in AI systems as they will evolve will be key---especially the marginal risk \dots more work also needs to be done to develop the AI governance workforce and reduce fragmentation in the AI community \dots policies that force closed development, while they may sound good in theory, would decimate the ecosystems that spring up around AI models.''

\paragraph{Content provenance for downstream use.}
Given that the most salient applications of
foundation models are generative AI systems, there is demand for content provenance techniques, such as watermarking, to
detect machine-generated content. Content
provenance could help with the tracking or
moderation of AI-generated content, such
as deepfakes, CSAM, and NCII. But akin
to liability, if foundation model developers
must ensure content provenance for downstream use, then these requirements may be
technically infeasible for open foundation
model developers.
The US Executive Order 14110, White House
voluntary commitments on AI, the Canadian Voluntary Code of Conduct, Chinese generative
AI regulations, and the G7 international code
of conduct all highlight content provenance.
However, today’s watermarking methods for
language models do not persist if models are
modified (\eg fine-tuned) and require that
users of a model follow certain protocols for
the watermarking guarantee to hold. Fundamentally, open foundation model developers
do not control how their models are modified
or used to generate content.

\paragraph{Liability for open data.}
Although foundation models do not require
the release of the underlying data used to
build the model, some developers choose to
release both the model weights and the training data. 
Multiple open foundation model developers tend to
release data openly. 
However, open release of
data exposes these entities to greater liability risk, as exemplified by lawsuits against
Stability AI based on its use of datasets from
the nonprofit Large-scale Artificial Intelligence Open Network (LAION) that allegedly included plaintiffs’ work. 
Although the
legality of training foundation models on
copyrighted data remains unclear across
many jurisdictions, the status quo presents
perverse incentives. 
Namely, model developers that transparently disclose and openly
provide data are subject to greater risk than
developers that conceal the data that they
use, even if the underlying facts are identical.
Considering this perverse incentive, government-mandated disclosure of training data
may be beneficial in some cases.

Public policy on AI should consider both open and closed
foundation model developers. 
When regulations directly address open foundation models, the precise definition used to identify these models and developers should be duly
considered.
Hinging regulation exclusively on open weights may not be appropriate given the gradient of release. 
And even when regulations do not directly address open
foundation models, they may have an adverse impact. 
Consequently, if policymakers are to implement such policies, they should directly consult the open foundation model community, with due consideration given to their interests.

\section{Compliance Pre-Assessment}
Evidence review organizes available evidence to best inform policy.
Alongside understanding the state of evidence, which can justify or undermine a proposed policy, policymakers also need to understand the state of industry practice.
What are companies currently doing and how do it deviate from what is proposed under a policy?

We introduce the method of \textit{compliance pre-assessment}: do company practices comply with a policy \textit{as if} the policy was in effect?
These pre-assessments are of incredible value to policymakers because they clarify two matters that policymakers often amorphously conceptualize given their incomplete understanding of market conditions.
First, what are current practices?
Second, how much change is required for companies to come into compliance?
Of course, while these pre-assessments are a useful simulation of what might change if a policy is enacted(namely under the assumptions that companies straightforwardly change their practices to directly comply with policy), many factors complicate the forecasting of policy impact (\eg attempts to circumvent compliance, confusion in policy interpretation, gaps in enforcement capacity).

We conduct a compliance pre-assessment on the most important AI policy to date, the European Union's Artificial Intelligence Act (the EU AI Act) \citep{bommasani2023eu-ai-act}.
The EU AI Act is a watershed moment in both the history of AI and the history of technology policy as it is the first comprehensive legislation to be enacted to govern artificial intelligence.
It, therefore, reflects remarkable prescience and commitment by the European Union.
The European Commission (the EU's executive body) penned the initial White Paper on the AI Act in February 2020 and introduced the first draft of the Act in April 2021.
The European Parliament and the Council of the EU (the EU's two legislative bodies), in conjunction with the Commission, achieved political agreement on the Act through the complex trilogue process in December 2023.
The AI Act was signed into law and entered force in August 2024.
13 independent chairs were recruited by the EU AI Office, the new body created to oversee and enforce the AI Act, to draft the Code of Practice to determine how the chapter of the Act related to foundation models would be interpreted and implemented.
At the time of writing, I serve as one of these chairs and, with Nuria Oliver, oversee the Code of Practice's sections on transparency, which is closely aligned with the work I discuss in this dissertation.
The EU AI Act sets the precedent for all subsequent AI policy: its efficacy will shape the future of artificial intelligence and society far beyond Europe.\footnote{The EU AI Act should also be remembered as a demonstration of the unbelievable impact of a single individual: Luca Bertuzzi. Luca's reporting on the AI Act is not only instrumental to all the impact I had, but the entire policy sphere remaining well-informed on the AI Act as it was negotiated. Researchers and journalists often share the same goals and essentially the same function of pursuing and communicating the truth. I leave this footnote as a reminder that there is immense untapped potential in tighter partnership between researchers and journalists.}

The pre-assessment I describe was a highly unusual yet highly influential injection of evidence into the high-stakes political process of the EU AI Act.
As context, each of the three entities involved in the legislative negotiation (Commission, Council, Parliament) first came to internal agreement within their multi-member body on their individual positions.
The Commission's position came first, given their unique role in the genesis of the Act, and the Council's position came second in late 2022.
Notably, as this happened, the technological world also changed greatly: OpenAI released ChatGPT in late 2022, prompting the Parliament to rethink their still-internal position.
Through a combination of public and private channels, and with the support of Marietje Schaake, given her extended public service in the European Parliament and familiarity with the Brussels political sphere, we received an advanced copy of the in-committee Parliament position on foundation models under the AI Act.
We conducted a compliance pre-assessment using this draft which, in parallel, passed through committee and was approved by a sweeping majority in the Parliament with a vote of 499 in favor, 28 against, and 93 abstentions on June 14, 2024.
We published our compliance pre-assessment the next day (June 15, 2024), which prompted extensive media coverage and policymaker engagement.

I narrate this story because it reveals both the unique opportunity for rigorous research of this form and the unique impact such research can have on a live political practice.
In particular, with extensive support from Marietje Schaake, this research was the basis for direct and extensive engagement with the highest policymakers on the EU AI Act (\ie the Parliament rapporteurs Brando Benifei and Dragos Tudorache, the Council Presidency's representative Carme Artigas).
And, through sustained dialogue with the staff of the key legislators, many of our findings and insights directly appear in the EU AI Act, including through consultation immediately prior to the famed 36-hour final negotiation in December 2023.
In short, under the right conditions, academics can influence the most important policies in our society: we should aspire for this level of impact and understand how to attain it by stepping out of our ivory tower.

\subsection{Methods}
To conduct a compliance pre-assessment, the two core primitives are (i) the policy requirements and (ii) the current practices.
Neither of these constructs are unambiguous in reality.
In particular, for the compliance pre-assessment we conducted, we needed to first interpret an in-flux legislative draft.
Even for a completely finalized piece of legislation or regulation, significant room remains for interpretation, and this is only amplified for an in-progress document.
Similarly, company practices are not clear.
We assess the \textit{stated} practices: for substantive requirements which require companies \textit{do} something, we at best assess whether they \textit{say they do} what is required, even if public transparency is not required.

\paragraph{Policy interpretation.}
We start with the Parliament draft as the policy of interest.
We extract 22 requirements directed towards foundation model providers from the Parliament draft. 
Of these requirements, we select a subset of 12 requirements to assess.
We chose this subset because we believe they could be meaningfully assessed using public information and they could be sensibly expected of companies prior to the policy being enacted (\eg they did not depend on infrastructure the government first would need to create for the companies to use to comply with the policy).
These 12 requirements can be organized into 4 categories: (i) 3 requirements on data resources (3), (ii) 2 requirements on compute resources, (iii) 4 requirements on the model itself, and (iv) 3 requirements on deployment practices. 
Many of these requirements center on transparency: for example, disclosure of what data was used to train the foundation model, how the model performs on standard benchmarks, and where it is deployed. 

\paragraph{Practice documentation.}
We then understood company practices at the time.
Since the policy set out expectations for all providers of foundation models, we did not restrict which entities we considered due to the policy itself.
Instead, we pragmatically decided to scope to 10 leading foundation model providers, including the two most influential EU providers at the time (Aleph Alpha, Mistral).
As with the Foundation Model Transparency Index, we assessed companies in relation to their flagship foundation model in our judgment, though companies in practice would need to comply for all their foundation models under the Parliament proposal.
For each of these models, we then systematically gathered information from public sources, following a procedure similar to what we did in the 2023 Foundation Model Transparency Index \citep{bommasani2023fmti}.

\paragraph{Assessment procedure.}
To assess practices in light of the policy, we designed an assessment procedure.
Since the 12 requirements were high-level, we mapped each requirement to a five-point rubric (\ie scores in the range 0--4).
Of course, our rubric could diverge from how legislators, enforcers, or companies interpreted the policy, but in doing so we also implicitly clarified the lower-level terrain in advance of clarity from policymakers working on the EU AI Act.
Even absent other contributions in the assessment, this in itself proved useful by priming EU policymakers to strive for greater precision, and thereby less confusion and ambiguity, in writing the EU AI Act.
Kevin Klyman and I each scored all 10 companies for all 12 requirements independently (Cohen’s $\kappa$ = 0.74), merging scores through panel discussion.
Therefore, each company received a single final score for each requirement, yielding overall compliance pre-assessment scores out of 48.

\subsection{Results}
\begin{figure}
\centering
\includegraphics[width=\textwidth]{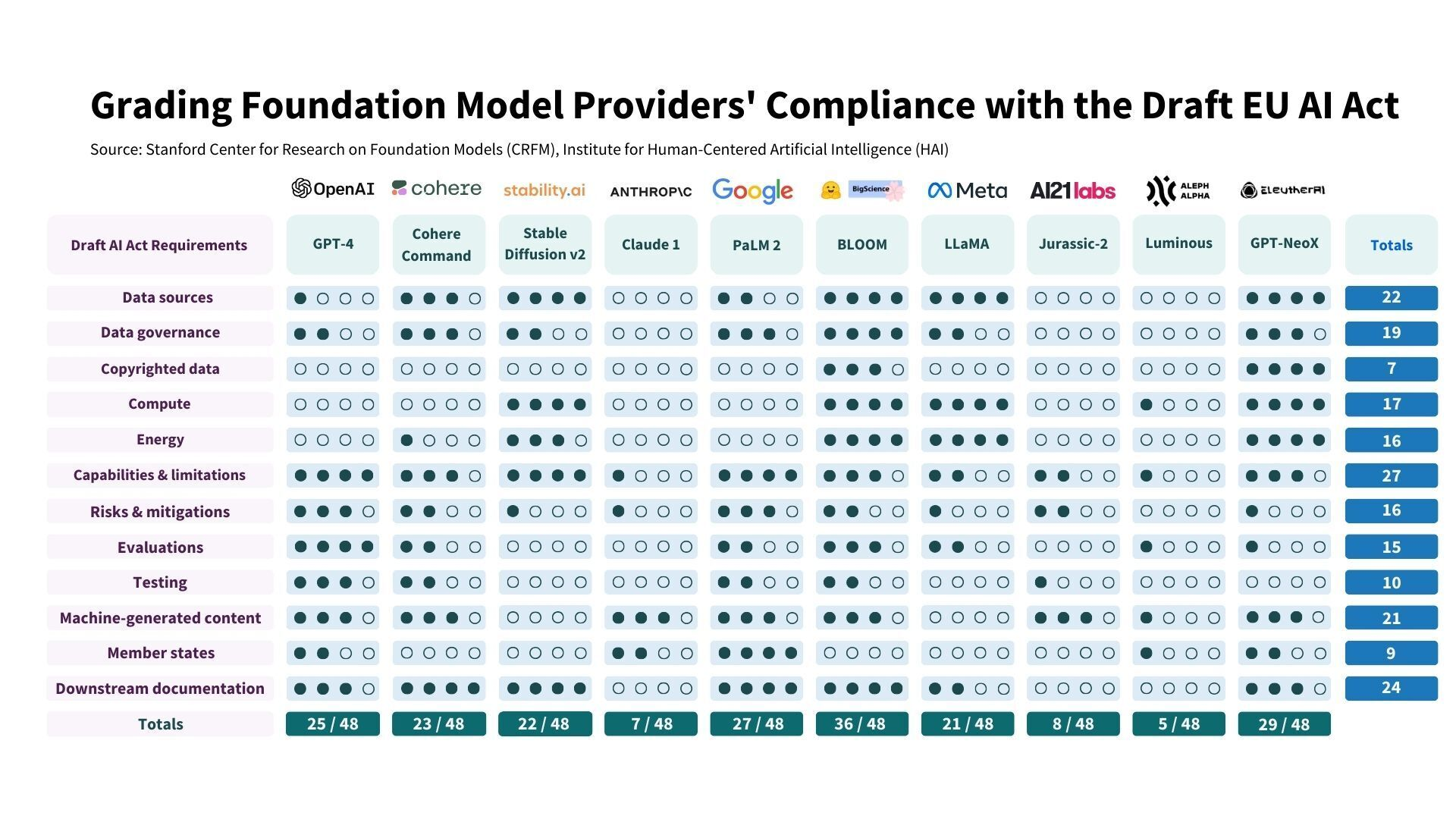}
\caption{\textbf{Compliance pre-assessment scores.} 
We assess 10 major foundation model providers (and their flagship models) for the 12 AI Act requirements on a scale from 0 (worst) to 4 (best). 
The best possible score is 48 as a result.
We present the final scores in the above figure with the justification for every grade made available. Our results demonstrate a striking range in compliance across model providers.
Even for the highest-scoring providers, there is still significant margin for improvement. This confirms that the Act (if enacted, obeyed, and enforced) would yield significant change to the ecosystem, making substantial progress towards more transparency and accountability.}
\label{fig:compliance}
\end{figure}

\autoref{fig:compliance} shows the results of the compliance pre-assessment. 
We see four areas where many organizations receive poor scores (generally 0 or 1 out of 4). 
They are (i) copyrighted data, (ii) compute/energy, (iii) risk mitigation, and (iv) evaluation/testing. 
These speak to established themes in the scientific literature.
In large part, my subsequent work on the Foundation Model Transparency Index has more rigorously and extensively confirmed these points.

\paragraph{Unclear liability due to copyright.}
Few providers disclose any information about the copyright status of training data. 
Many foundation models are trained on data that is curated from the Internet, of which a sizable fraction is likely copyrighted. 
The legal validity of training on this data as a matter of fair use, especially for data with specific licenses, and of reproducing this data, remains unclear.
In particular, a key concern is whether the existence of EU law on this subject would have cross-jurisdictional impacts (\eg on active US-based litigation).

\paragraph{Uneven reporting of energy use.} 
Foundation model providers inconsistently report energy usage, emissions, their strategies for measurement of emissions, and any measures taken to mitigate emissions. 
How to measure the energy required to train foundation models is contentious \citep{strubell2019energy, patterson2021carbon}.
Regardless, the reporting of these costs proves to be unreliable, in spite of many efforts that have built tools to facilitate such reporting \citep{lacoste2019quantifying, henderson2020towards, luccioni2023counting}.

\paragraph{Inadequate disclosure of risk mitigation/non-mitigation.} The risk landscape for foundation models is immense, spanning many forms of malicious use, unintentional harm, and structural or systemic risk. 
While many foundation model providers enumerate risks, relatively few disclose the mitigations they implement and the efficacy of these mitigations. 
The Act also requires that providers describe “non-mitigated risks with an explanation on the reason why they cannot be mitigated”, which none of the providers we assess do.

\paragraph{Absence of evaluation standards/auditing ecosystem.} Foundation model providers rarely measure models’ performance in terms of intentional harms such as malicious use or factors such as robustness and calibration. 
Many in the community have called for more evaluations, but standards for foundation model evaluation (especially beyond language models) remain a work-in-progress. 
Fortunately, since we conducted this pre-assessment, evaluation practices have improved some but still remain unsatisfying \citep{weidinger2025evaluationsciencegenerativeai}.

\paragraph{Open vs. closed} 
We find a clear dichotomy in compliance as a function of release strategy, or the extent to which foundation model providers make their models publicly available. 
Open releases generally achieve strong scores on resource disclosure requirements (both data and compute), with EleutherAI receiving 19/20 for these categories. 
However, such open releases make it challenging to monitor or control deployment, with closed releases scoring better on deployment-related requirements. 
For instance, Google’s PaLM 2 receives 11/12 for deployment. 

The relationship between release and area-specific compliance to some extent aligns with our intuitions. 
Open releases are often conducted by organizations that emphasize transparency, leading to a similar commitment to disclosing the resources required to build their foundation models. 
Closed releases, by contrast, often coincide with models that power the provider’s flagship products and services, meaning the resources underlying the model may be seen as more of a competitive advantage (\eg specific data sources) or a liability (\eg copyrighted data). 
In addition, open-sourcing a model makes it much more difficult to monitor or influence downstream use, whereas APIs or developer-mediated access provide easier means for structured access.

Most importantly, our compliance pre-assessment spoke to the viability of compliance with the proposed policy.
We found that no foundation model provider achieves a perfect score, with ample room for improvement in most cases.
But could organizations fully comply with all requirements?
While sufficiently powerful incentives (\eg large fines for noncompliance) could change company conduct, even absent strong regulatory pressure, we concluded meaningful change should be readily available.
That is, many providers could reach total scores in the high 30s or 40s through meaningful, but plausible, changes. 
To be concrete, the entry-wise maximum across OpenAI and Hugging Face/BigScience is 42 (almost 90\% compliance).
While progress in each of these areas requires some work, in many cases we believe this work is minimal relative to building and providing the foundation model and should be seen as a prerequisite for being a responsible and reputable model provider.
We conclude that enforcing the 12 assessed requirements would bring substantive change while remaining within reach for providers.

\section{Policy Design}
To design policy, the most fundamental questions are (i) which entities should comply with the policy and (ii) what they should be expected to do?
In my mind, these two components resemble two standard levels of analysis of language in linguistics.
Namely, the first part on scoping (\ie who complies) is the syntax, structuring the policy.
And the second part is the semantics, providing the substance of the policy, where here I specifically focus on policies that share the quality of generating evidence.
In this way, this section also showcases that policymakers should not just expectantly wait for evidence and sit idle, but instead proactively accelerate the rate of evidence generation.

\subsection{Scoping}
Well-designed regulation is proportionate: The obligations government imposes on an actor are commensurate with the associated risk to ensure regulatory burdens are such that organizations have the resources to comply and not so burdensome as to curb innovation. The United States, along with 29 other governments, committed to adequately considering a proportionate governance approach by signing the Bletchley Declaration at the 2023 U.K. AI Safety Summit: It states that ``\dots countries should consider the importance of a pro-innovation and proportionate governance and regulatory approach that maximises the benefits and takes into account the risks associated with AI. This could include making, where appropriate, classifications and categorisations of risk\dots''.

In this section, I describe my work on scope that I developed in \citet{bommasani2023tiers} and then made more legible to policymakers in \citet{nelson2024ntia} and \citet{bommasani2025ca}.
How should policymakers structure regulation to tailor the obligations to the entities? Thresholds are a standard mechanism for determining the scope of regulation. Thresholds distinguish entities to allow for heightened obligations to be applied to some entities without being applied to others. We define a threshold as a simple decision rule where (i) a quantity of interest, such as model training compute measured in floating point operations (FLOP), is compared to a value, such as $10^{26}$ FLOP such that (ii) entities that surpass the threshold are subject to more regulatory scrutiny than those that do not. Thresholds often exempt certain entities from burdensome compliance to foster innovation. For example, businesses with less than \$500,000 in annual gross revenue do not meet the enterprise coverage threshold, exempting them from recordkeeping obligations under the US Fair Labor Standards Act. Thresholds also constrain regulatory attention to a smaller set of entities that warrant greater focus or scrutiny. For example, US banks with \$250 billion or more in total consolidated assets are designated as ``systemically important'' financial institutions, subjecting them to heightened scrutiny by the Federal Reserve under the Dodd-Frank Wall Street Reform and Consumer Protection Act.

While thresholds are ubiquitous across nearly every policy domain and jurisdiction, their design and application to AI policy remains nascent. As of January 2025, AI-specific thresholds have primarily been based on the computational cost of model training, which is measured in computational operations (\eg, FLOP). Noteworthy examples of compute thresholds include $10^{26}$ FLOP in the Biden Executive Order 14110 on Safe, Secure, and Trustworthy Development and Use of Artificial Intelligence and $10^{25}$ FLOP in the European Union’s AI Act. Other approaches have been considered: California’s proposed and vetoed SB 1047 Safe and Secure Innovation for Frontier Artificial Intelligence Models Act used the monetary cost of training, setting the threshold at \$100 million.

In general, thresholds are imperfect: Most risks vary continuously, whereas thresholds set discrete boundaries to distinguish among different entities. Thresholds are also imperfect proxies: By relying on specific quantifiable metrics, they necessarily center what can be measured even where the underlying risk cannot be fully measured. Useful thresholds for frontier AI governance, therefore, must strike pragmatic compromises between (i) the theoretical constructs that should yield greater regulatory scrutiny and (ii) the practical operationalizations of what can be measured. To this end, thresholds will likely need to be updated on the basis of new scientific evidence to keep pace with the evolving technological state of affairs.

\paragraph{Thresholds for foundation models.}
To proportionately regulate foundation models and the AI supply chain, policymakers must understand the many axes of variation to inform how they set thresholds. 
Supply chain monitoring \citep{bommasani2023ecosystem, cma2023ai} documents significant variation in the organizational profile of model developers, including their corporate status, employee headcount, and geographic location. These developers expend very different amounts of resources to build their models, referring to costs in money, data, computation, and energy. 
The resulting models also differ greatly in fundamental properties like model architectures (\eg, dense models where all model parameters are active vs. sparse models where a small subset are active) and modalities (\eg, Meta’s unimodal text-to-text Llama 3.3 vs. Google’s highly multimodal Gemini 2.0 that can take text, code, images, audio, and/or video as inputs and generate text, audio, and/or images as output). They vary even more in their capabilities and associated risks when evaluated on standard benchmarks. Once development is completed, models all vary greatly in how they are released (\eg, EleutherAI’s fully open Pythia, Meta’s open-weight Llama 3.3, OpenAI’s API-based GPT-4, Google’s fully closed Gopher). 
For all of these reasons and other business reasons, foundation models vary immensely in how they are used downstream and their ultimate impact on the public.

To define a threshold to distinguish foundation model developers, we see four natural approaches:
\begin{enumerate} 
\item \textbf{Developer-level properties:} 
For example, thresholding based on the number of employees could reduce obligations for small companies that may lack the personnel to comply. As an example from another policy domain, the Occupational Safety and Health Administration Recordkeeping and Reporting Occupational Injuries and Illnesses Standard partially exempts businesses with at most 10 employees from maintaining certain logs of work-related injuries and illnesses.
\item \textbf{Cost-level properties:} 
For example, thresholding based on the compute-related cost to develop a model could restrict regulatory attention to capital-intensive models given concerns of barriers to competition. As an example from another policy domain, the Clean Air Act’s program for the Prevention of Significant Deterioration requires permitting for facilities that emit, or have the potential to emit, either 100 tons per year of regulated air pollutant in specific source categories (\eg, chemical plants) or 250 tons per year of regulated air pollutant in all other source categories.
\item \textbf{Model-level properties:} 
For example, thresholding based on the model’s performance on benchmarks that involve identifying software vulnerabilities could identify models that may enable cyberattacks. As an example from another policy domain, the Consumer Product Safety Improvement Act uses laboratory testing for lead presence to limit lead to 100 parts per million of total lead content in any accessible part of children's products intended for children 12 years old and under.
\item \textbf{Impact-level properties:} 
For example, thresholding based on the number of commercial users of a model could identify models with broad dependence that may pose systemic risk. As an example from another policy domain, the European Union’s Digital Services Act imposes greater scrutiny on Very Large Online Platforms and Very Large Online Search Engines, defined as those with at least 45 million monthly active users in the EU.
\end{enumerate}

To determine the appropriate approach for setting a threshold in a given policy context, we recommend that policymakers adopt the default approach of using thresholds that align with the regulatory intent. For example, if the intent is to encourage up-and-coming startups, then shielding them via exemptions for small businesses may be appropriate: This intent could be expressed via a threshold based on business-level head count or revenue. If the intent is to discourage the misuse of certain dual-use capabilities, then categorizing entities based on evaluation results for their models for the associated capabilities may be appropriate. And if the intent is to place greater scrutiny on efforts likely generating large-scale disruption to societal operation and/or large-scale harm from the adoption of their technologies, then an impact-level designation may be appropriate.

While the principle of aligning the regulatory structure (\ie, the methods for thresholding) with the regulatory intent should be the starting point, practical concerns should also be considered. 
In practice, some of the quantities we describe above may be proxies for others.
In several cases, policymakers have already used training compute (a cost-level metric) as a proxy for some combination of greater capabilities, greater misuse risks, and/or greater downstream harm.

While a number of additional considerations influence threshold design \citep[see][]{bommasani2023tiers, heim2024trainingcomputethresholdsfeatures}, we highlight three:
\begin{enumerate}
\item \textbf{Determination time:} 
Different quantities first become measurable at different times. While a foundation model developer may estimate training compute even before training for internal budgeting purposes, the evaluation results for a model can only be determined after development and, still further, the downstream impact for a model can only be determined after deployment. In some cases, estimates can be prepared in advance, but these temporal effects have two key consequences. First, developers can better prepare to comply with policy if they are aware earlier that they will be in-scope. And second, certain obligations are difficult to impose if they require intervention before the determination time: It is difficult to expect a developer to implement certain governance practices for training data if they only determine they exceed a threshold after deployment.
\item \textbf{Measurability:}
Different quantities have differing complexity in how they are measured, spanning both the intrinsic difficulty of measuring the quantity and overarching standardization in how to measure the quantity. In particular, absent significant agreement on how to measure the quantity of interest, progress in implementation may stall as stakeholders disagree on the specifics of measurement.
\item \textbf{External verifiability:}
Different quantities vary in whether external parties of any kind are able to measure them. Prior AI policy in New York highlights that if the regulated entity is the one to unilaterally determine if they are in scope of the policy, then the policy may fail to have the desired effect due to widespread null compliance (Wright et al., 2024). For example, risk evaluation results may be externally reproduced to confirm developer’s designation, whereas training compute may require the ability to monitor the compute resources developers use, and training data size may be entirely unobservable given the lack of access to training data. 
\end{enumerate}

Since policy may have different regulatory intents and existing thresholds vary in their profiles of determination time, measurability, and external verifiability, we agree with \citet{nelson2024ntia} that ``a one-size-fits-all approach or a single threshold metric is inadequate for governance because different AI systems and their outputs present unique challenges and risks.'' To this end, we point to the European Union’s AI Act, which designates models trained with $10^{25}$ FLOP as posing systemic risk as of March 2025 as the default criteria. 
However, the AI Act in Annex XIII affords the regulator flexibility to also consider alternatives metrics, such as the number of parameters, size of the data set, estimated cost or time of training, estimated energy consumption, benchmarks and evaluations of capabilities of the model, and whether the model has a high impact on the internal market due to its reach (either due to at least 10,000 registered business users or the number of registered end-users). Further, to capture fast-moving scientific developments, the AI Act creates a scientific panel that is empowered to issue qualified alerts to identify models that may pose systemic risk even if they are not captured by predefined quantitative thresholds.

Overall, we emphasize that irrespective of which combination of metrics is most appropriate in the present, policymakers should ensure that mechanisms exist not only to update specific quantitative values, given the rapid pace of technological and societal change in AI, but also to change the metrics altogether.

\paragraph{Principles.}
\textit{Generic developer-level thresholds seem to be generally undesirable given the current AI landscape.} Since many small entities can develop hugely influential and potentially risky foundation models, as demonstrated by the Chinese company DeepSeek, the use of thresholds based on developer-level properties may inadvertently ignore key players. 
In fact, major players in the space, such as Anthropic, OpenAI, and xAI, may be relatively small according to some conventional metrics of businesses (\eg, head count). 
At the same time, these approaches may bring into scope massive, established companies in other industries that are simply exploring the use of AI since thresholds based on properties of companies may not distinguish between the entire business and the AI-specific subset. 
Therefore, we caution against the use of customary developer-level metrics that do not consider the specifics of the AI industry and its associated technology. \\

\noindent \textit{Compute thresholds are currently the most attractive cost-level thresholds, but they are best combined with other metrics for most regulatory intents.} Training compute is by far the most common metric for thresholds in existing AI policy given its clear strengths: (i) Fairly authoritative methods exist for measuring compute \citep{fmf2024flops}; (ii) Compute can be estimated prior to the development of the model; and (iii) Compute may be externally verifiable by entities that monitor or operate compute facilities. However, established concerns around (i) the large disparities across modalities for training compute and (ii) the limited predictive power of compute for proxying downstream risk given it does not factor in how models are distributed and adopted in the economy have been cited in critiques of compute \citep{nelson2024ntia, bommasani2023tiers, hooker2024limitationscomputethresholdsgovernance}. 
New developments in AI present further concerns given (iii) the shift from training-time compute allocation to inference-time compute allocation and (iv) the complex accounting involved for synthetic data in training data.

On this basis, we conclude that training compute still appears better than other cost-level metrics that are more complex to calibrate across different developers (\eg, dataset size, monetary cost). As a result, compute thresholds may be best positioned for use as initial filters to cheaply screen for entities that may warrant greater scrutiny, operating alongside other metrics and thresholds in an overall governance framework, aligning with the recommendations of \citep{nelson2024ntia, bommasani2023tiers, heim2024trainingcomputethresholdsfeatures, hooker2024limitationscomputethresholdsgovernance}. \\

\noindent \textit{Thresholds based on risk evaluation results and observed downstream impact are promising for safety and corporate governance policy, but they have practical issues.} Policy that aims to manage risk by influencing developer practices often addresses one or both of (i) misuse of foundation models by malicious actors and (ii) sociotechnical harms from widespread deployment. Therefore, these intents align directly with evaluations of risk (\eg, model performance at generating child sexual abuse material or cyberattacks) as well as the downstream footprint of models (\eg, if models are integrated into high-stakes decision-making systems for hiring or benefits determination). 

However, the question of which risk evaluations are sufficiently trustworthy looms in the air: The evaluations conducted by the US AI Safety Institute and UK AI Security Institute may provide some guidance on this. In addition, industry-led safety frameworks, which articulate explicit thresholds of risk based on evidence of measured capabilities, may offer a pathway for industry consensus. Agreement on which risks should be tracked and how they can be measured through evaluations, ideally harmonized to level the playing field through third-party auditing, offers an industry-inspired pathway that can inform governance.

On downstream impact, model developers may not have this information, especially in the case of open models, so other mechanisms for market monitoring may be required instead. Overall, policy can actively develop the infrastructure to turn these approaches into fully viable options in threshold design.

\subsection{Evidence-generating policy}
As the International Scientific Report notes, policymakers often make critical AI policy decisions with limited scientific evidence, posing an “evidence dilemma”: acting preemptively might lead to ineffective or unnecessary measures, while waiting for stronger evidence could leave society vulnerable. In response, policymakers should not remain idle: policy can actively accelerate the generation of evidence that can best-inform future policy decisions. By clarifying standards for evidence generation and evidence sharing, policy can reward developers with excellent safety practices while simultaneously strengthening industry competitiveness and building public trust. This balanced approach can create a positive feedback loop where transparency becomes a competitive advantage, incentivizing the continuous improvement of safety measures rather than treating safety and innovation as competing priorities.

To that end, I describe specific mechanisms policymakers should pursue to grow the evidence base and serve as the foundation of evidence-based AI policy based on my work in \citet{bommasani2025evidence}, which I made more accessible to policymakers in \citet{bommasani2025ca}.

\paragraph{Incentivize pre-release evaluation.}
Prior to releasing a model, model developers and external parties can evaluate the model to better understand how its release will impact society. Model developers should proactively measure risks prior to deployment: such evaluations can clarify the extent to which models pose marginal risks. 
Currently, many leading developers have signed onto the Frontier AI Safety commitments, which include a commitment to pre-release evaluation. 
While many developers do conduct these evaluations and publish results, recent reporting suggests the quality of these evaluations may be degrading (\eg the time afforded to internal evaluation given intense pressure to rapidly release models).\footnote{\url{https://www.ft.com/content/8253b66e-ade7-4d1f-993b-2d0779c7e7d8}}

Policymakers should incentivize the evaluation of models prior to release. 
To supplement their internal testing, developers should work with trusted external entities to provide independent assessments of these risks. 
To ensure this evidence can be appropriately interpreted, developers and external testers should publicly clarify the conditions involved in testing (\eg if testers are paid by developers or are restricted in their ability to disclose unfavorable results). 

\paragraph{Increase information sharing.}
Pre-release capability and safety evaluations provide insight into model capabilities and risks but are alone inadequate for assessing the societal impact of AI. More generally, key information about AI and its societal impact is siloed within AI companies. For example, the 2024 Foundation Model Transparency Index scores leading AI companies for their transparency and finds that companies score on average 31\% for publicly sharing information on how they mitigate risk \citep{bommasani2024fmti}. 
In light of these evidence gaps, policymakers in several jurisdictions have proposed or enacted transparency requirements. 
California’s AB 2013 requires developers of generative AI systems or services to publicly disclose information on training data. 
The EU AI Act requires providers of general-purpose AI models to document a broader set of information, including a public summary of training data. 
Critically, many policies globally emphasize companies sharing safety-related information specifically with governments. 
While this approach is valuable and there are many valid reasons to restrict information sharing to trusted actors \citep{kolt2024responsiblereportingfrontierai}, public transparency is essential for true accountability. 
In practice, it is often citizens, journalists, civil society organizations and academics who are at the forefront of identifying sociotechnical harm.

Policymakers should require leading AI companies to disclose more information about their safety practices to governments and, especially, to the public. 
First, transparency obligations should reflect informational needs: disclosing safety frameworks clarifies the steps developers take internally to mitigate risk, and disclosing evaluation results clarifies the current level of measured risk. 
Second, transparency obligations should reflect who will best use information: given the status quo of where accountability may come from, governments should prioritize improving public transparency in many cases. 
Finally, transparency obligations should be proportionate to not impose undue burdens on developers: criteria used to differentiate obligations should have specified processes for how they will update and avoid absolute reliance on fraught proxies such as training-time compute or monetary costs \citep{bommasani2023tiers, hooker2024limitationscomputethresholdsgovernance}.

\paragraph{Monitor post-deployment impacts.} 
Once AI systems are deployed, especially at scale, they impact society in a variety of ways. Yet transparency about these impacts are most opaque: the 2024 Foundation Model Transparency Index highlights that leading AI companies score just 15\% on average for what they publicly disclose on their models’ post-deployment impact.
Yet, these very impacts clarify how general-purpose technologies like today’s AI models shape society in specific ways. 
Recent approaches, such as the Anthropic Economic Index \citep{handa2025economictasksperformedai}, make partial inroads to clarify these impacts.
Beyond the efforts of individual companies, adverse event reporting databases, such as those proposed by the US National AI Advisory Committee \citep{naiac2023aers}, are critical to grow the collective evidence base by documenting concrete instances of harm in practice. 
While initial attempts like the AI Incident Database and the OECD AI Incidents Monitor provide coverage of adverse events, precise standards for (i) which entities are responsible for reporting, (ii) what constitutes an adverse event, and (iii) which parties need to be informed of an adverse event do not yet exist.

Policymakers should increase post-deployment monitoring of AI harms. 
Related domains, like cybersecurity, can guide how to design and implement post-deploying harm monitoring for AI \citep{stein2024rolegovernmentsincreasinginterconnected}. 
For example, the US Cybersecurity and Infrastructure Security Agency administers an incident reporting database under the Cyber Incident Reporting for Critical Infrastructure Act of 2022. 
Initiatives like this successfully address challenges that arise in AI such as how to coordinate disclosure of a vulnerability to many affected model developers and system providers \citep{longpre2025inhouse}, especially given issues with AI models and systems may generalize across different models \citep{wallace2019universal, zou2023universal}.

\paragraph{Protect third-party research.}
Important information about AI is often siloed within companies that develop and deploy the technology, but relevant expertise is more widely distributed. 
Third-party research by independent parties is an indispensable form of evidence \citep{raji2022audit, longpre2025inhouse}, given the independence from commercial incentives and, thereby, the greater trust it may confer. 
To conduct research on released AI models and deployed AI systems, third-party researchers require access to these technologies. 
While most major foundation models are available to researchers in some form, usually via an API (\eg OpenAI’s GPT-4o) or via their weights (\eg Meta’s Llama 3.3), current practices that govern access inhibit third party research. 
The terms of service for many leading developers include clauses that, potentially inadvertently, suppress third-party research. 
To rigorously measure risks, researchers may violate the terms of service, which is intended to discourage malicious use, in turn risking platform bans or legal action \citep{klyman2024fas, klyman2024acceptableusepoliciesfoundation}. 

Policymakers should create shields to protect good-faith third-party AI research. 
In May 2024, 350 leading AI researchers and advocates signed an open letter that advocated for a safe harbor to protect such research.\footnote{\url{https://sites.mit.edu/ai-safe-harbor/}}  
The proposed safe harbor draws inspiration from cybersecurity, where similar safe harbor provisions exist at the federal level in the United States. 
In general, if researchers follow established rules of conduct and responsibly disclose issues with AI technologies to advance the public interest, then indemnification from legal liability would create the appropriate incentives to grow the evidence base \citep{longpre2024safe}.

\paragraph{Prioritize well-evidenced interventions.}
Successfully mitigating the risks of AI requires sociotechnical approaches which recognize the role of humans, organizations, and technology in a defense-in-depth approach. Defense-in-depth means layering technical model-level interventions with a broader suite of societal interventions elsewhere in the supply chain. Critically, for different risks, the evidence for whether intervening at a specific point in the supply chain should inform where policy targets interventions.

Policymakers should strengthen societal defenses, especially given clear evidence of unmitigated risk even absent AI capabilities. 
For many threat vectors surrounding malicious use, AI capabilities are exploited as an intermediary step in a more complex process (\eg synthesizing disinformation that is then disseminated via social media networks, obtaining information that is then used to build bioweapons). 
To address potential marginal risk from AI for these threat vectors, downstream interventions may be effective while also reducing pre-existing non-AI risk for the same threat vectors. 
For example, prevalent cybersecurity practices are insufficient according to experts: reinforcing codebases would not only address these long-standing issues, but also better prepare society for AI-related cyberattacks.

\paragraph{Bridge fragmented subcommunities.}
Currently, the AI community is fractured with many divergent views on how to approach risk and policy. At present, the degree of consensus or the lack thereof remains unclear for many critical questions (\eg what forms of evidence are credible for supporting the claim that frontier models pose marginal risk in enabling bioweapons development?). 
Forging consensus will be difficult given the striking divides in the AI community on core issues (\eg the rate of technological progress) but currently the deliberative processes to even attempt consensus formation do not exist.

Policymakers should catalyze the formation of scientific consensus. Scientific consensus, including on areas of uncertainty or immaturity, is a powerful primitive for better AI policy. 
Existing efforts like the International Scientific Report, United Nations High-level Advisory Body on Artificial Intelligence, and global network of AI Safety Institutes are key initial steps to build international consensus. 
Close policymaker-scientist partnerships can accelerate this process and other industries provide useful historical precedent. 
For example, the United Nations High-level Advisory Body on Artificial Intelligence recommends forming an International Scientific panel on AI, which could prepare consensus reports akin to the efforts of the Intergovernmental Panel on Climate Change (IPCC). 

\section{Conclusion}
This chapter concludes the substantive research covered in this dissertation.
At its core, it imagines a more robust research-policy interface by concretizing work in both directions.
In the research-to-policy direction, the focus of my work has been on generating, synthesizing and communicating useful evidence as required for evidence-based AI policy.
A critical future direction is to build robust scientific consensus: while I have done this in some of my works \citep{kapoor2024societal, bommasani2025evidence, bommasani2025ca}, more robust scientific consensus formation processes will need to be built for AI policy to emulate the rigor of evidence-based AI policy seen in other domains (\eg the IPCC's work on climate policy).
In the policy-to-research direction, the focus of my work has been to articulate how policy can accelerate evidence generation and stimulate scientific consensus formation.
A critical future direction is to inexorably install evidence-based AI policy as a non-optional determinant for policy,\footnote{The bipartisan Evidence Act of 2018 may prove to be critical for this outcome.} especially to avoid policy that is poorly founded in evidence \citep{guha2023ai}.
However, more than anything else, the success of this research-policy interface will hinge on mutual trust, repeated engagement, and improved outcomes, which I will continue to build with many fellow travelers in the coming years.

\chapter{Conclusion}\label{chapter:conclusion}
\chaptermark{\small Conclusion}
I began my PhD in September 2020 at a time of crisis, where policymakers had to rise to the occasion of addressing challenges at the local (California wildfires), national (racial protests) and global (COVID-19 pandemic) levels.
At the time, I did not expect to interact with a single policymaker during my PhD.
After all, computer science has been disinterested in and divorced from public policy throughout its history.
Yet, what has transpired over the past five years, both in my PhD and in the world, was entirely unforeseeable.

As I end my PhD in June 2025, I have worked directly with policymakers in each of those jurisdictions.
Policymakers must rise to the occasion yet again, now finding common cause in the pursuit of AI governance.
As one of a small handful of academics materially improving AI policy, I believe a strong
research-policy interface that bridges these often distant worlds is the critical path
for better outcomes.
This dissertation puts forth this vision, one rooted in productive multidisciplinary collaboration, to show new avenues for academic computer science to be catalytic in producing better public outcomes.
I close this dissertation by challenging future academics to pursue this vision for scholarship, unencumbered by the silos of disciplinary expectations and uncompromising in bettering our collective society.
\newpage

\appendix

\bibliographystyle{acl_natbib}
\bibliography{references, references-1, references-2,main,all}

\end{document}